\documentclass{sig-alternate_mod}

\usepackage{url,mathrsfs,algorithm}
\usepackage{amsmath,wrapfig,color,bigfoot}
\usepackage{amsfonts}
\usepackage{graphicx}
\usepackage{subfigure}

\begin{document}

\title{Min-Max Kernels}

\numberofauthors{1}
\author{
\alignauthor
{Ping Li}\\
       \affaddr{Department of Statistics and Biostatistics}\\
       \affaddr{Department of Computer Science}\\
       \affaddr{Piscataway, NJ 08854, USA}\\
       \email{pingli@stat.rutgers.edu}
}

\maketitle
\begin{abstract}

The min-max kernel is a generalization of the popular \textbf{resemblance} kernel (which is designed for  binary data). In this paper, we demonstrate, through an extensive classification study using kernel machines, that the min-max kernel often provides an effective measure of similarity for nonnegative data. As the min-max kernel is nonlinear and might be difficult to be used for industrial applications with massive data, we show that the min-max kernel can be linearized via hashing techniques. This allows practitioners to apply min-max kernel to large-scale applications using well matured linear algorithms such as linear SVM or logistic regression.

\vspace{0.08in}

\noindent The previous  remarkable work on {\em consistent weighted sampling (CWS)} produces samples in the form of ($i^*, t^*$) where the $i^*$ records the location (and in fact also the weights) information analogous to the samples produced by classical minwise hashing on binary data. Because the $t^*$ is theoretically unbounded, it was not immediately clear how to  effectively implement CWS for building large-scale linear classifiers. In this paper, we provide a simple solution by  discarding $t^*$ (which we refer to as the ``0-bit'' scheme). Via an extensive empirical study, we show that this 0-bit scheme does not lose essential information. We then apply the ``0-bit'' CWS for building linear classifiers to approximate min-max kernel classifiers, as extensively validated on a wide range of publicly available classification datasets.

\vspace{0.08in}

\noindent We expect this work will generate interests among data mining practitioners who would like to efficiently utilize the nonlinear information of  non-binary and nonnegative data.

\end{abstract}

\section{Introduction}

Nonnegative data are  common in practice and  the existence of negative entries in a dataset is often  due to  shifting or normalization. In this paper we show that  the  \textbf{min-max kernel} can provide an effective measure of  similarity for nonnegative data and  should be useful for building effective large-scale data mining tools via   hashing techniques.

\vspace{0.08in}

Given two nonnegative data vectors,  $u,v\in\mathbb{R}^D$, we define
\begin{align}\label{eqn_MM}
\textbf{min-max}:\hspace{0.1in} K_{MM}(u,v) = \frac{\sum_{i=1}^D \min\{u_i,\ v_i\}}{\sum_{i=1}^D \max\{u_i,\ v_i\}}
\end{align}
which is a generalization of the well-known {\rm resemblance}:
\begin{align}\label{eqn_resemblance}
\textbf{resemblance}:\hspace{0.0in} K_{R}(u,v) = \frac{\sum_{i=1}^D 1\{u_i>0 \text{ and }  v_i>0\}}{\sum_{i=1}^D 1\{u_i>0 \text{ or } v_i>0\}}
\end{align}
The resemblance is a popular measure of similarity for binary data~\cite{Proc:Broder,Proc:Li_Konig_WWW10}. The prior work~\cite{Proc:HashLearning_NIPS11} used the term ``resemblance kernel'' because the resemblance can be written as the (expectation) of an inner product (and hence it is a positive definite kernel).  It will be soon clear that $K_{MM}$ (\ref{eqn_MM}) can also be written as the expectation of an inner product.

Readers (e.g., those from computer vision) probably have realized that the min-max kernel defined in (\ref{eqn_MM}) is  related to the following so-called {\em intersection kernel~\cite{Maji_CVPR08}}:
\begin{align}\label{eqn_KI}
\textbf{intersection}:\hspace{0.1in} &K_{I}(u,v) = \sum_{i=1}^D \min\{u_i,\ v_i\},\\\notag
&\sum_{i=1}^D u_i = 1,\ \ \ \sum_{i=1}^D v_i = 1
\end{align}
In this paper, we will extensively compare the min-max kernel with the intersection kernel in the context of kernel machines for classification. Interestingly, for most  datasets in our experimental study, the min-max kernel outperforms the intersection kernel, and in some cases significantly so. Of course, another advantage of the min-max kernel is the existence of hashing techniques~\cite{Report:Manasse_CWS10,Proc:Ioffe_ICDM10} to approximate this nonlinear kernel by linear kernel (at least conceptually). \\

The sum-to-one normalization in the definition of intersection kernel (\ref{eqn_KI})  appears natural, since the data vectors (e.g., $u$ and $v$) were treated as histograms when   the intersection kernel was designed. For our curiosity, we also define, what we call, the ``normalized min-max kernel'' as follows:
\begin{align}\label{eqn_NMM}
\textbf{n-min-max}:\hspace{0.1in} &K_{NMM}(u,v) = \frac{\sum_{i=1}^D \min\{u_i,\ v_i\}}{\sum_{i=1}^D \max\{u_i,\ v_i\}}\\\notag
&\sum_{i=1}^D u_i = 1,\ \ \ \sum_{i=1}^D v_i = 1
\end{align}
Our experiments will show that, for most datasets, this normalization step  only affects the classification accuracies very marginally, although there are also exceptions.  In this paper, we often use ``\textbf{min-max kernel\textbf{s}}'' to refer to both the min-max kernel and the n-min-max kernel. Note that the normalization step is conducted before applying hashing, which means that these two kernels are no different as far as the research on hashing is concerned.  \\

It is worth mentioning that  the above three kernels (min-max, intersection, and n-min-max) have no tuning parameters. Thus, it is often  possible to   further improve the performance by, for example, using multiple kernels or kernels combined in a special fashion (e.g., the  CoRE kernels~\cite{Proc:Li_UAI14} by multiplying resemblance with correlation).

We will compare these three types of parameter-free kernels with the basic (tuning-free) kernel:
\begin{align}
\textbf{linear}:\hspace{0.1in} &K_\rho(u,v) = \sum_{i=1}^D u_i v_i,\\\notag
& \sum_{i=1}^D u_i^2 = 1,\ \ \ \sum_{i=1}^D v_i^2 = 1
\end{align}
For convenience, we enforce the normalization (to unit length) because in practice  (e.g., when running linear SVM) the normalization step is typically recommended.

\vspace{0.08in}

The min-max kernel was sparsely discussed in the literature~\cite{Report:Manasse_CWS10,Proc:Ioffe_ICDM10}. In contrast, the resemblance kernel (\ref{eqn_resemblance}) has been widely used in practice on binary (or binarized) data~\cite{Proc:Broder,Proc:Broder_WWW97,Article:Urvoy08,Proc:Das_WWW07,Proc:Pandey_WWW09,Proc:Chierichetti_KDD09,Proc:Cherkasova_KDD09,Article:Forman09,Proc:Cormode_SIGMOD05,Proc:Koudas_SIGMOD06,Proc:Henzinger_SIGIR06,Proc:Bayardo_WWW07}. For example, \cite{Proc:HashLearning_NIPS11} demonstrated the  use of $b$-bit minwise hashing~\cite{Proc:Li_Konig_WWW10} for training large-scale (resemblance kernel) SVM and logistic regression.

\vspace{0.08in}

\noindent\textbf{Summary of our contributions:}\hspace{0.2in} This paper aims at addressing several interesting and important issues regarding the use of min-max kernels for data mining applications:
\begin{enumerate}
\item {\em Why using min-max kernels?}\hspace{0.1in}  Table~\ref{tab_KernelSVM} and Figures~\ref{fig_KernelSVM_1} to~\ref{fig_KernelSVM_3} provide an extensive empirical study of kernel SVMs for classification on a sizable collection of public datasets, for comparing linear kernel, min-max kernel, n-min-max kernel, and intersection kernel. The results illustrate the advantages of the min-max kernels over the linear kernel as well as the intersection kernel.
\item {\em The ``0-bit'' CWS hashing for min-max kernels.}\hspace{0.1in}\\ The remarkable prior work on {\em consistent weighted sampling (CWS)} provides a recipe to sample min-max kernels (i.e., the collision probability of the samples is the min-max kernel), in the form of ($i^*, t^*)$. Because $t^*$ is theoretically unbounded, it was not immediately clear how to effectively implement a ``$b$-bit'' version of CWS which is needed in order to apply the method for large-scale industrial applications. We provide a (surprisingly) simple solution by completely discarding $t^*$ (after hashing), which we refer to as the ``0-bit'' scheme and is validated by a large set of experiments.
\item {\em Large-scale learning with (modified) CWS hashing}.\hspace{0.1in} In light of our contributions 1 and 2, we apply the proposed 0-bit CWS hashing  for efficiently building large-scale linear classifiers approximately in the space of min-max kernels, as verified by extensive experiments.
\end{enumerate}

\section{Kernel SVM Experiments}\label{sec_KernelSVM}

In this section, we present an experimental study for classification using kernel machines based on the four types of kernels we have introduced: the linear kernel, the min-max kernel,  n-min-max kernel, and the intersection kernel. To simplify the experimental procedure, we use LIBSVM {\em pre-computed kernel} functionality and $l_2$-regularization. Table~\ref{tab_KernelSVM} summarizes the test classification accuracies.

\begin{figure}[h!]
\begin{center}
\mbox{
\includegraphics[width=1.75in]{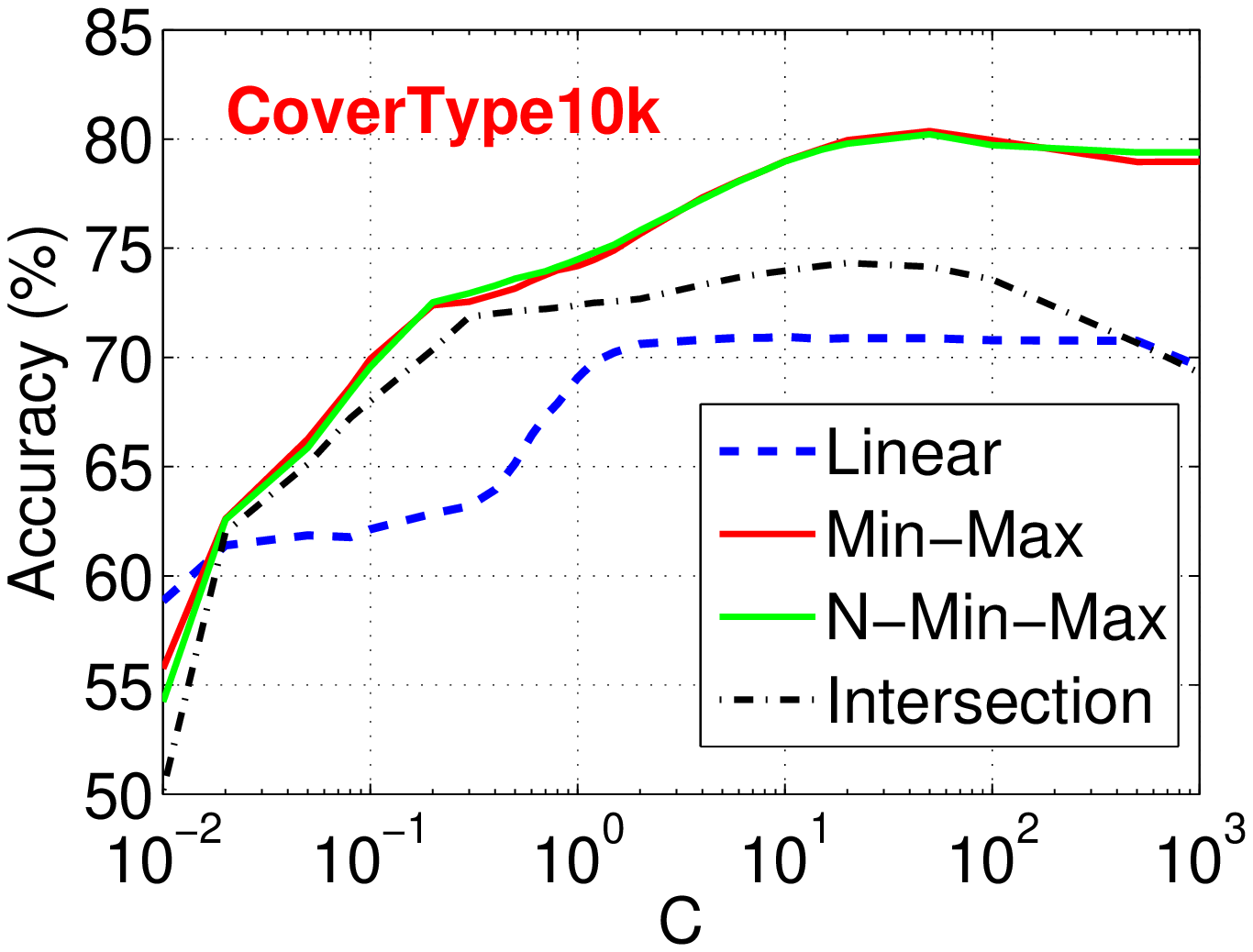}\hspace{-0.14in}
\includegraphics[width=1.75in]{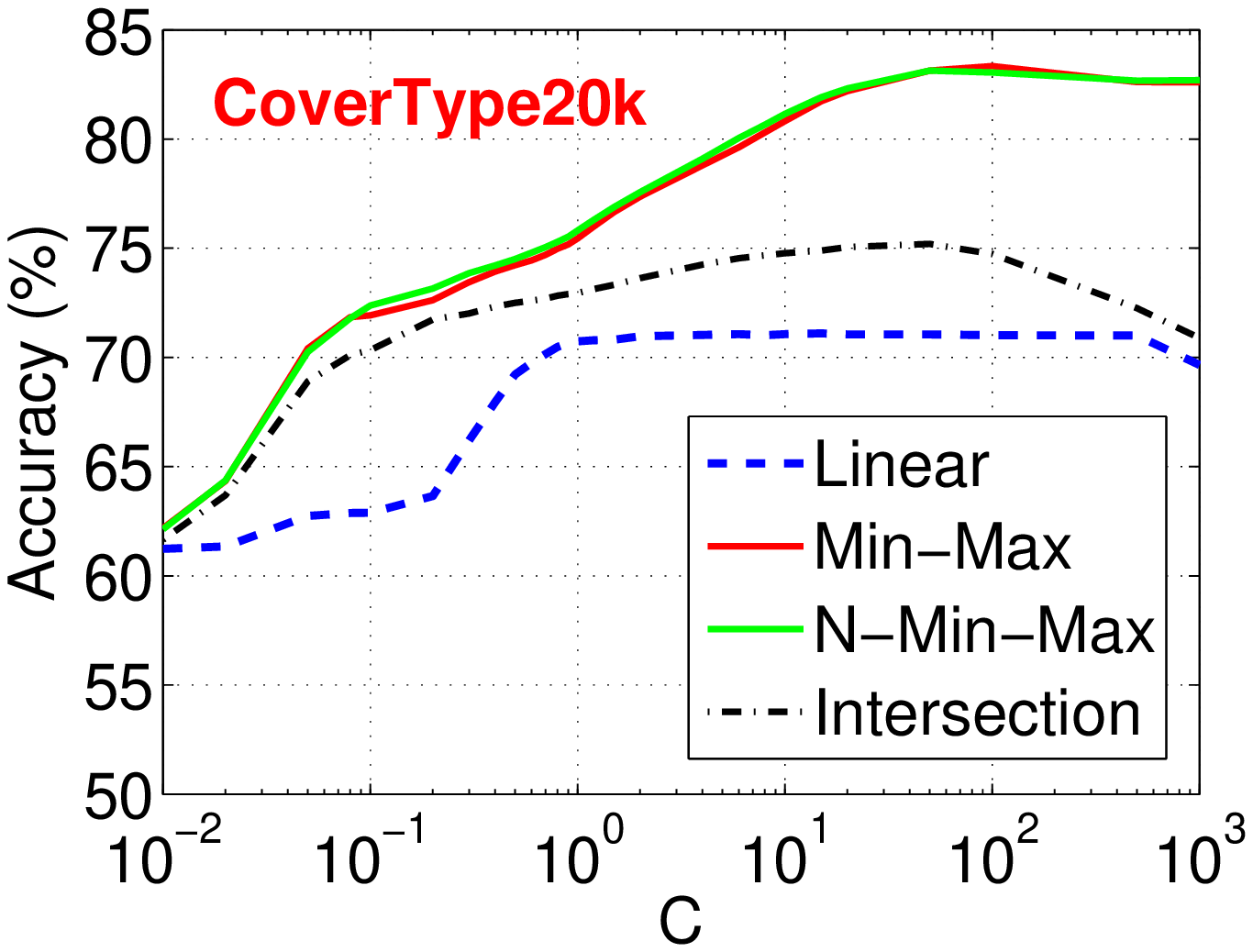}
}
\mbox{
\includegraphics[width=1.75in]{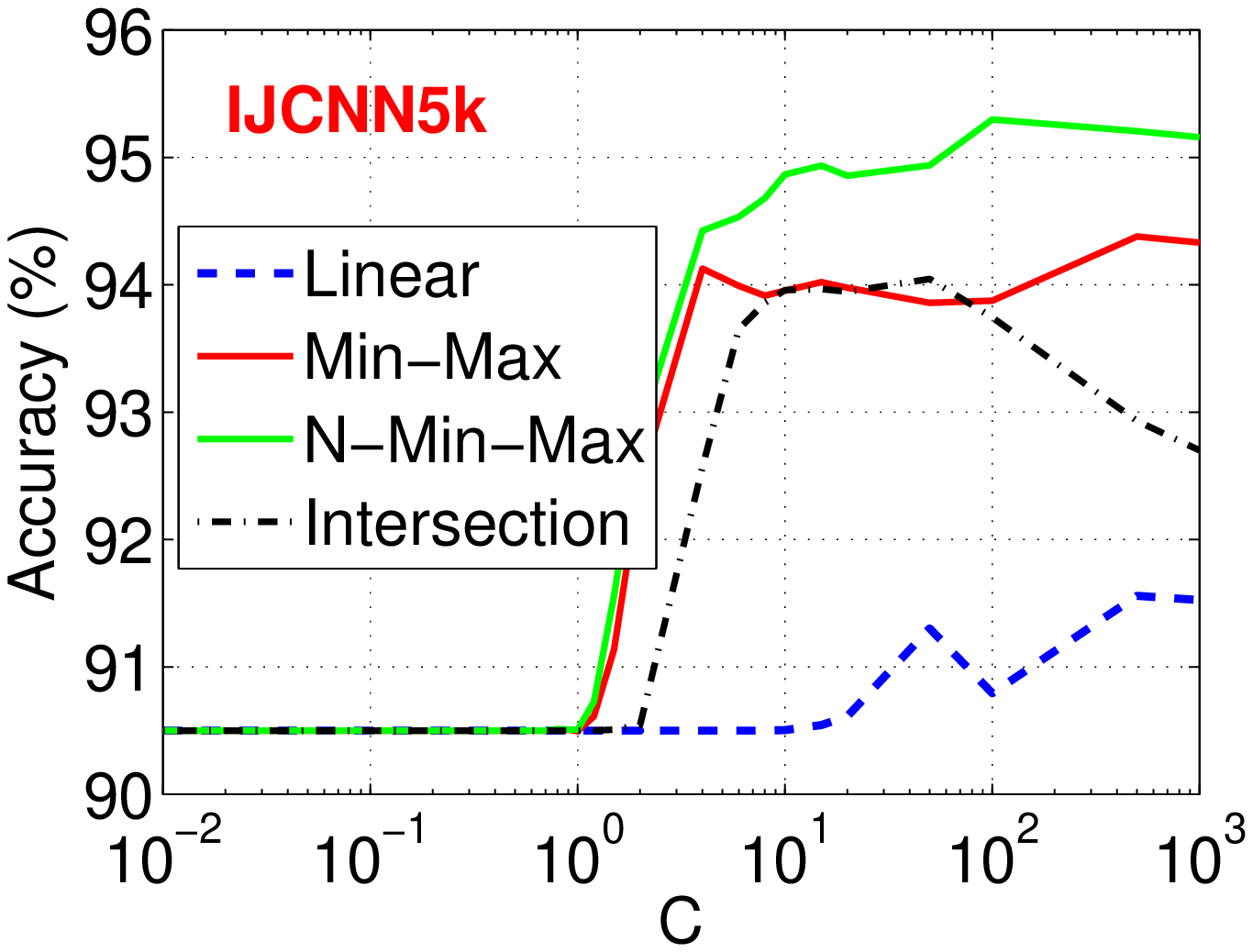}\hspace{-0.14in}
\includegraphics[width=1.75in]{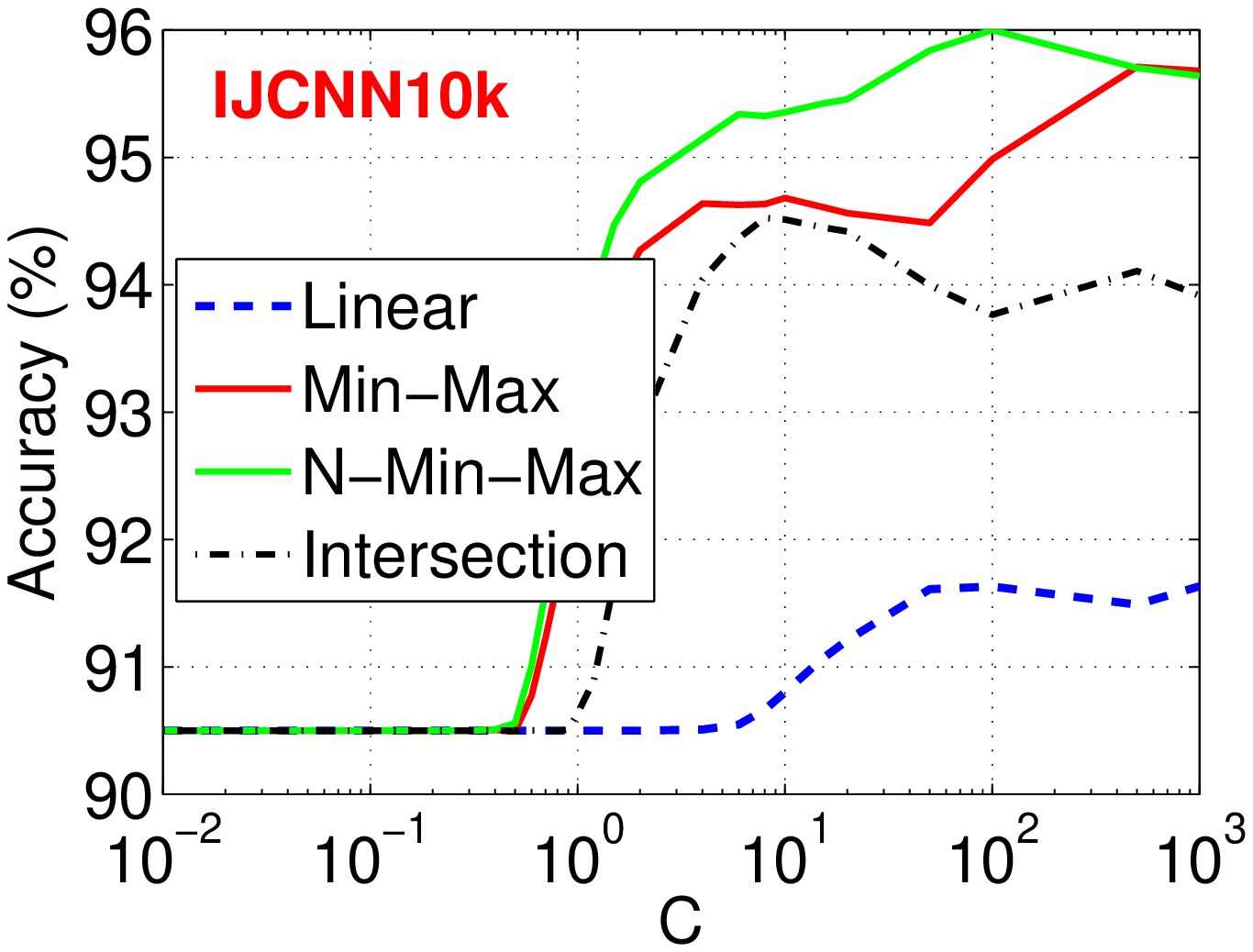}
}
\mbox{
\includegraphics[width=1.75in]{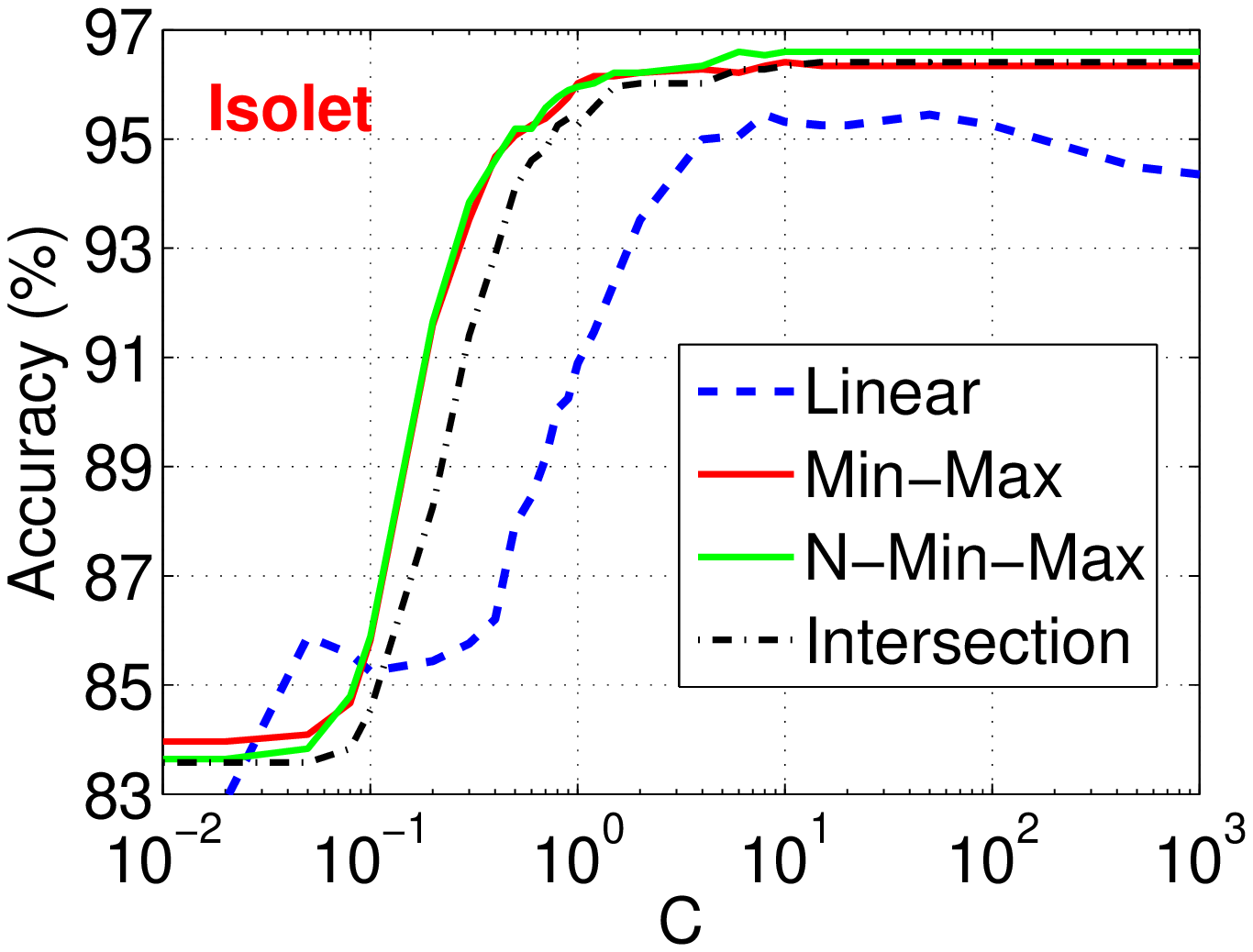}\hspace{-0.14in}
\includegraphics[width=1.75in]{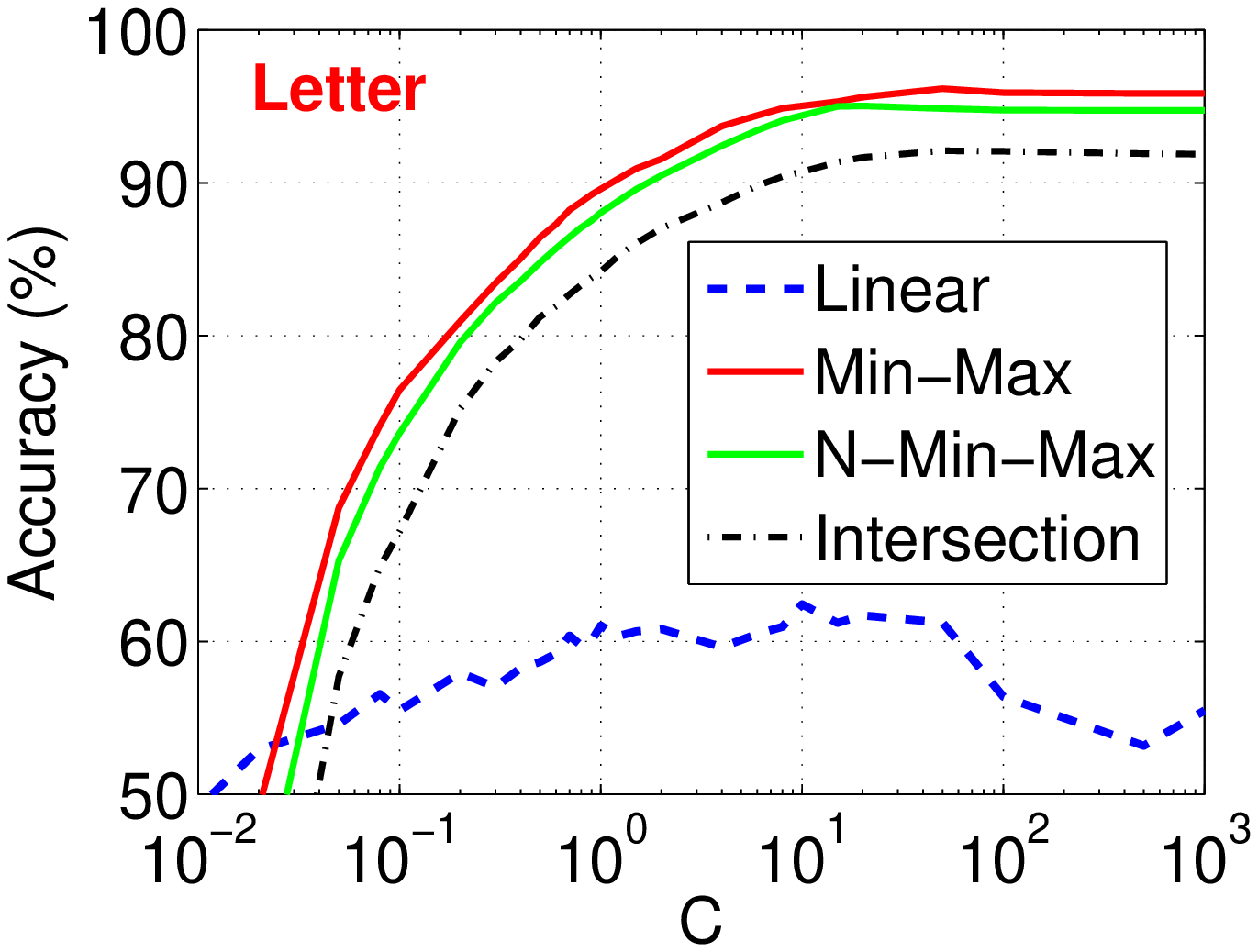}
}

\mbox{
\includegraphics[width=1.75in]{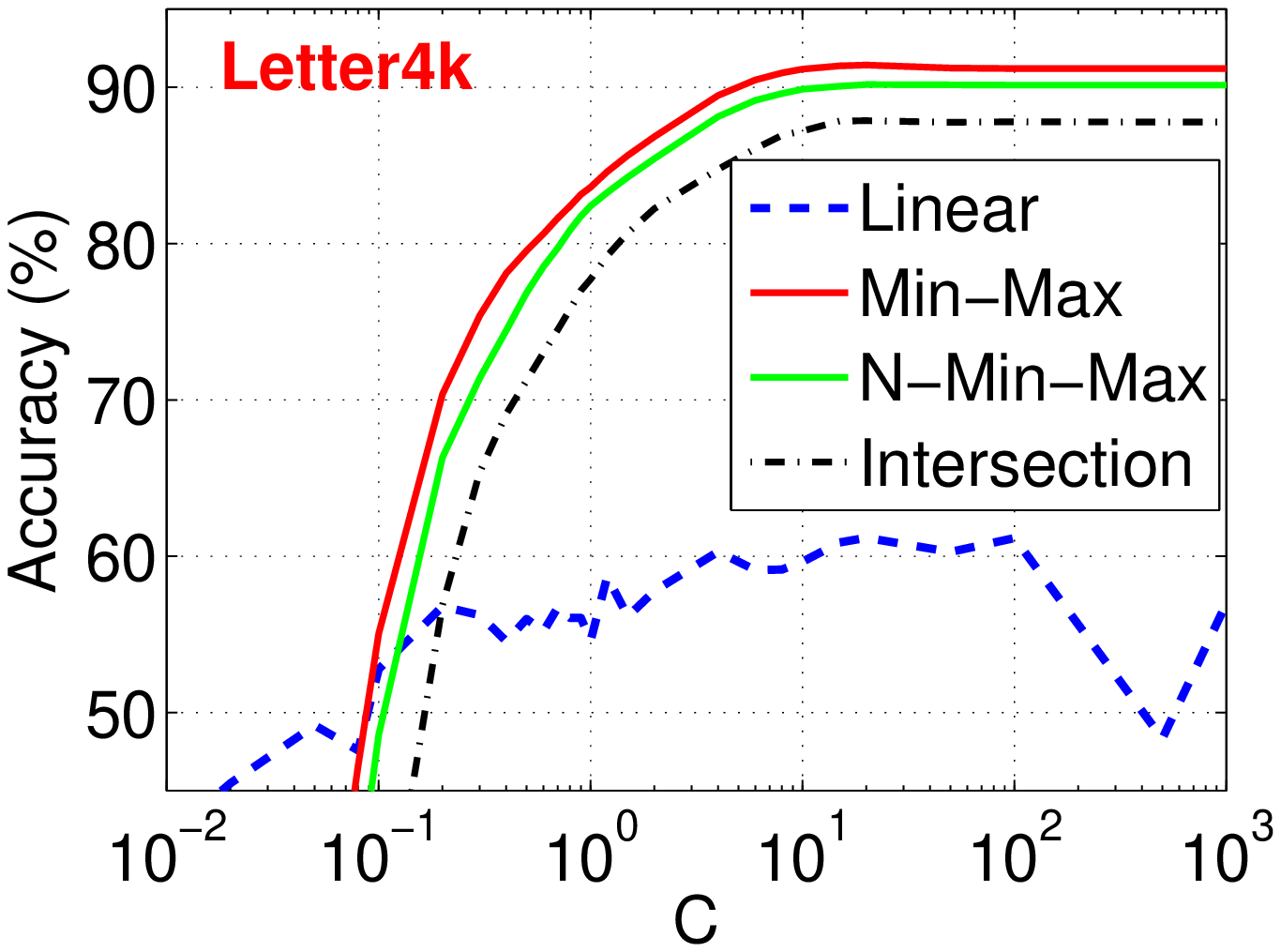}\hspace{-0.14in}
\includegraphics[width=1.75in]{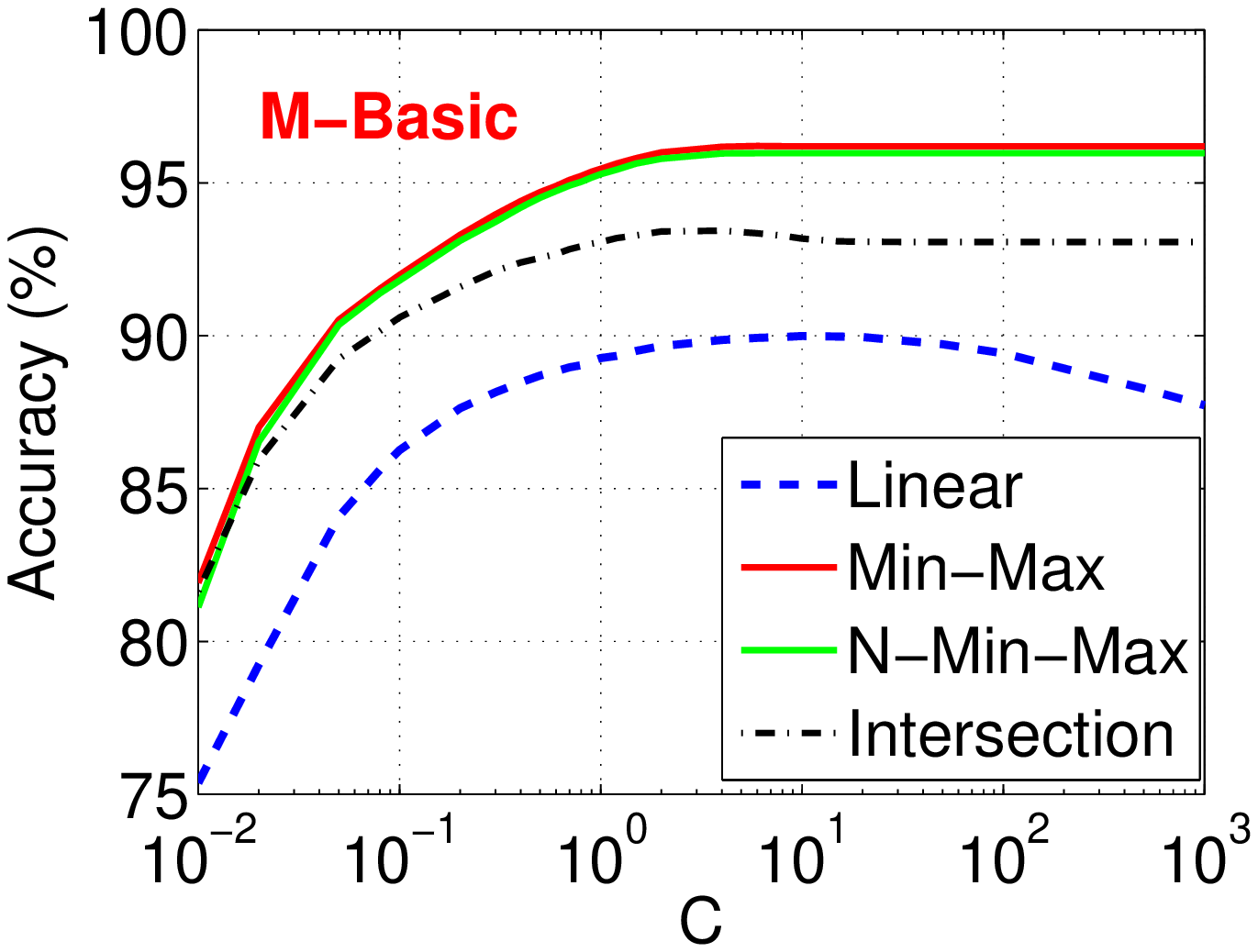}
}

\mbox{
\includegraphics[width=1.75in]{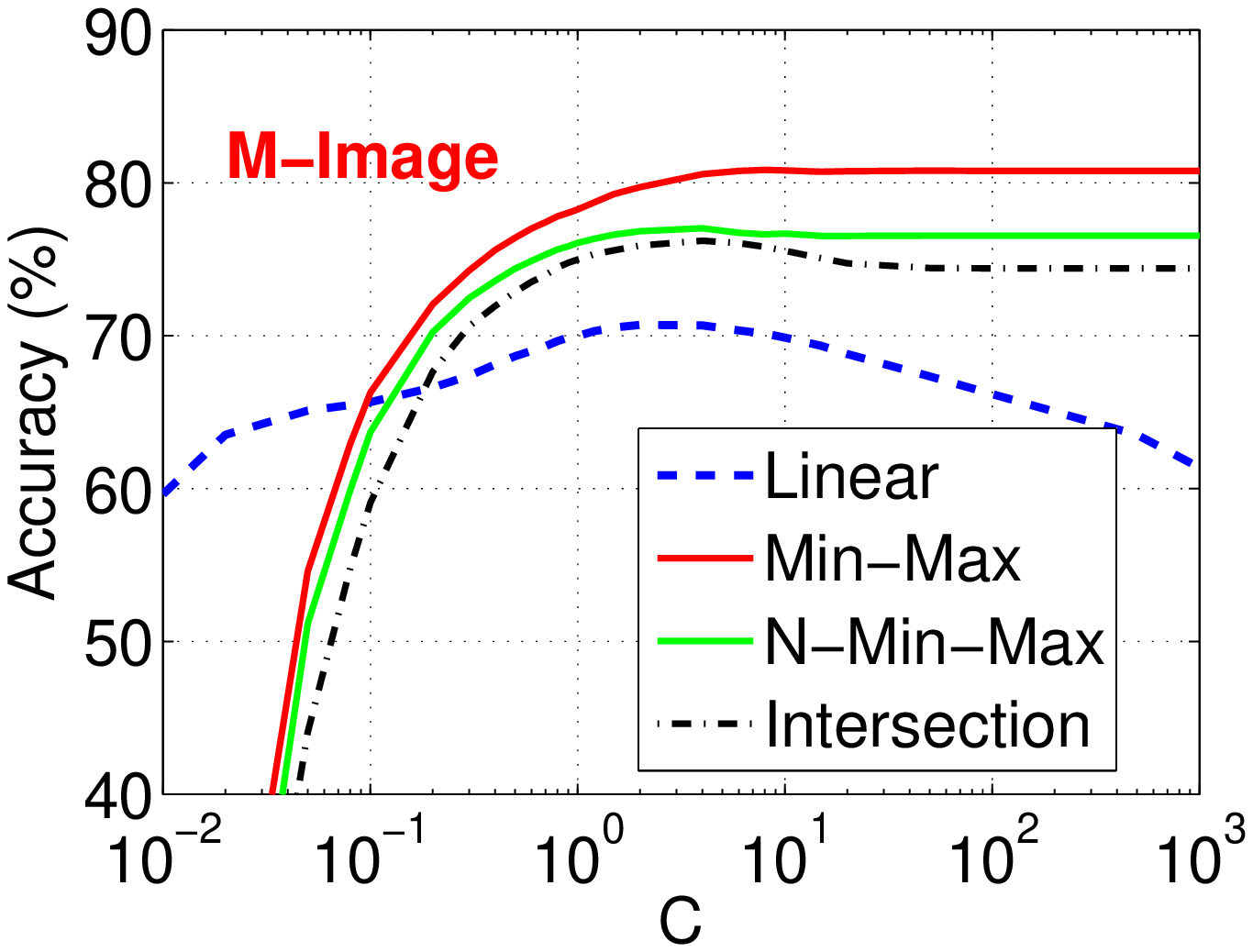}\hspace{-0.14in}
\includegraphics[width=1.75in]{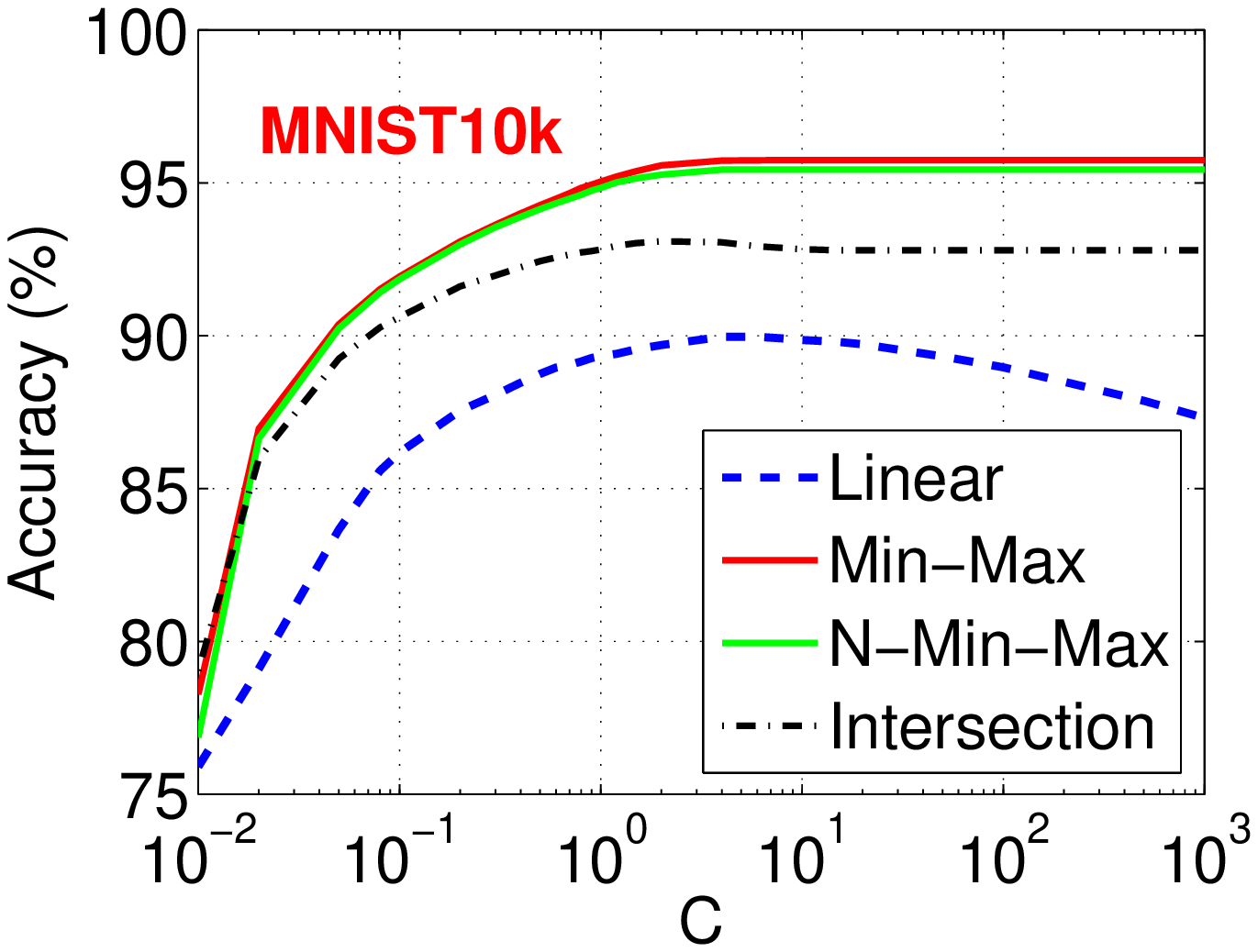}
}

\end{center}
\vspace{-0.3in}
\caption{Test classification accuracies  for four types of kernels using $l_2$-regularized  SVM (with a tuning parameter $C$, i..e, the x-axis.). Each panel presents the results for one particular dataset (see more data information in Table~\ref{tab_KernelSVM}). The two solid curves represent the min-max kernel (red, if color is available) and the  n-min-max kernel (green, if color is available). The dashed curve (blue) and the dot dashed (black) curve represent, respectively, the linear kernel and the intersection kernel. See Figures~\ref{fig_KernelSVM_2} and~\ref{fig_KernelSVM_3} for the results on more datasets.     }\label{fig_KernelSVM_1}
\end{figure}

\begin{figure}[t]
\begin{center}

\mbox{
\includegraphics[width=1.75in]{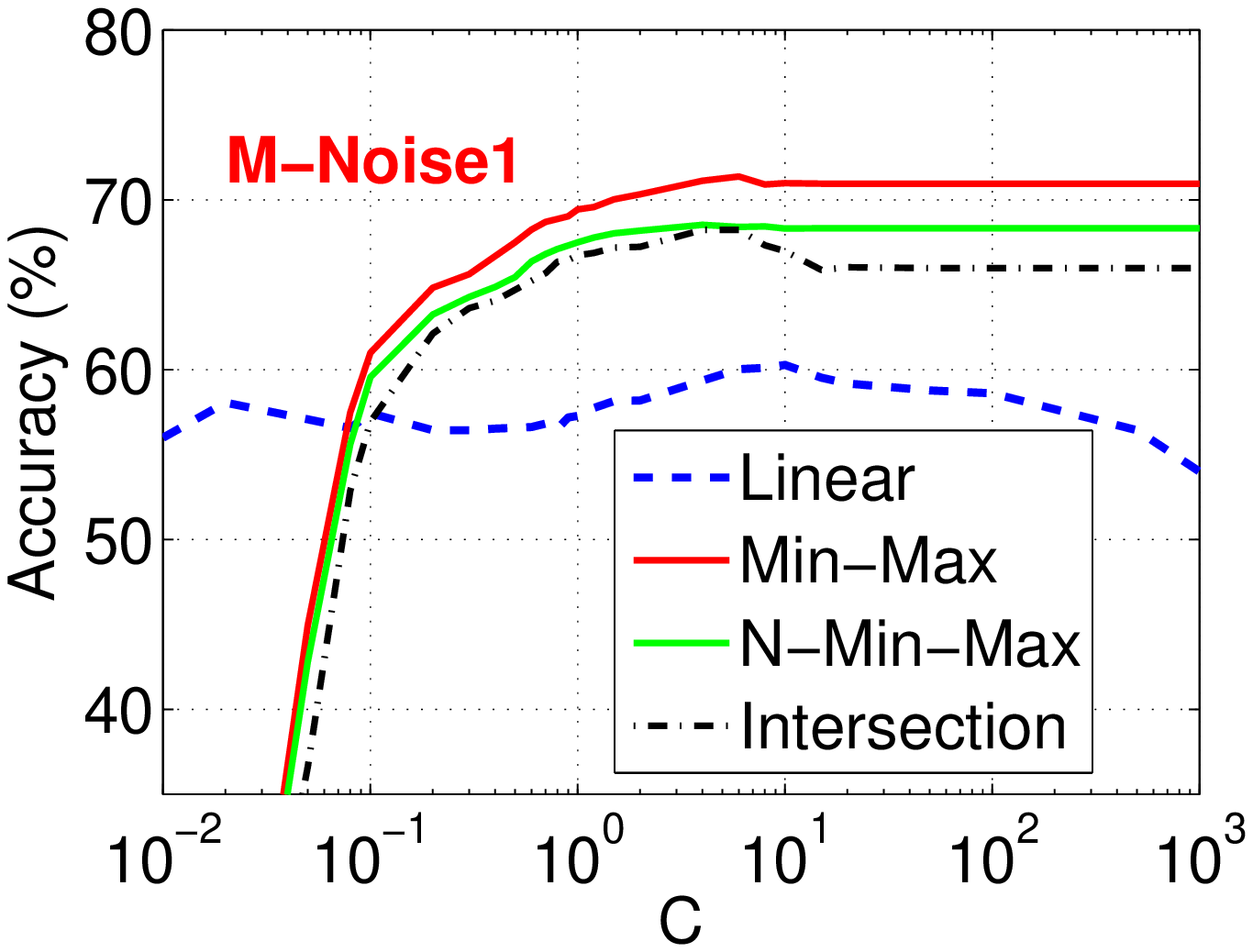}\hspace{-0.14in}
\includegraphics[width=1.75in]{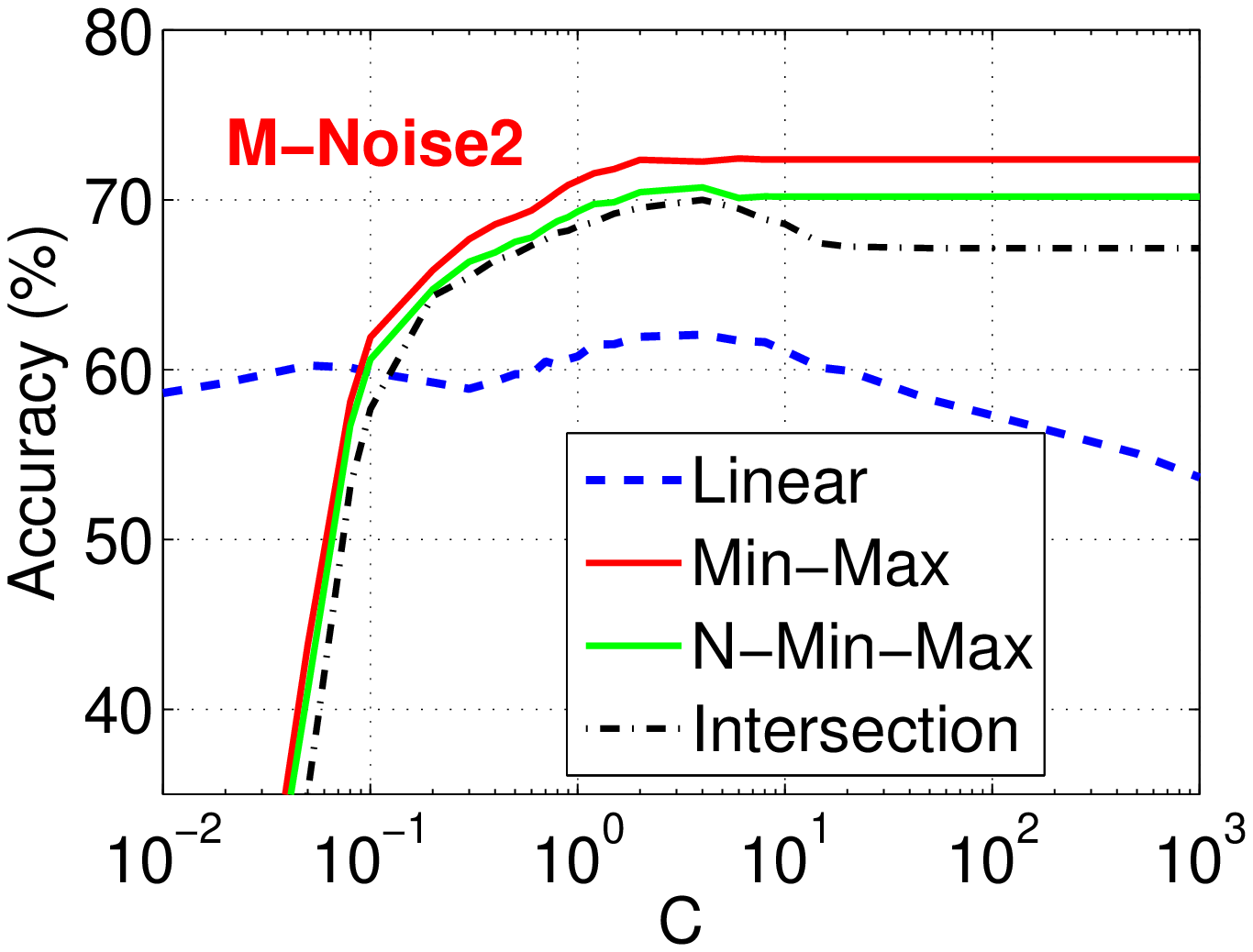}
}
\mbox{
\includegraphics[width=1.75in]{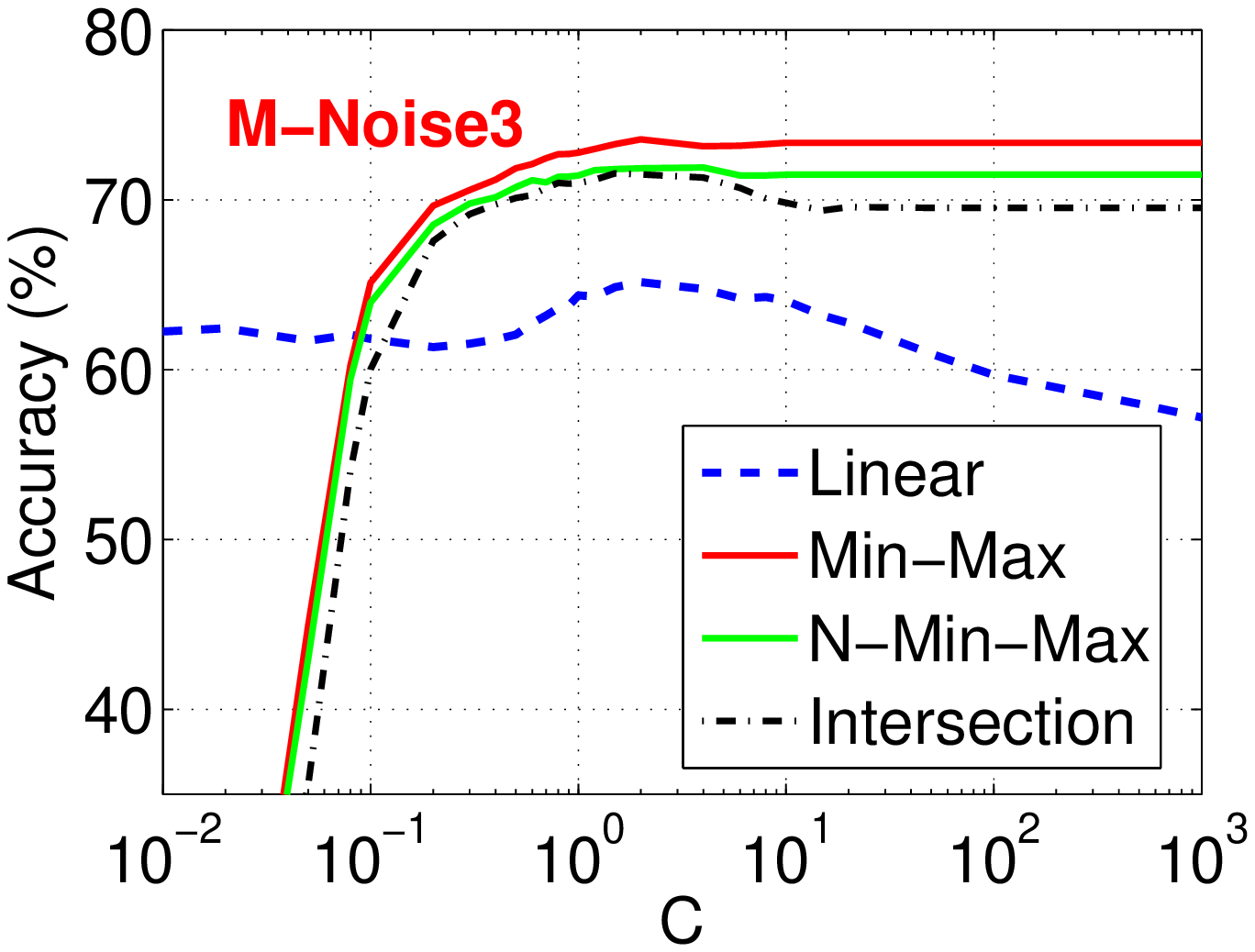}\hspace{-0.14in}
\includegraphics[width=1.75in]{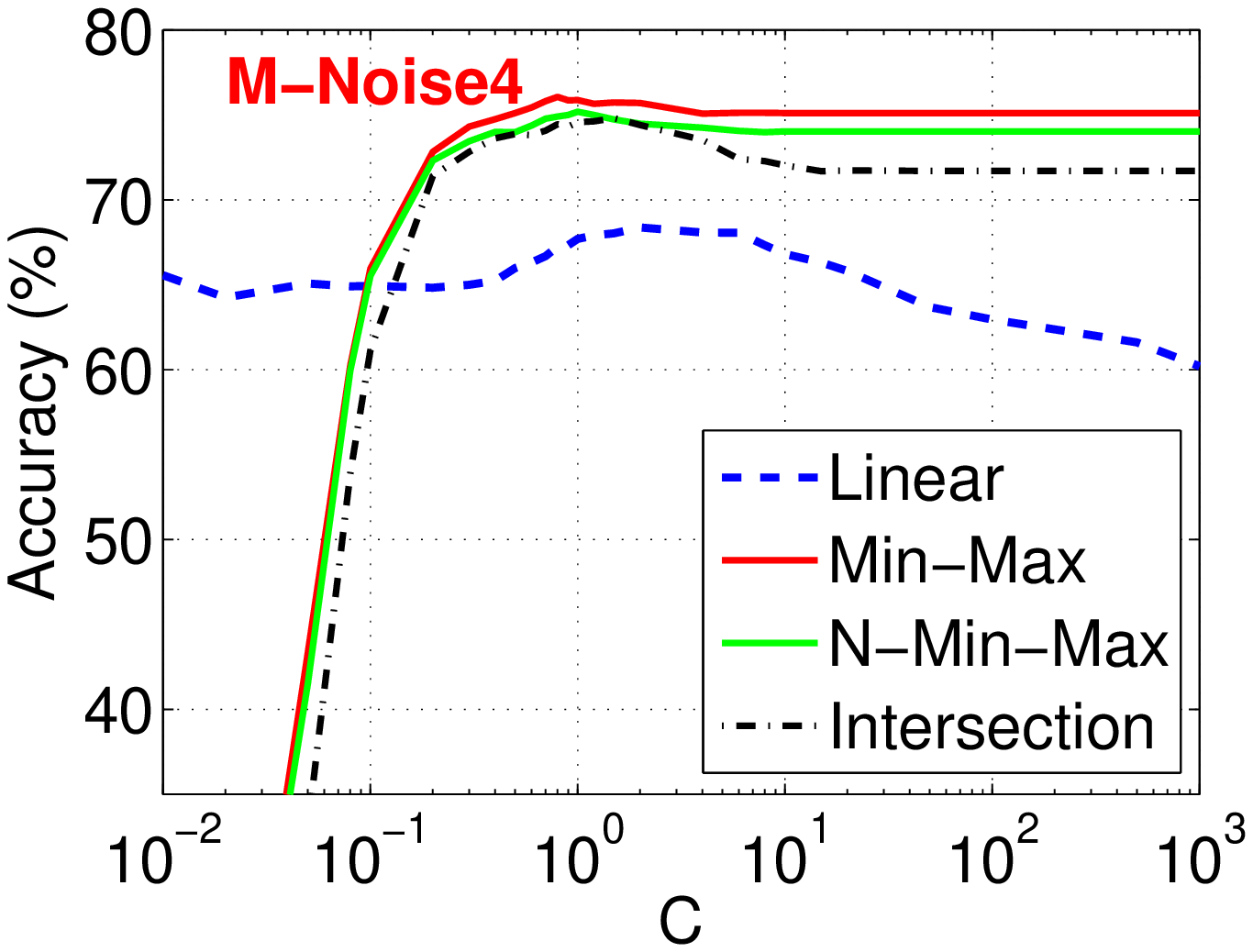}
}
\mbox{
\includegraphics[width=1.75in]{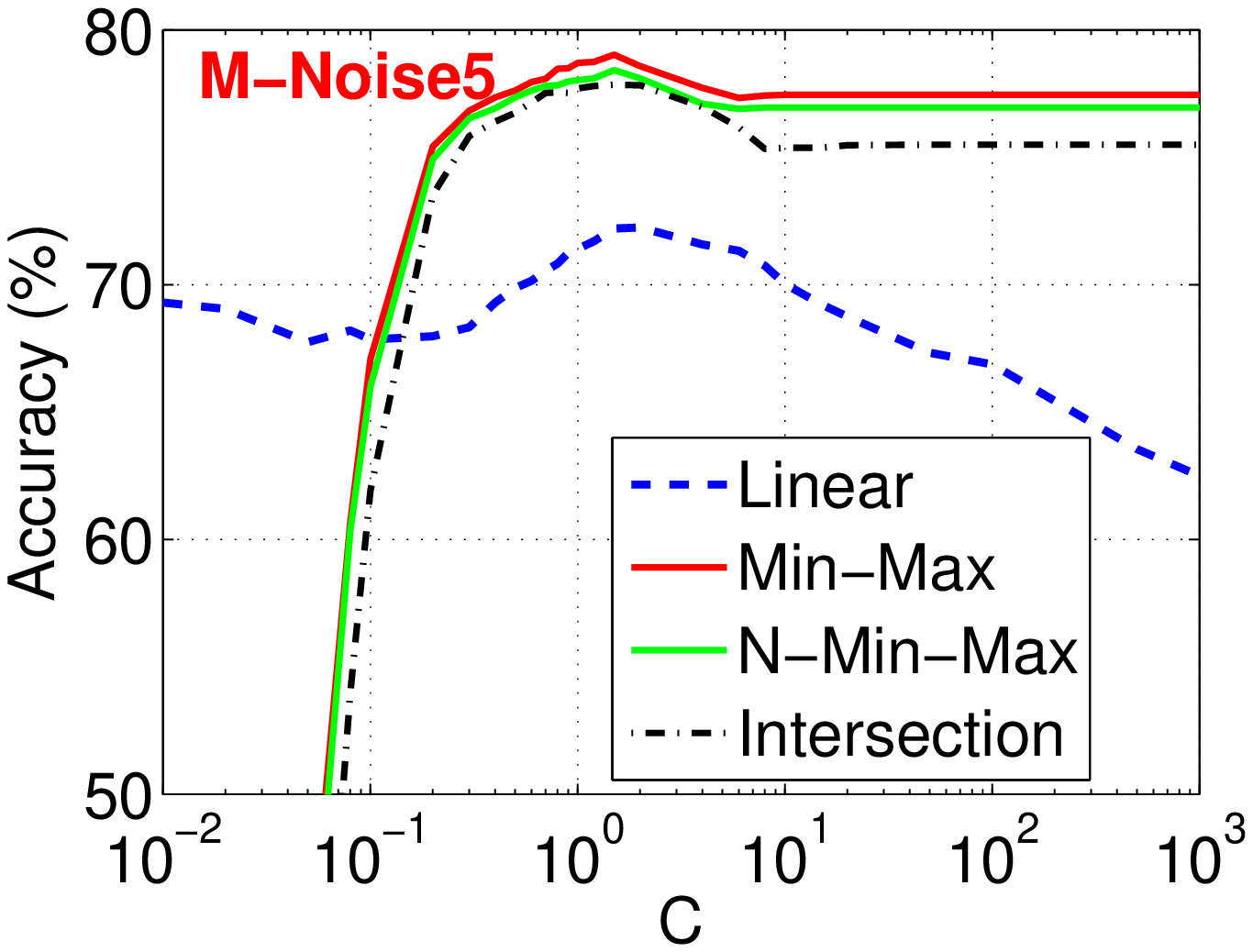}\hspace{-0.14in}
\includegraphics[width=1.75in]{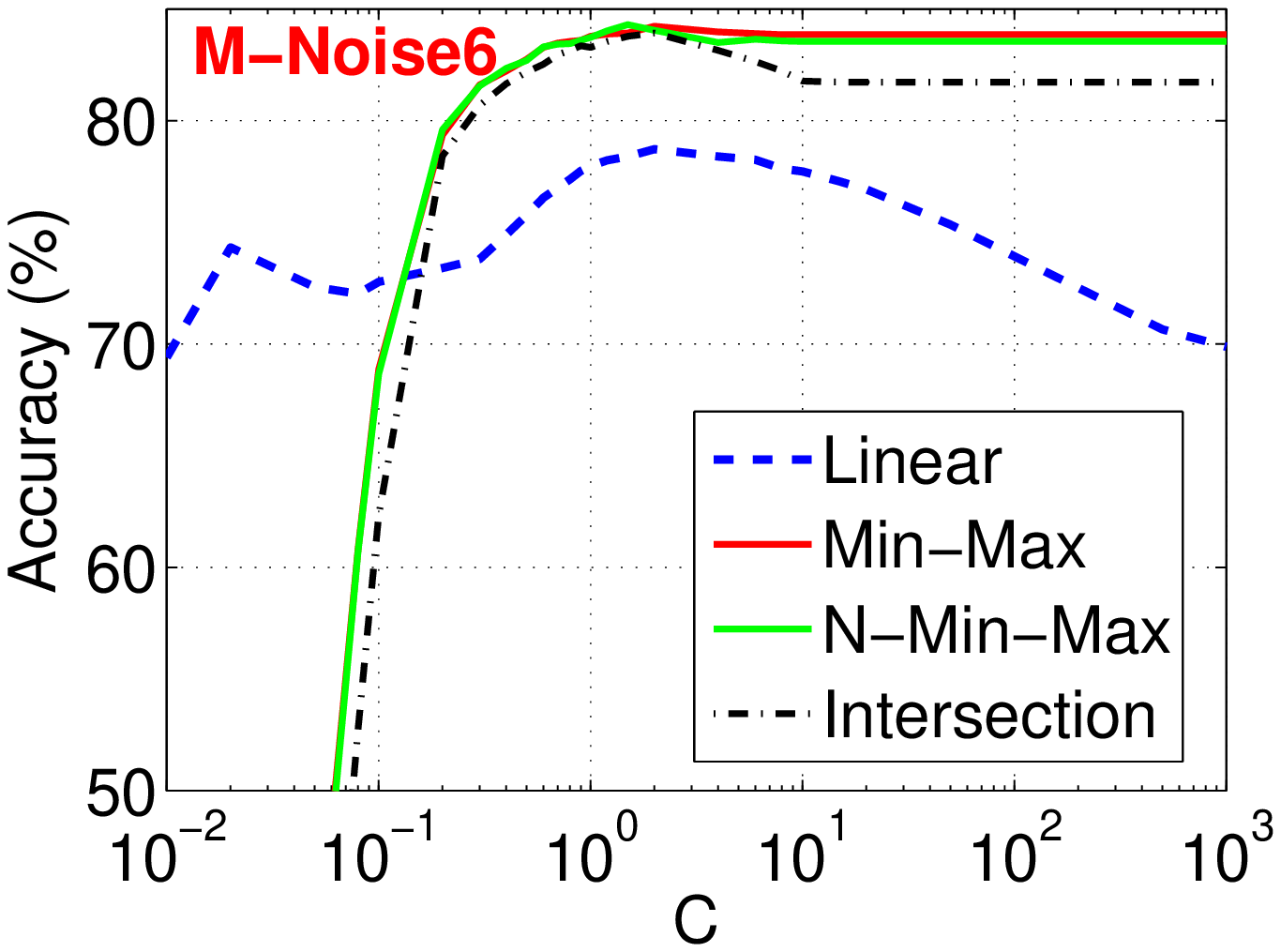}
}
\mbox{
\includegraphics[width=1.75in]{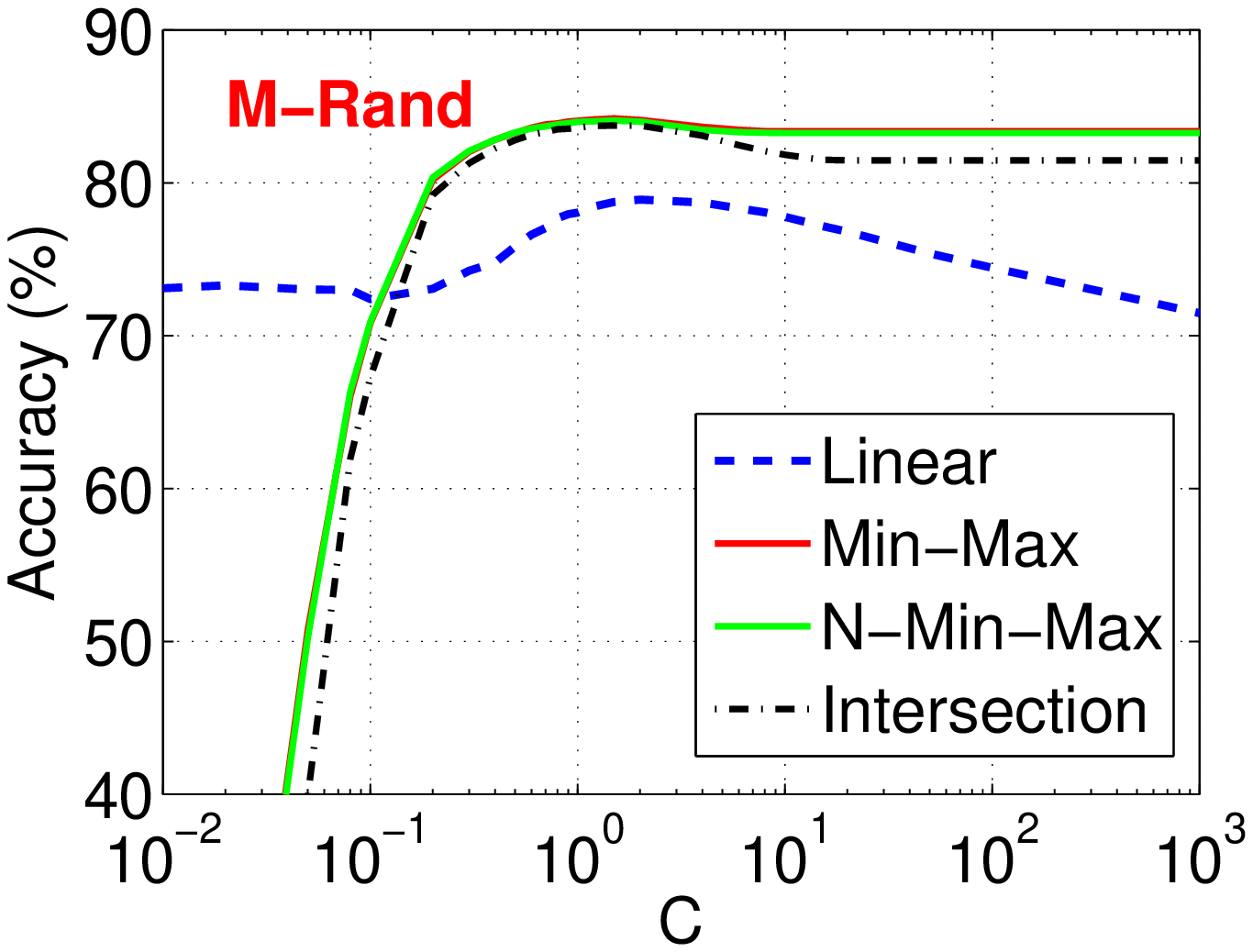}\hspace{-0.14in}
\includegraphics[width=1.75in]{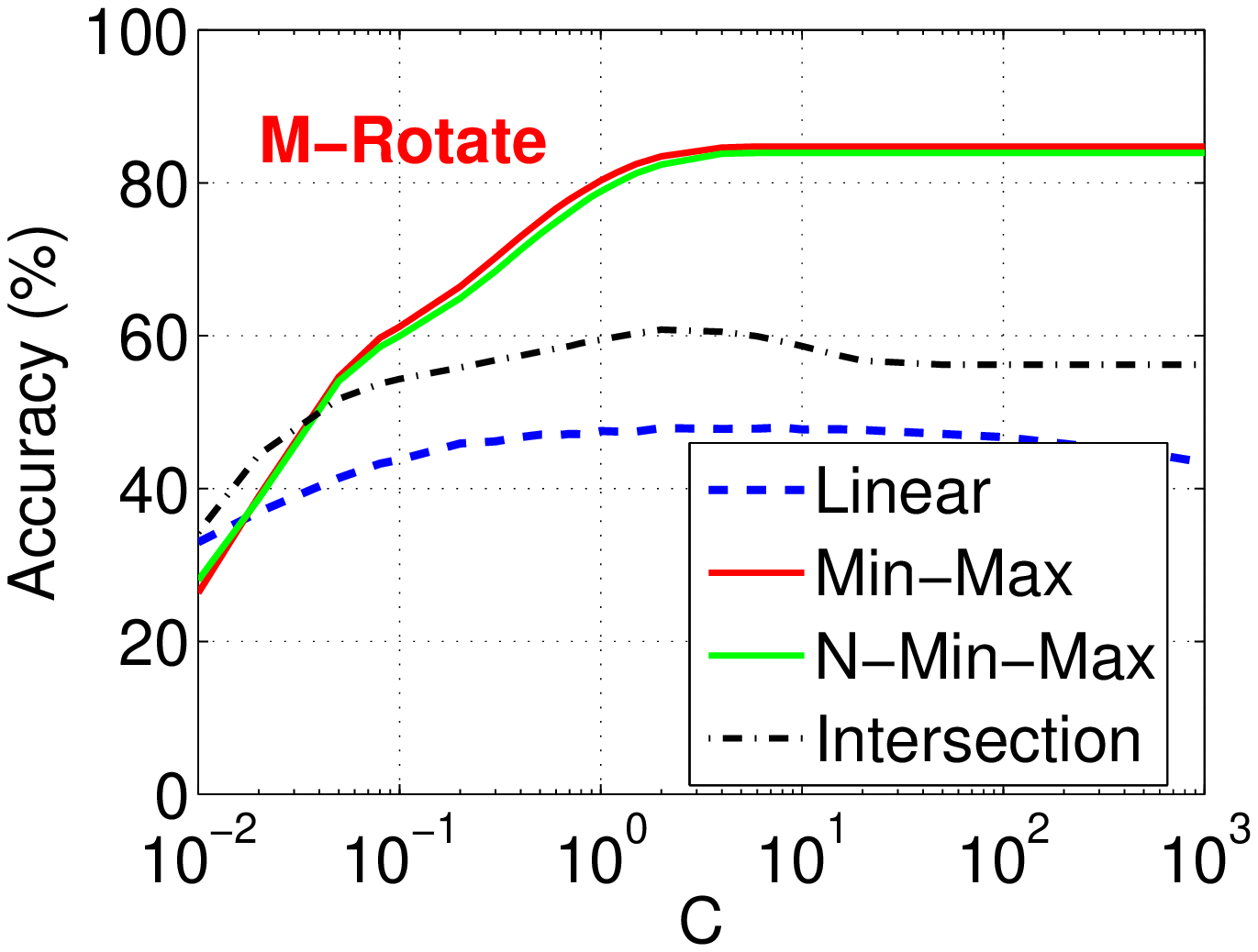}
}
\mbox{
\includegraphics[width=1.75in]{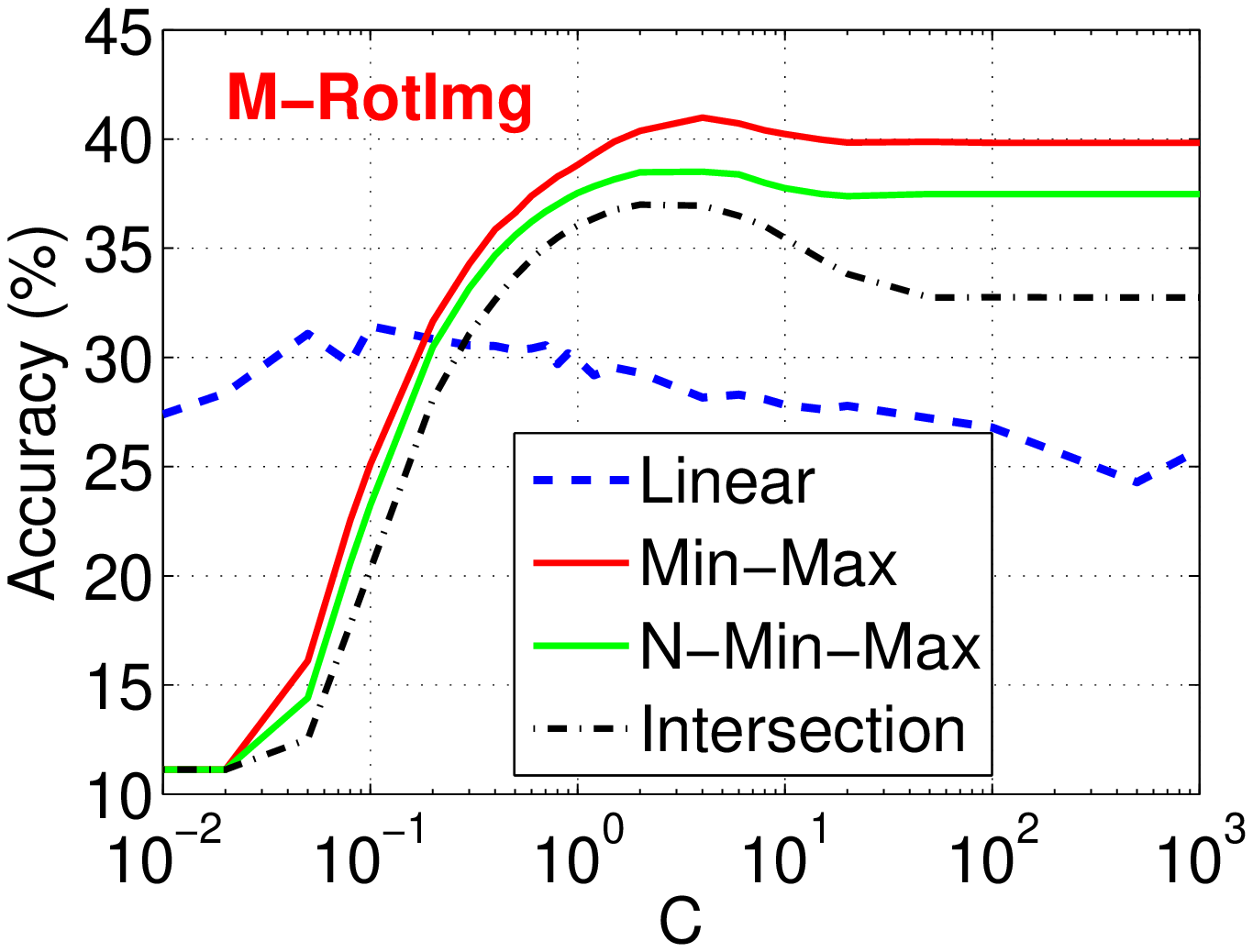}\hspace{-0.14in}
\includegraphics[width=1.75in]{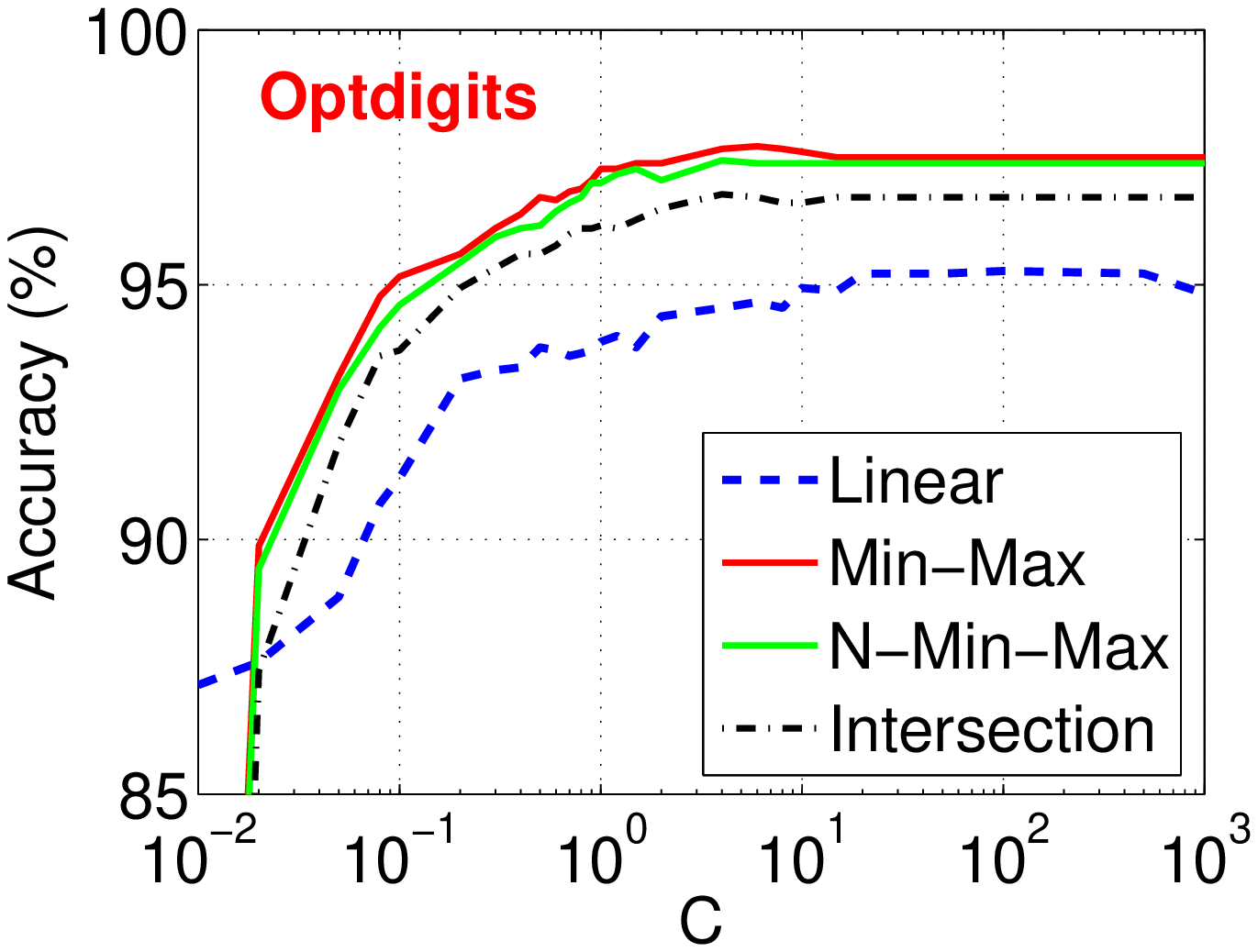}
}
\mbox{
\includegraphics[width=1.75in]{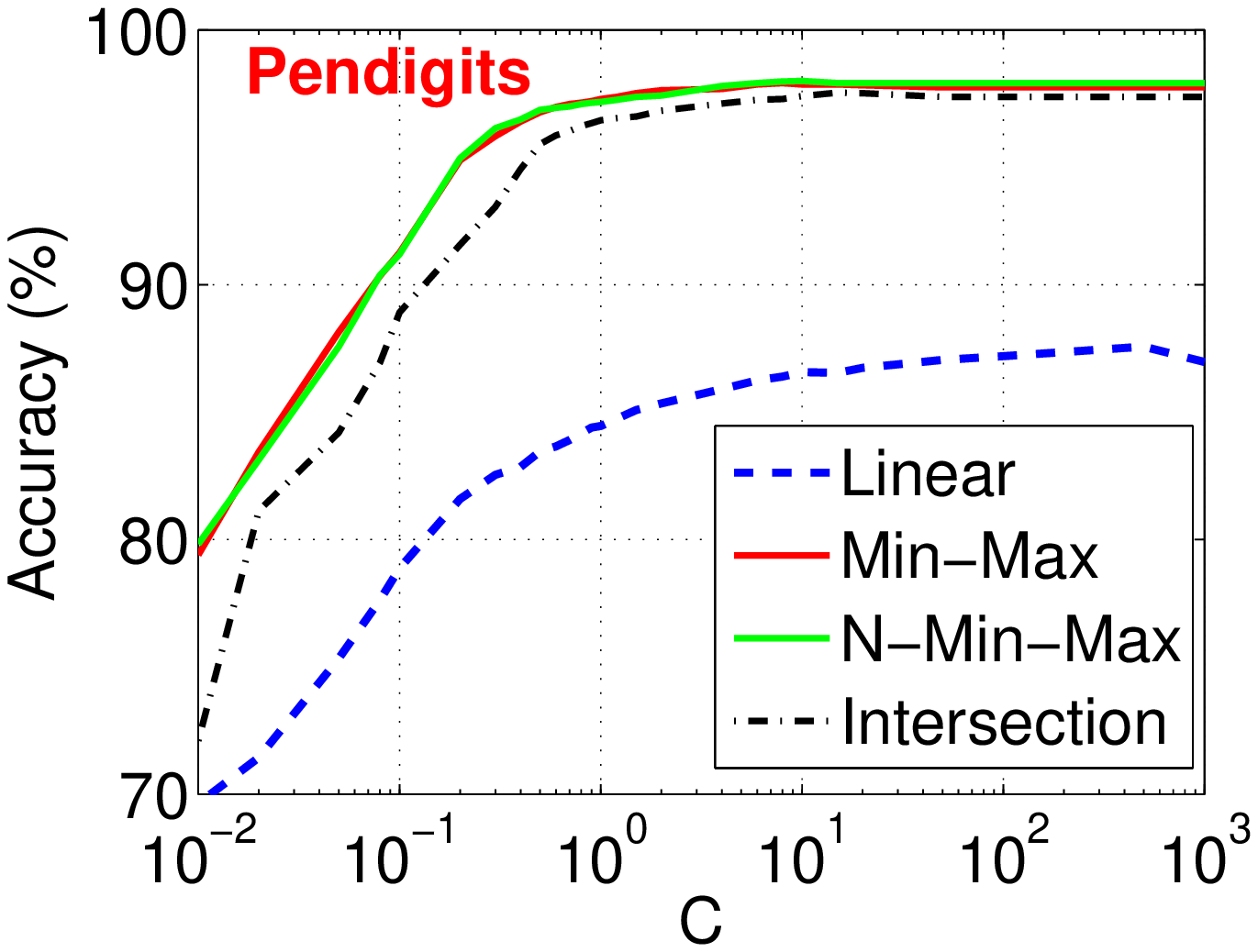}\hspace{-0.14in}
\includegraphics[width=1.75in]{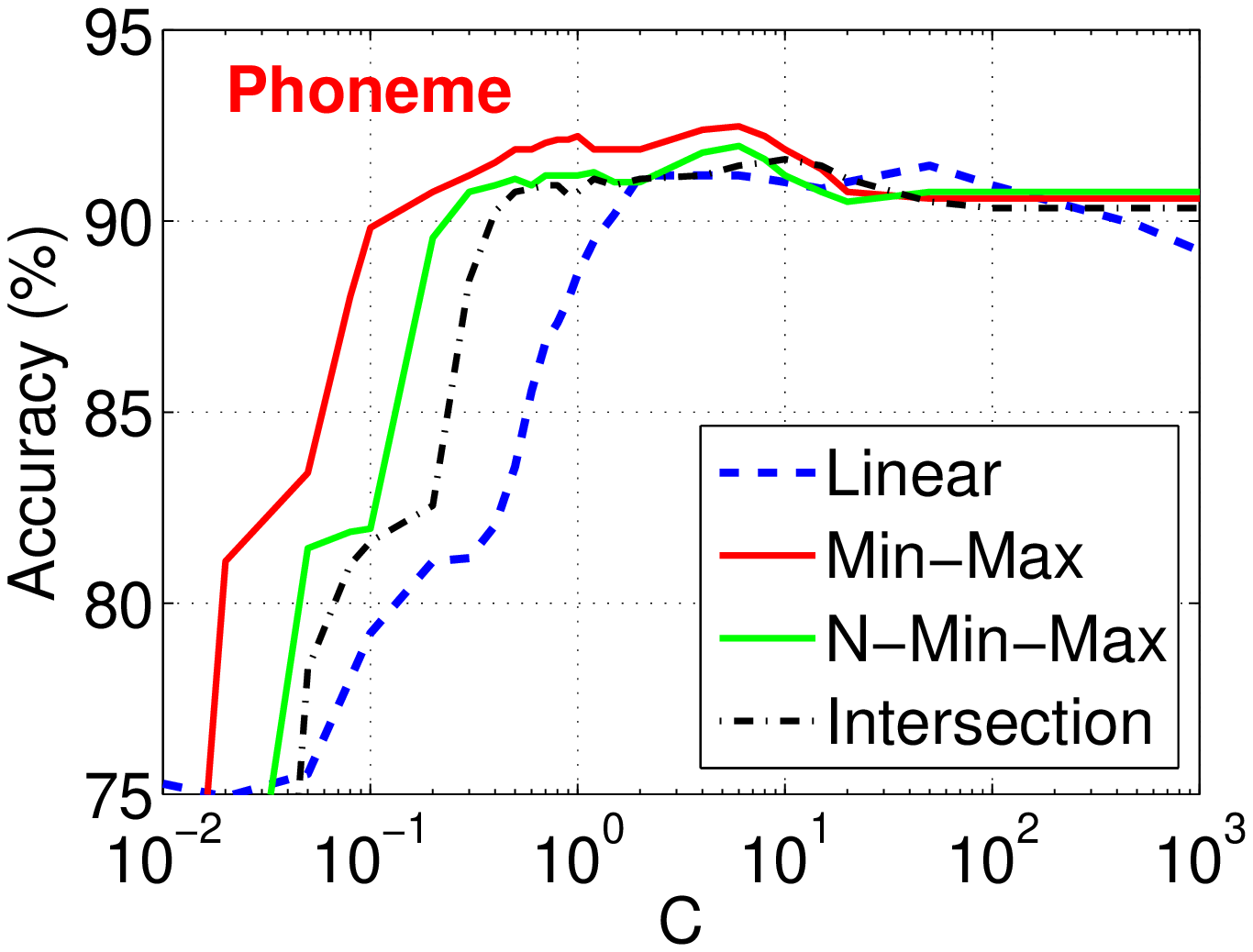}
}

\end{center}
\vspace{-0.2in}
\caption{Test classification accuracies  for four types of kernels using $l_2$-regularized  SVM.}\label{fig_KernelSVM_2}
\end{figure}

\begin{figure}[t]
\begin{center}

\mbox{
\includegraphics[width=1.75in]{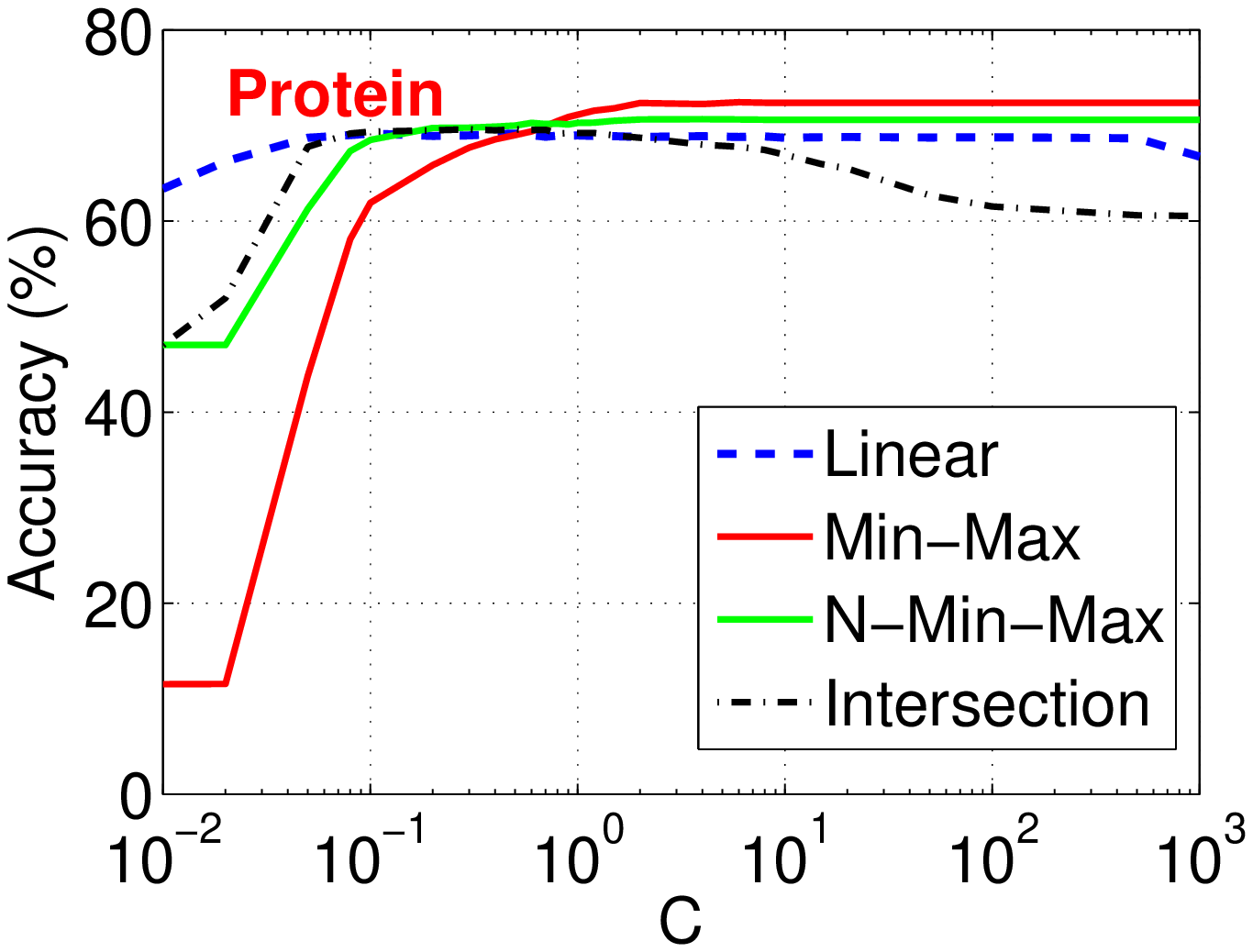}\hspace{-0.14in}
\includegraphics[width=1.75in]{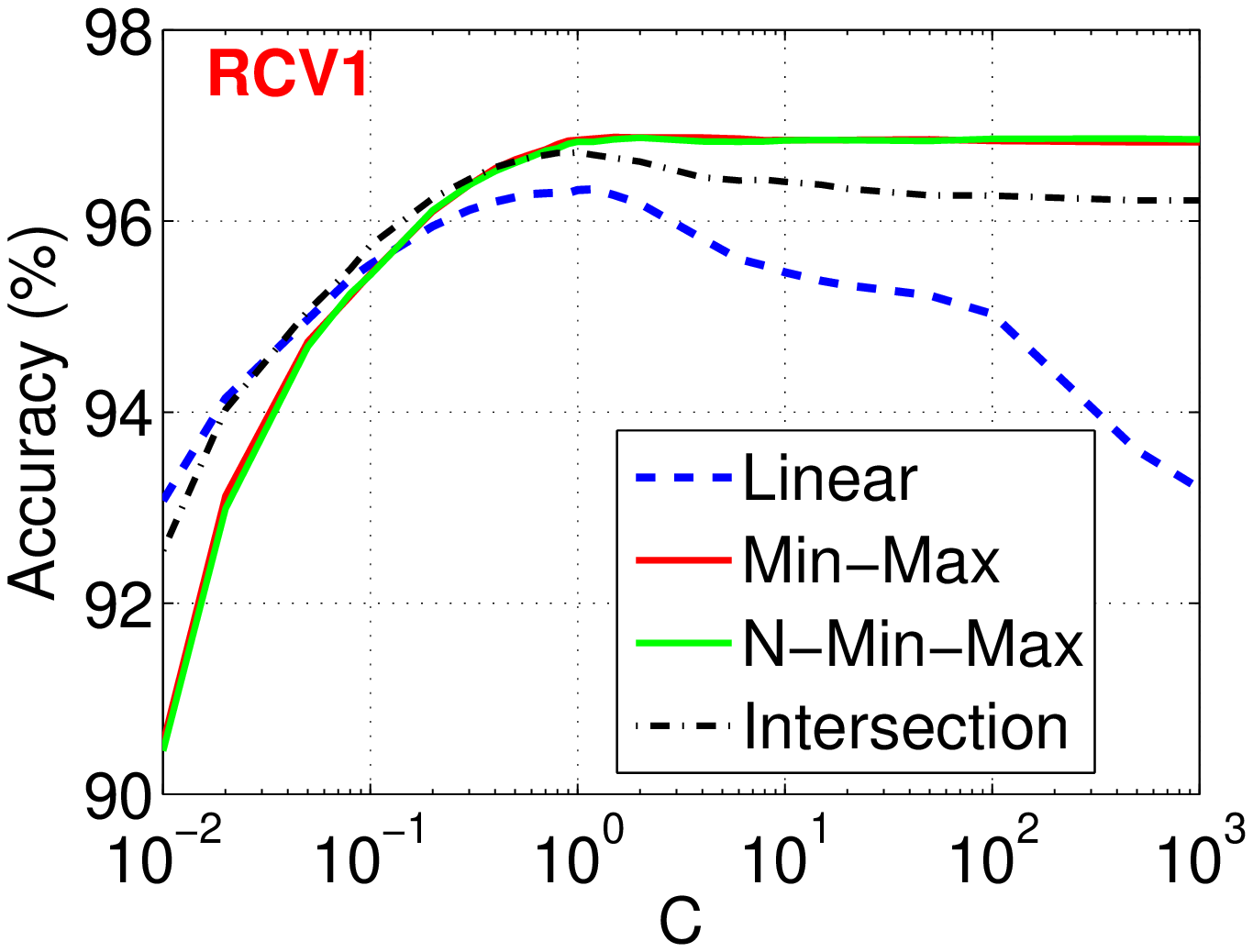}
}

\mbox{
\includegraphics[width=1.75in]{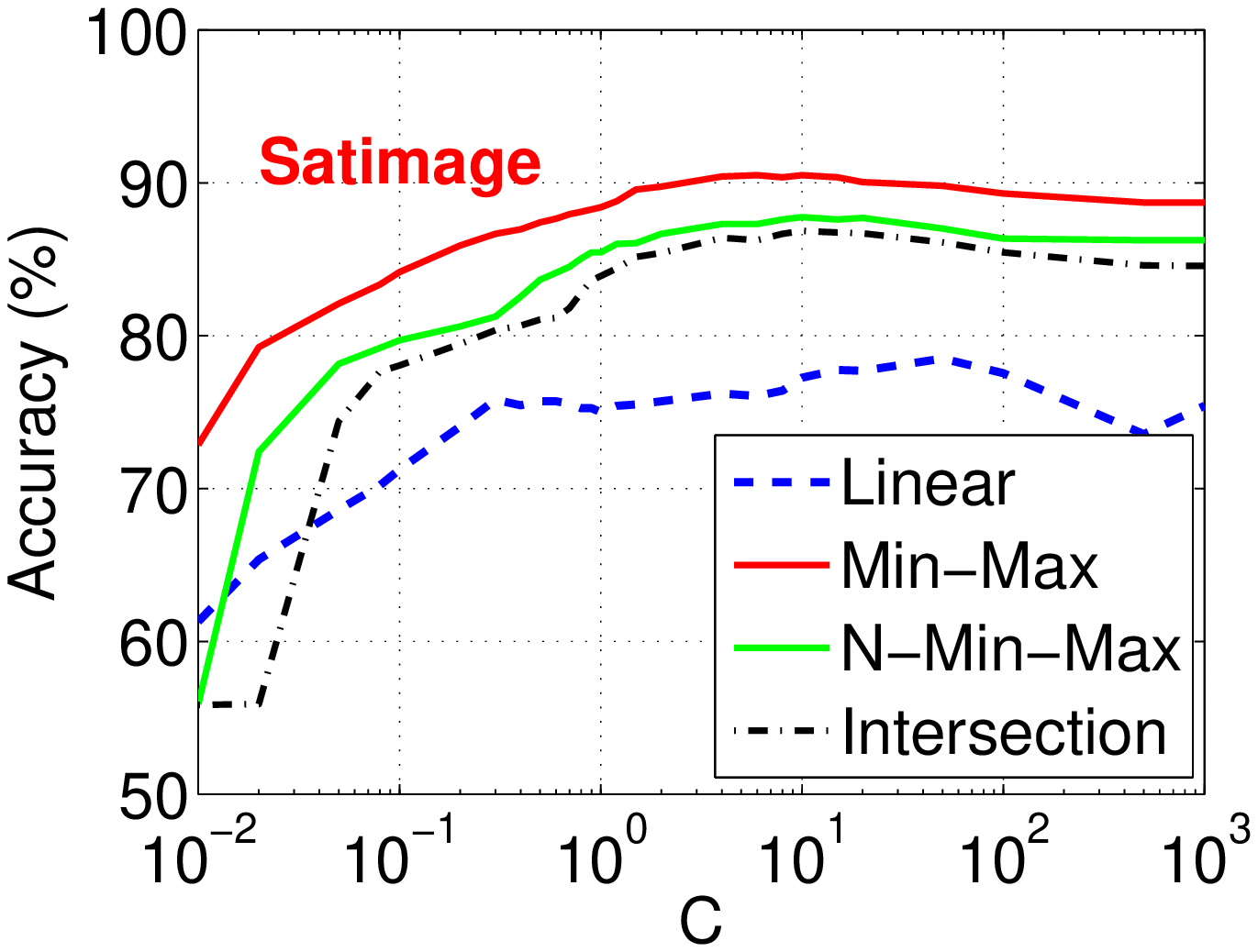}\hspace{-0.14in}
\includegraphics[width=1.75in]{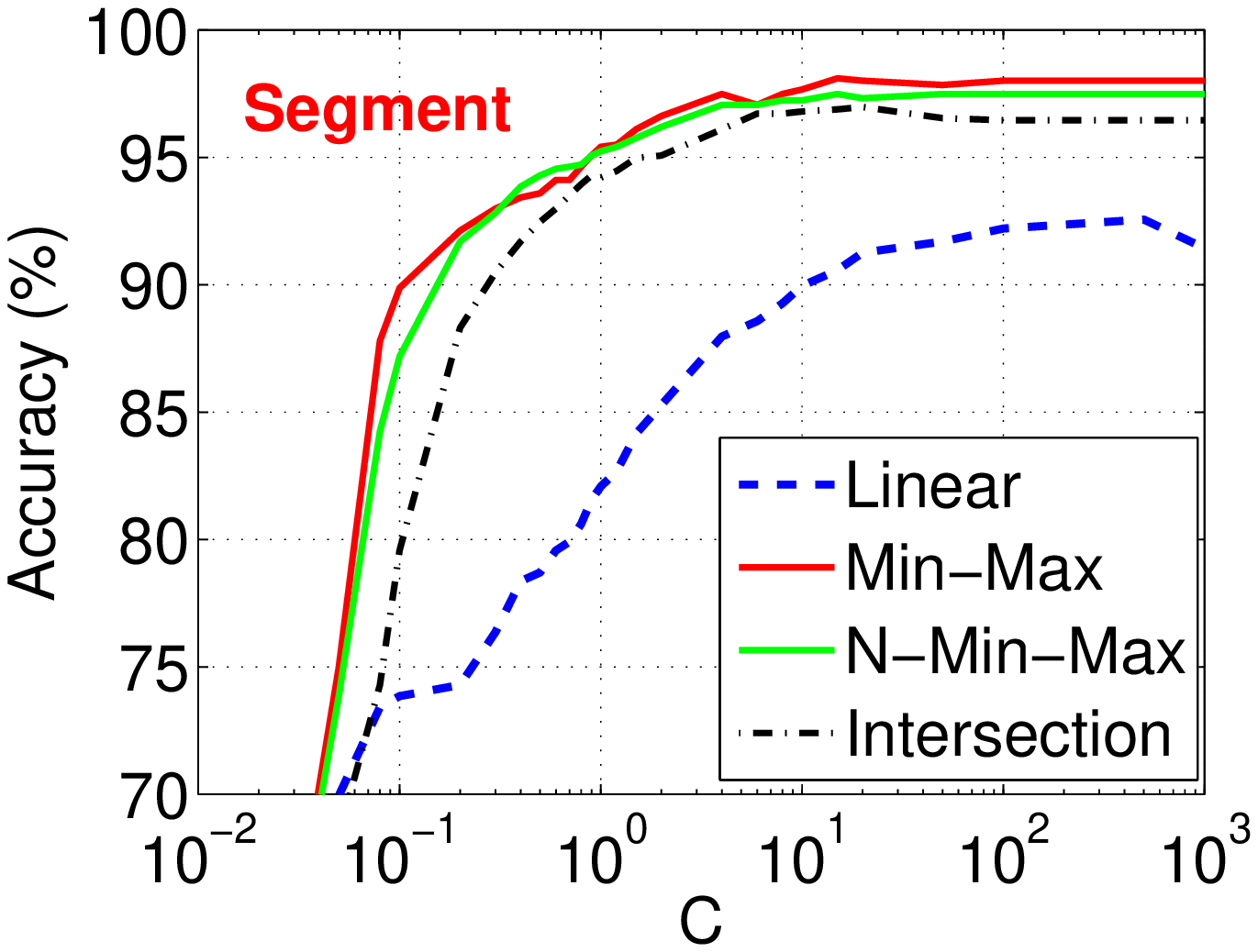}
}
\mbox{
\includegraphics[width=1.75in]{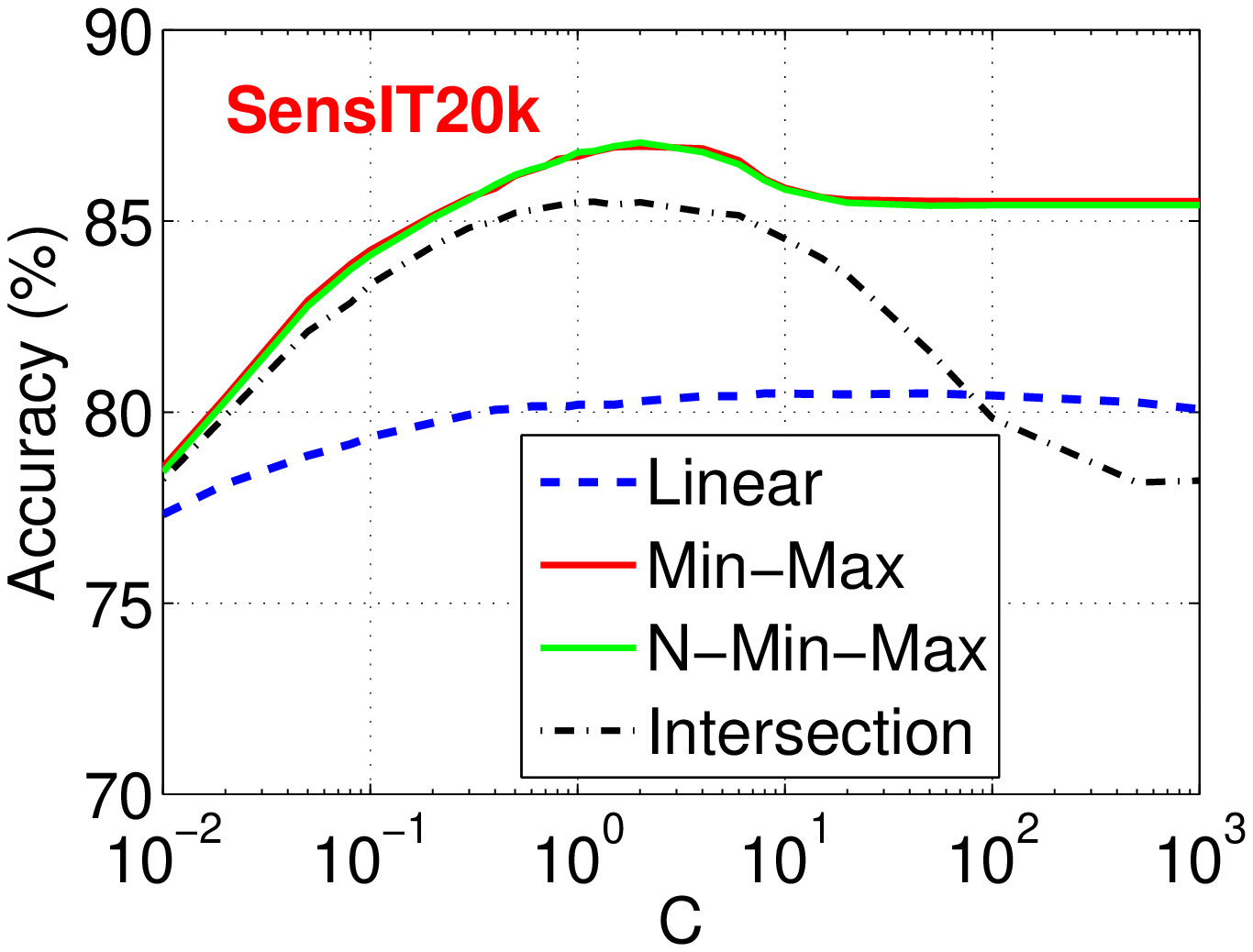}\hspace{-0.14in}
\includegraphics[width=1.75in]{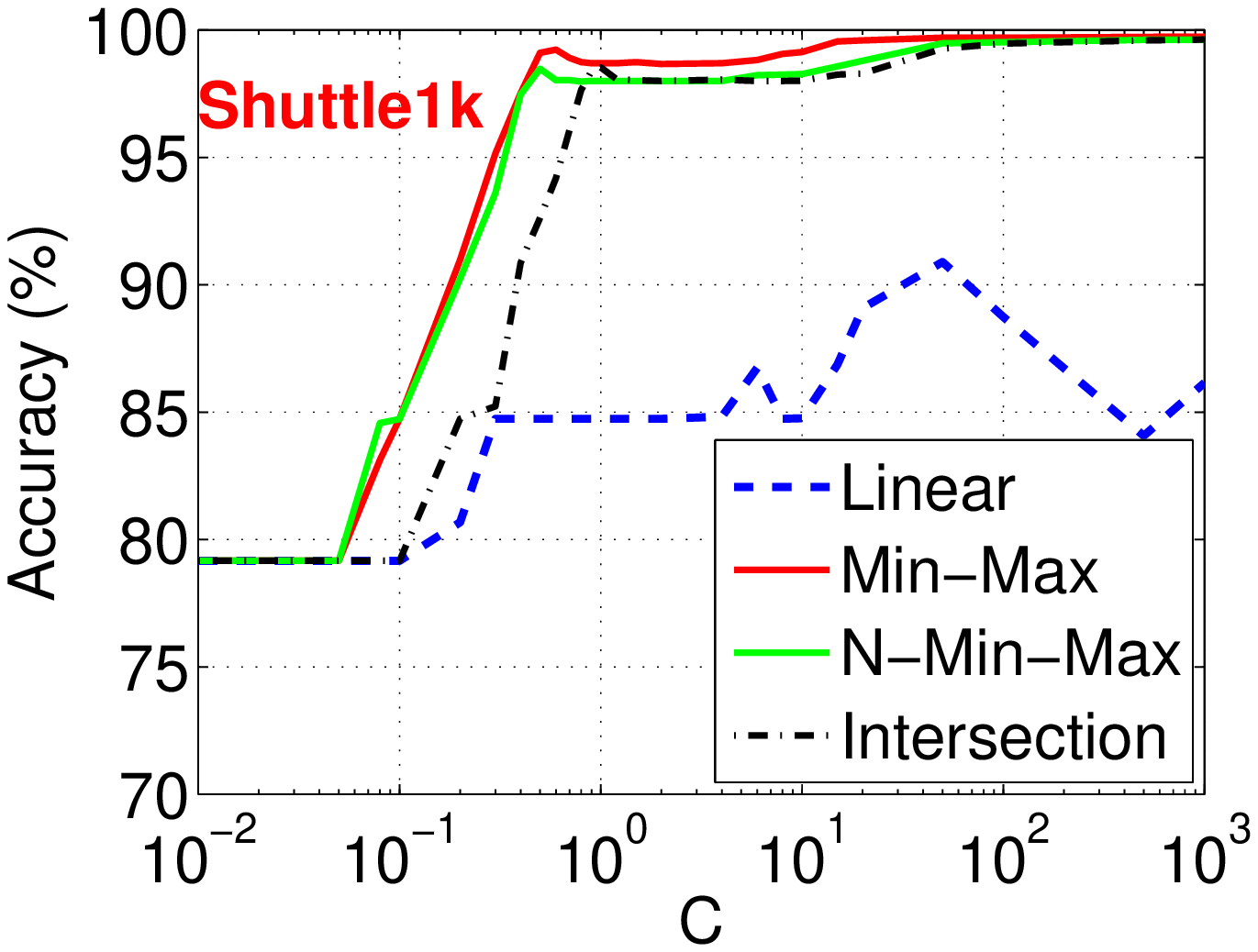}
}
\mbox{
\includegraphics[width=1.75in]{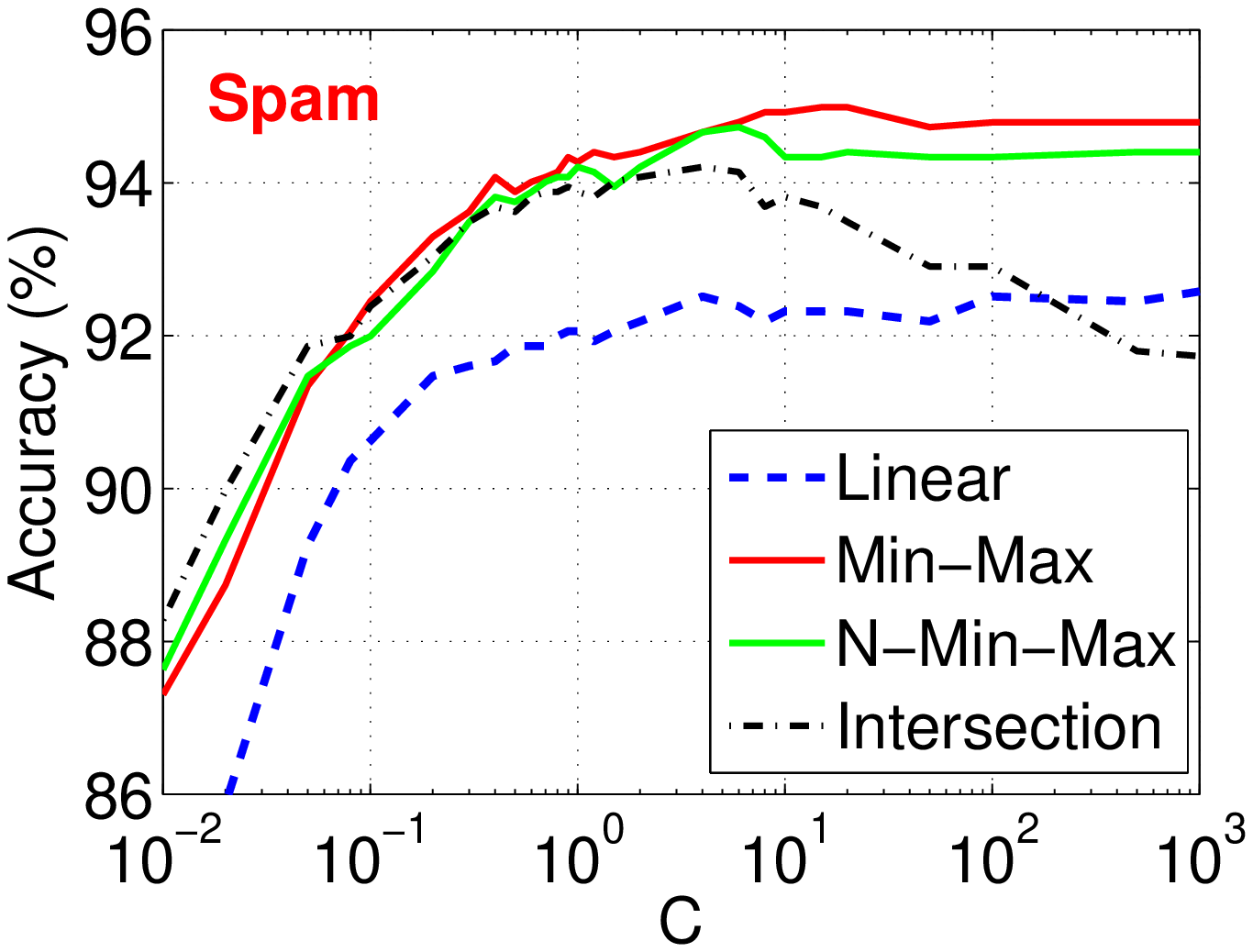}\hspace{-0.14in}
\includegraphics[width=1.75in]{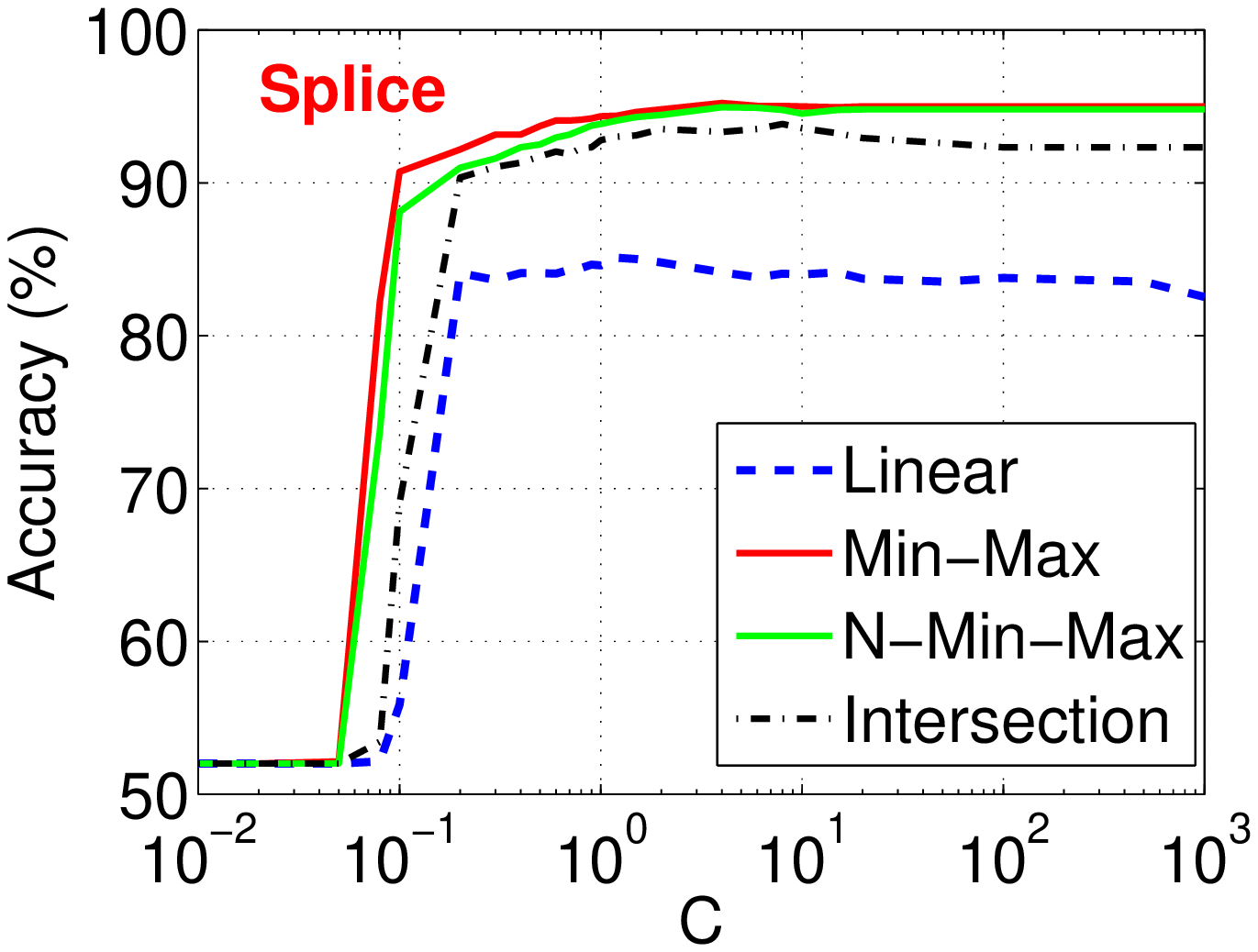}
}
\mbox{
\includegraphics[width=1.75in]{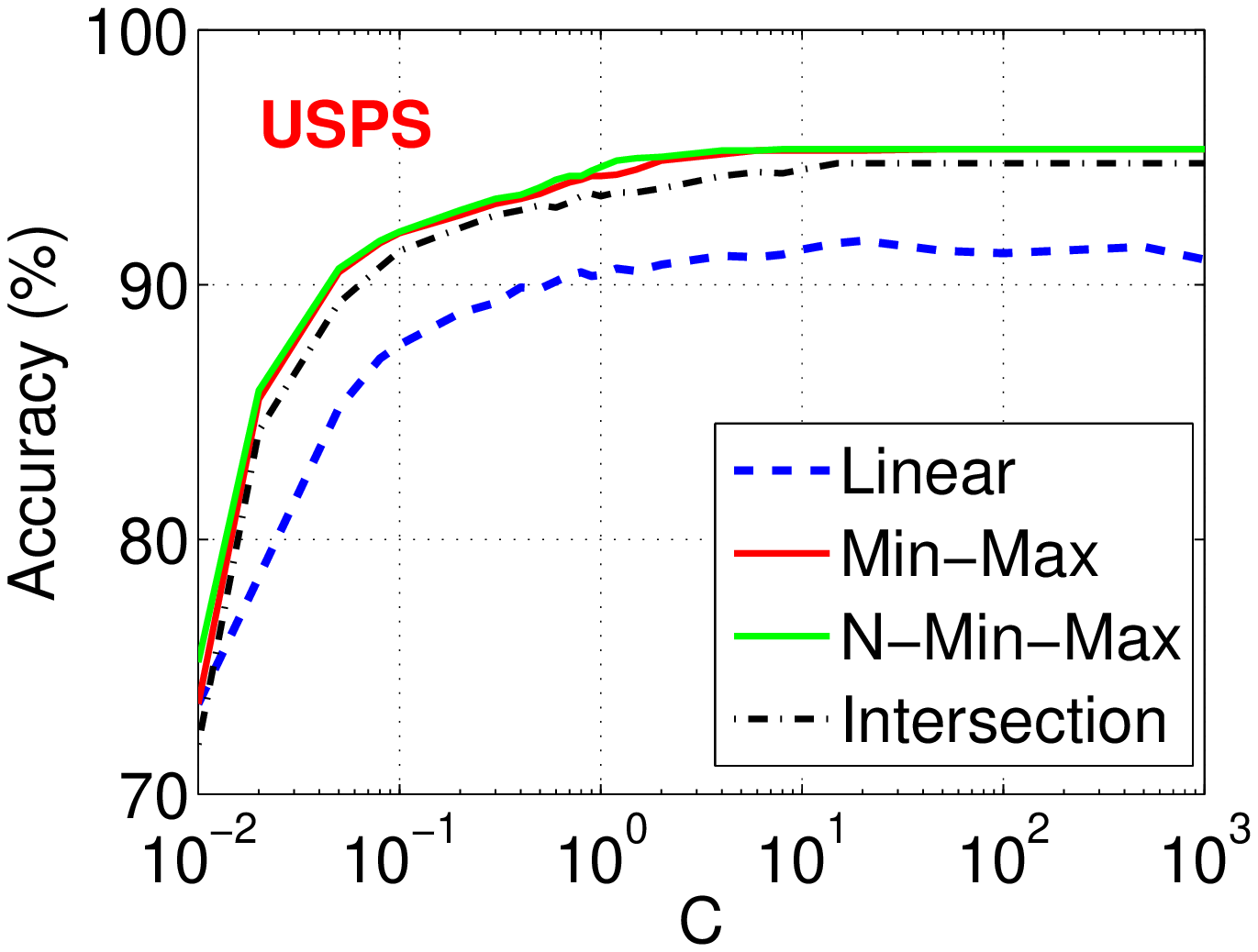}\hspace{-0.14in}
\includegraphics[width=1.75in]{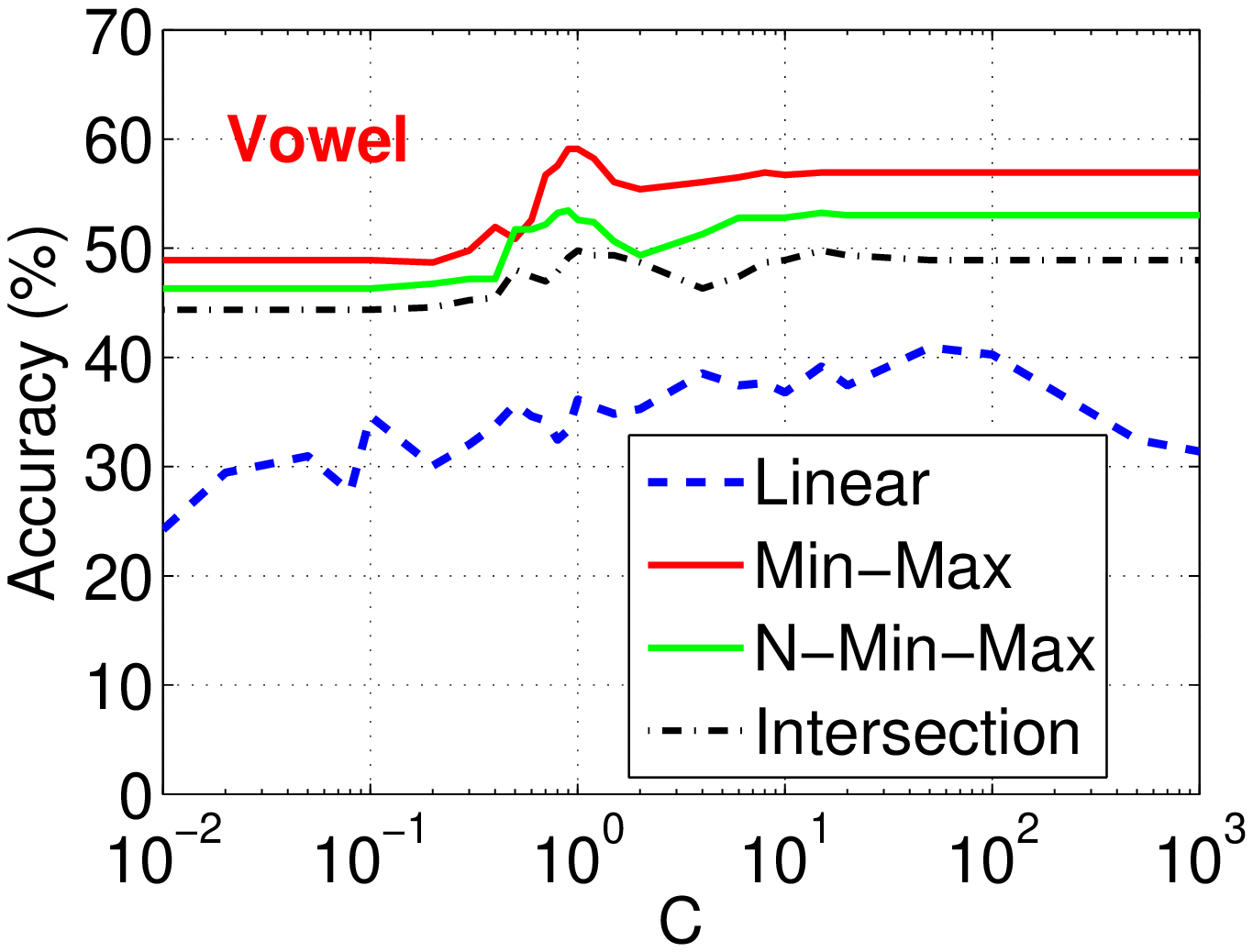}
}

\mbox{
\includegraphics[width=1.75in]{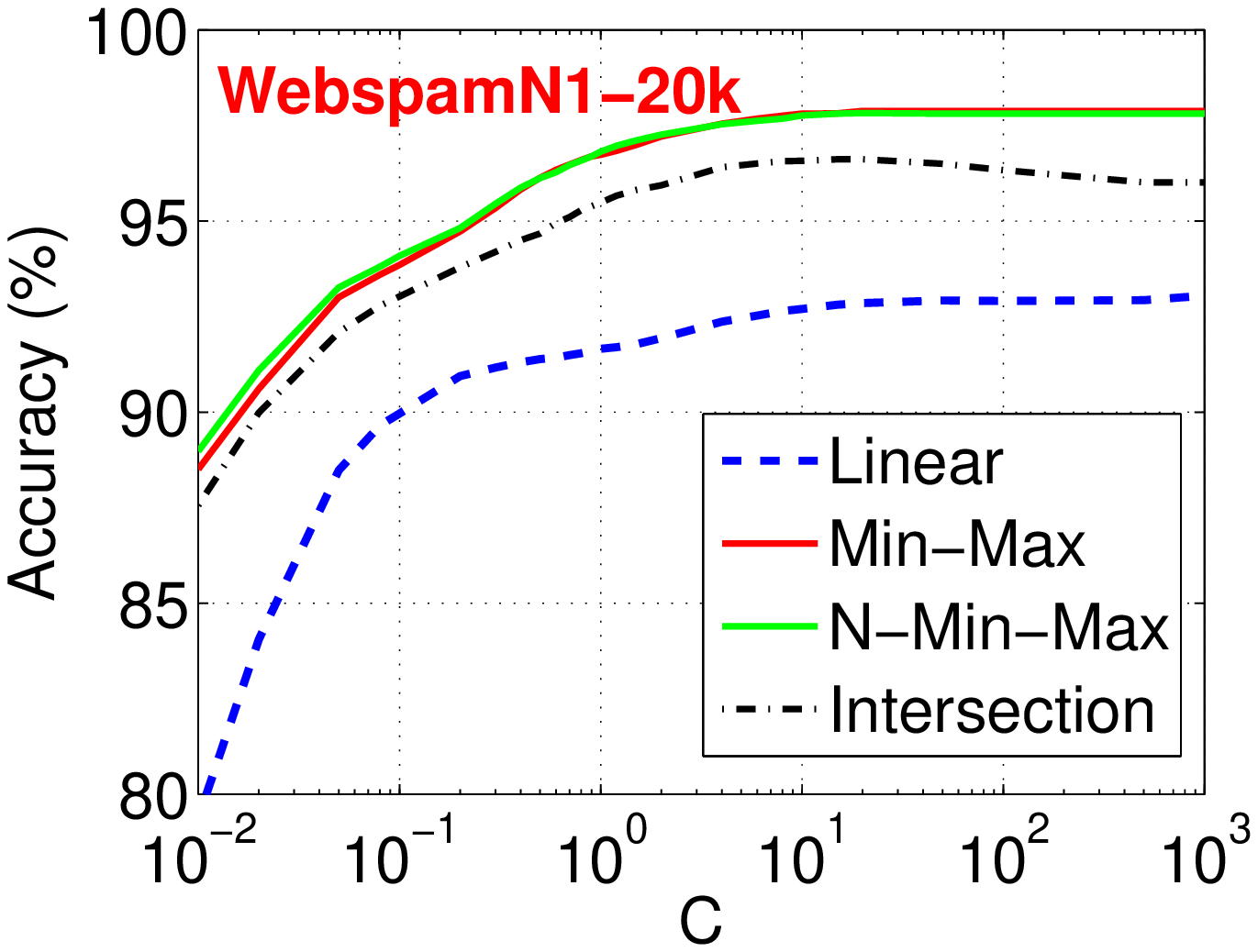}\hspace{-0.14in}
\includegraphics[width=1.75in]{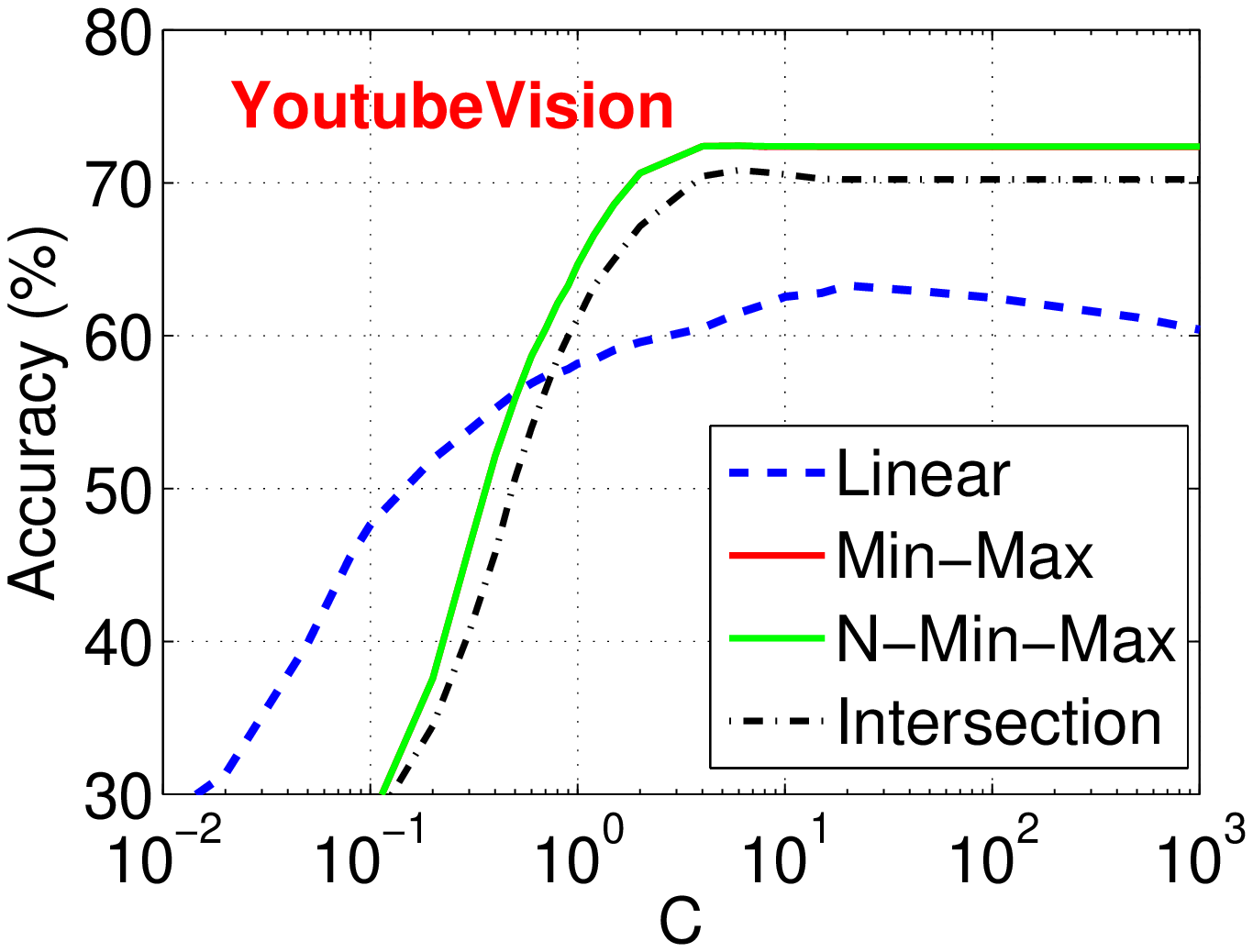}
}

\end{center}
\vspace{-0.2in}
\caption{Test classification accuracies  for four types of kernels using $l_2$-regularized  SVM.}\label{fig_KernelSVM_3}
\end{figure}

While these kernels do not have tuning parameters, there is a regularization parameter $C$ for $l_2$-regularized SVM. To ensure repeatability, we report the test classification accuracies for a wide range of $C$ values from $10^{-2}$ to $10^{3}$ with a fine grid, in Figures~\ref{fig_KernelSVM_1} to~\ref{fig_KernelSVM_3}. The accuracies  reported in Table~\ref{tab_KernelSVM} are the (individually) highest points on the curves.

The results in Table~\ref{tab_KernelSVM} and Figures~\ref{fig_KernelSVM_1} to~\ref{fig_KernelSVM_3} confirm that using min-max kernels typically result in better classification performance compared to linear kernel as well as intersection kernel. This experimental study, to an extent, help justify  the use of min-max kernels in learning applications.


\clearpage\newpage

\begin{table*}[t]
\caption{Classification accuracies (in \%) for using 4 different kernels. We use LIBSVM ``pre-computed'' kernels and $l_2$-regularized kernel SVM (which has a tuning parameter $C$). The reported test classification accuracies (i.e., the rightmost 4 columns) are the best accuracies from a wide range of $C$ values; see Figures~\ref{fig_KernelSVM_1} to~\ref{fig_KernelSVM_3} for more details.  The datasets are all public (and mostly well-known), from various sources including the UCI repository, the LIBSVM web site, the book web site of~\cite{Book:Hastie_Tib_Friedman}, and the  two papers~\cite{Proc:Larochelle_ICML07,Proc:ABC_UAI10} which compared deep nets, boosting and trees, kernel SVMs, etc. (Also see {http://hunch.net/?p=1467} for interesting discussions.)\vspace{0.1in} Whenever possible, we use the conventional partitions of training and testing sets.  We have made  efforts to ensure the repeatability of our experiments by using pre-computed kernels and reporting the results for a very wide range of $C$ values. However, this strategy also limits the scale of our experiments because most workstations do not have sufficient memory to store the kernel matrix for datasets of even moderate  sizes (for example, a merely $60,000 \times 60,000$ kernel matrix has $3.6\times 10^9$ entries). Therefore, for the sake of repeatability, for  a few datasets we only use a subset of the samples.  Please feel free to contact the author if more information is needed in order to reproduce the experiments. Several special notes about the datasets:\vspace{0.1in} (i) Whenever possible, we always use the data ``as they are'' from the sources. Although we agree it is a very important research task to study how to transform the data to favor certain type of similarities, it is not the focus of our paper (and may hurt the repeatability of the experiments if we try to alter the data).   (ii) Several datasets downloaded from the LIBSVM site were already scaled to [-1, 1]. To make use of these datasets, we simply transform them by $(z+1)/2$, where $z$ is the original feature value. \vspace{0.3in}
}
\begin{center}{
{\begin{tabular}{l r r c c c c }
\hline \hline
Dataset     &\# train samples  &\# test samples   &linear   &min-max &n-min-max &intersection \\
\hline
Covertype10k   &10,000 &50,000 &70.9   &\textbf{80.4}  &80.2 &74.3\\
Covertype20k   &20,000 &50,000 &71.1   &{\bf 83.3}  &83.1 &75.2 \\
IJCNN5k &5,000 &91,701 &   91.6   &94.4   &{\bf95.3}  &94.0\\
IJCNN10k &10,000 &91,701 & 91.6   &95.7   &{\bf96.0}   &94.5\\
Isolet &6,238 &1,559 &95.4   &96.4  &{\bf96.6}   &96.4\\
Letter &16,000 &4,000 &62.4   &{\bf96.2}   &95.0   &92.1\\
Letter4k &4,000 &16,000 &61.2 &{\bf91.4} &90.2 &87.9\\
M-Basic   &12,000 &50,000 &90.0   &{\bf96.2}   &96.0   &93.4\\
M-Image &12,000 &50,000 &70.7   &{\bf80.8}   &77.0   &76.2\\
MNIST10k &10,000 &60,000 &90.0   &{\bf95.7}  &95.4   &93.1\\
M-Noise1 &10,000 &4,000 &60.3   &{\bf71.4}   &68.5   &68.2\\
M-Noise2 &10,000 &4,000 &62.1   &{\bf72.4}  &70.7   &70.0\\
M-Noise3 &10,000 &4,000 &65.2   &{\bf73.6}   &71.9   &71.6\\
M-Noise4 &10,000 &4,000 & 68.4  &{\bf76.1}  &75.2   &74.8\\
M-Noise5 &10,000 &4,000 &72.3   &{\bf79.0}   &78.4   &77.9\\
M-Noise6 &10,000 &4,000 &78.7   &84.2   &{\bf84.3}   &83.9\\
M-Rand &12,000 &50,000 &78.9   &{\bf84.2}   &84.1   &83.7\\
M-Rotate &12,000 &50,000    &48.0   &{\bf84.8}   &83.9   &60.8  \\
M-RotImg &12,000 &50,000 &31.4   &{\bf41.0}   &38.5   &37.0\\
Optdigits &3,823 &1,797 &95.3   &{\bf97.7}  &97.4  &96.8\\
Pendigits &7,494 &3,498 &87.6   &97.9   &{\bf98.0}  &97.5\\
Phoneme &3,340 &1,169 &91.4   &{\bf92.5}   &92.0   &91.6\\
Protein &17,766 &6,621 & 69.1   &{\bf72.4}  &70.7   &69.6\\
RCV1 &20,242 &60,000 &96.3   &{\bf96.9}   &{\bf96.9}   &96.7\\
Satimage &4,435 &2,000 & 78.5   &{\bf90.5}   &87.8   &86.9\\
Segment &1,155 &1,155 &92.6   &{\bf98.1}  &97.5  &97.0\\
SensIT20k &20,000 &19,705 &80.5   &86.9   &{\bf87.0}   &85.5\\
Shuttle1k &1,000 &14,500 &90.9   &{\bf99.7}   &99.6   &99.6\\
Spam  &3,065 &1,536 & 92.6   &{\bf95.0} & 94.7   &94.2\\
Splice &1,000 &2,175 &85.1   &{\bf95.2}   &94.9  &93.8\\
USPS   &7,291 &2,007     &91.7  &{\bf95.3}   &{\bf95.3} &94.8\\
Vowel &528 &462 &40.9   &{\bf59.1}  &53.5   &49.8\\
WebspamN1-20k (1-gram) &20,000 &60,000 &93.0  &{\bf97.9}   &97.8   &96.6\\
YoutubeVision &11,736 &10,000 &63.3 &{\bf72.4}   &{\bf72.4}   &70.8\\
\hline\hline
\end{tabular}}
}
\end{center}\label{tab_KernelSVM}

\end{table*}

\newpage\clearpage

The purpose of this experimental study  on kernel SVMs is not try to show that min-max kernels achieve the best classification accuracies. In fact, compared to trees or deep nets~\cite{Proc:Larochelle_ICML07,Proc:ABC_UAI10}, simply using min-max kernels usually does achieve the best accuracies, although the results are  close. Since min-max kernels have no tuning parameters, we can expect to boost the performance by using additional parameters or by combining multiple the same (or different) types of kernels. For example, using the idea from CoRE kernels~\cite{Proc:Li_UAI14}, we can multiply min-max kernel with chi-square kernels (which can be hashed by sign cauchy projections~\cite{Proc:Li_NIPS13}).

For large-scale industrial applications, typically  it is difficult to directly use (nonlinear) kernels. Fortunately, with CWS (consistent weighted sampling), we can linearize the min-max kernel. In other words, it is possible to achieve the good performance of min-max kernels at the cost of linear kernels.   In this paper, we will show how to do CWS better.

\section{Hashing Min-Max Kernel}\label{sec_CWS}

The classification experiments reported in Table~\ref{tab_KernelSVM} and Figures~\ref{fig_KernelSVM_1} to~\ref{fig_KernelSVM_3} have demonstrated the effectiveness of  min-max kernels in terms of prediction accuracies. However, in order to make min-max kernels practical for large-scale data mining tasks, we need to resort to hashing techniques to (approximately) transform nonlinear kernels into linear kernels.

It is well understood~\cite{Book:Bottou_07} that computing kernels are expensive and the kernel matrix, if fully materialized, does not fit in memory even for relatively small applications. In contrast, highly efficient linear algorithms, e.g.,~\cite{Proc:Joachims_KDD06,Proc:Shalev-Shwartz_ICML07,URL:Bottou_SGD,Article:Fan_JMLR08}, have been widely used in practice for truly large-scale applications such as click predictions in online advertising~\cite{Proc:McMahan_KDD13}.

\subsection{Consistent Weighted Sampling (CWS)}

The prior efforts~\cite{Report:Manasse_CWS10,Proc:Ioffe_ICDM10} have lead to the method called  ``consistent weighted sampling (CWS)'' for hashing min-max kernels. Here, we adopt the beautiful description of CWS in~\cite{Proc:Ioffe_ICDM10} as shown in Alg.~\ref{alg_CWS}.

\begin{algorithm}{\small
\textbf{Input:} Data vector $u$ = ($u_i\geq 0$, $i=1$ to $D$)

\textbf{Output:} Consistent uniform sample ($i^*$, $t^*$)

\vspace{0.08in}

For $i$ from 1 to $D$

\hspace{0.25in}$r_i\sim Gamma(2, 1)$, \ \ $c_i\sim Gamma(2, 1)$,  $\beta_i\sim Uniform(0, 1)$

\hspace{0.2in} $t_i\leftarrow \lfloor \frac{\log u_i }{r_i}+\beta_i\rfloor$, \ $y_i\leftarrow \exp(r_i(t_i - \beta_i))$,\  $a_i\leftarrow c_i/(y_i \exp(r_i))$

End For

$i^* \leftarrow arg\min_i \ a_i$,\hspace{0.3in}  $t^* \leftarrow t_{i^*}$
}\caption{Consistent Weighted Sampling (CWS)}
\label{alg_CWS}
\end{algorithm}

Given a data vector $u\in\mathbb{R}^D$, Alg.~\ref{alg_CWS} provides the procedure for generating one CWS sample $(i^*, t^*$). In order to generate $k$ such samples, we have to repeat the procedure  $k$ times using an independent set of random numbers $r_i$, $c_i$, $\beta_i$. For clarity, we denote the samples for data vectors $u$ and $v$  as
\begin{align}
\left(i^*_{u,j}, t^*_{u,j}\right)\ \text{ and }\  \left(i^*_{v,j}, t^*_{v,j}\right),\  \  j = 1, 2, ..., k
\end{align}
Basically we need to generate 3 matrices: $\{r\}$, $\{c\}$, and $\{\beta\}$, of size $D \times k$. All the data vectors will use the same 3 matrices.  This is essentially the same cost as random projections (which however approximate linear kernels).

The basic theoretical result of CWS says the ``collision probability'' is exactly $K_{MM}$:
\begin{align}\label{eqn_CWS_Prob}
\mathbf{Pr}\left\{\left(i^*_{u,j}, t^*_{u,j}\right) =  \left(i^*_{v,j}, t^*_{v,j}\right)\right\} = K_{MM}(u,v)
\end{align}
Thus, it is clear that, at least conceptually, we can express $K_{MM}(u,v)$ as the expectation of an inner product and hence $K_{MM}$ is positive definite, just like how~\cite{Proc:HashLearning_NIPS11} showed the resemblance is a type of positive definite kernel.

\subsection{Drawback of CWS for Data Mining}

Although the basic probability result (\ref{eqn_CWS_Prob}) says conceptually we can use CWS for building linear classifiers (approximately in the space of min-max kernels), it is not immediately clear how it can be implemented efficiently.

\cite{Proc:Ioffe_ICDM10} briefly mentioned that one can ``uniformly map'' the sample space $(i^*, t^*)$ to a space $b$ bits: $\{0, 1, 2, ..., 2^b-1\}$. This however can not be (easily) achieved. While $i^*$ is bounded by $D$, $t^*$ is actually unbounded (see Alg.~\ref{alg_CWS}). Also note that space of samples is very large. If we represent $i^*$ by $b_i$ bits and $t^*$ (approximately) by $b_t$ bits, the space will be $2^{b_i+b_t}$. Thus, we must find an efficient representation of CWS samples in order to use this nice method effectively for machine learning and data mining applications.

\subsection{Our ``0-bit'' Proposal for CWS}

It is now known how to use $b$-bit minwise hashing to approximate the resemblance kernel and use it for large-scale applications~\cite{Proc:Li_Konig_WWW10,Proc:HashLearning_NIPS11}. Therefore, in this paper, we focus on representing $t^*$. Perhaps surprisingly, our proposal is simple: just ignore $t^*$ in the sample $(i^*, t^*$), i.e., the ``0-bit'' scheme.

If we examine Alg.~\ref{alg_CWS}, we can see that $i^*$ has already encoded the information about the weights of the data. A rigorous proof however turns out to be a  difficult probability problem, which is outside the scope of this paper. Here, we try to empirically demonstrate the following observation:
\begin{align}\label{eqn_CWS_Prob_Approx}
\mathbf{Pr}\left\{ i^*_{u,j} =  i^*_{v,j}\right\} \approx  \mathbf{Pr}\left\{\left(i^*_{u,j}, t^*_{u,j}\right) =  \left(i^*_{v,j}, t^*_{v,j}\right)\right\}
\end{align}

We call our proposal the ``0-bit'' scheme only to mean that we use 0 bit for coding $t^*$.  We also call the original proposal as the ``full'' scheme since it stores all  the bits needed for $t^*$.

\subsection{An Experimental Study on ``0-bit'' CWS}

\vspace{-0.15in}
\begin{table}[h]
\caption{\small Information of the 13 pairs of English words. For example, ``HONG''  refers to the  vector of occurrences of the word ``HONG'' in $2^{16}$ documents.  $f_1$ and $f_2$ are the numbers of nonzeros in word 1 and word 2 respectively. For each  pair, we include the numerical values for both the resemblance (``$R$'') and the min-max kernel ($MM$).
 }
\begin{center}{\scriptsize
\begin{tabular}{l l r r  c c}
\hline \hline
Word 1 & Word 2 &$f_1$  &$f_2$  &$R$ &$MM$ \\\hline
A  & THE &39063       &42754     & 0.6444    &0.3543\\
ADDICT &PRICELESS &77 &77 & 0.0065    &0.0052\\
AIR &DOCTOR &3159 &860 &0.0439 &0.0248\\
CREDIT & CARD &2999 &2697 &0.2849 &0.2091\\
GAMBIA &KIRIBATI &206   &186  &0.7118 &0.6070\\
HONG & KONG &940 &948 &0.9246 &0.8985\\
OF & AND &37339   &36289 &0.7711 &0.6084\\
PAPER &REVIEW & 1944    &3197 &0.0780  &0.0502\\
PIPELINE &FLUSH &139 &118 &0.0158 &0.0143\\
SAN &FRANCISCO &3194 &1651 &0.4758 &0.2885\\
THIS &TODAY &27695   &5775 &0.1518 &0.0658\\
TIME & JOB &37339   &36289 &0.1279 &0.0794\\
UNITED &STATES &4079 &3981 &0.5913 &0.5017\\
\hline\hline

\end{tabular}
}
\end{center}
\label{tab_word_pairs}\vspace{-0.2in}
\end{table}

Table~\ref{tab_word_pairs} lists 13 pairs of English words. Each word represents a vector of occurrences of that word in a total of $2^{16}$ documents. This is a typical example of heavy-tailed data in that the weights vary dramatically. In common machine learning applications, the weights often do not vary as much (at least at the point when we are prepared to compute distances/similarites from  data). In that sense, we are actually testing our ``0-bit'' proposal in a more challenging setting.

\begin{figure}[h!]
\begin{center}
\mbox{
\includegraphics[width=1.75in]{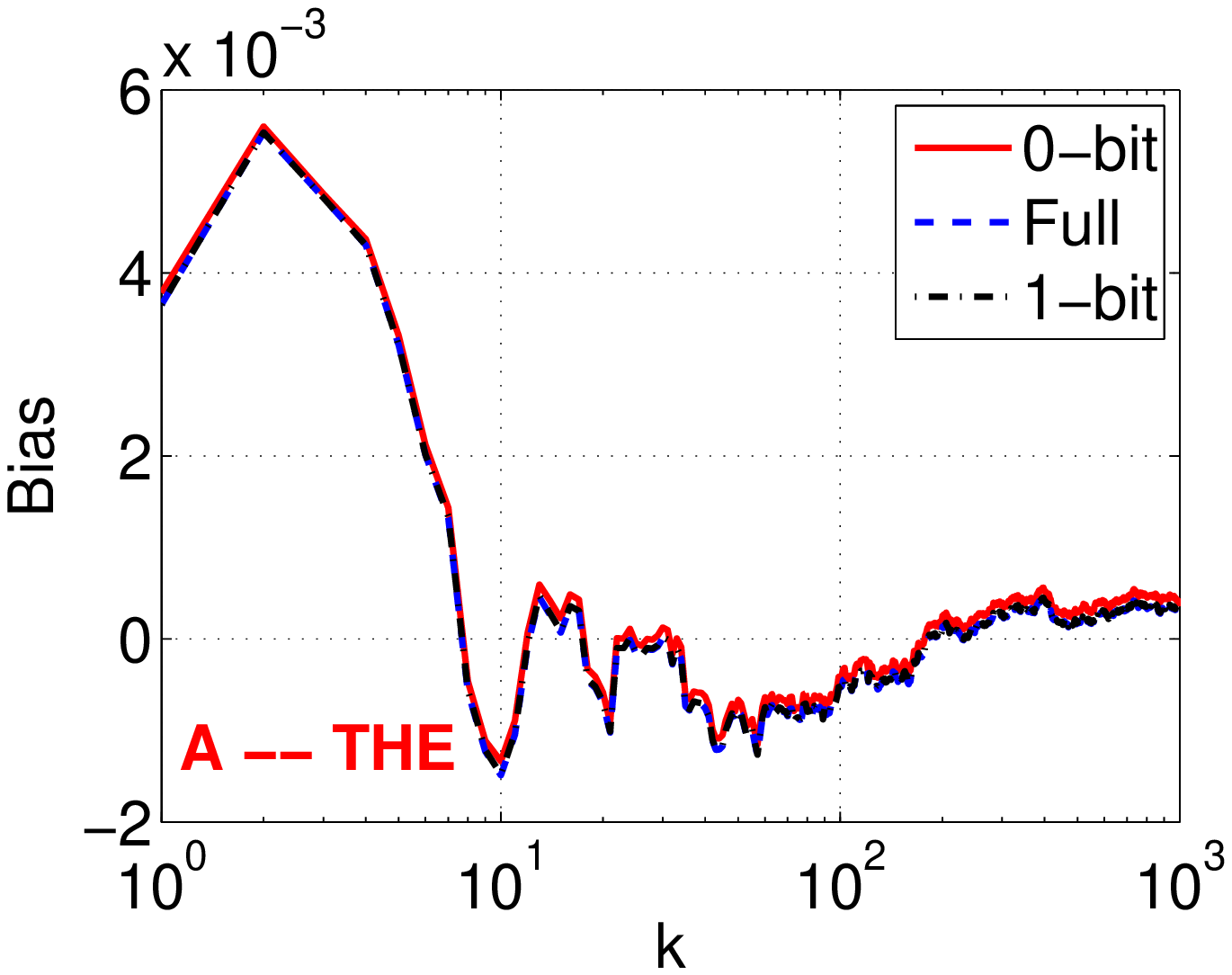}\hspace{-0.14in}
\includegraphics[width=1.75in]{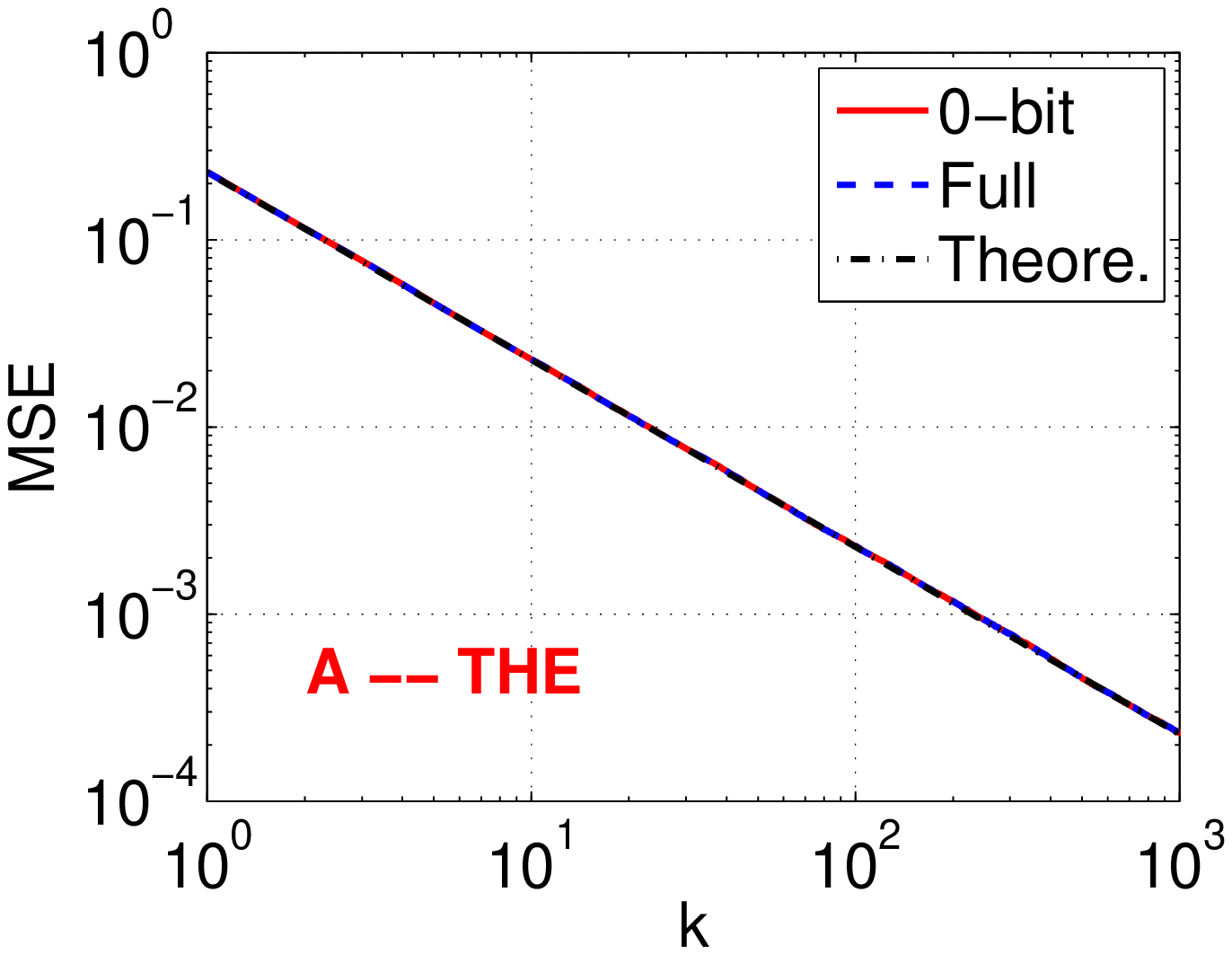}
}
\vspace{-0.1in}

\mbox{
\includegraphics[width=1.75in]{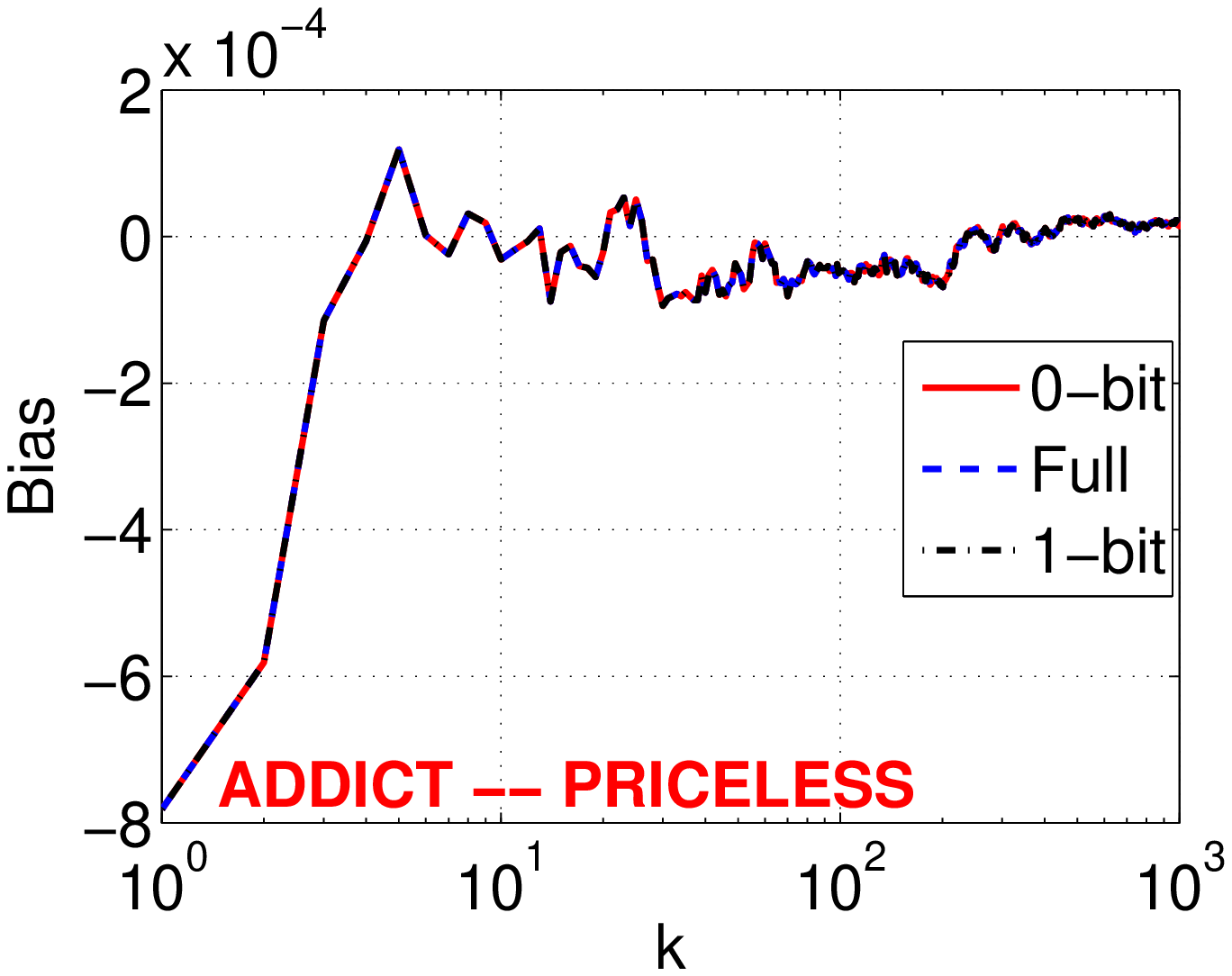}\hspace{-0.14in}
\includegraphics[width=1.75in]{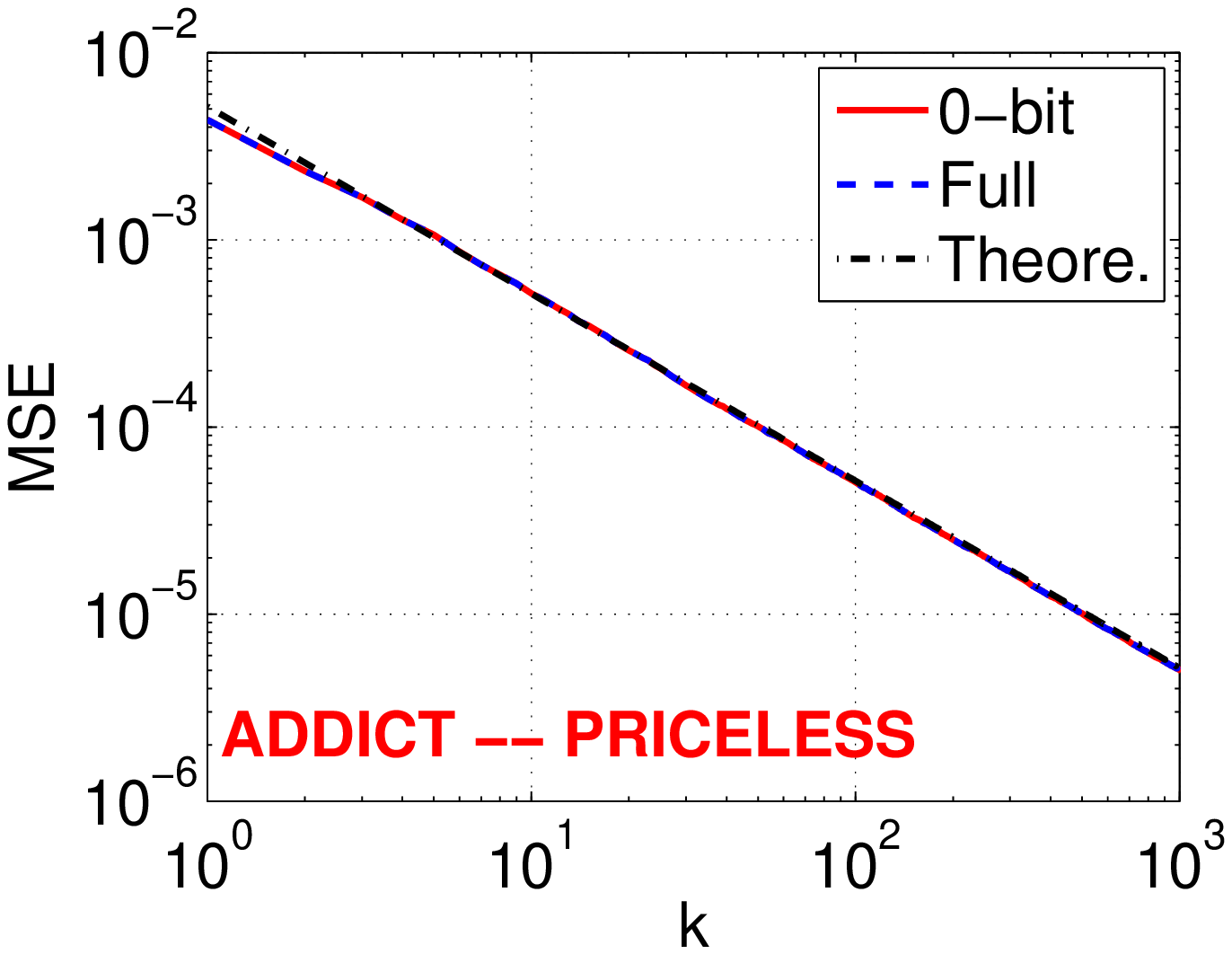}
}

\vspace{-0.1in}

\mbox{
\includegraphics[width=1.75in]{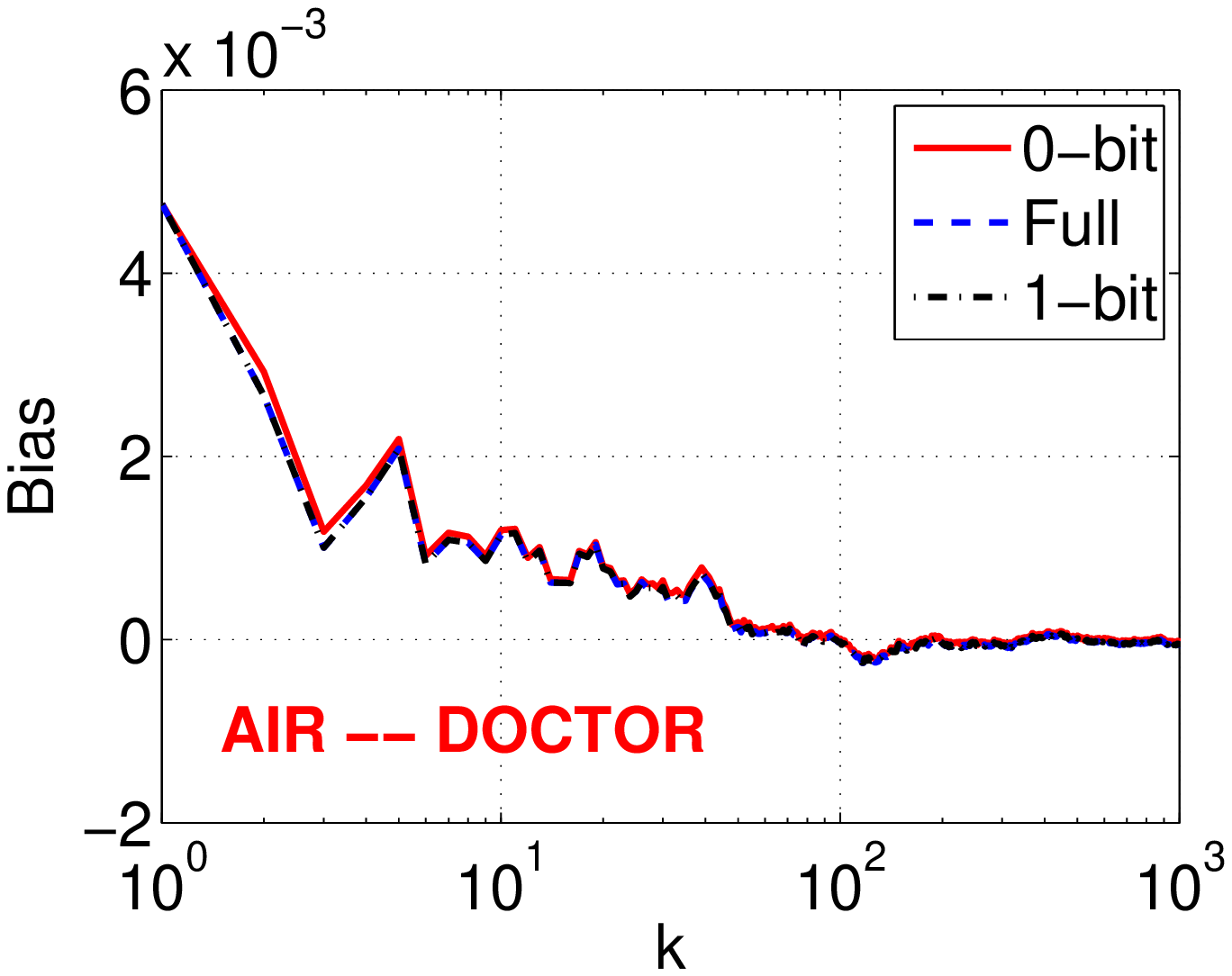}\hspace{-0.14in}
\includegraphics[width=1.75in]{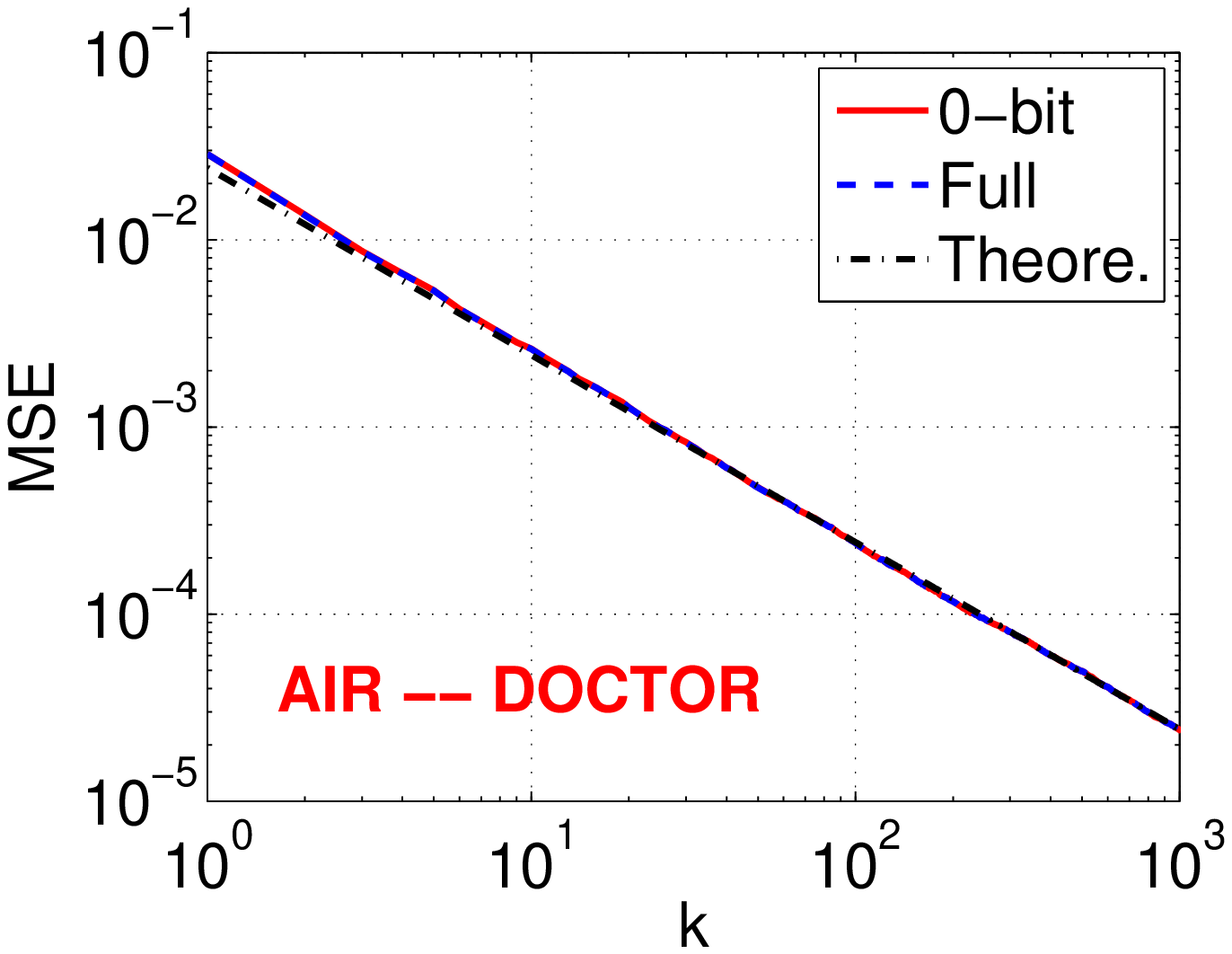}
}

\vspace{-0.1in}

\mbox{
\includegraphics[width=1.75in]{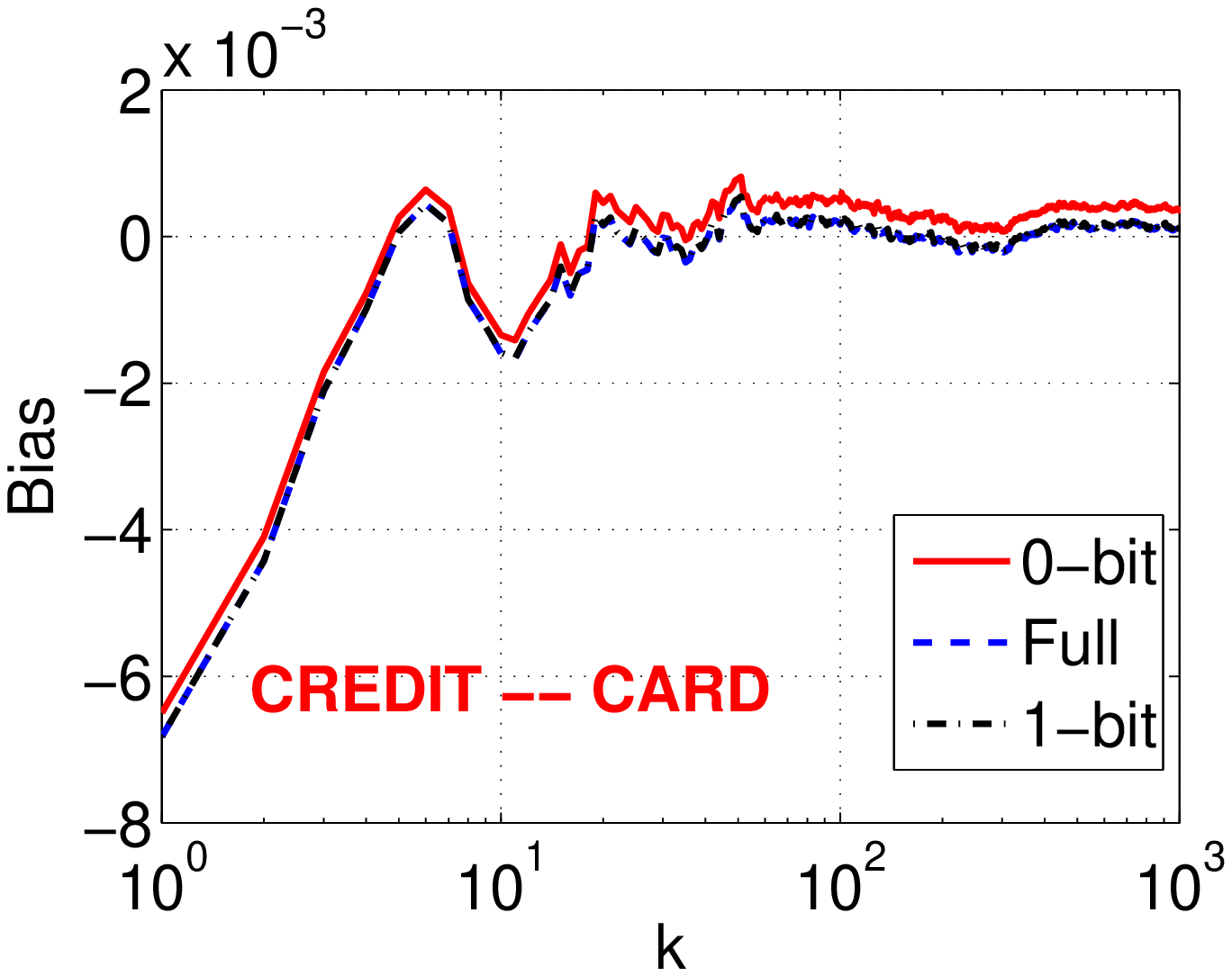}\hspace{-0.14in}
\includegraphics[width=1.75in]{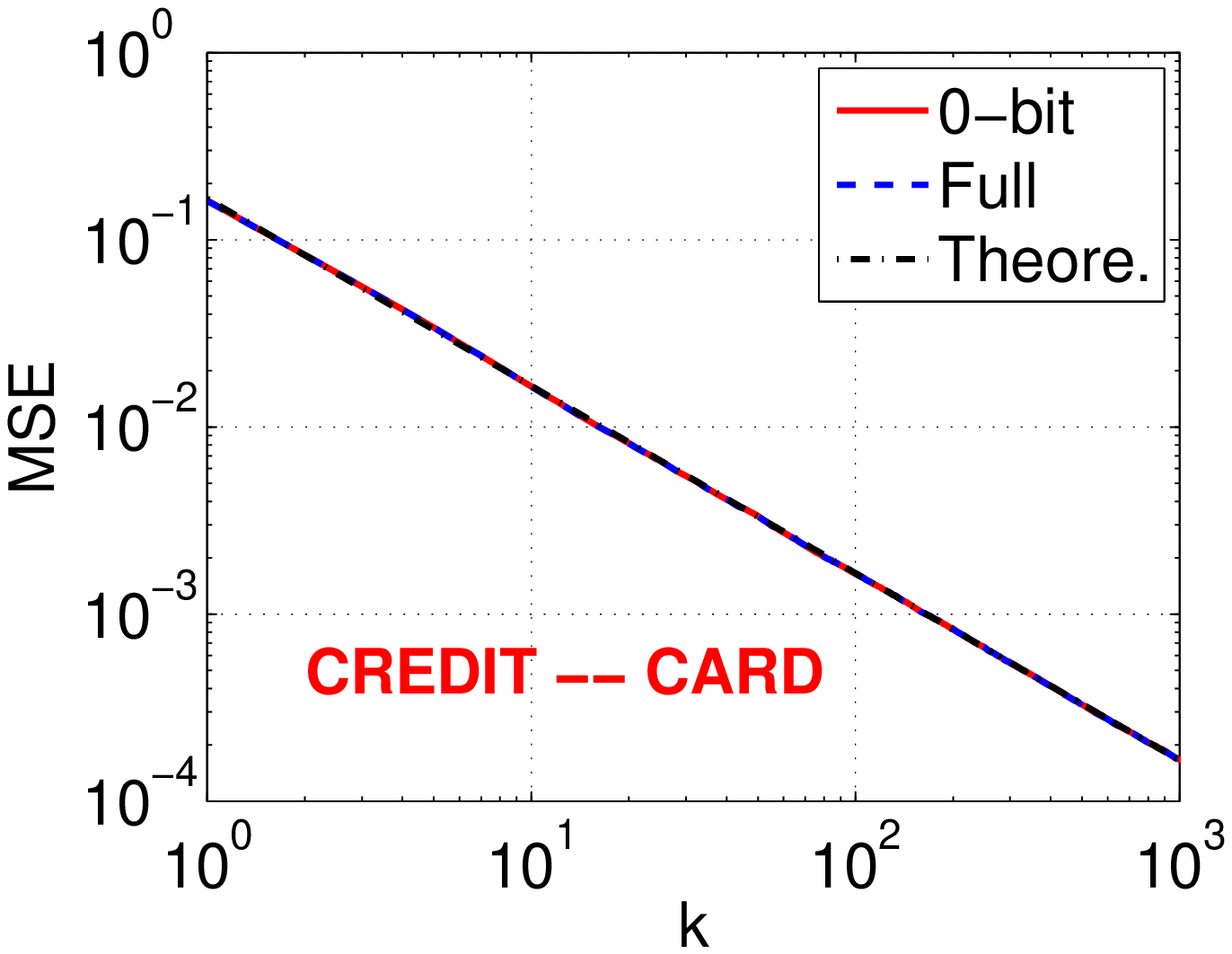}
}

\vspace{-0.1in}

\mbox{
\includegraphics[width=1.75in]{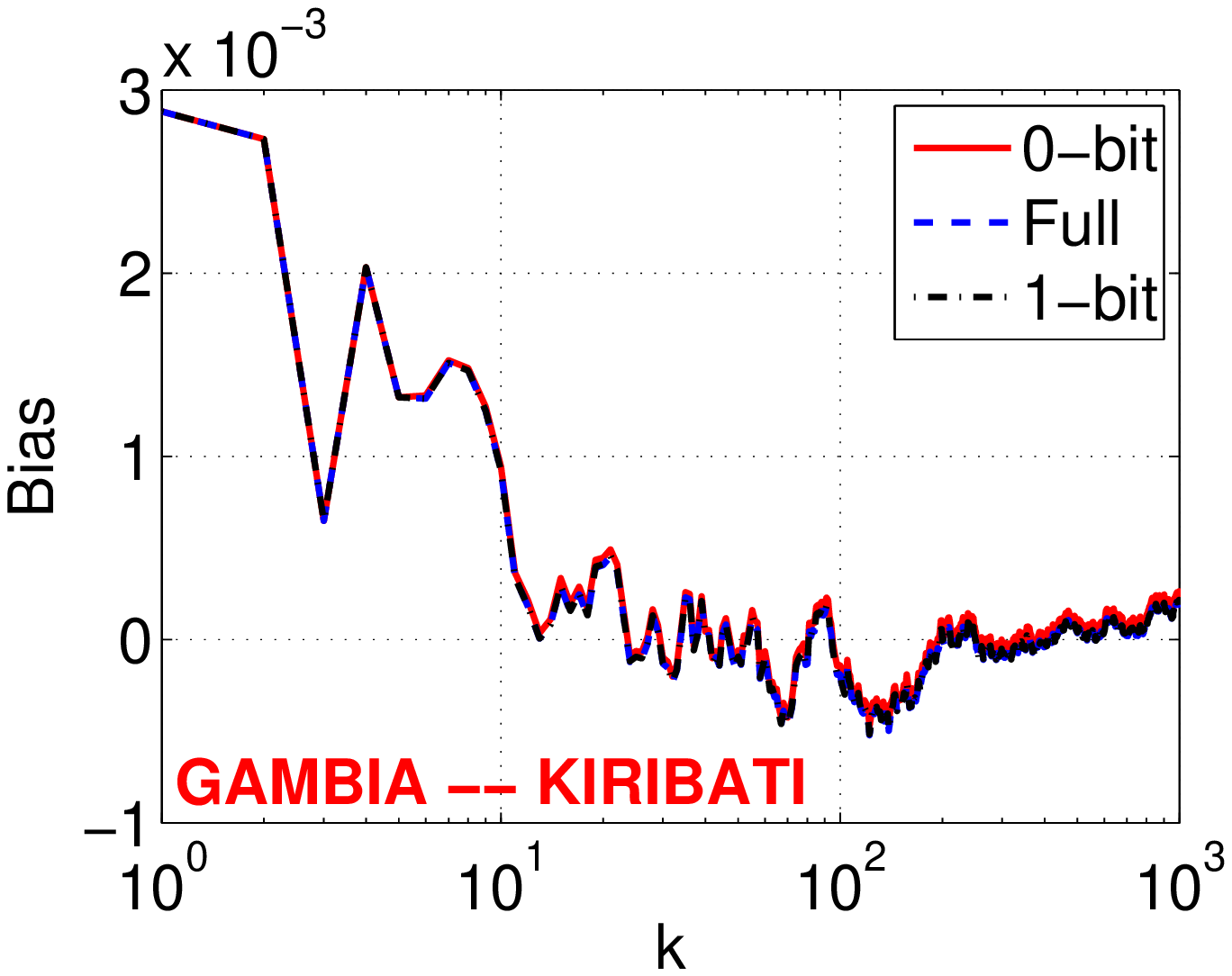}\hspace{-0.14in}
\includegraphics[width=1.75in]{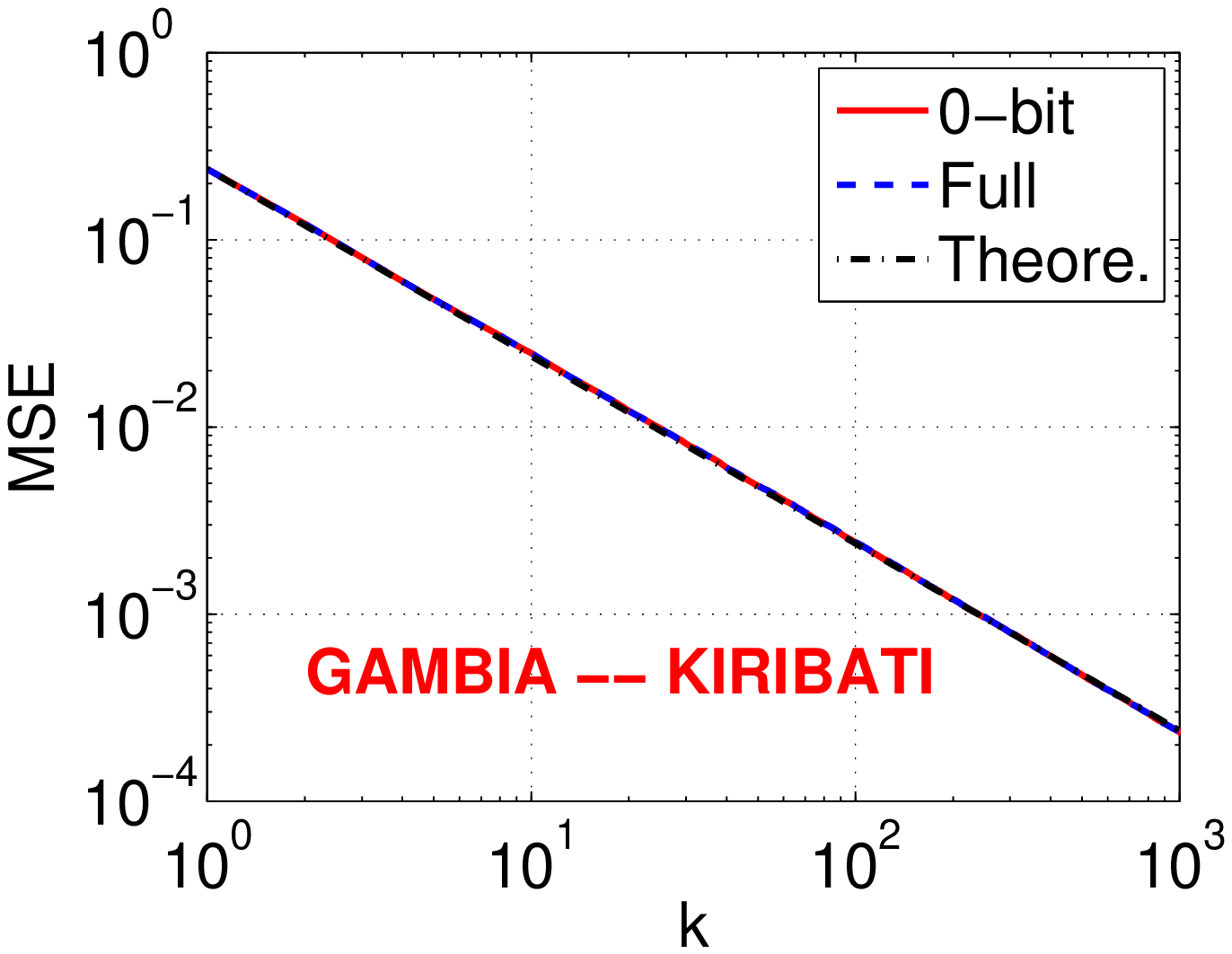}
}

\vspace{-0.1in}

\mbox{
\includegraphics[width=1.75in]{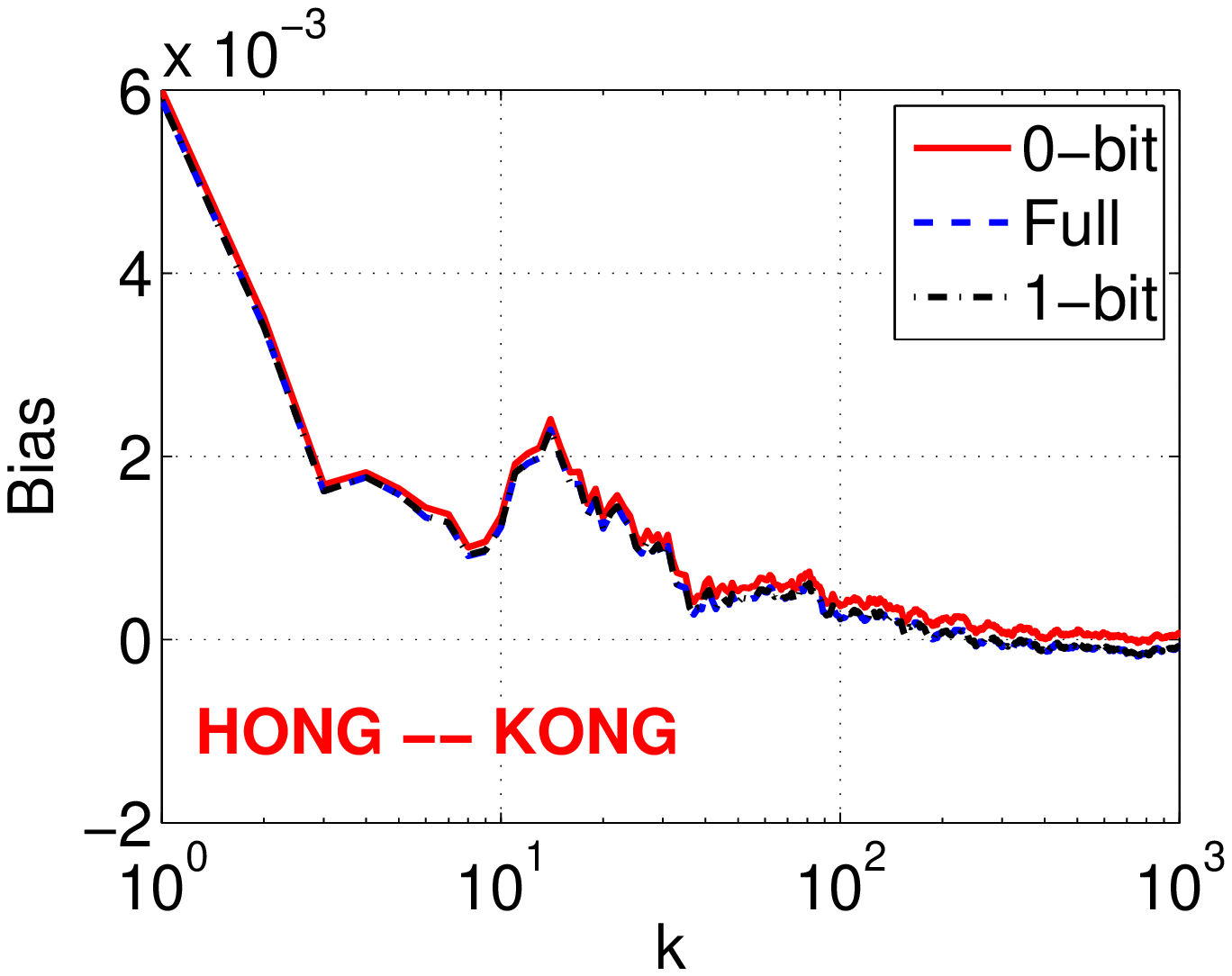}\hspace{-0.14in}
\includegraphics[width=1.75in]{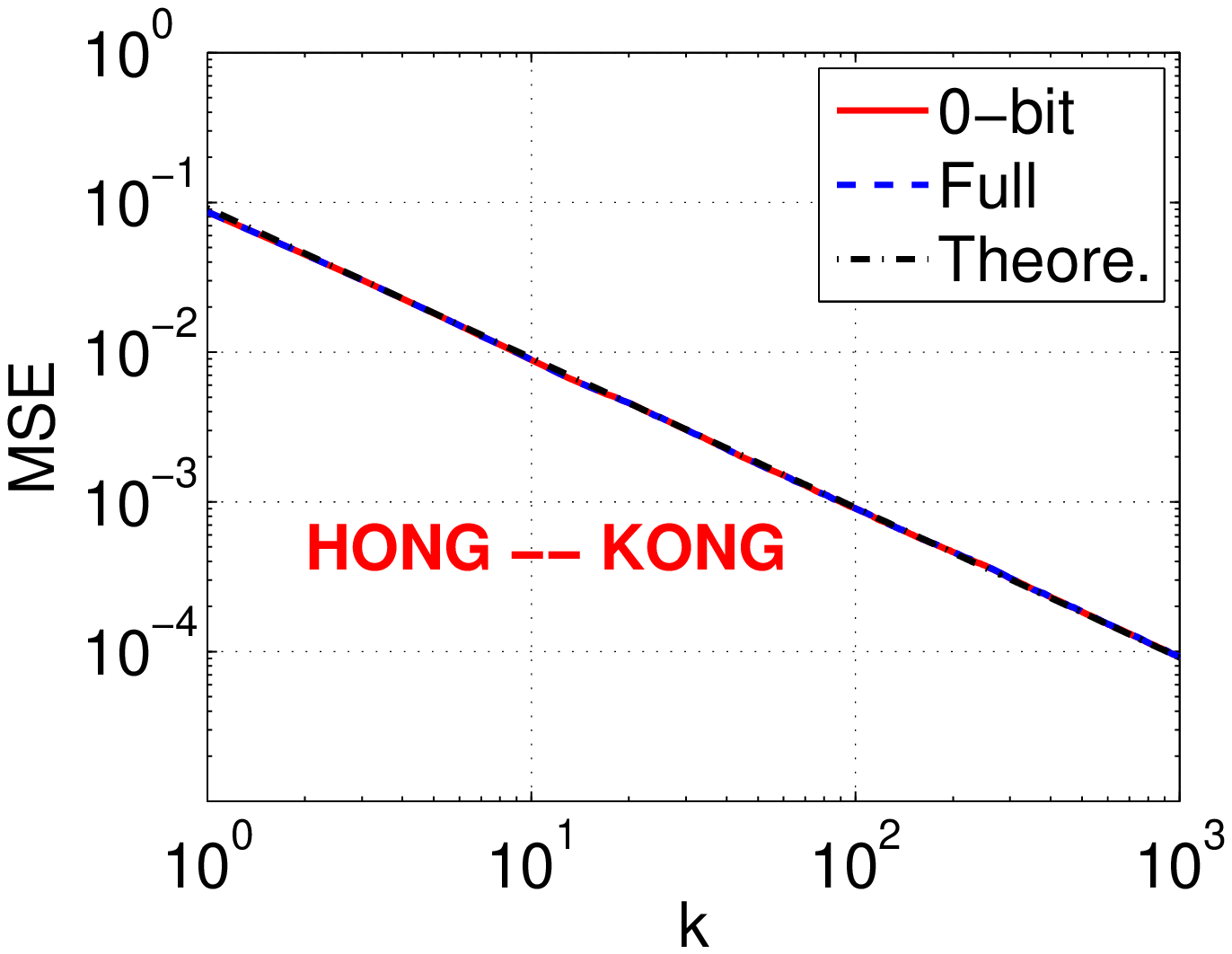}
}

\end{center}
\vspace{-0.3in}
\caption{\small Results for estimating min-max kernels using the ``full'' scheme by recording all the bits of ($i^*, t^*$) and the ``0-bit'' scheme by  discarding $t^*$. For each word pair and  $k$, we conducted simulations 10,000 times to compute the mean square errors (MSE) and the biases.  The empirical MSE curves (right column) show that both the 0-bit  and the full scheme match the theoretical variance. The empirical biases (left column) present a magnified view of  errors. For a few pairs (also see Figure~\ref{fig_CWS_Words_2}), the estimations by the 0-bit scheme have  noticeable ($\ll 10^{-4}$) biases. By using the ``1-bit'' scheme (i.e., by recording whether $t^*$ is even or odd), these biases vanish.   }\label{fig_CWS_Words_1}
\end{figure}

\begin{figure}[h!]
\begin{center}
\mbox{
\includegraphics[width=1.75in]{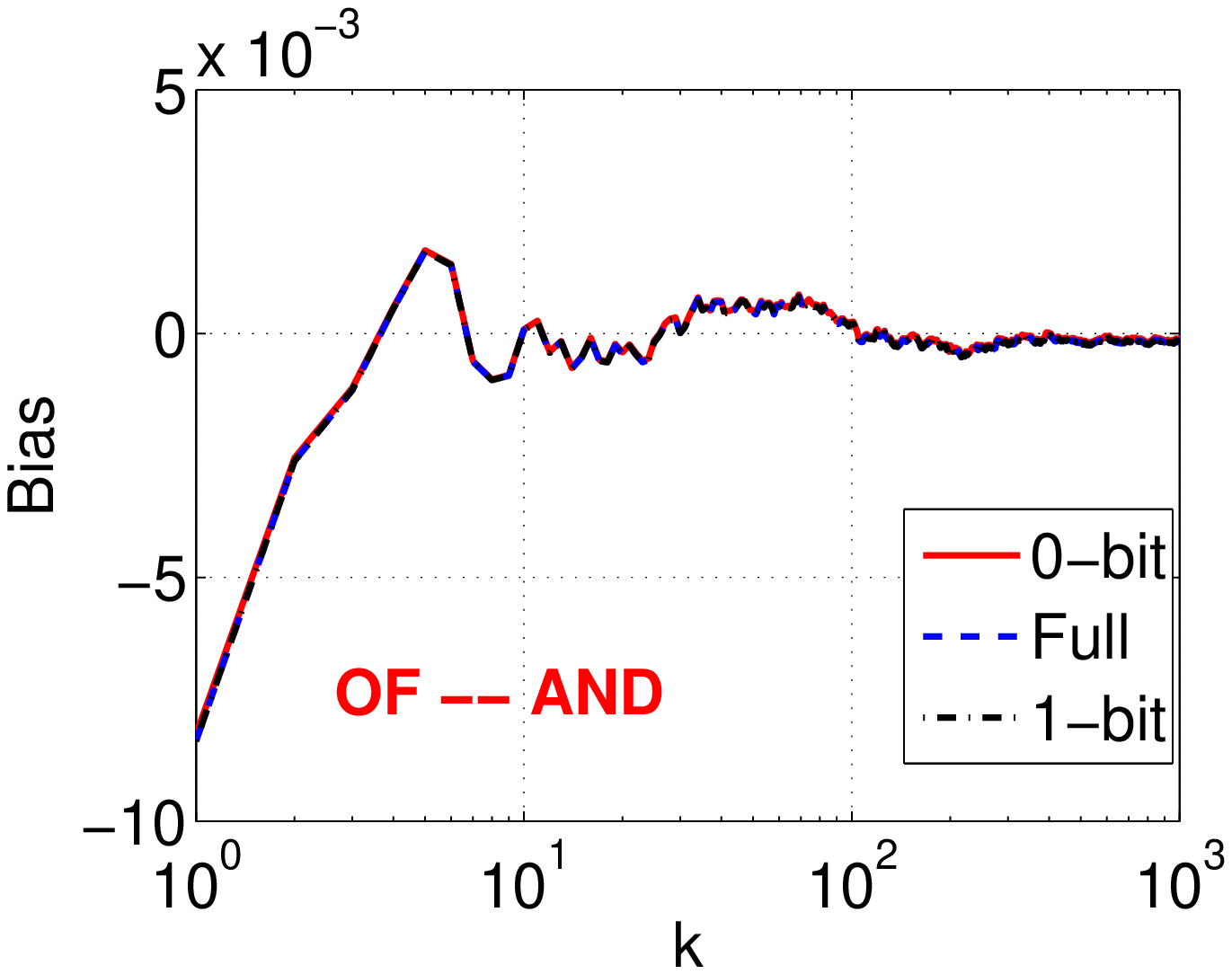}\hspace{-0.14in}
\includegraphics[width=1.75in]{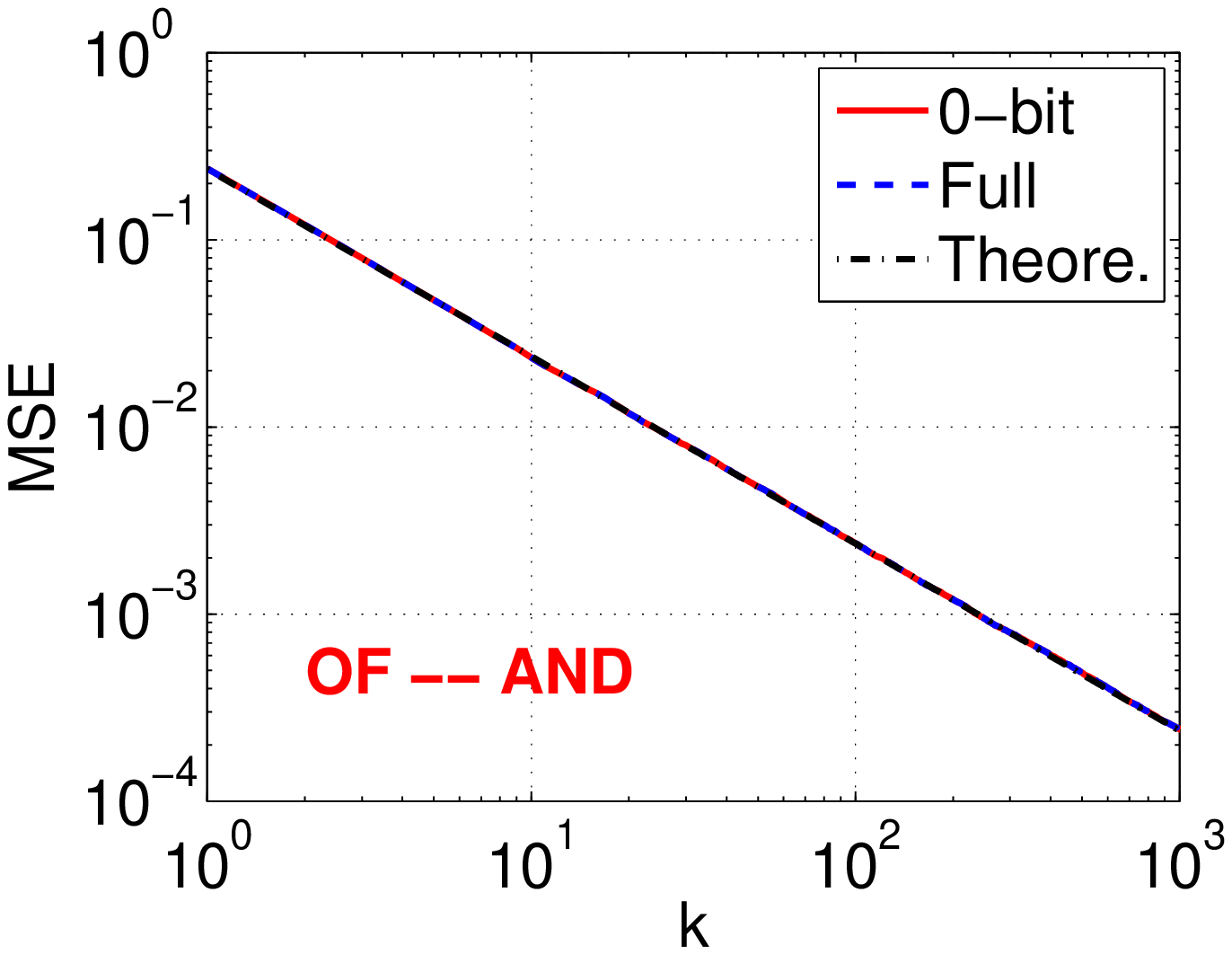}
}

\vspace{-0.1in}

\mbox{
\includegraphics[width=1.75in]{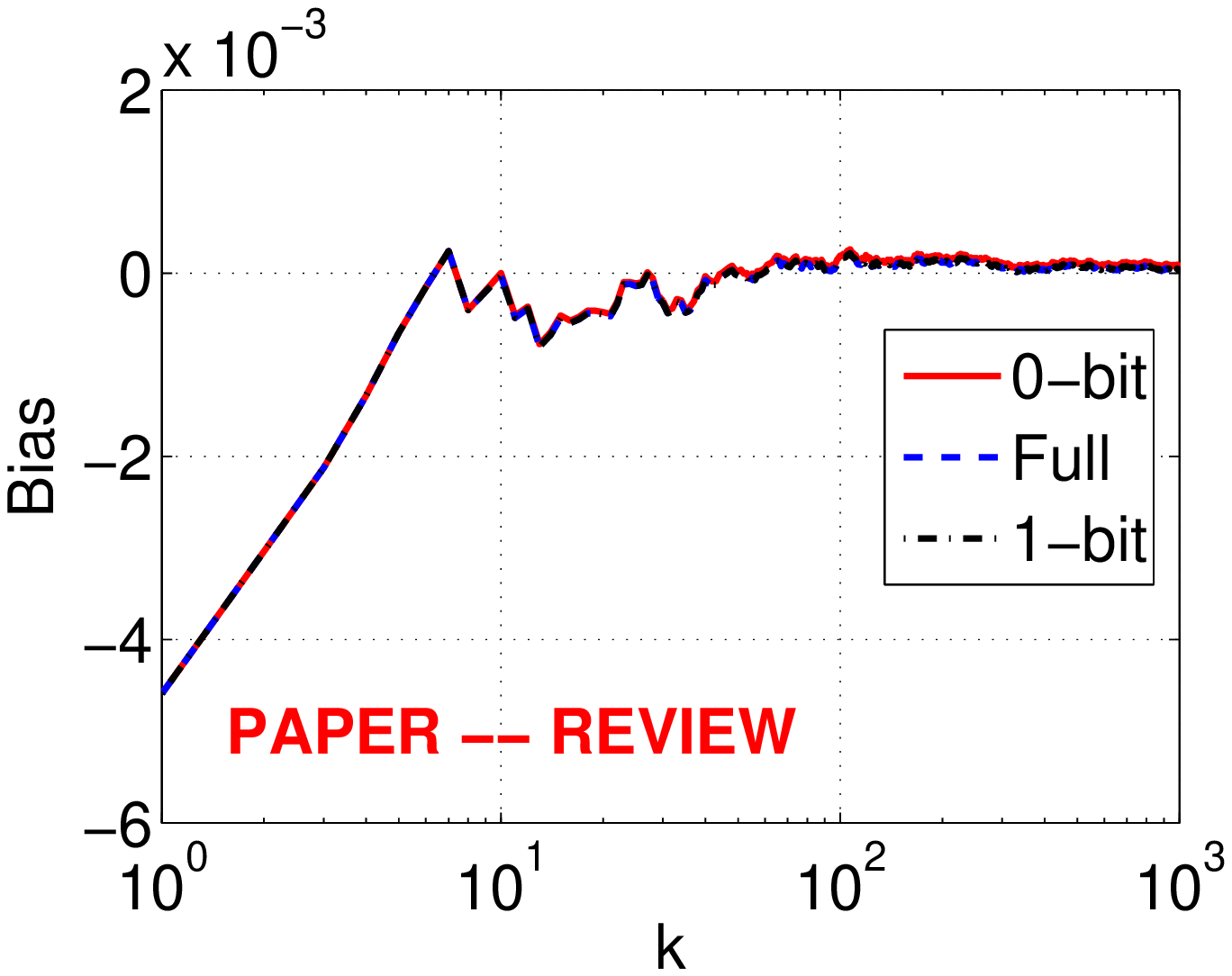}\hspace{-0.14in}
\includegraphics[width=1.75in]{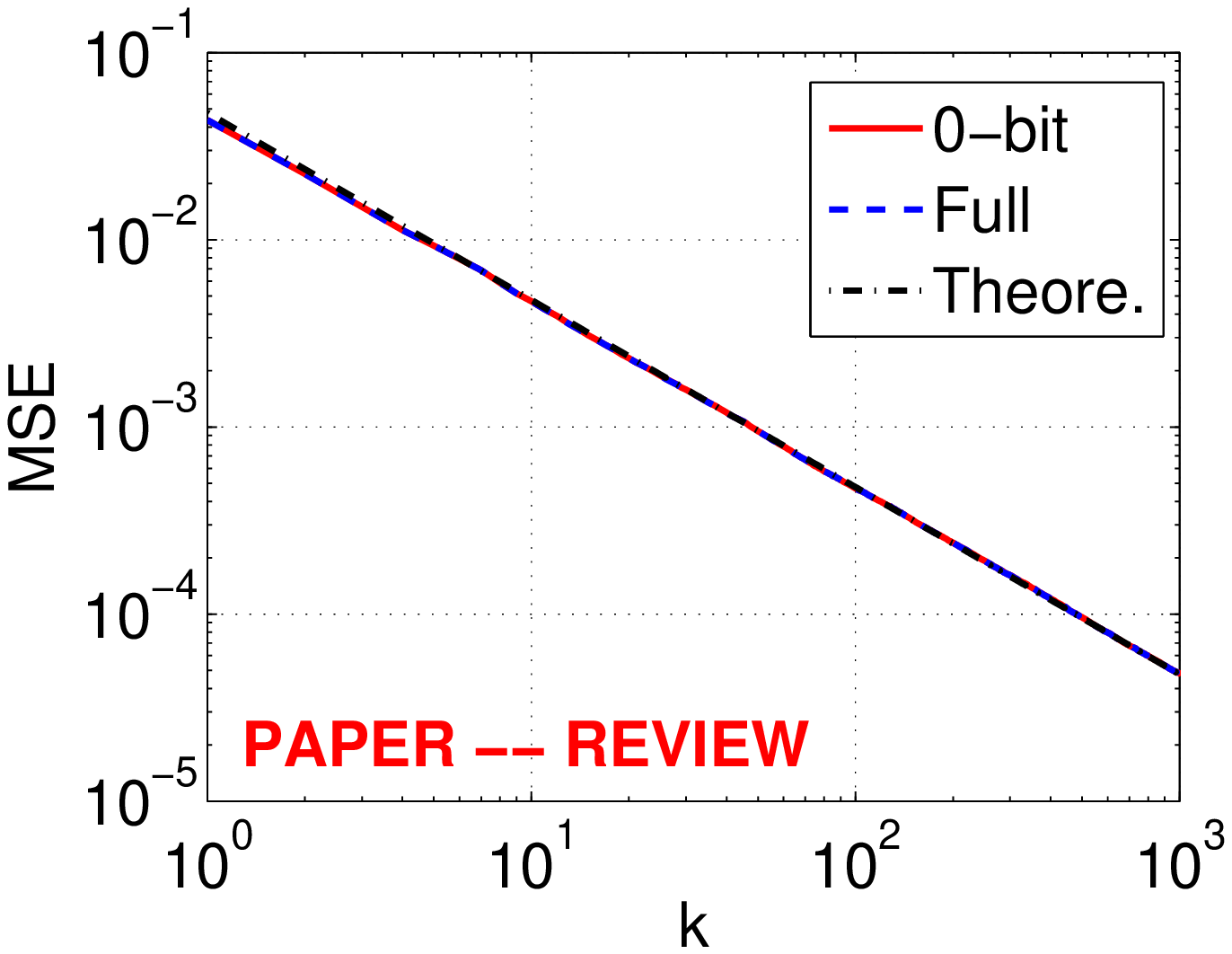}
}

\vspace{-0.1in}

\mbox{
\includegraphics[width=1.75in]{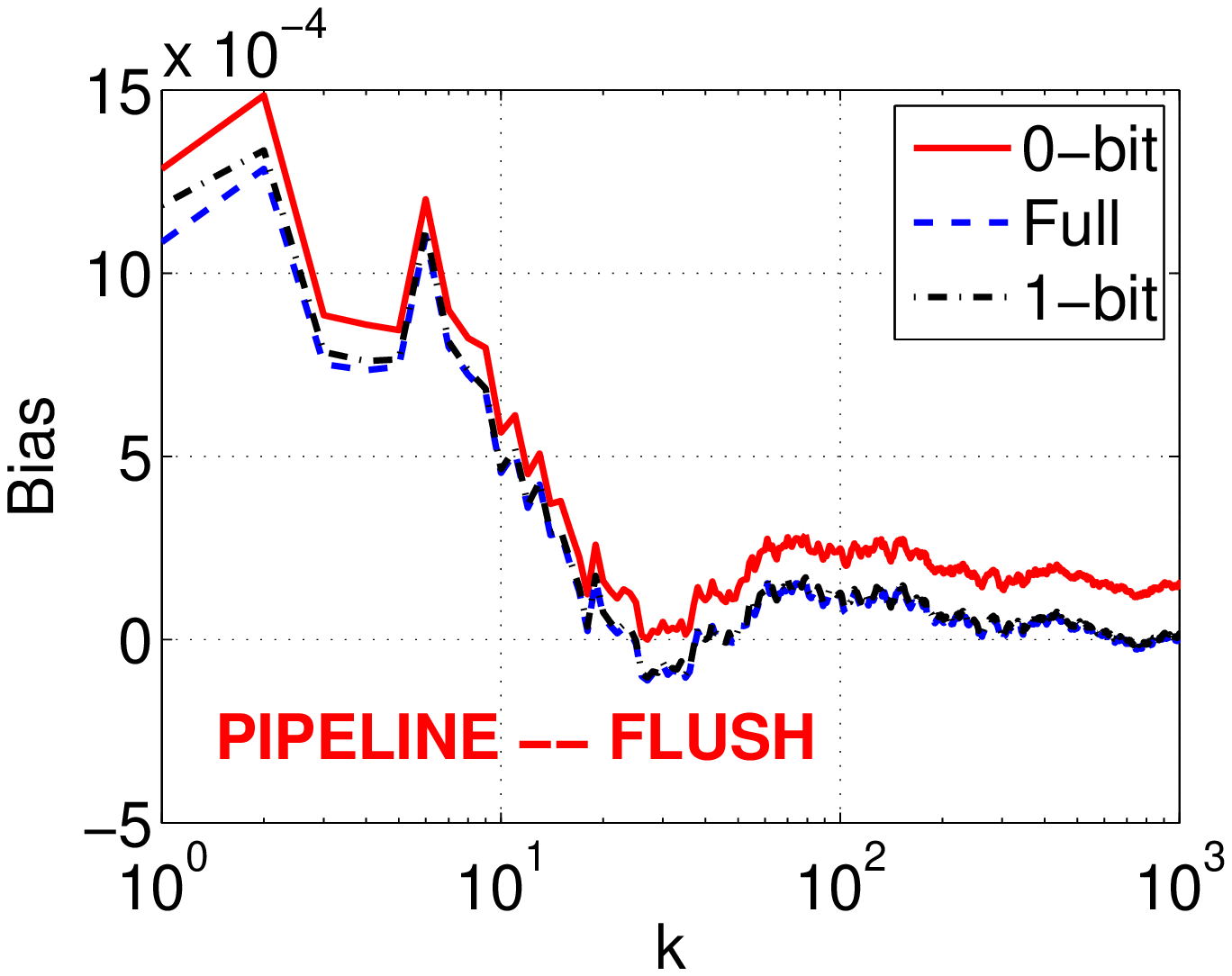}\hspace{-0.14in}
\includegraphics[width=1.75in]{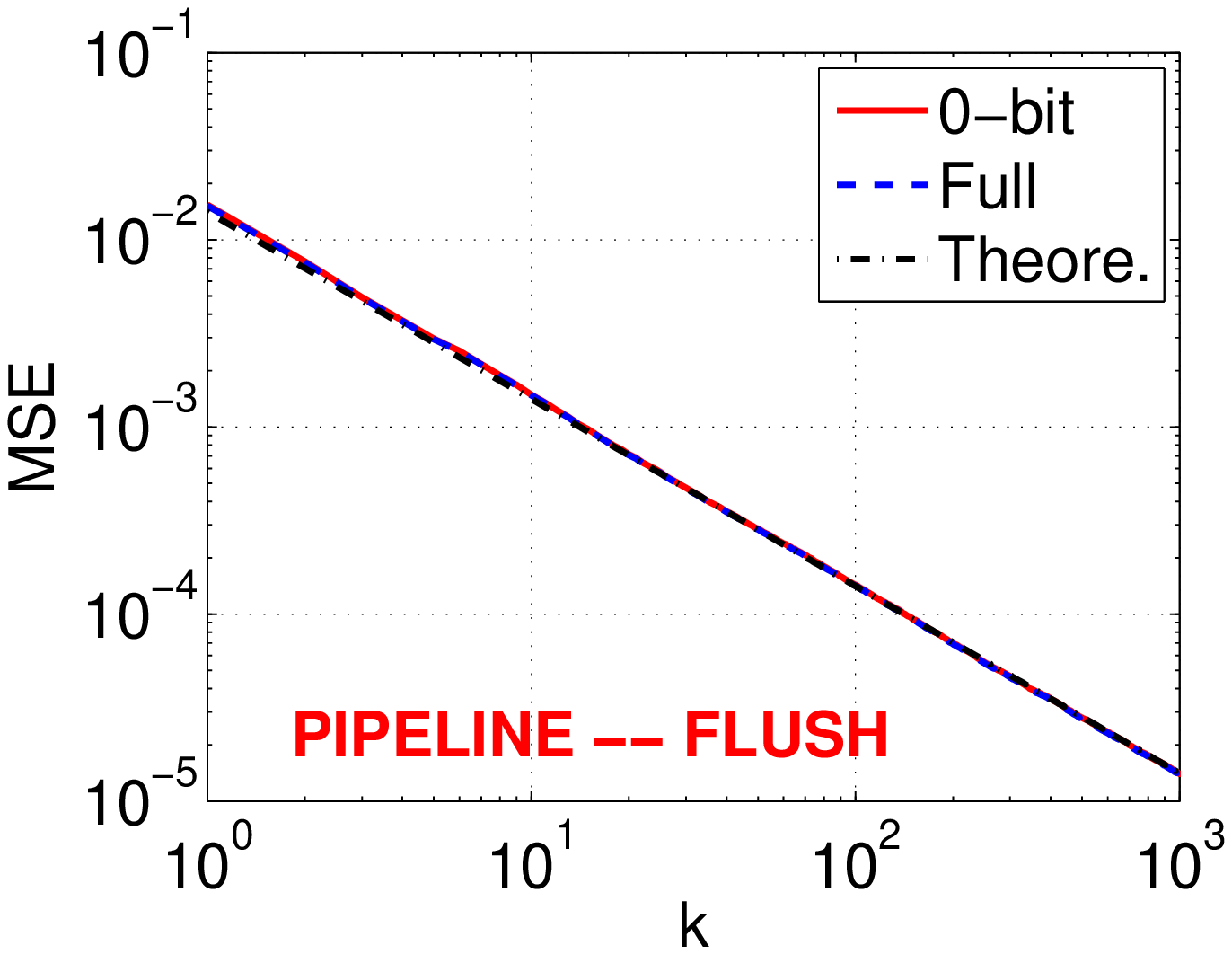}
}

\vspace{-0.1in}

\mbox{
\includegraphics[width=1.75in]{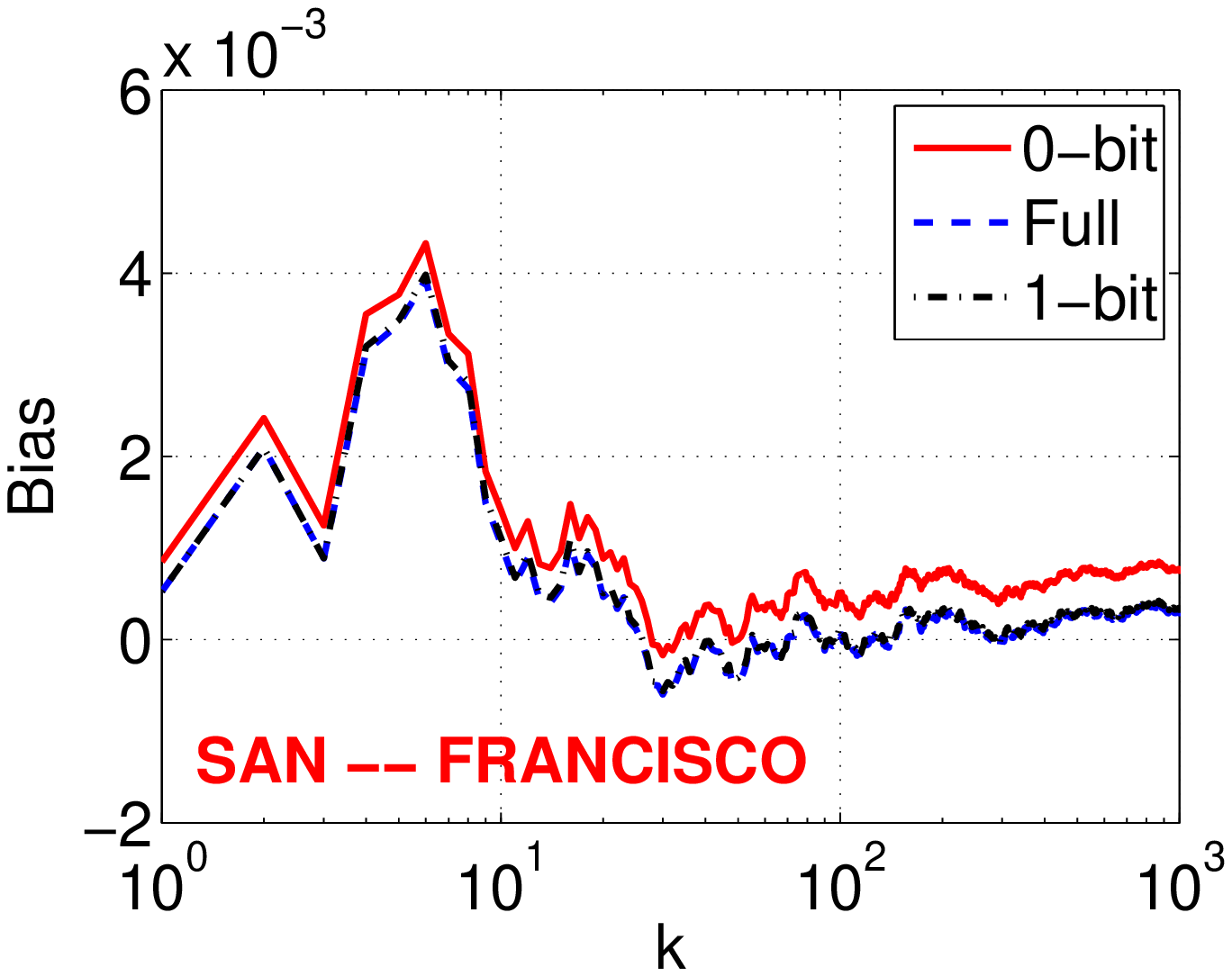}\hspace{-0.14in}
\includegraphics[width=1.75in]{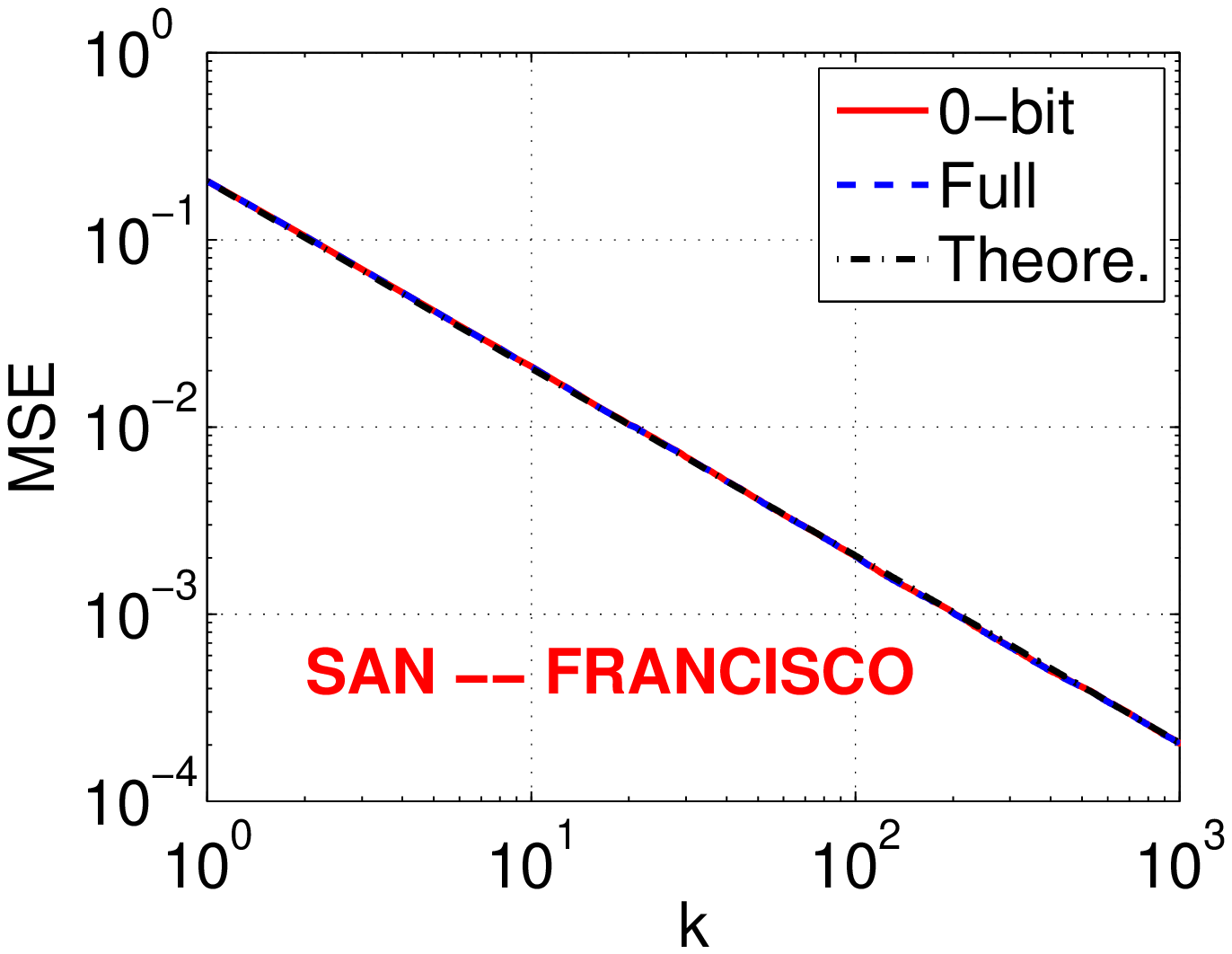}
}

\vspace{-0.1in}

\mbox{
\includegraphics[width=1.75in]{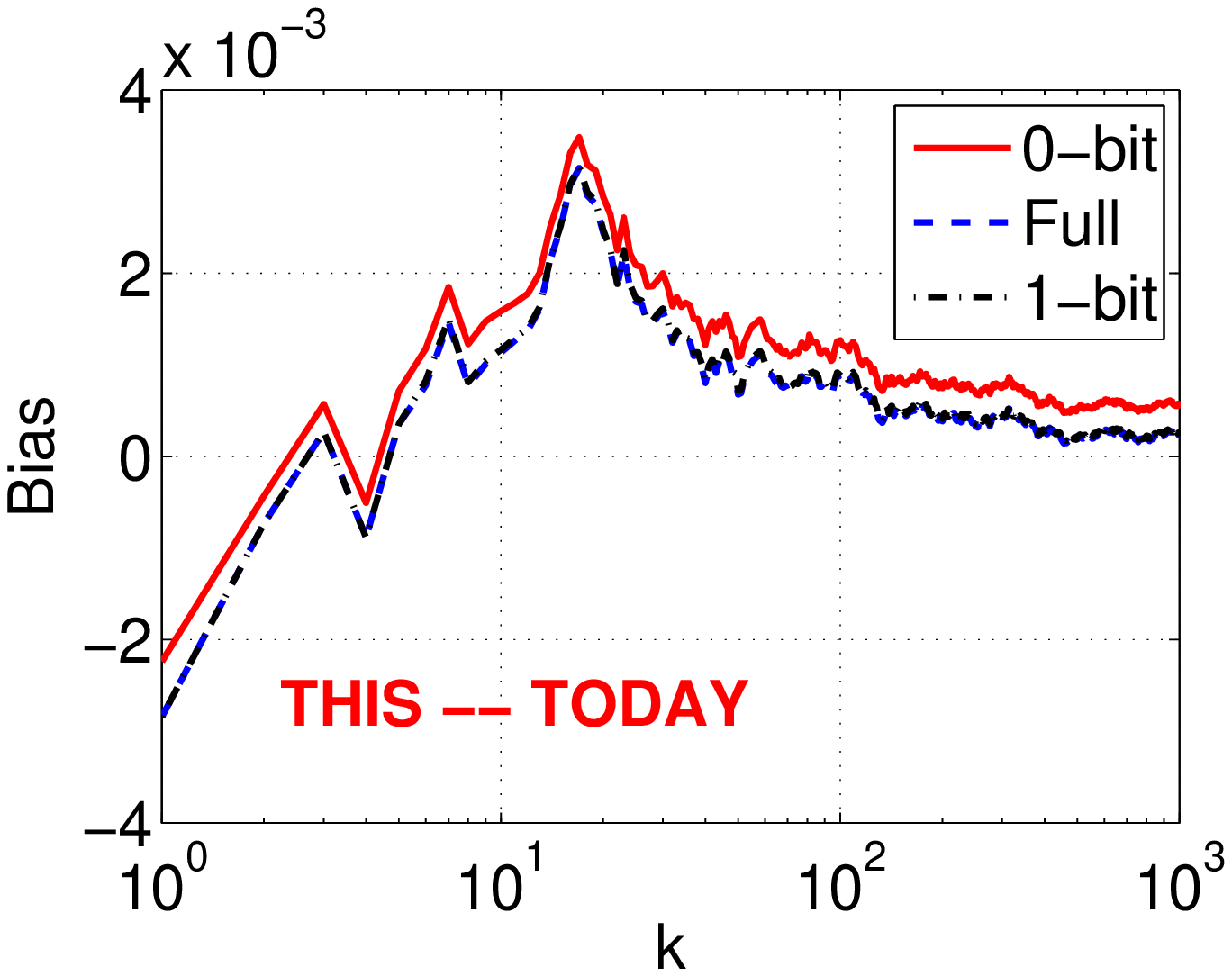}\hspace{-0.14in}
\includegraphics[width=1.75in]{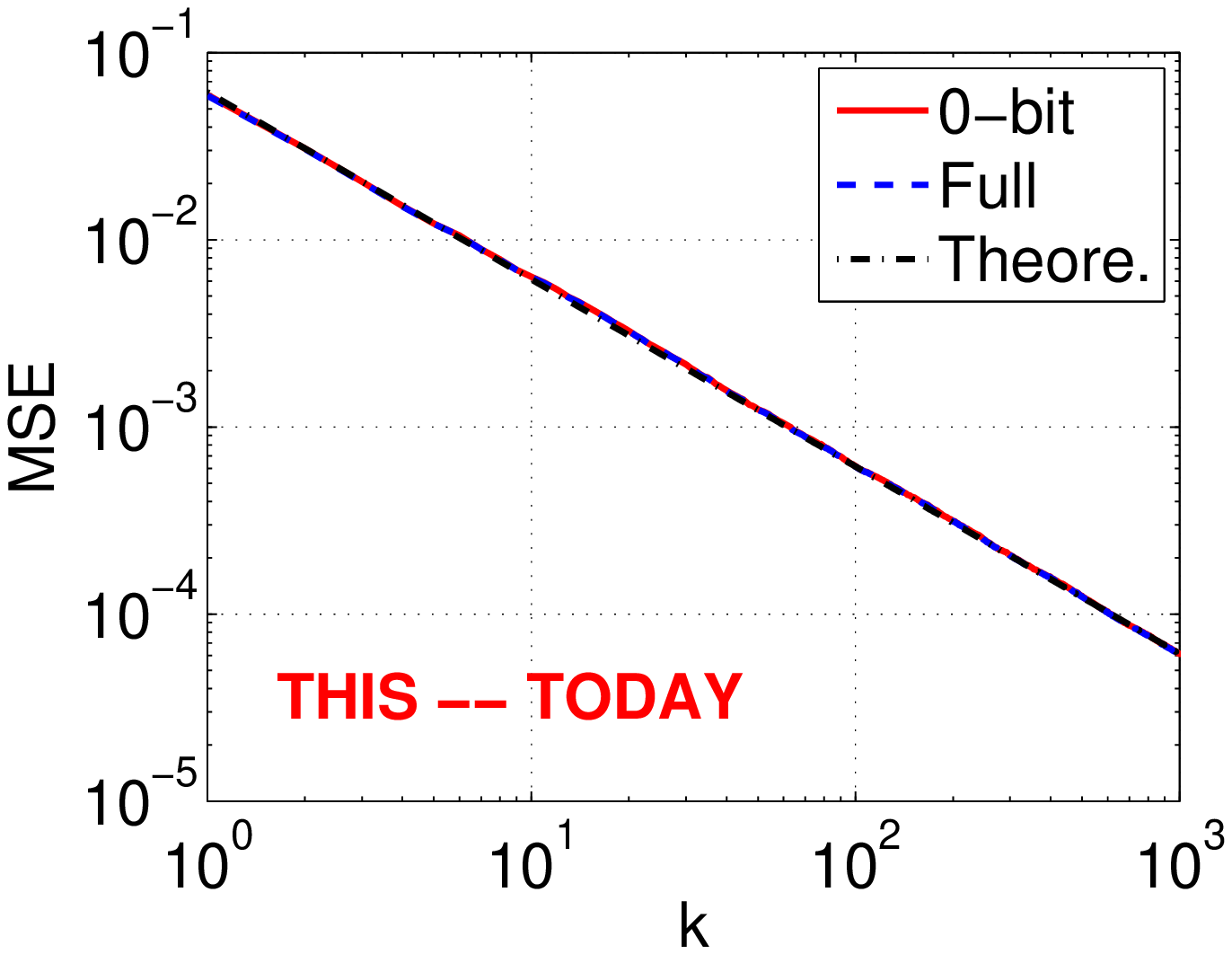}
}

\vspace{-0.1in}

\mbox{
\includegraphics[width=1.75in]{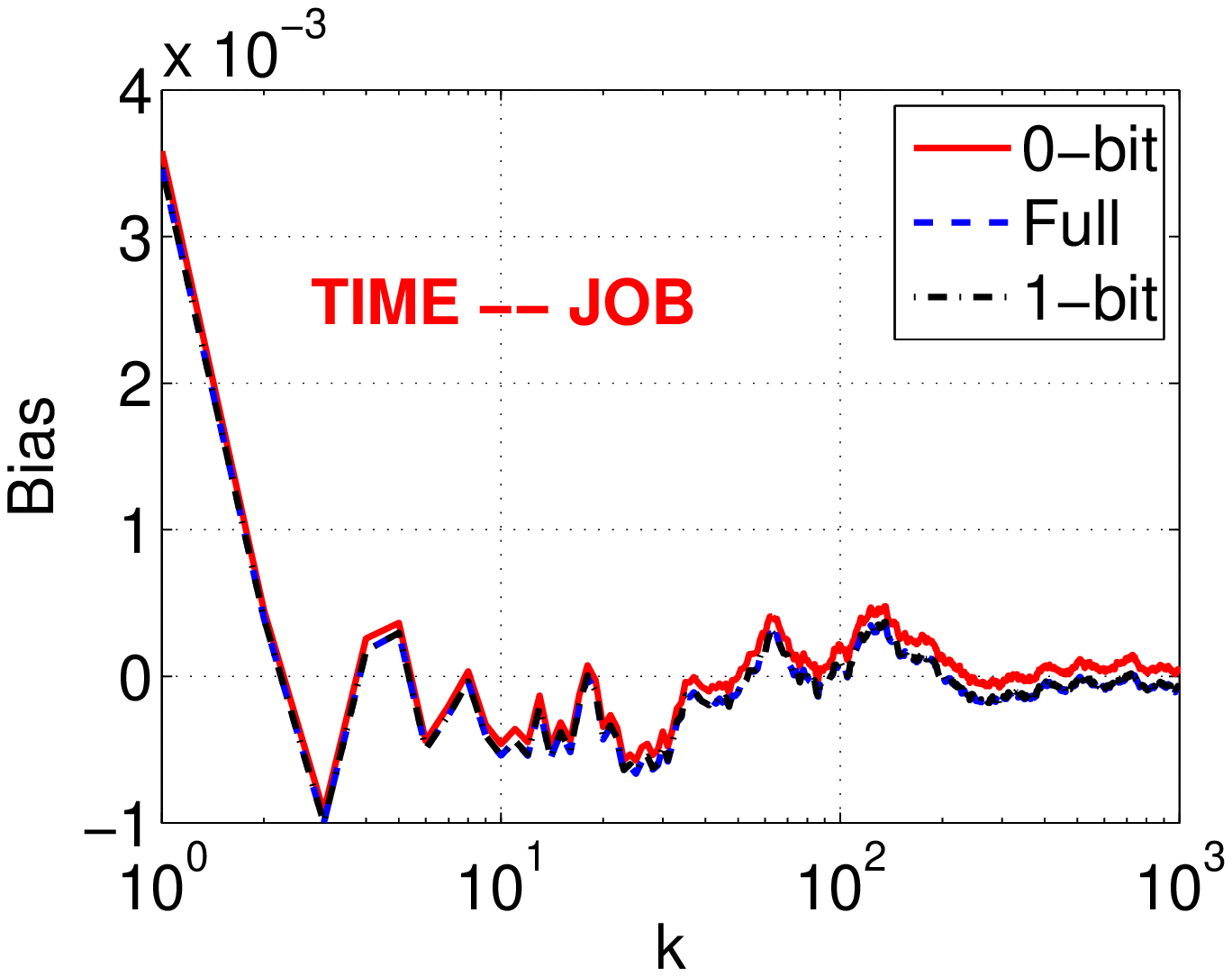}\hspace{-0.14in}
\includegraphics[width=1.75in]{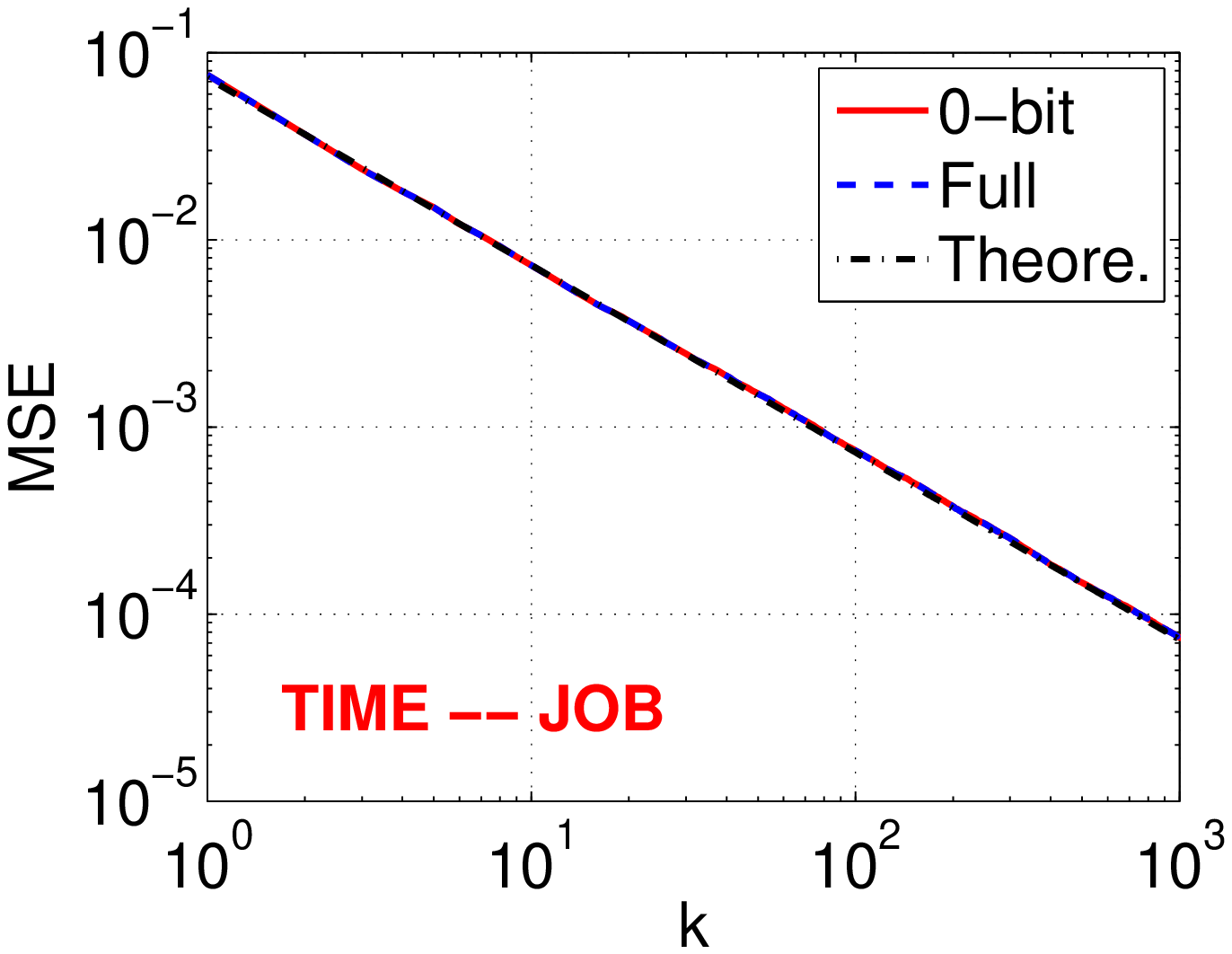}
}

\vspace{-0.1in}

\mbox{
\includegraphics[width=1.75in]{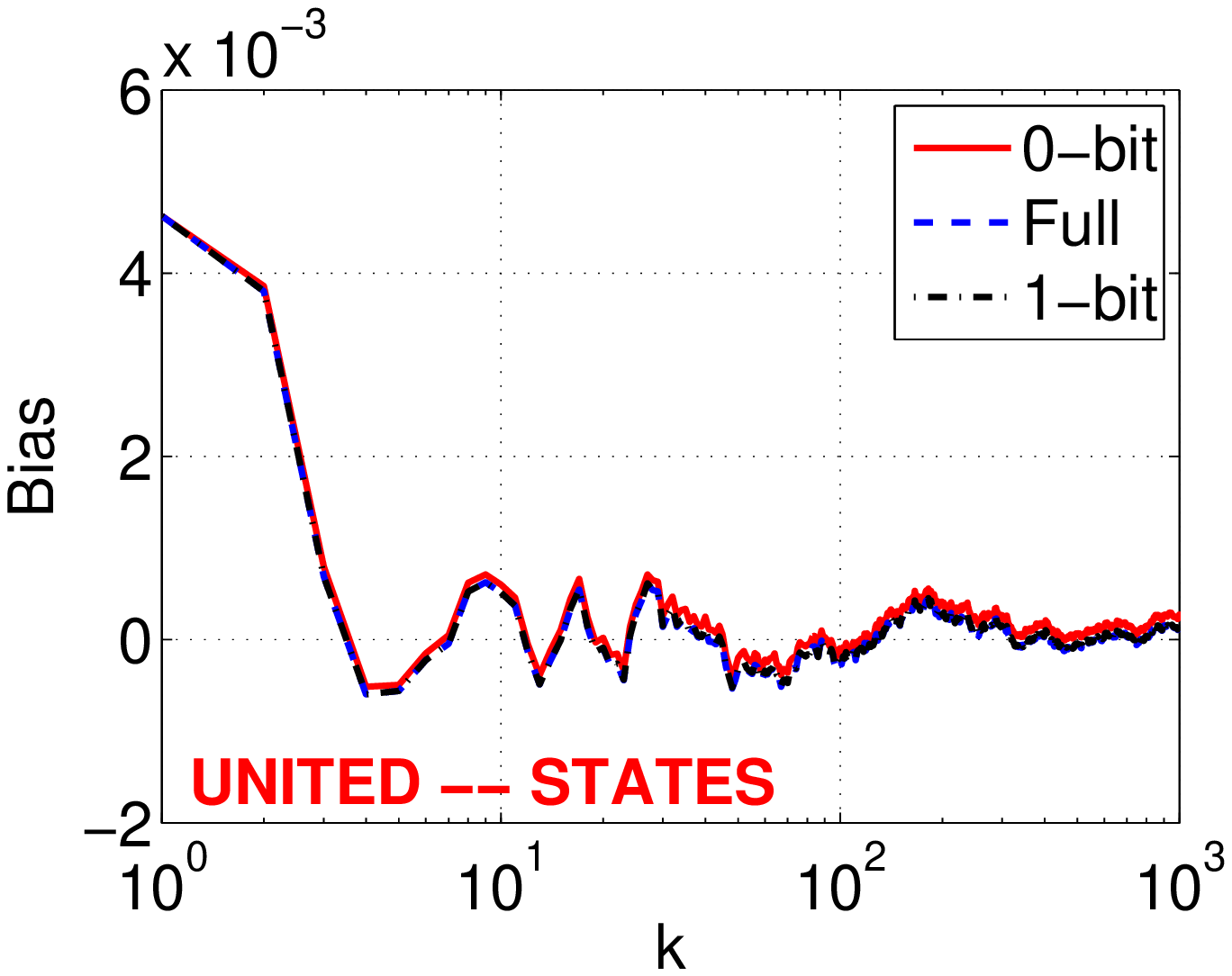}\hspace{-0.14in}
\includegraphics[width=1.75in]{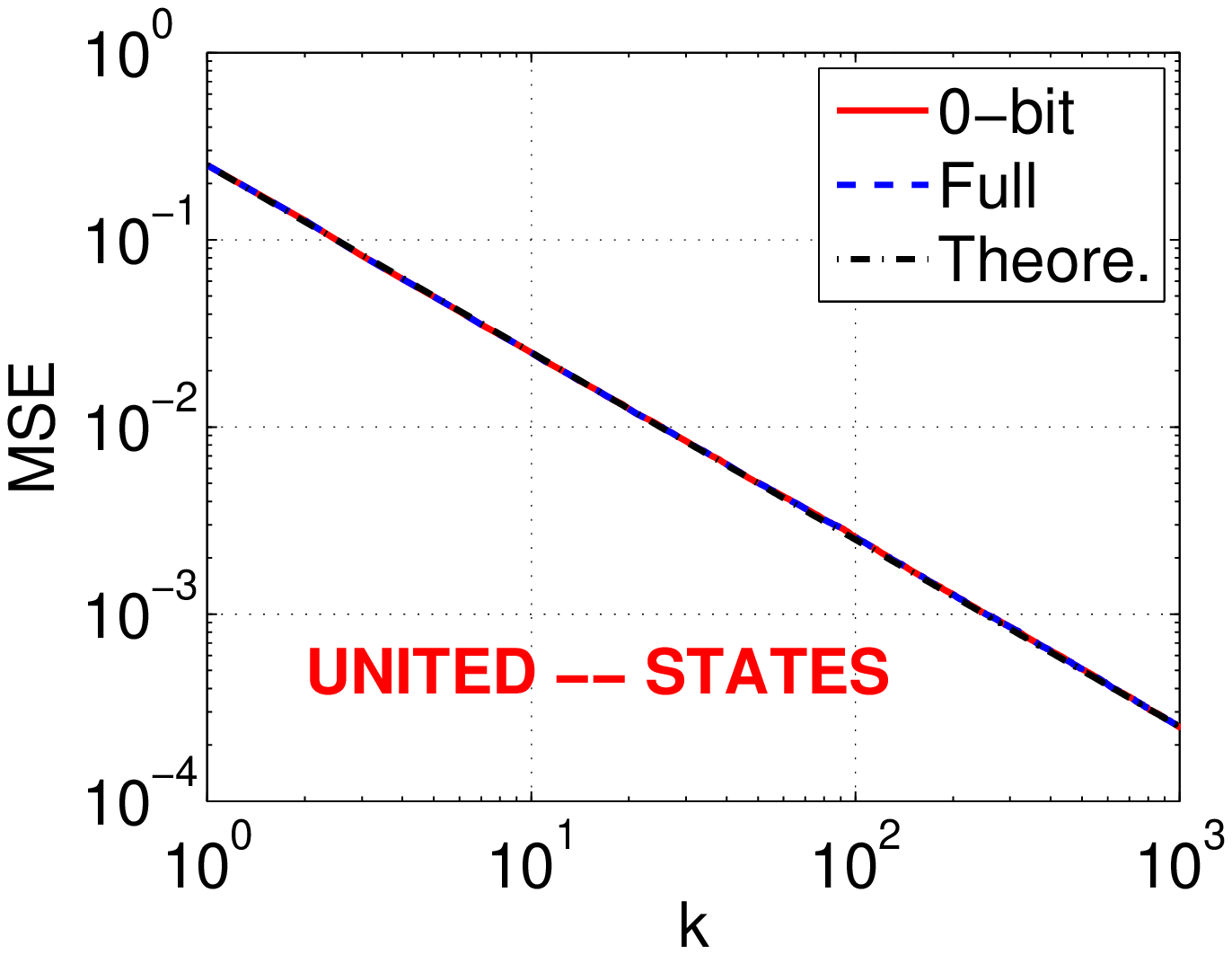}
}

\end{center}
\vspace{-0.25in}
\caption{Simulations for estimating min-max kernels. See the caption of Figure~\ref{fig_CWS_Words_1} for more details.}\label{fig_CWS_Words_2}
\end{figure}

We have experimented with many more pairs of words than these 13 pairs but the results look essentially the same, i.e., no practical difference between the 0-bit scheme and the full scheme, as can be shown in Figures~\ref{fig_CWS_Words_1} to~\ref{fig_CWS_Words_2}.

In the experiment, we let $k$ vary from 1 to 1000 and  estimate $K_{MM}$ from $k$ measurements $(i^*_{,j},t^*_{,j})$, $j=1$ to $k$. With the full scheme, we keep all the bits of $t^*$. With the 0-bit scheme, we completely discard $t^*$. For each $k$, we repeat the simulations $10,000$ times to reliably compute the empirical mean square error (MSE) and the bias for each pair.

The right columns of Figures~\ref{fig_CWS_Words_1} and~\ref{fig_CWS_Words_2} plot the empirical MSEs, together with the theoretical variance: $K_{MM}(1-K_{MM})/k$ (i.e., the variance of binomial). Because the curves for the 0-bit scheme and the full scheme overlap the theoretical variances, we can conclude, at least for these data, that our proposed 0-bit scheme is essentially unbiased and the variance matches the theoretical variance of the full scheme.

\vspace{0.05in}

To avoid many ``boring'' figures, we let $k$ be as small as 1 (while typical simulations would use a much large number such as 10 to start with). Nevertheless, these MSE curves are still quite boring since all the curves essentially overlap.

To make the presentations   somewhat more interesting, we also present the empirical biases in the left columns of the two figures. Now we can see some discrepancies between the two schemes typically on the order of $\ll10^{-4}$ (in the stabilized zone, i.e., when $k$ is not too small). While such small biases (at the 4th or 5th decimal points)  would not make any practical differences, they do serve the purpose to remind us that the 0-bit scheme is indeed an approximation.

To make the plots even more interesting, we add the curves for the ``1-bit'' scheme (i.e., by recording whether $t^*$ is even or odd). For ``CREDIT-CARD'', ``PIPELINE-FLUSH'', ``SAN-FRANCISCO'', and ``THIS-TODAY'', we can observe  (very small) differences between the 0-bit scheme and the full-scheme. The differences vanish once we use the ``1-bit'' scheme.

\vspace{0.05in}

From Table~\ref{tab_word_pairs}, we can see that binarizing the data usually lead to very different similarities (i.e., the last two columns, i.e., $R$ and $MM$, differ significantly). The 0-bit scheme, which only uses $i^*$, still very well approximates the original min-max kernel instead of the resemblance kernel. This confirm that, even though our samples (i.e., $i^*$) in the same format as  samples from minwise hashing (for example, both are integers bounded by $D$), they are statistically very different samples. In other words, our 0-bit scheme is not the same as simply doing the original minwise hashing.

\vspace{0.05in}

\begin{figure}[h!]
\begin{center}

\mbox{
\includegraphics[width=1.6in]{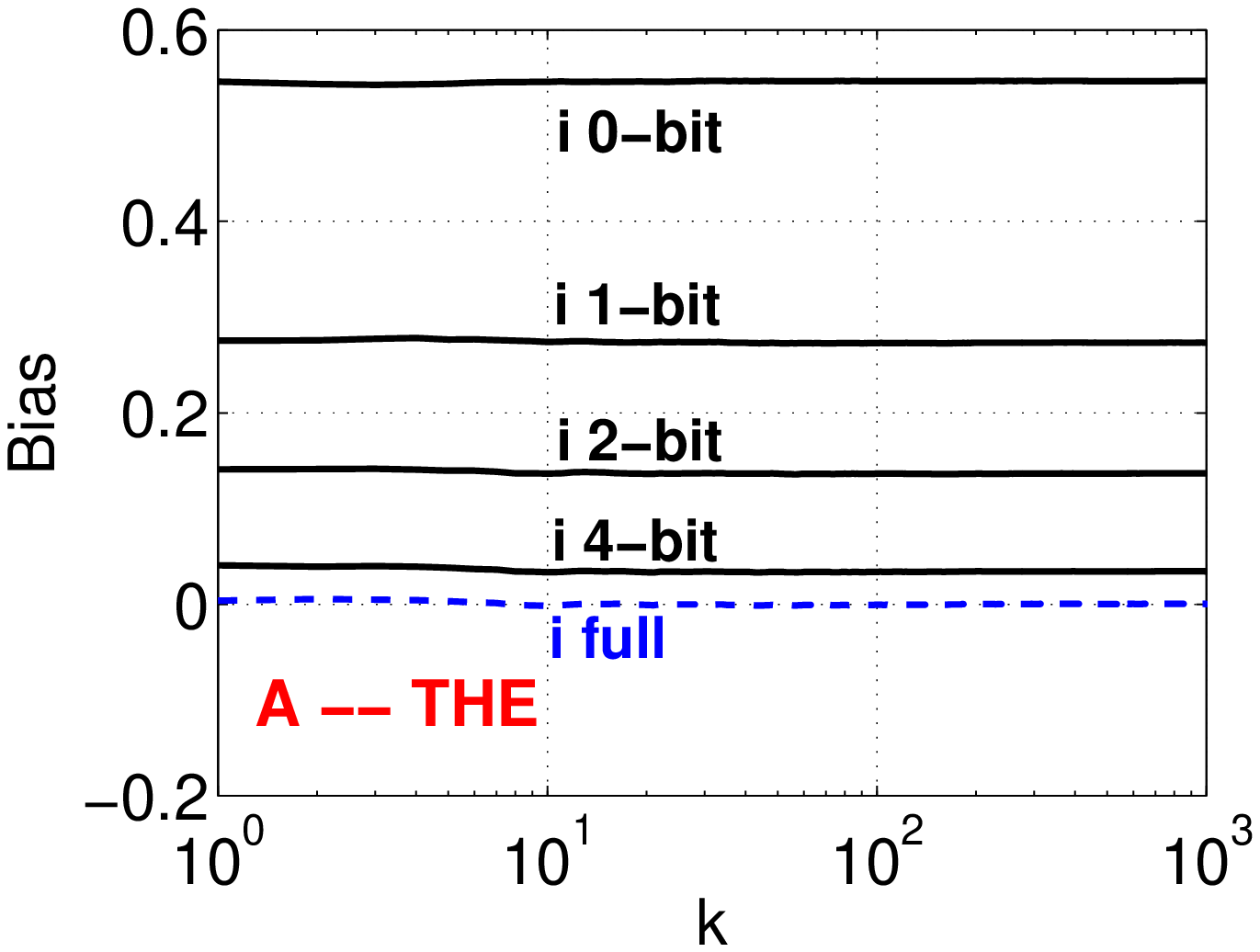}\hspace{-0.05in}
\includegraphics[width=1.6in]{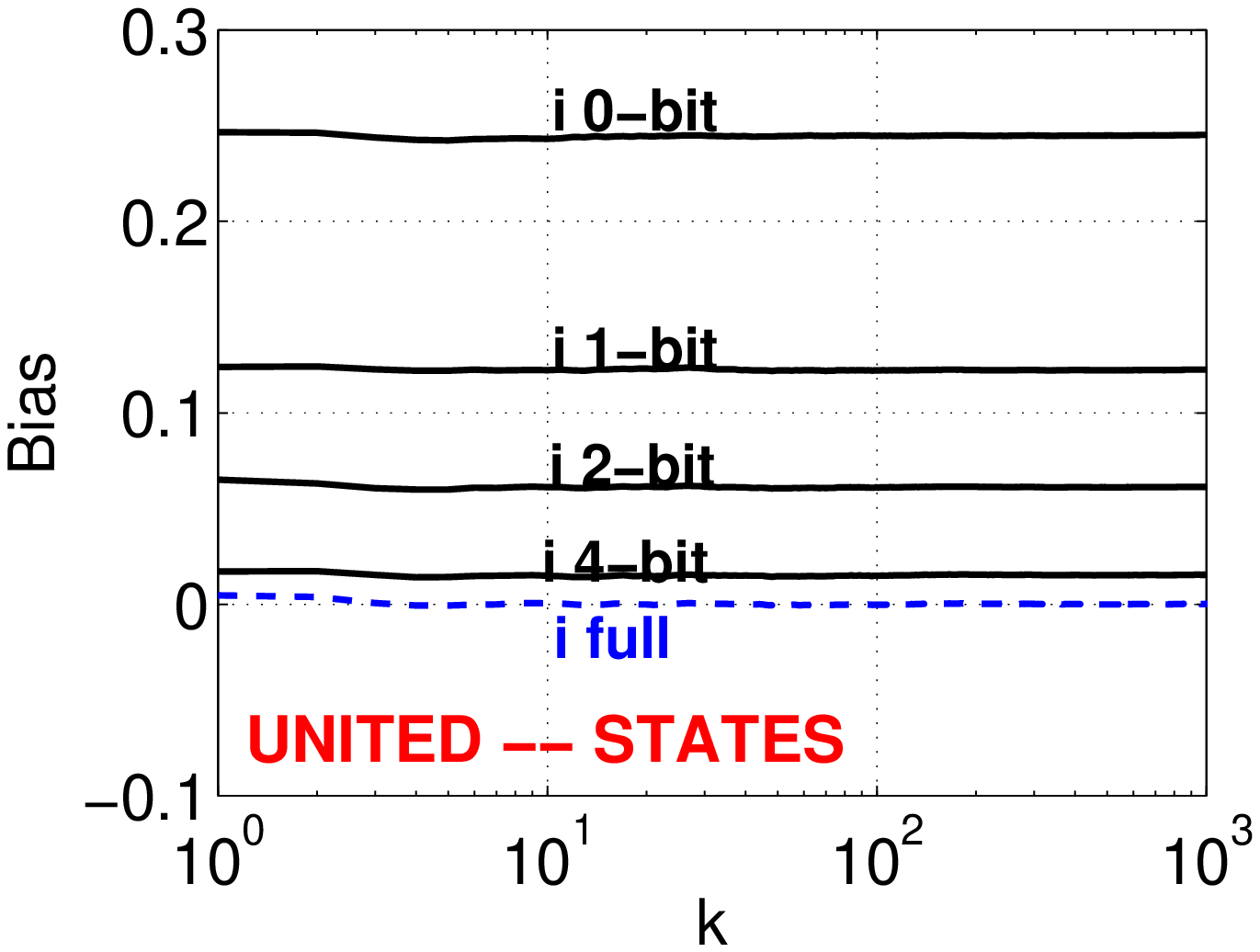}
}

\end{center}
\vspace{-0.3in}
\caption{The biases by  using full information of $t^*$ and only a few (0, 1, 2,  or 4) bits of $i^*$. }\label{fig_CWS_Words_3}\vspace{-0.2in}
\end{figure}

Finally, to entertain  readers, we add Figure~\ref{fig_CWS_Words_3} to report the bias results by keeping all the bits of $t^*$ and only a few (0,1,2,4) bits of $i^*$. Clearly, only using $t^*$ or $t^*$ with a few bits of $i^*$ will not lead to good estimate of the min-max kernel.

\section{ Kernel SVM with Modified CWS}\label{sec_LearningCWS}

We conduct a set of experiments by using ``0-bit'' CWS  for approximately training min-max kernel SVMs. Basically, for each dataset, we apply CWS hashing  for $k$ up to 4096 and, after hashing, we discard $t^*$ only keep a matrix of $\{i^*\}$, which has the same of number of rows as the number of examples in the dataset and $k$ columns. We then use the popular LIBLINEAR package~\cite{Article:Fan_JMLR08} for training a linear SVM on the data generated by $\{i^*\}$, following the scheme proposed by~\cite{Proc:HashLearning_NIPS11}.


\clearpage\newpage


\begin{figure*}
\begin{center}

\mbox{
\includegraphics[width=1.65in]{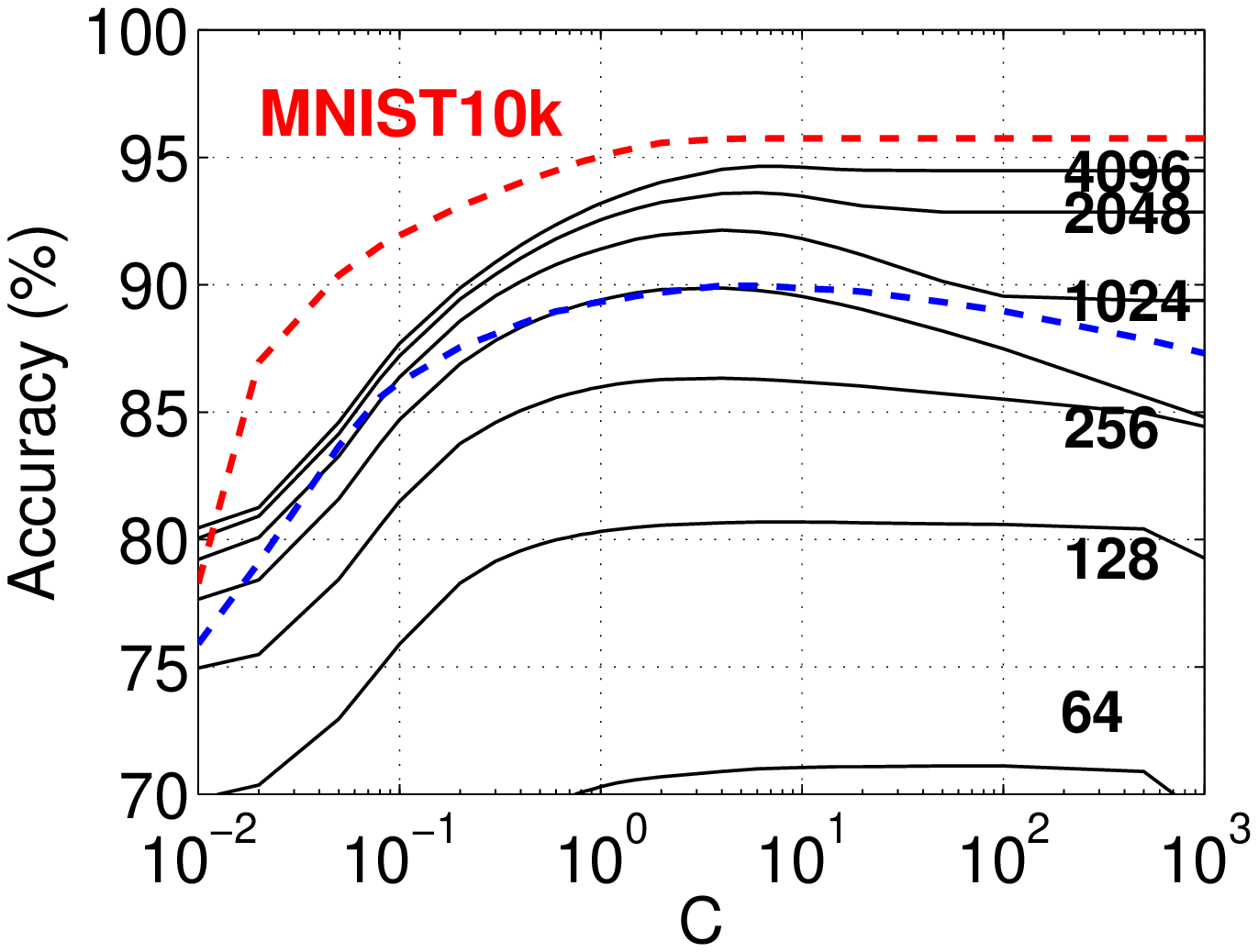}\hspace{-0.05in}
\includegraphics[width=1.65in]{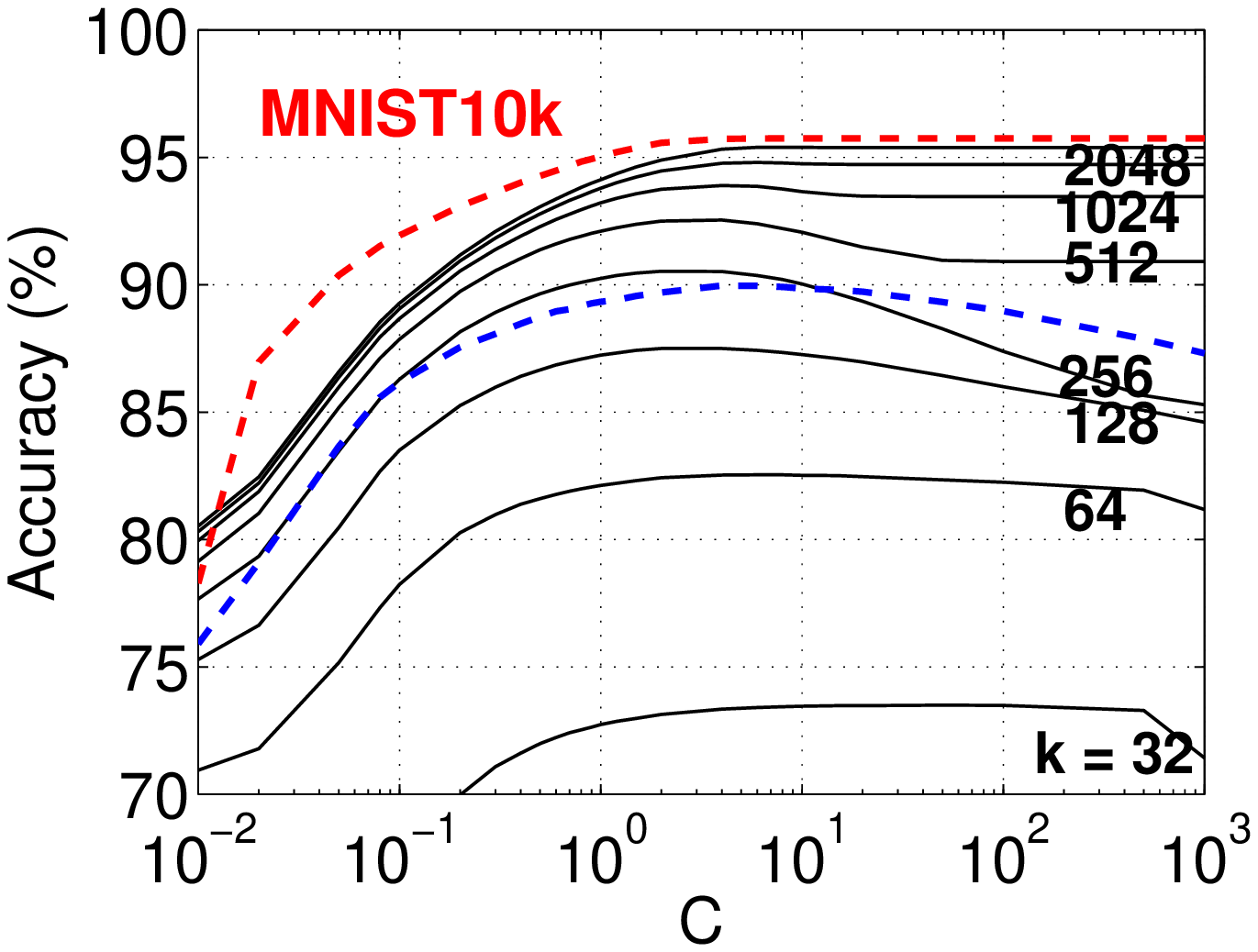}\hspace{-0.05in}
\includegraphics[width=1.65in]{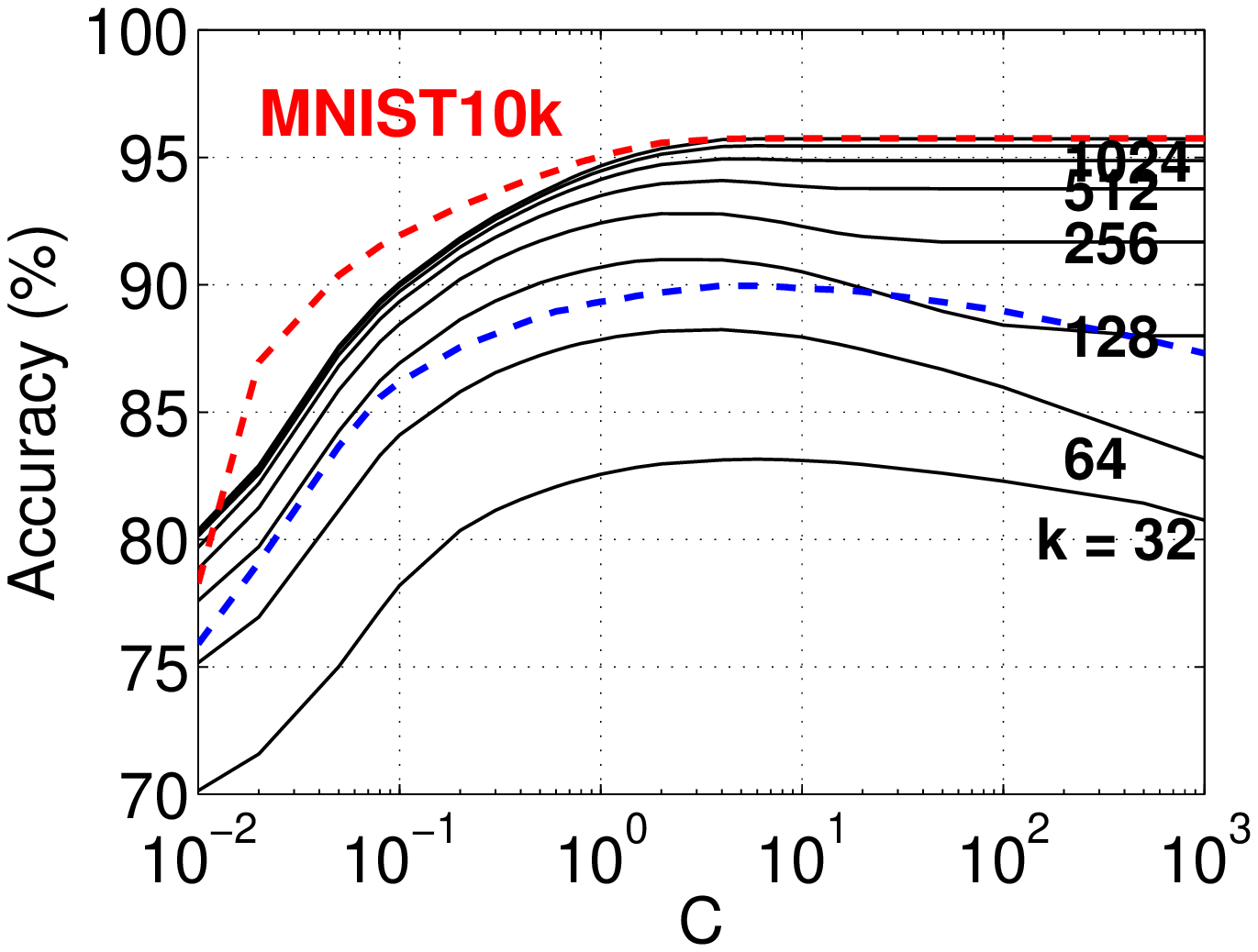}\hspace{-0.05in}
\includegraphics[width=1.65in]{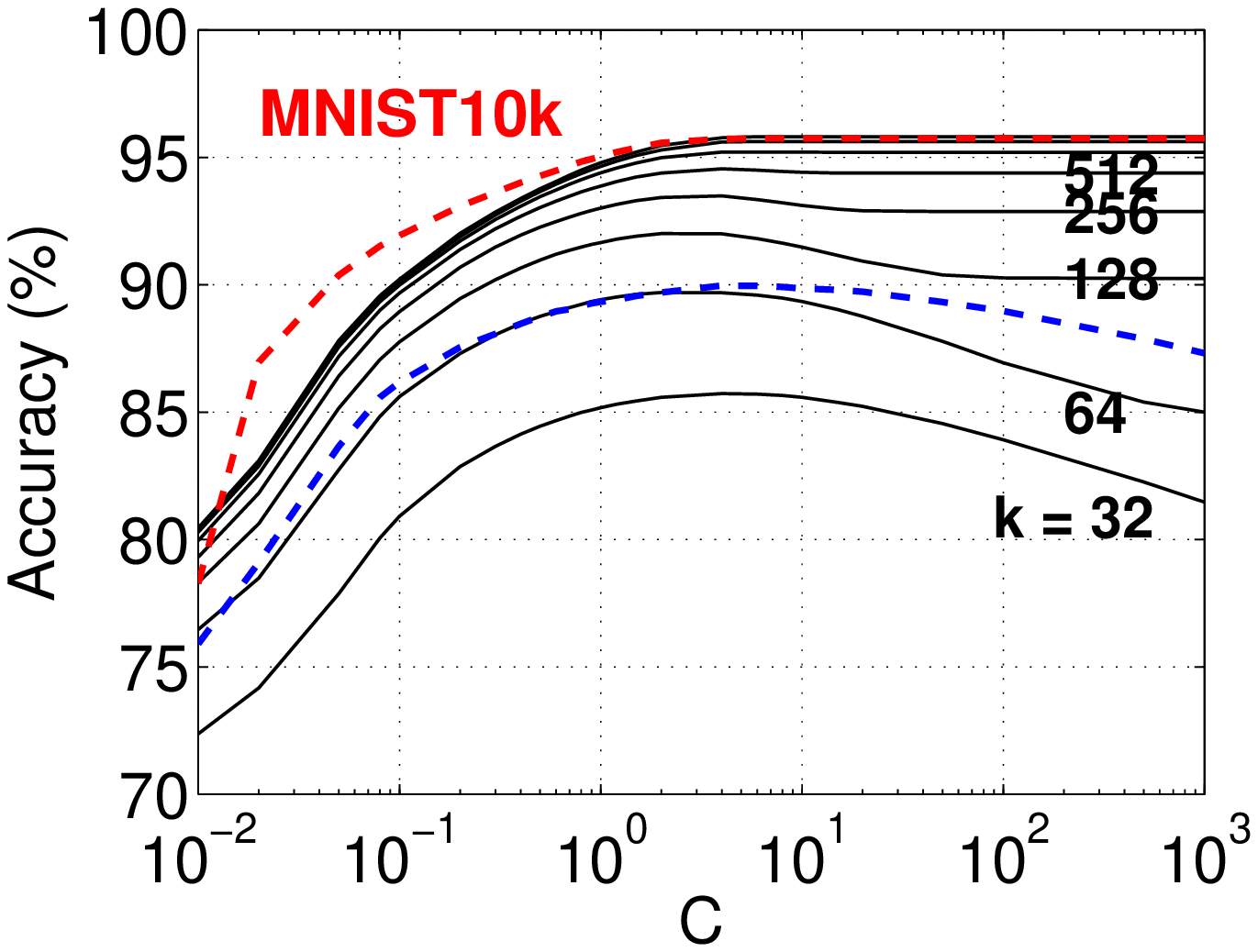}
}

\vspace{-0.11in}

\mbox{
\includegraphics[width=1.65in]{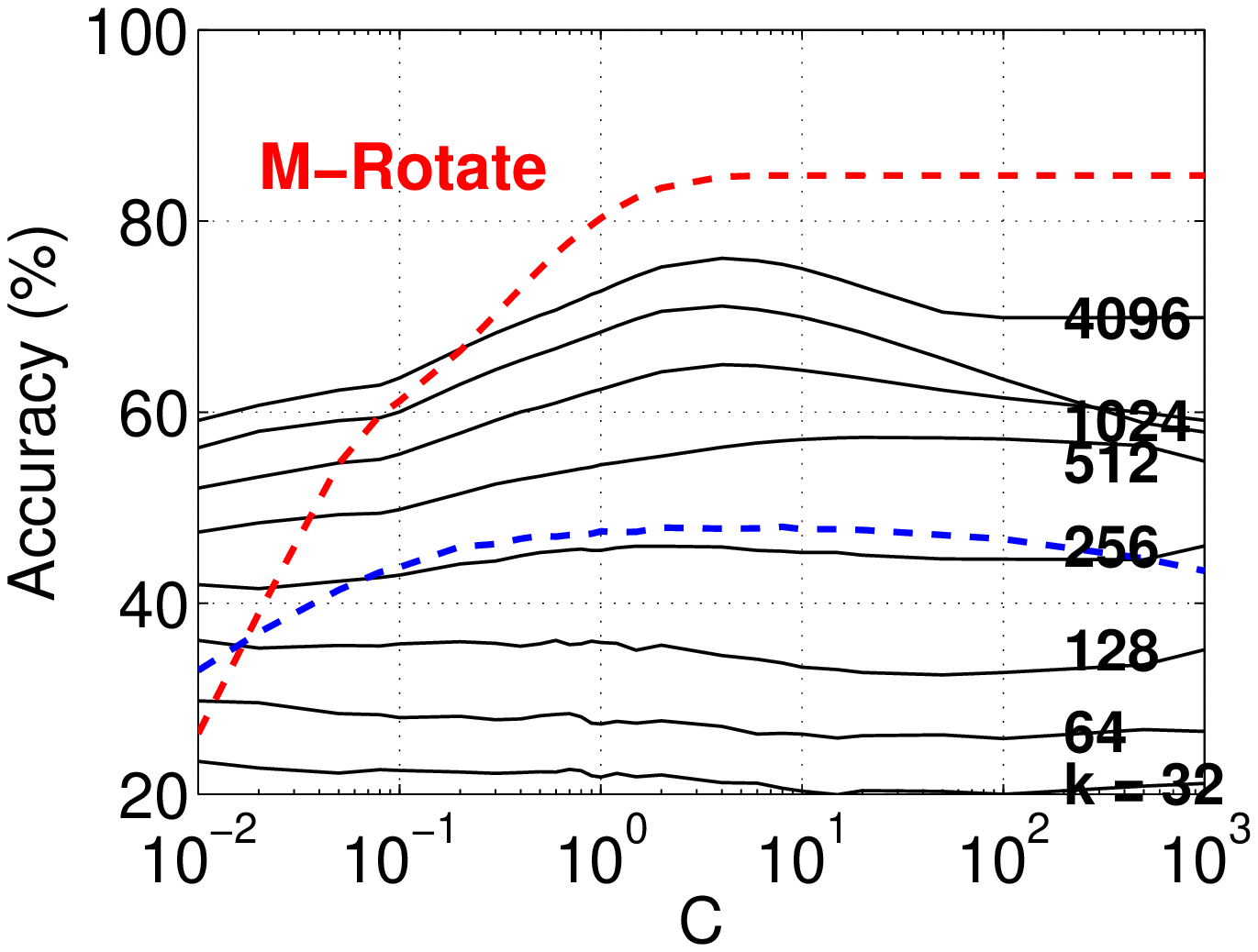}\hspace{-0.05in}
\includegraphics[width=1.65in]{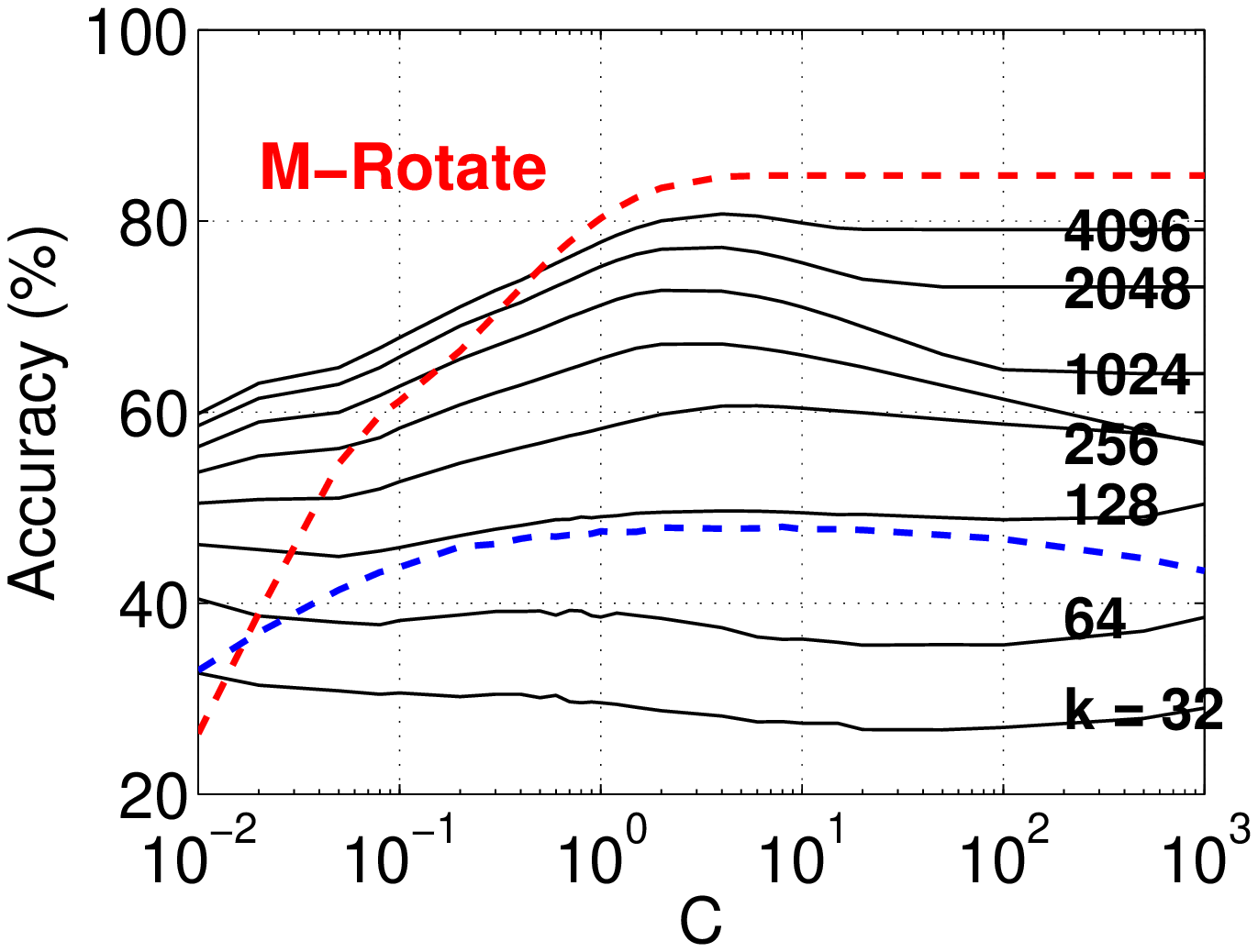}\hspace{-0.05in}
\includegraphics[width=1.65in]{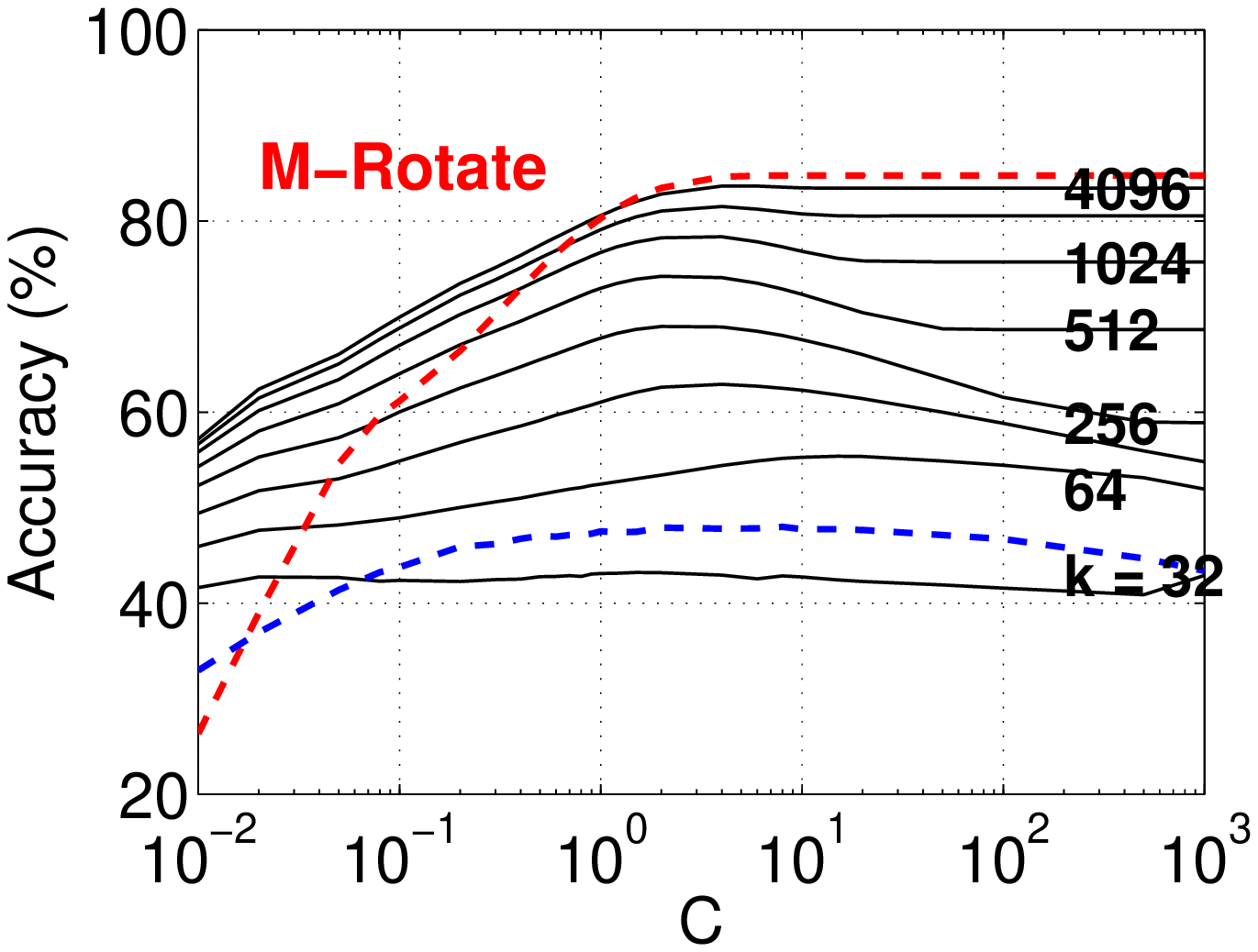}\hspace{-0.05in}
\includegraphics[width=1.65in]{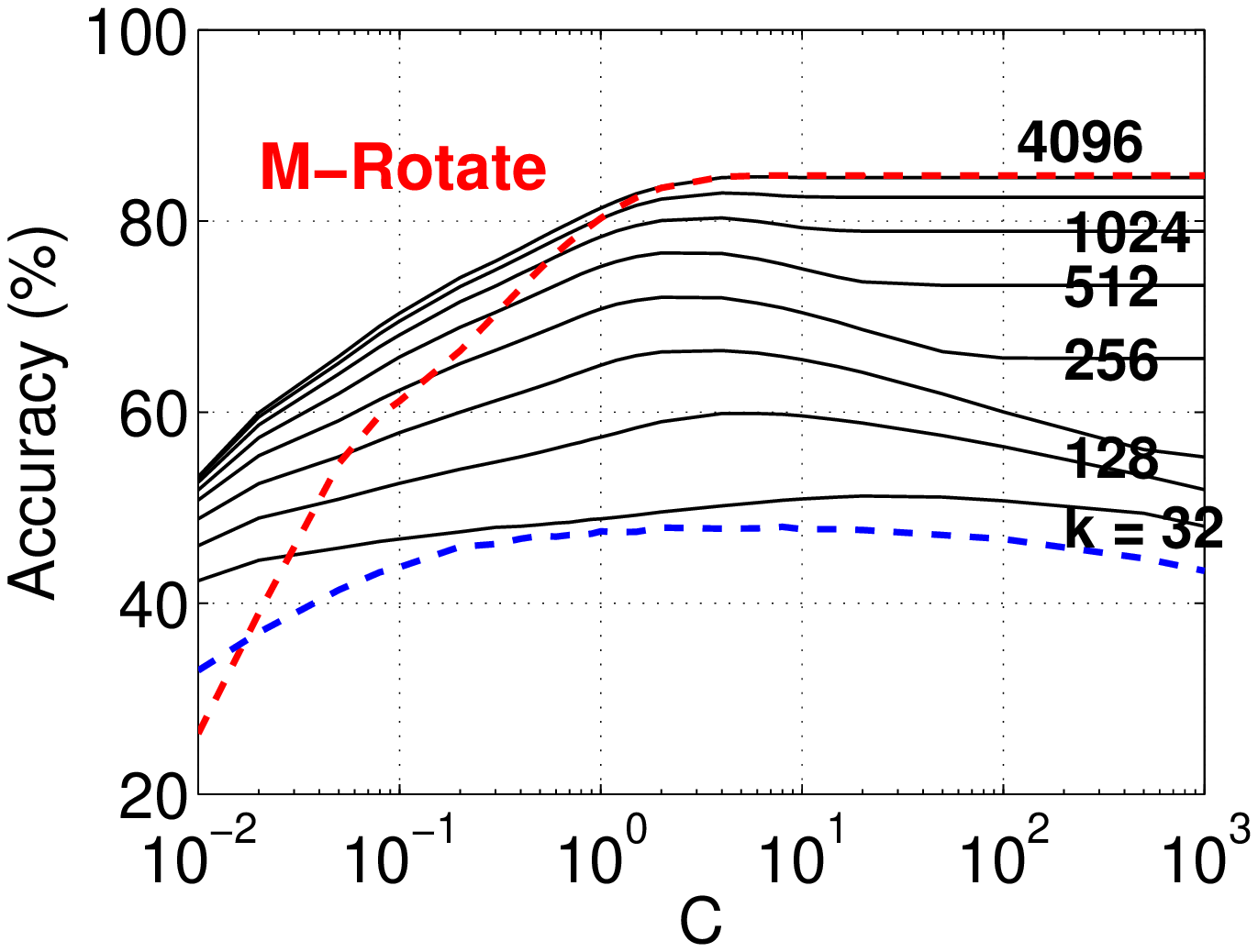}
}

\vspace{-0.11in}

\mbox{
\includegraphics[width=1.65in]{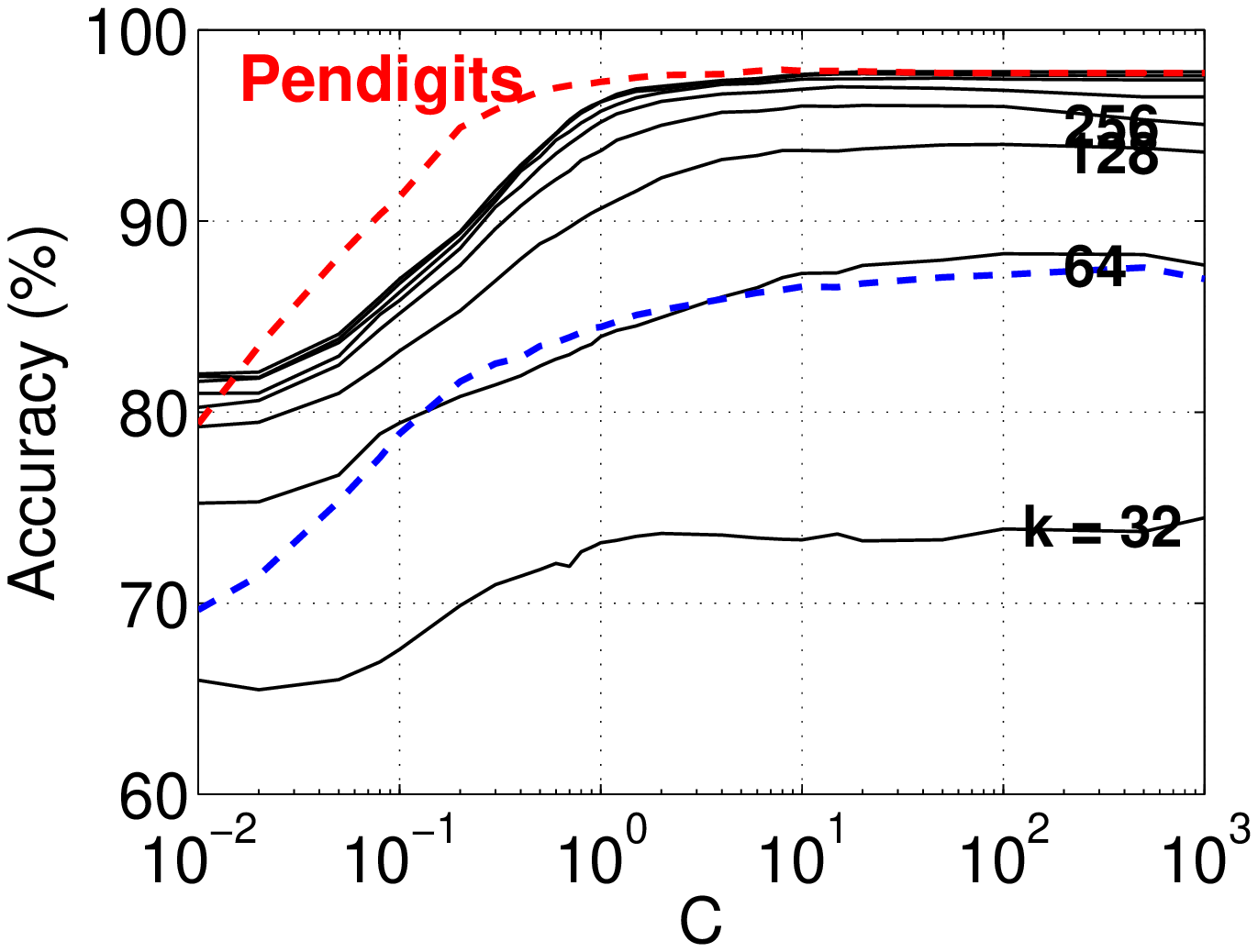}\hspace{-0.05in}
\includegraphics[width=1.65in]{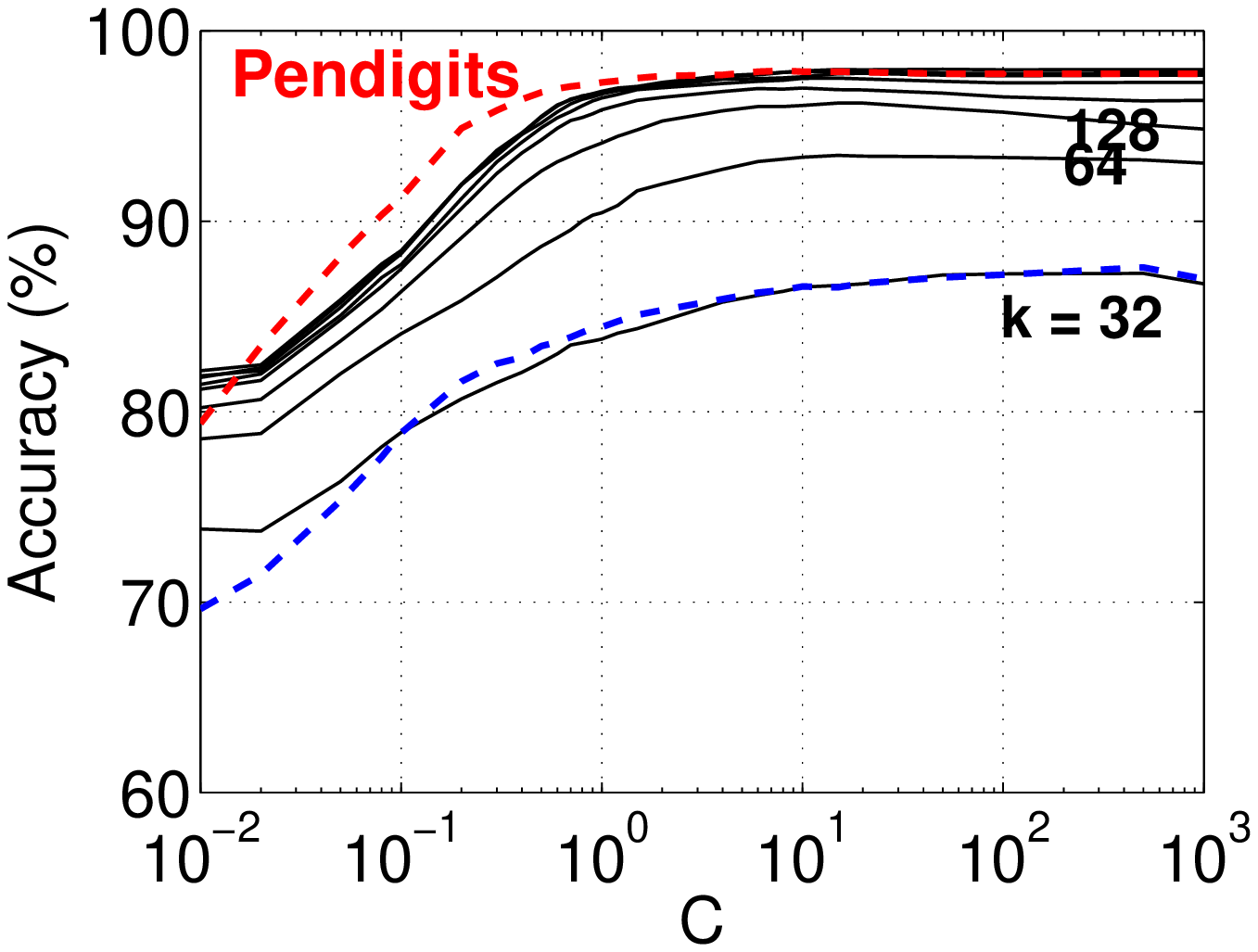}\hspace{-0.05in}
\includegraphics[width=1.65in]{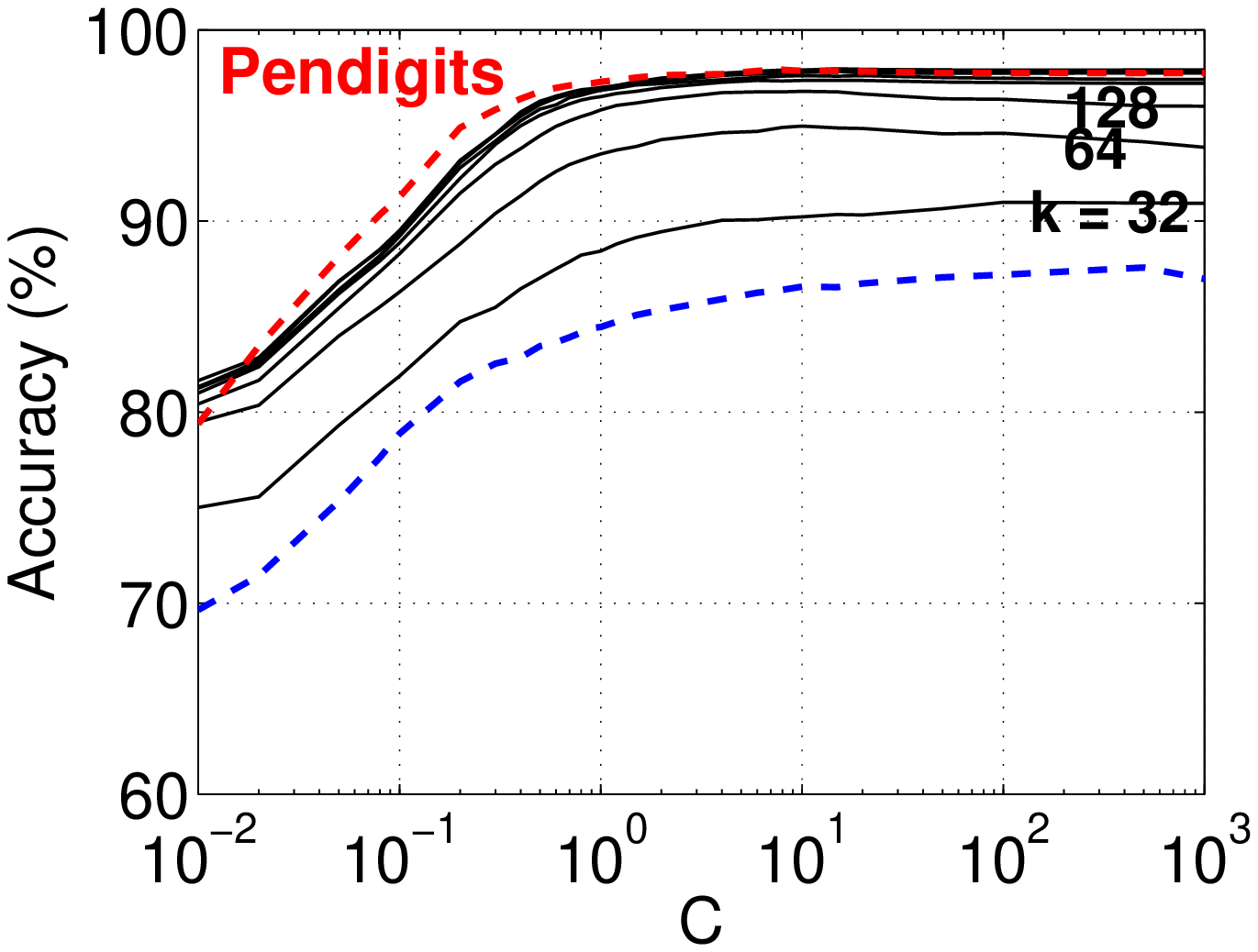}\hspace{-0.05in}
\includegraphics[width=1.65in]{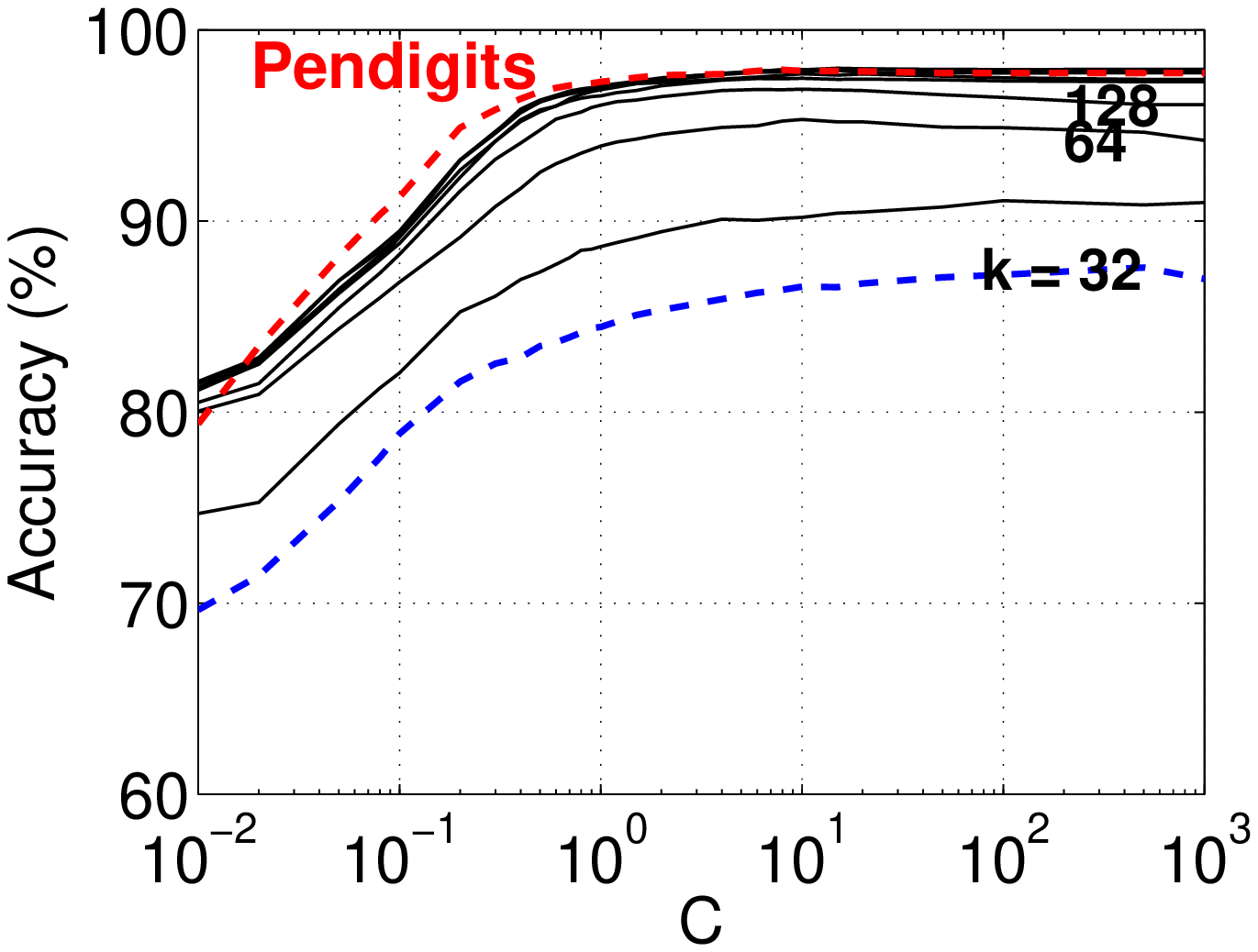}
}

\vspace{-0.11in}

\mbox{
\includegraphics[width=1.65in]{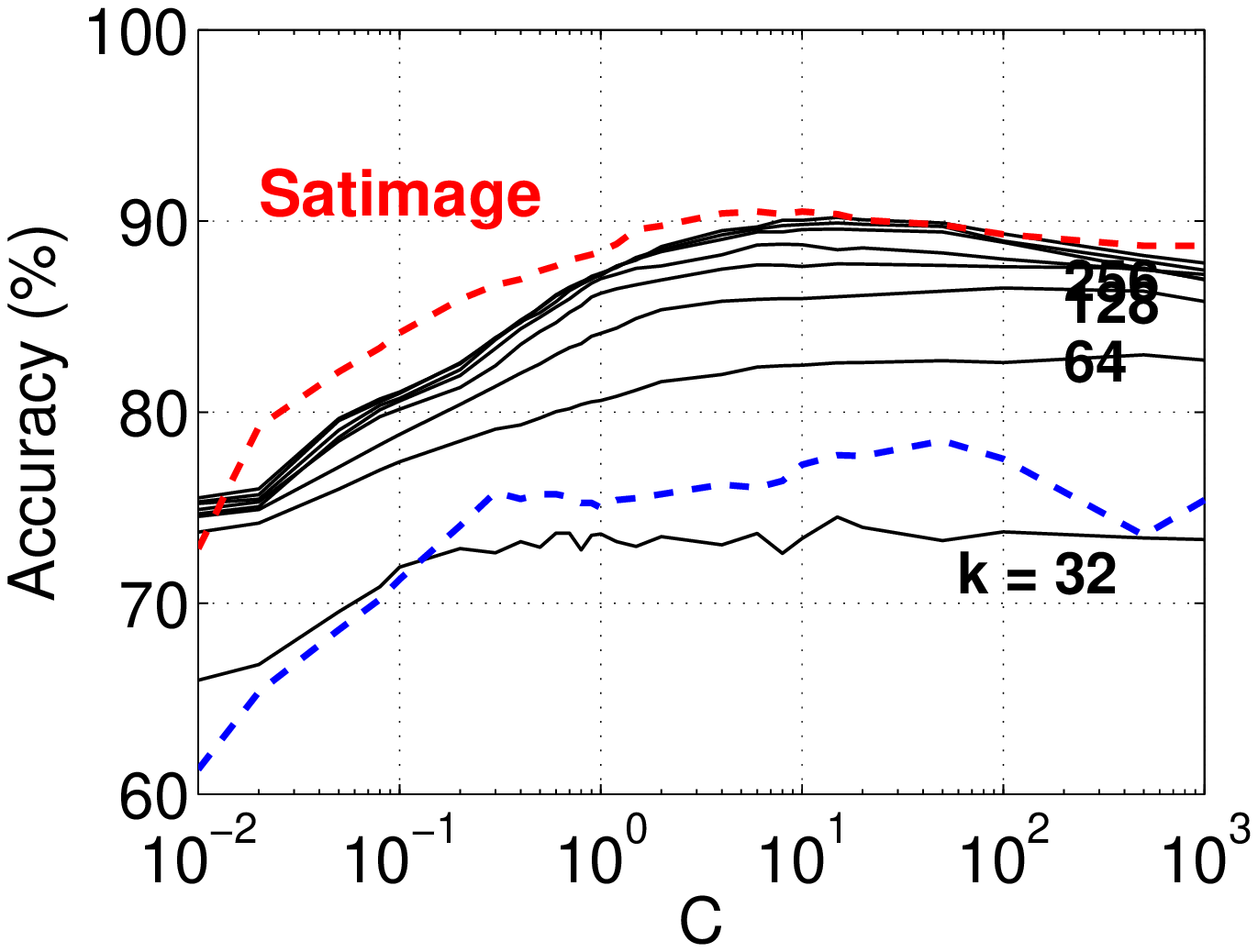}\hspace{-0.05in}
\includegraphics[width=1.65in]{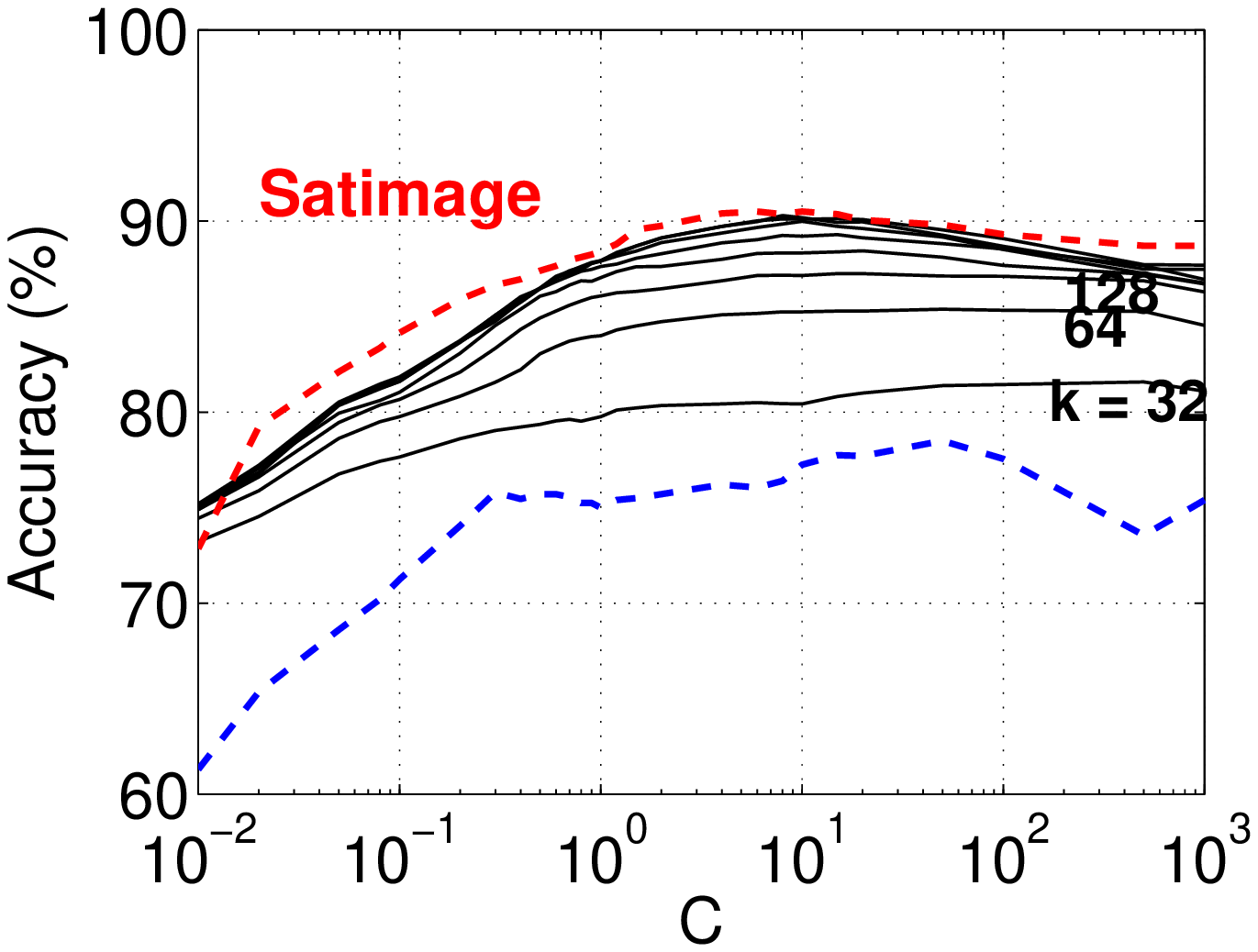}\hspace{-0.05in}
\includegraphics[width=1.65in]{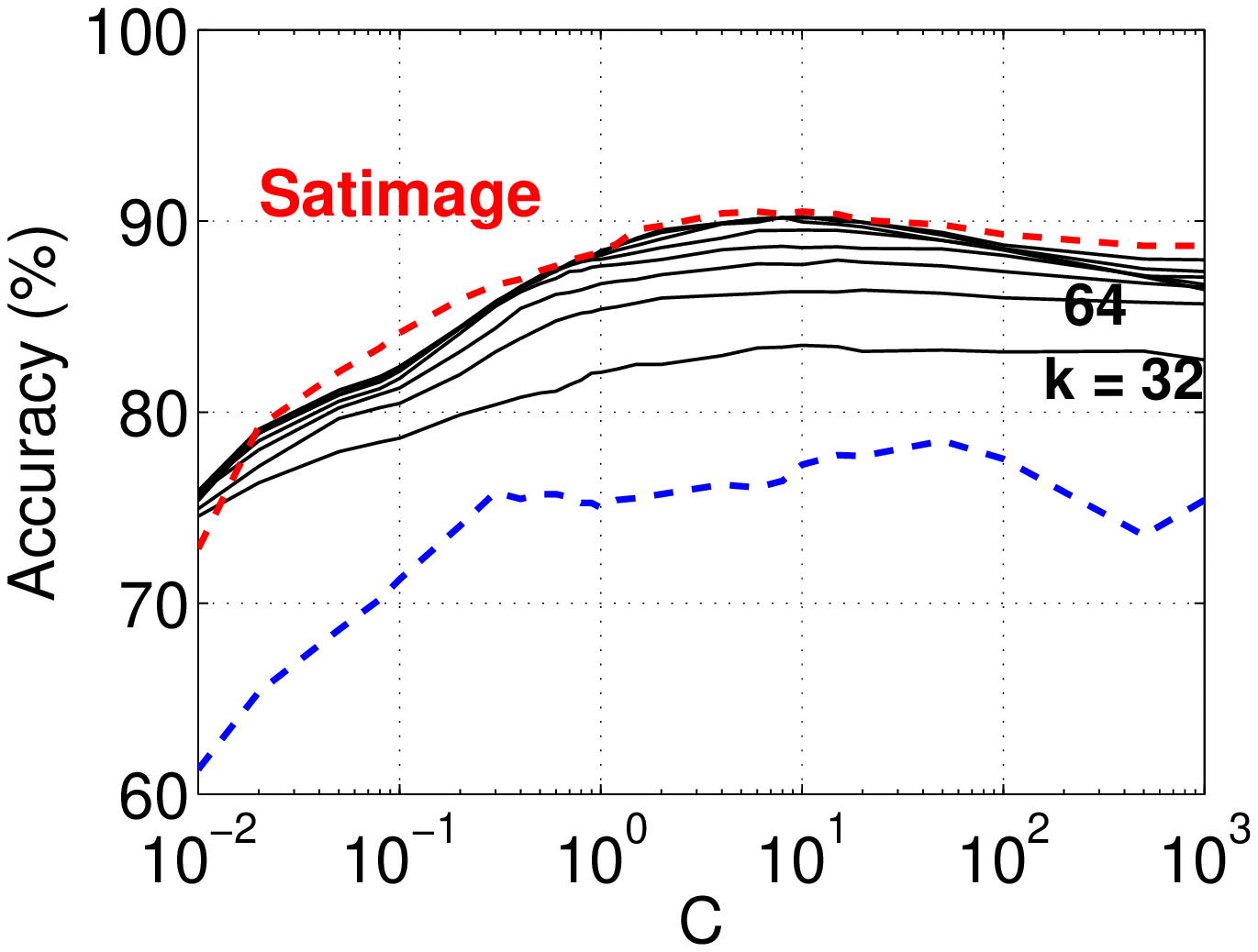}\hspace{-0.05in}
\includegraphics[width=1.65in]{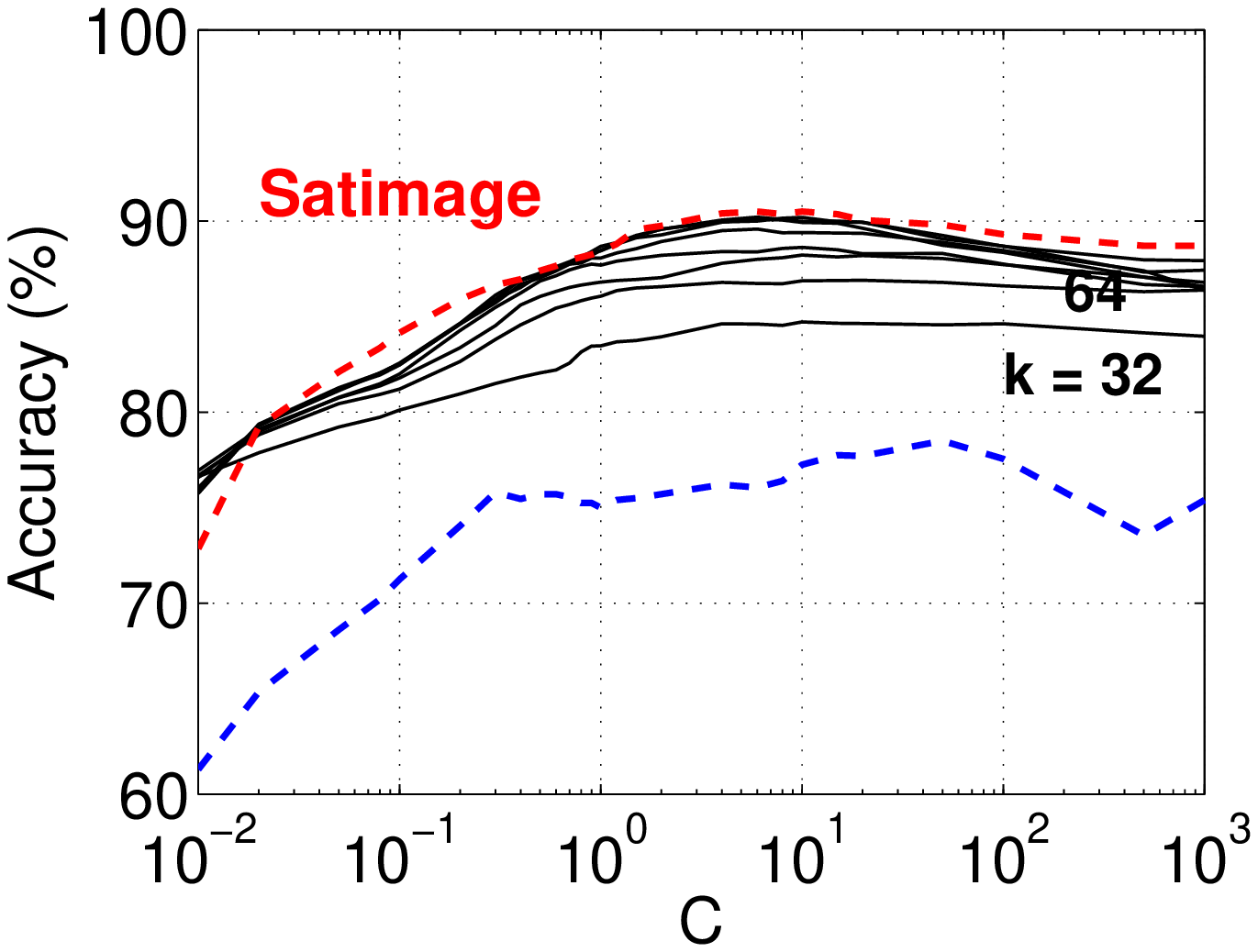}
}

\vspace{-0.11in}

\mbox{
\includegraphics[width=1.65in]{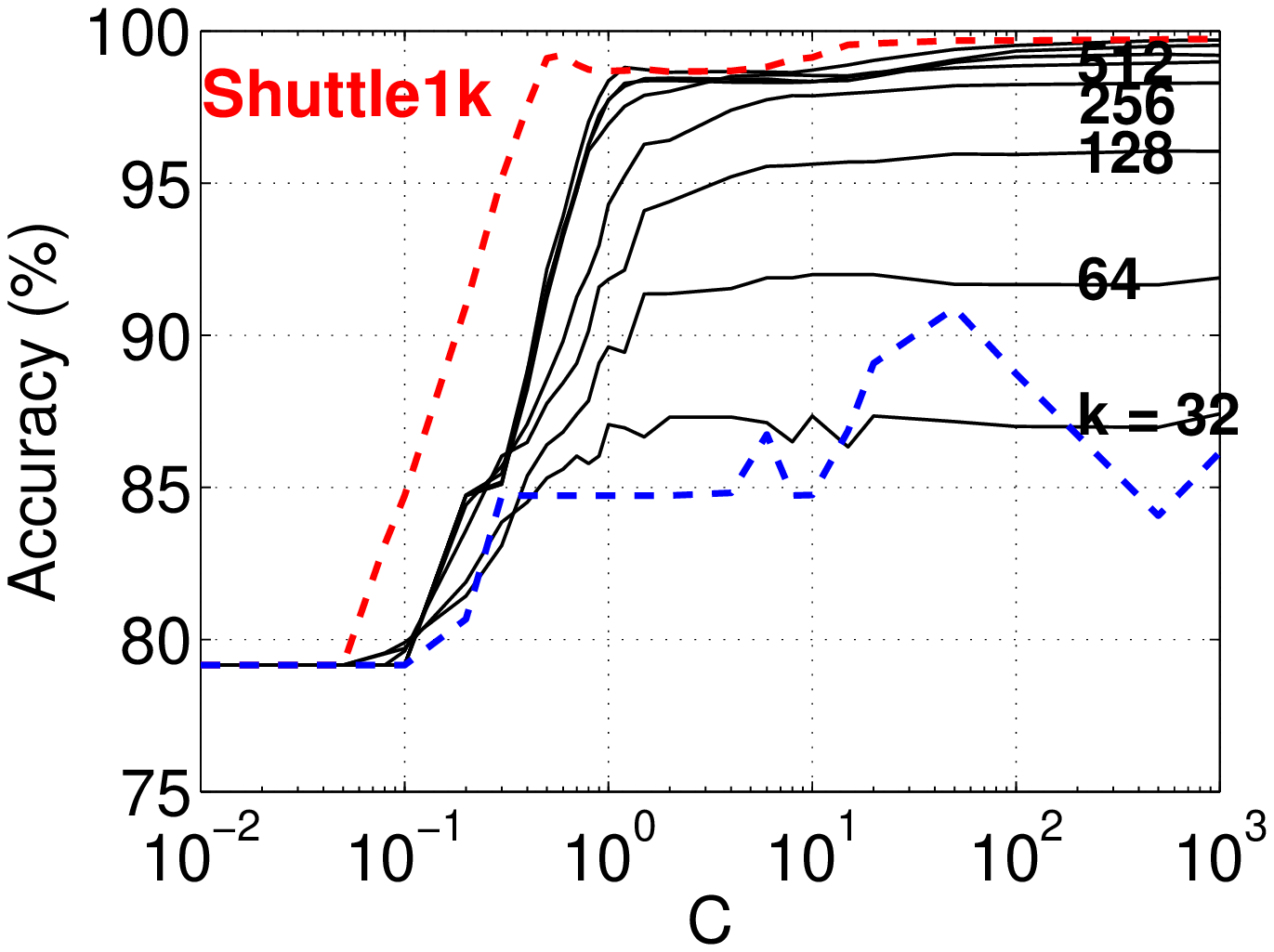}\hspace{-0.05in}
\includegraphics[width=1.65in]{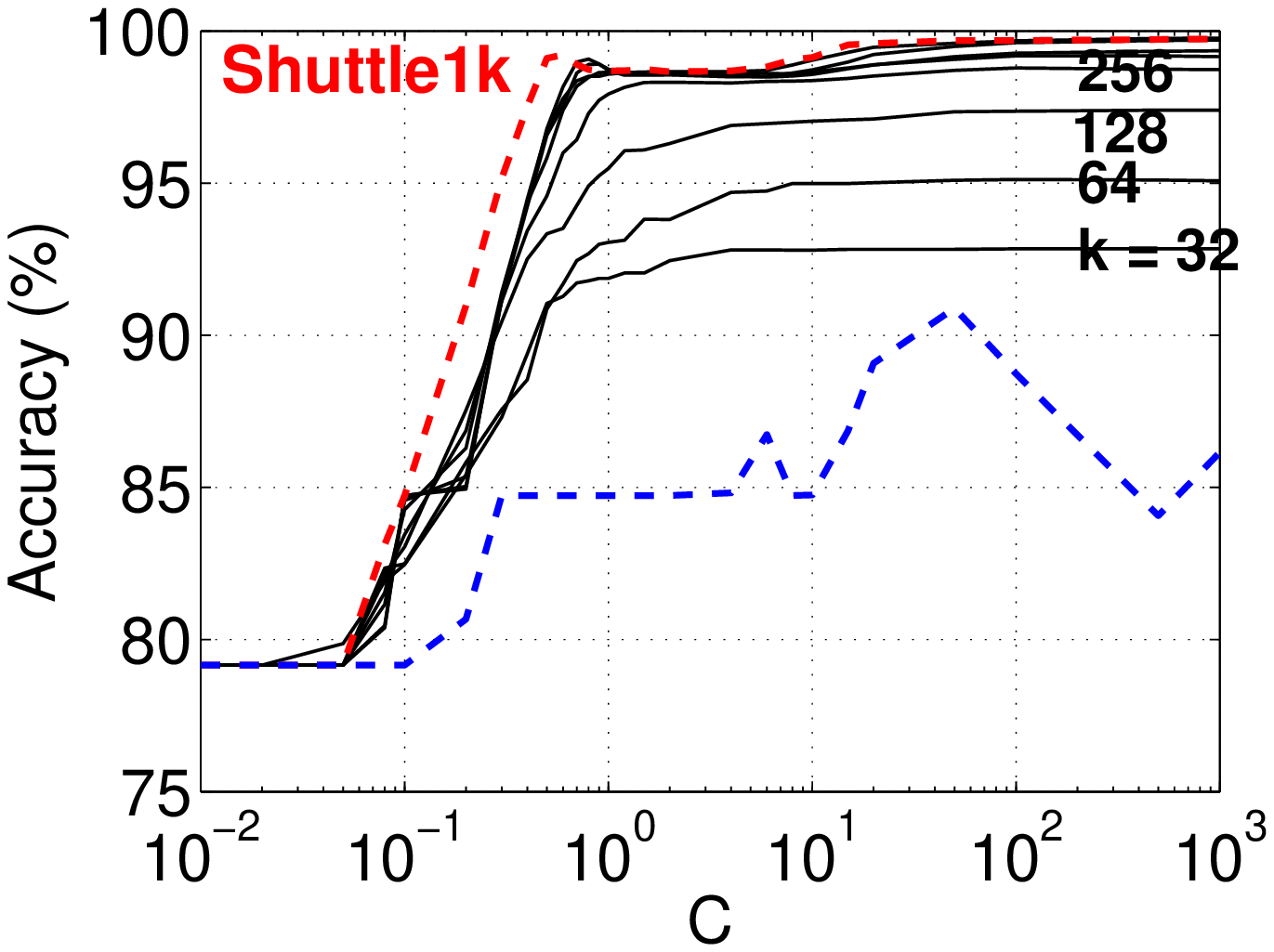}\hspace{-0.05in}
\includegraphics[width=1.65in]{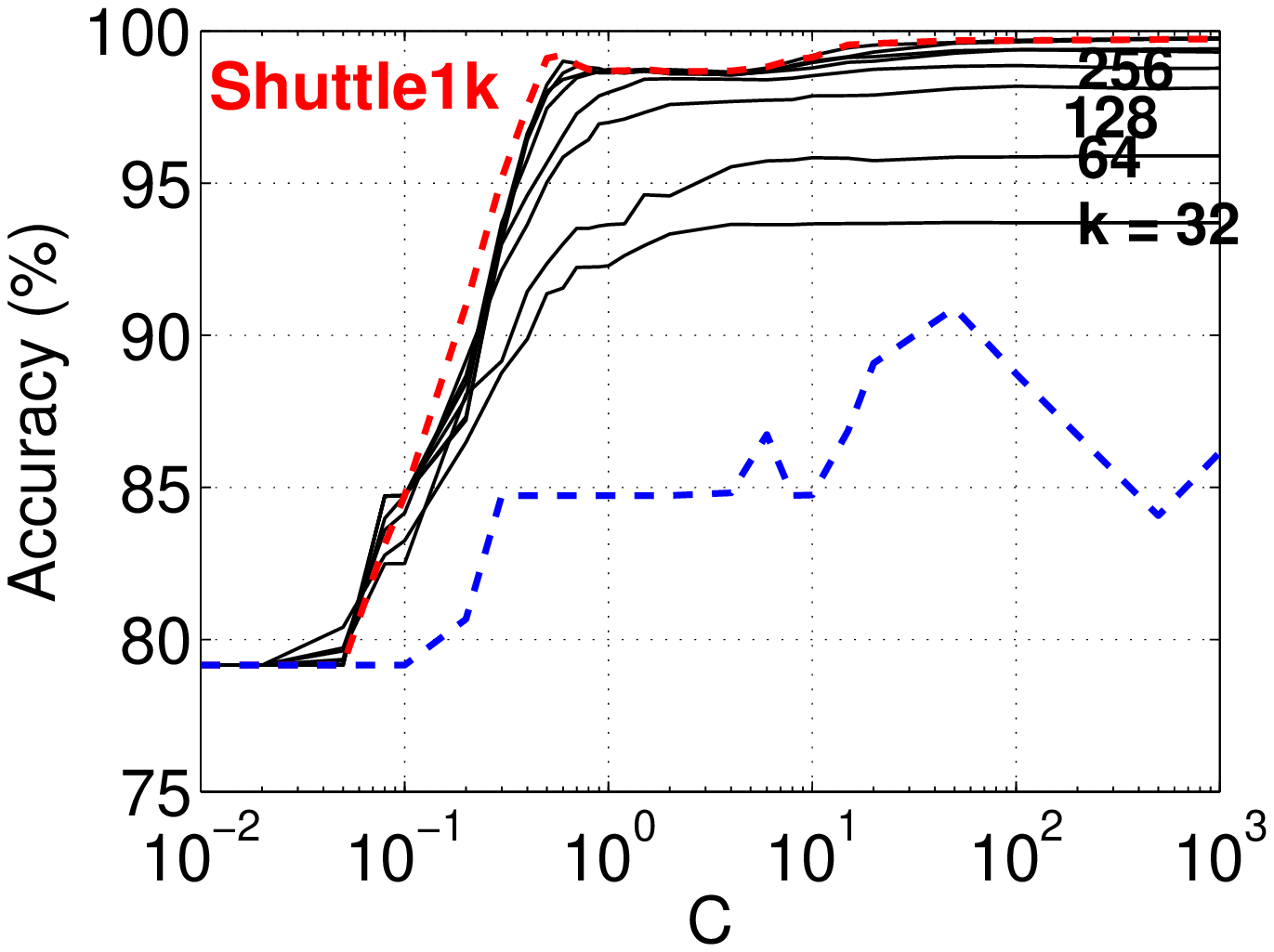}\hspace{-0.05in}
\includegraphics[width=1.65in]{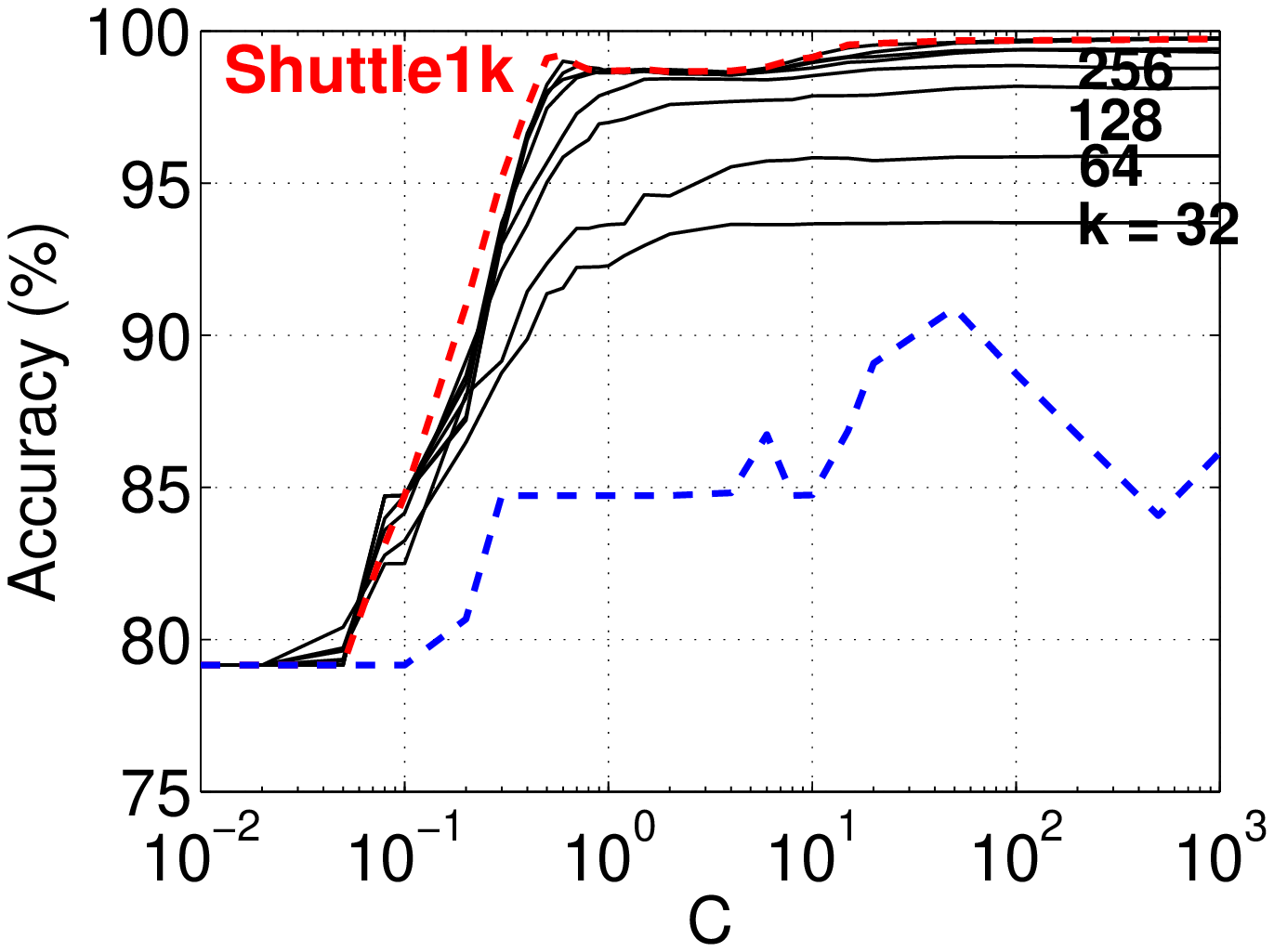}
}

\vspace{-0.11in}

\mbox{
\includegraphics[width=1.65in]{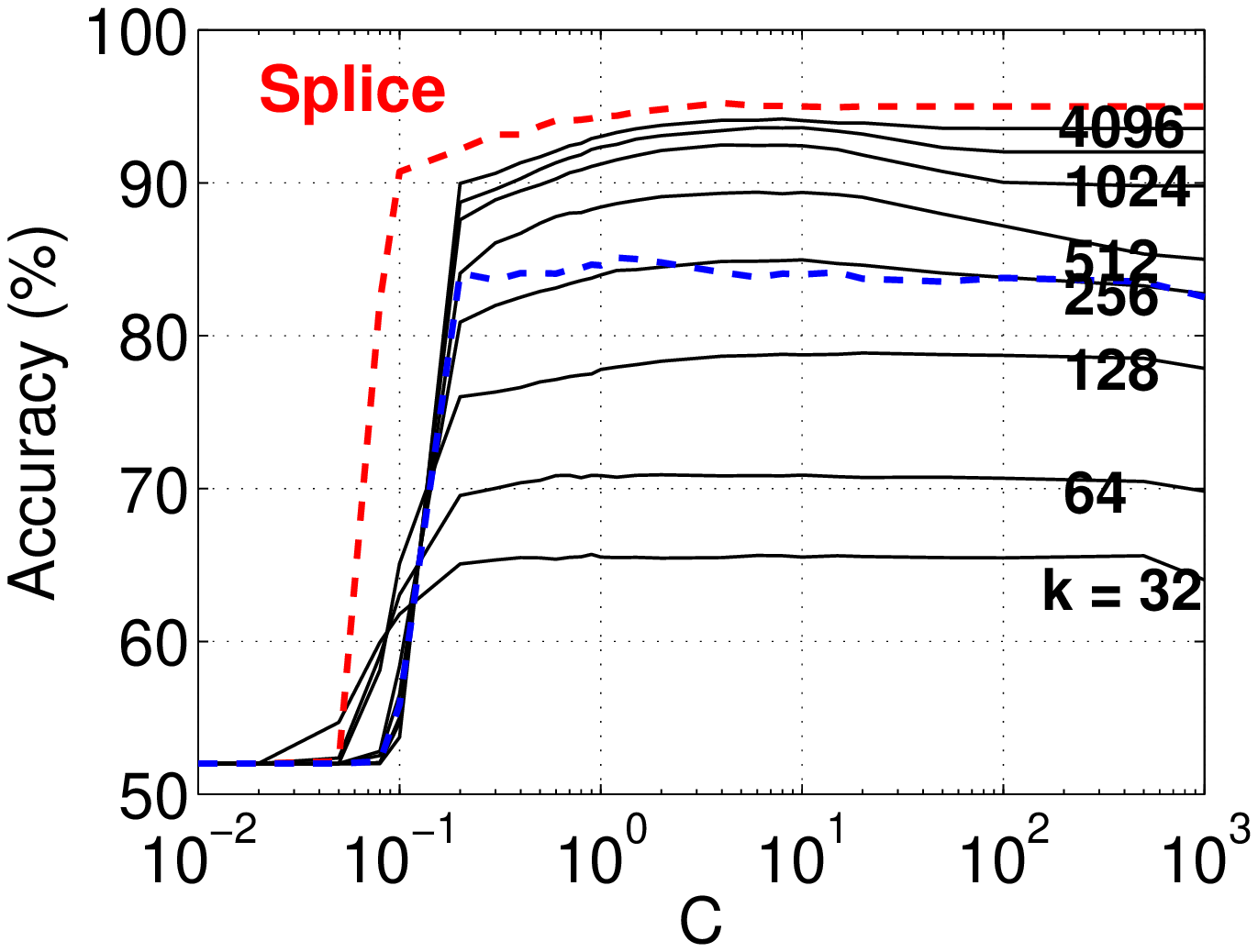}\hspace{-0.05in}
\includegraphics[width=1.65in]{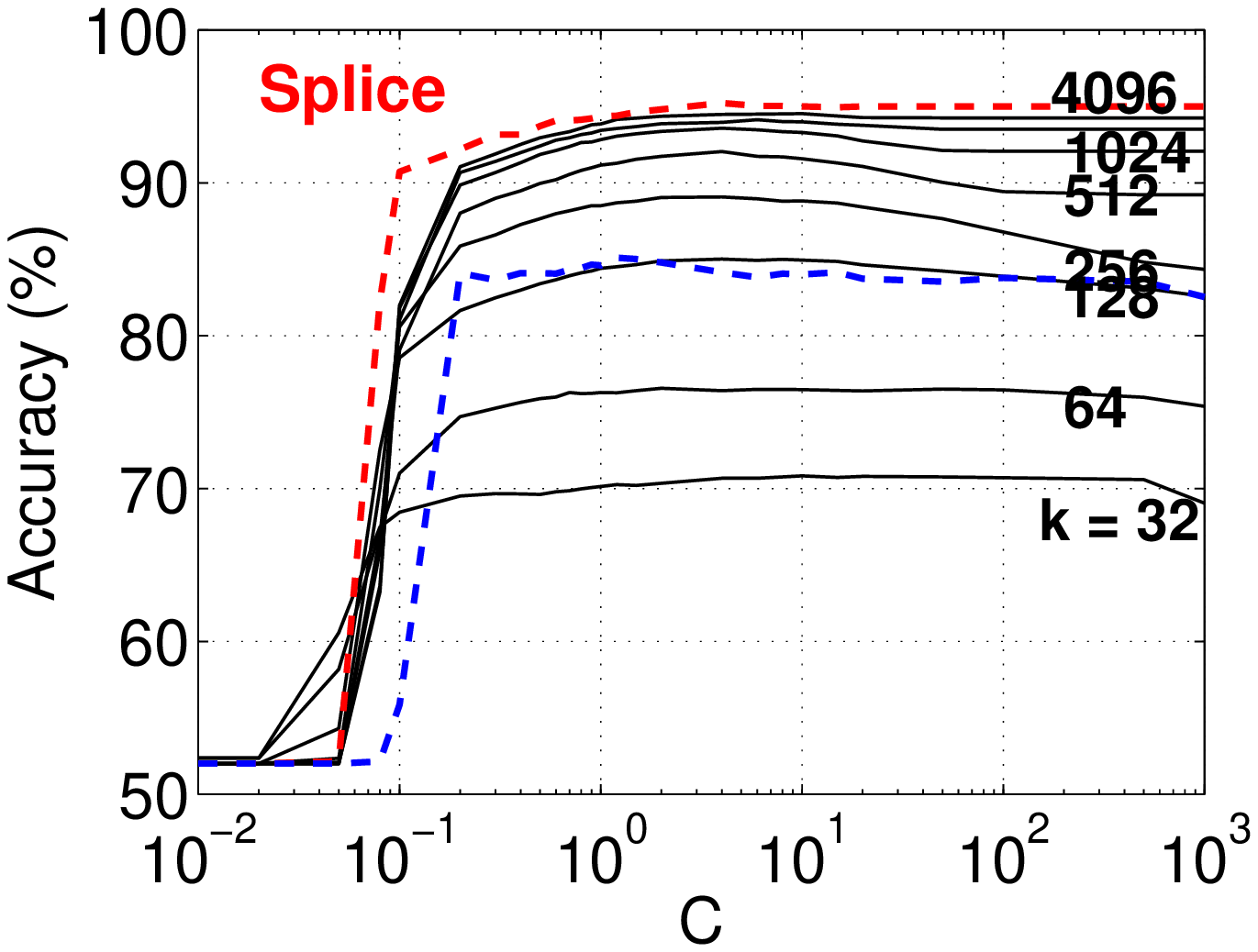}\hspace{-0.05in}
\includegraphics[width=1.65in]{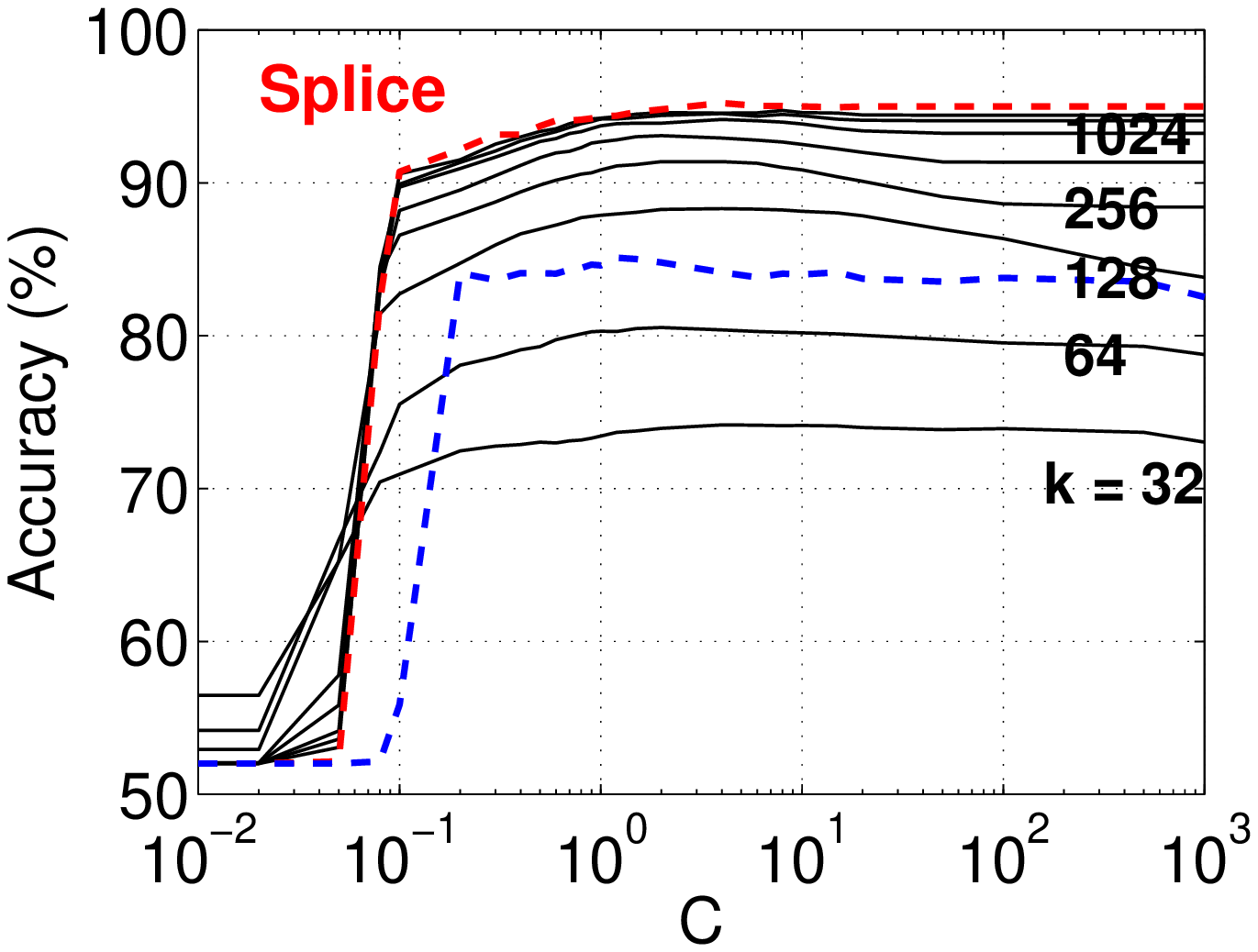}\hspace{-0.05in}
\includegraphics[width=1.65in]{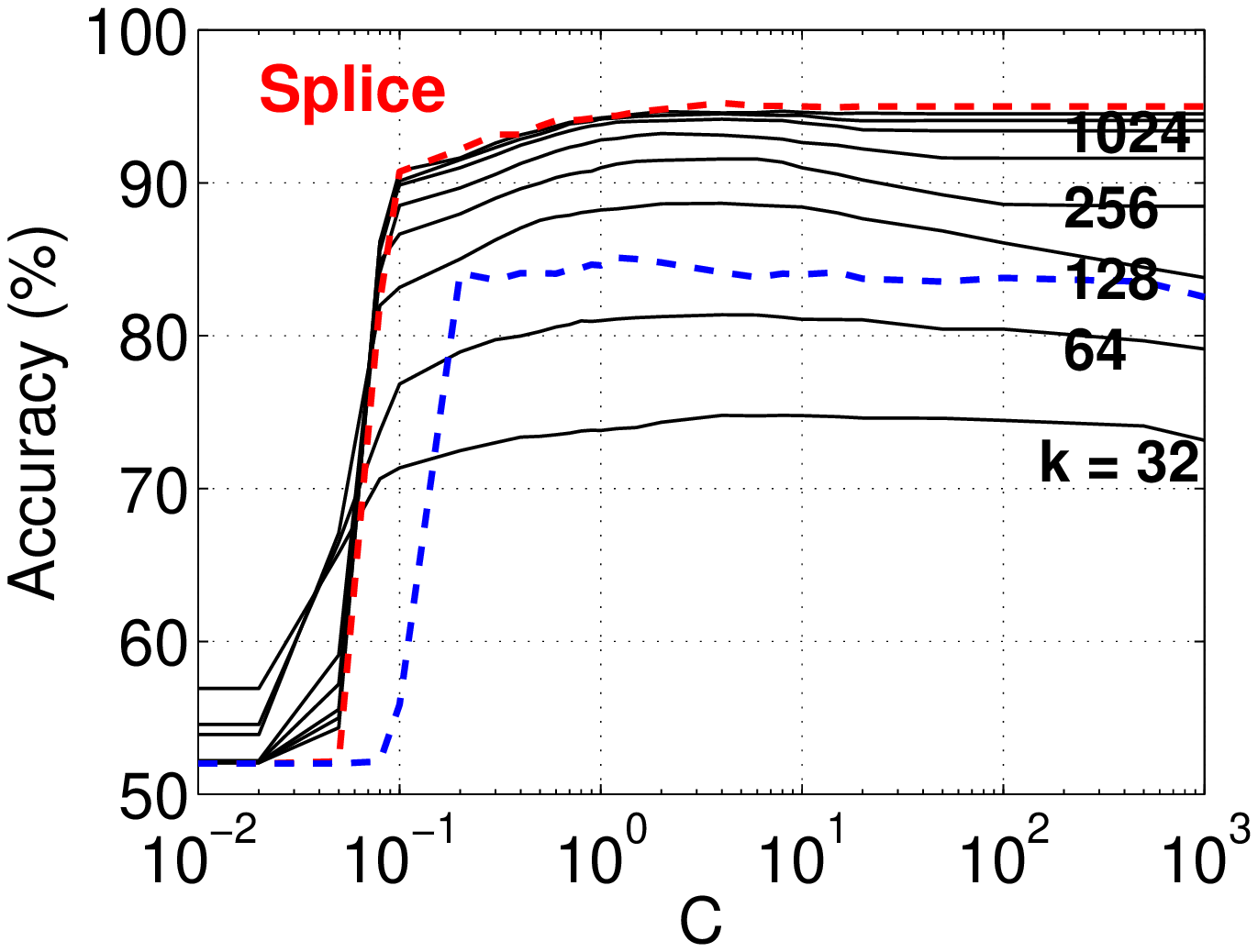}
}

\vspace{-0.11in}

\mbox{
\includegraphics[width=1.65in]{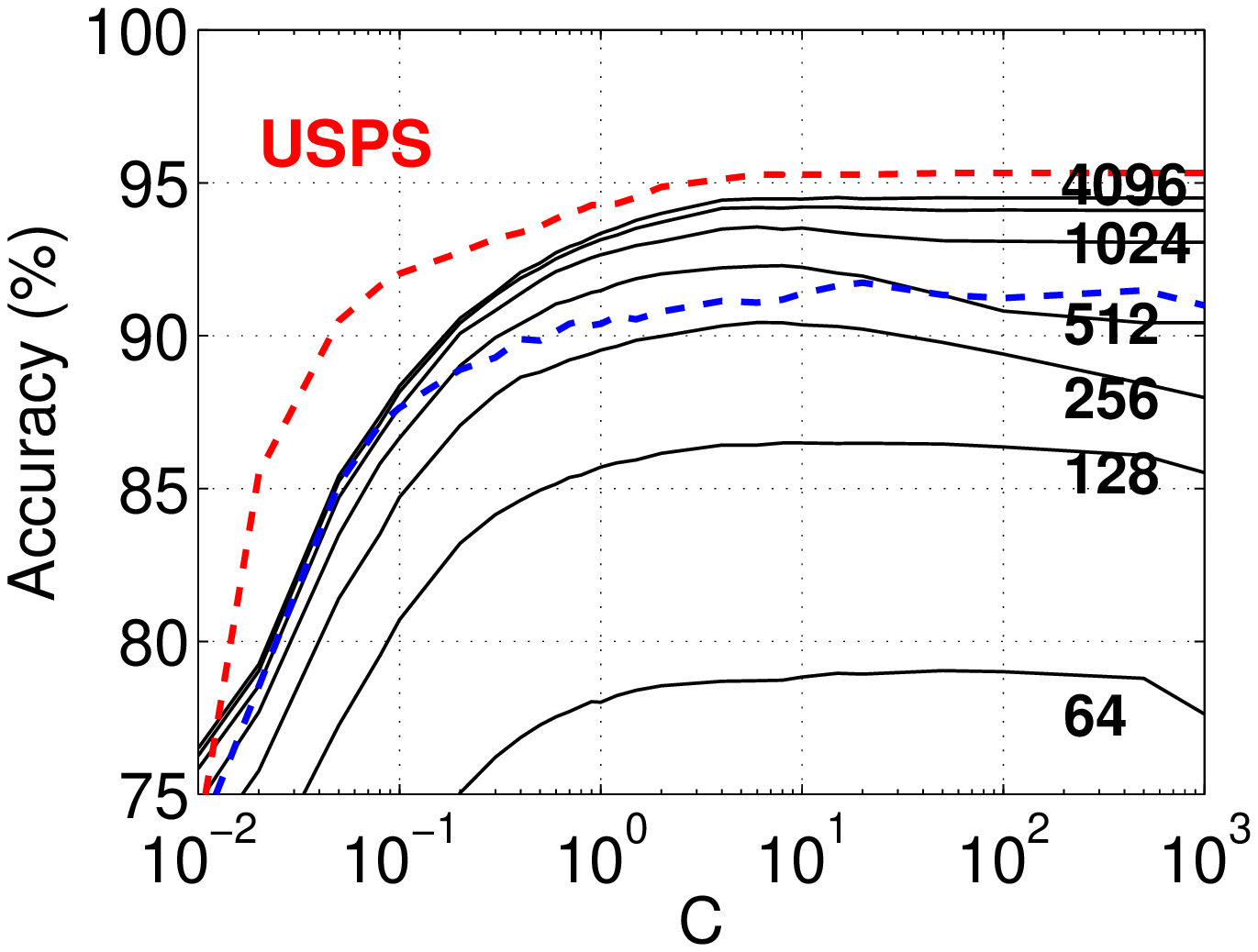}\hspace{-0.05in}
\includegraphics[width=1.65in]{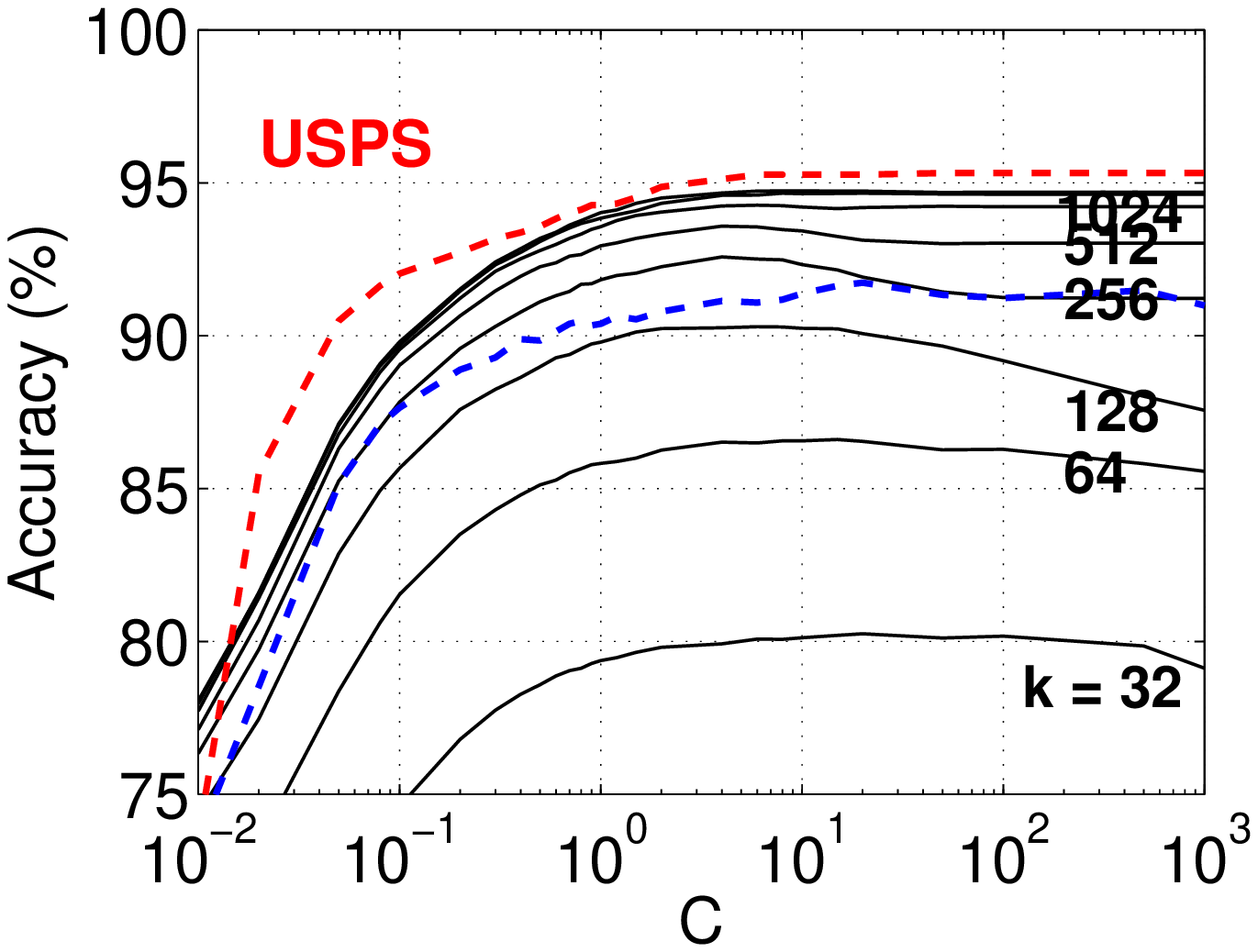}\hspace{-0.05in}
\includegraphics[width=1.65in]{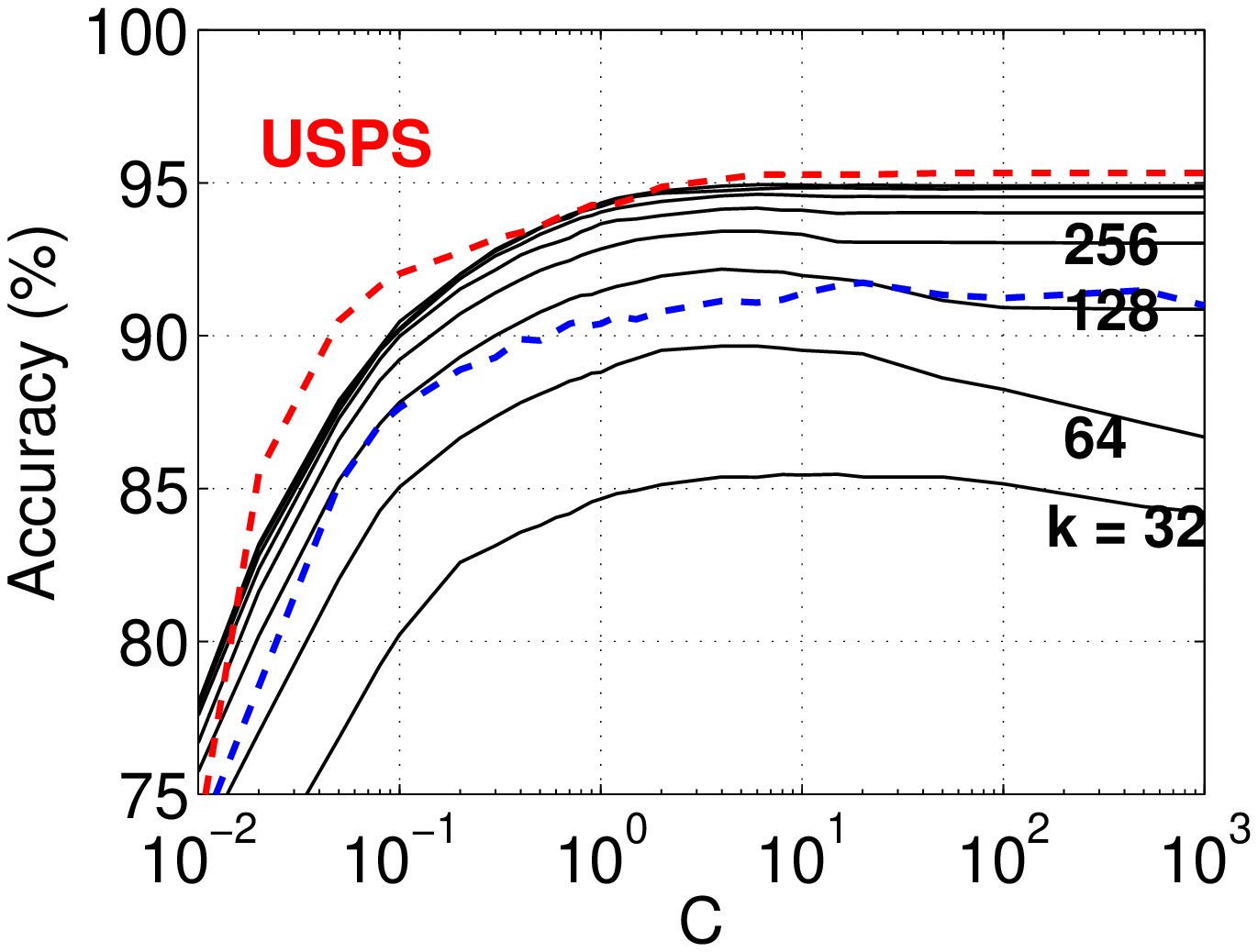}\hspace{-0.05in}
\includegraphics[width=1.65in]{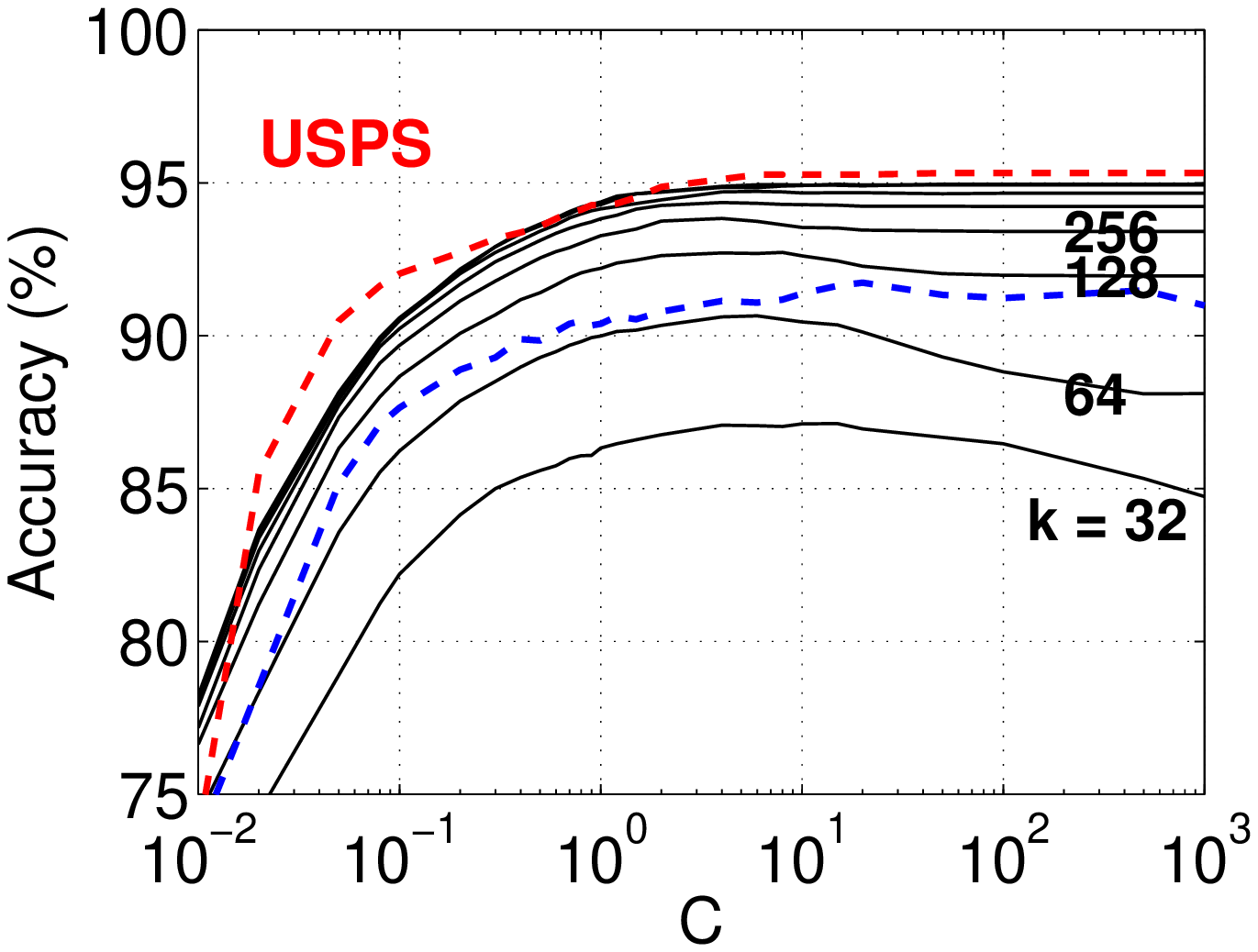}
}

\vspace{-0.11in}

\mbox{
\includegraphics[width=1.65in]{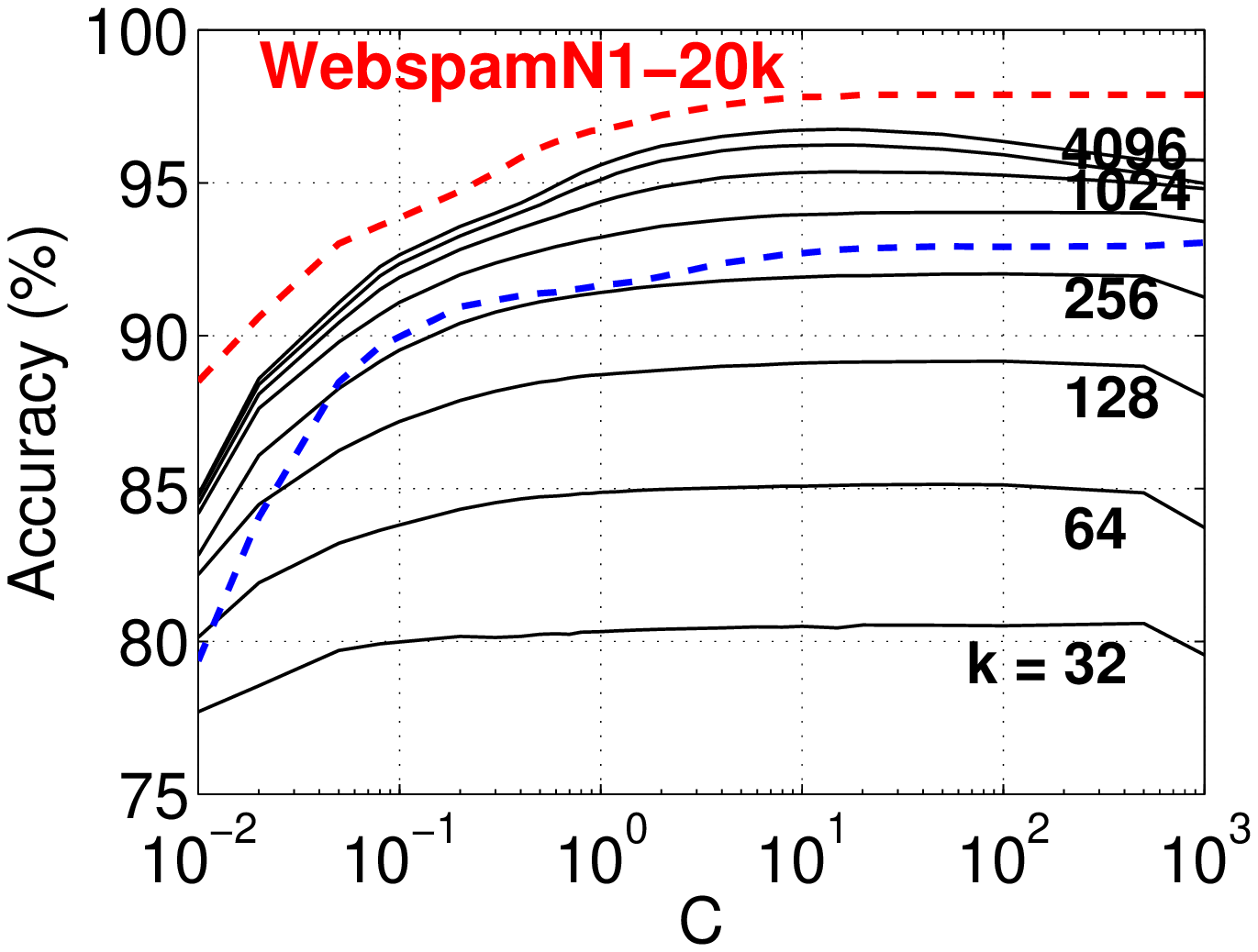}\hspace{-0.05in}
\includegraphics[width=1.65in]{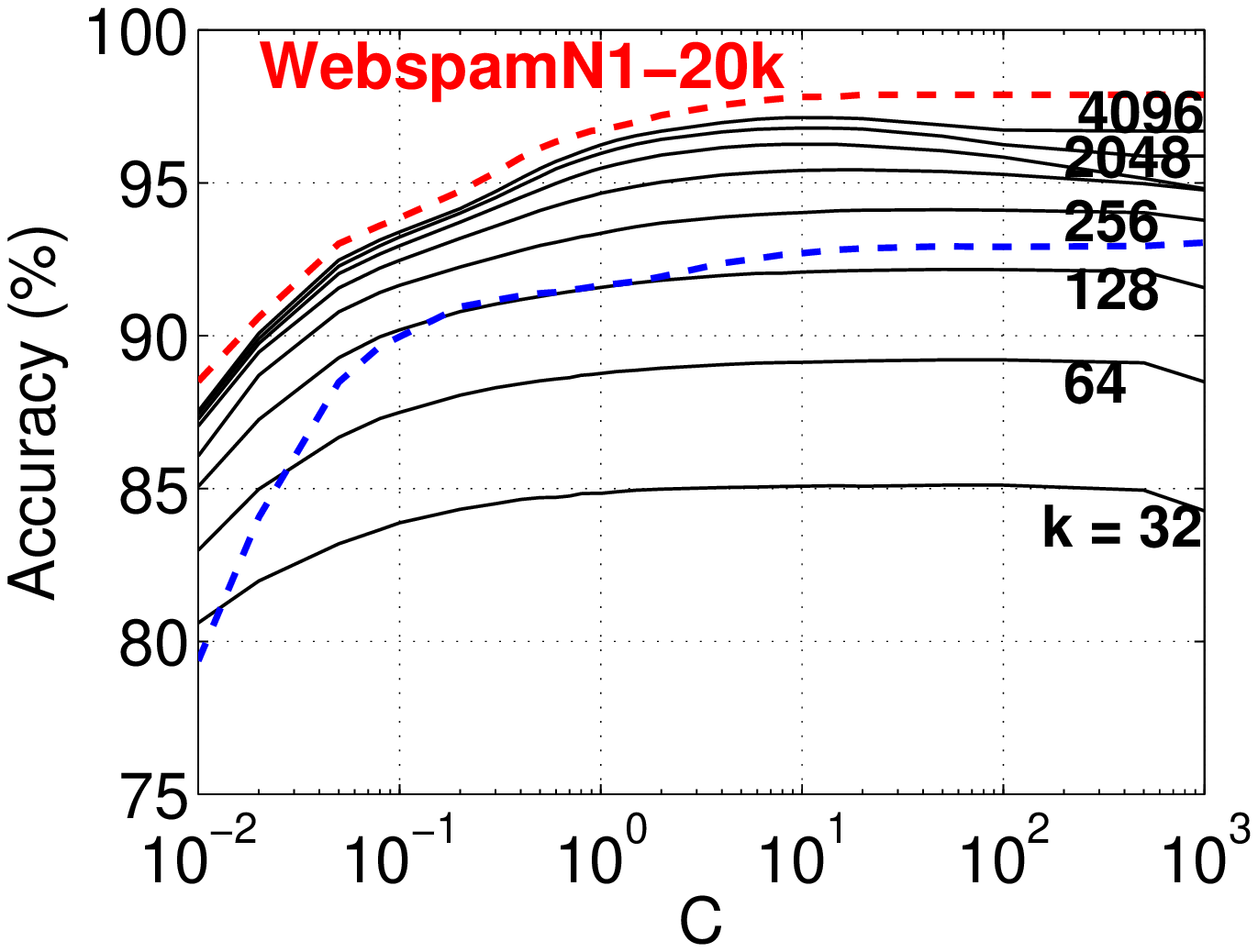}\hspace{-0.05in}
\includegraphics[width=1.65in]{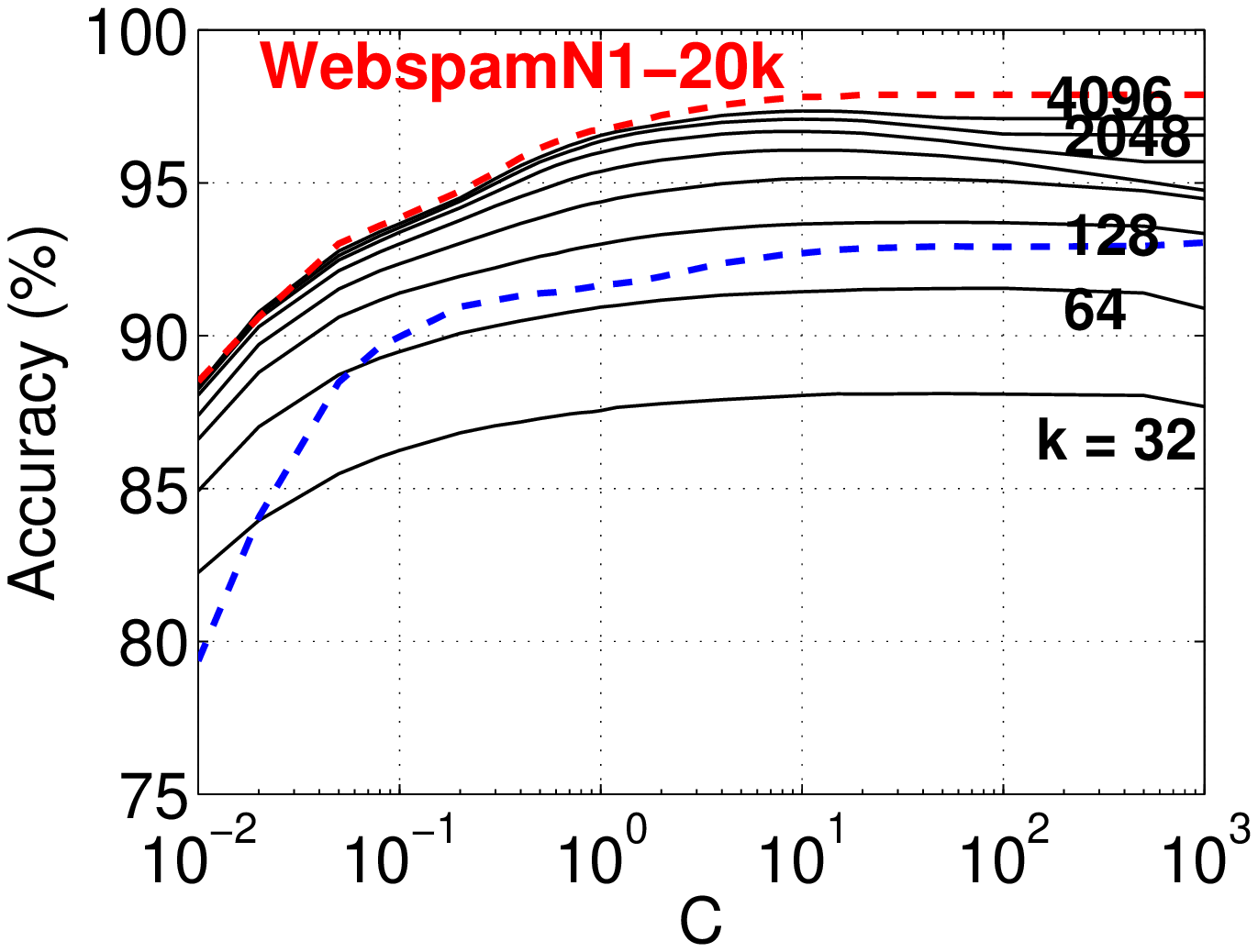}\hspace{-0.05in}
\includegraphics[width=1.65in]{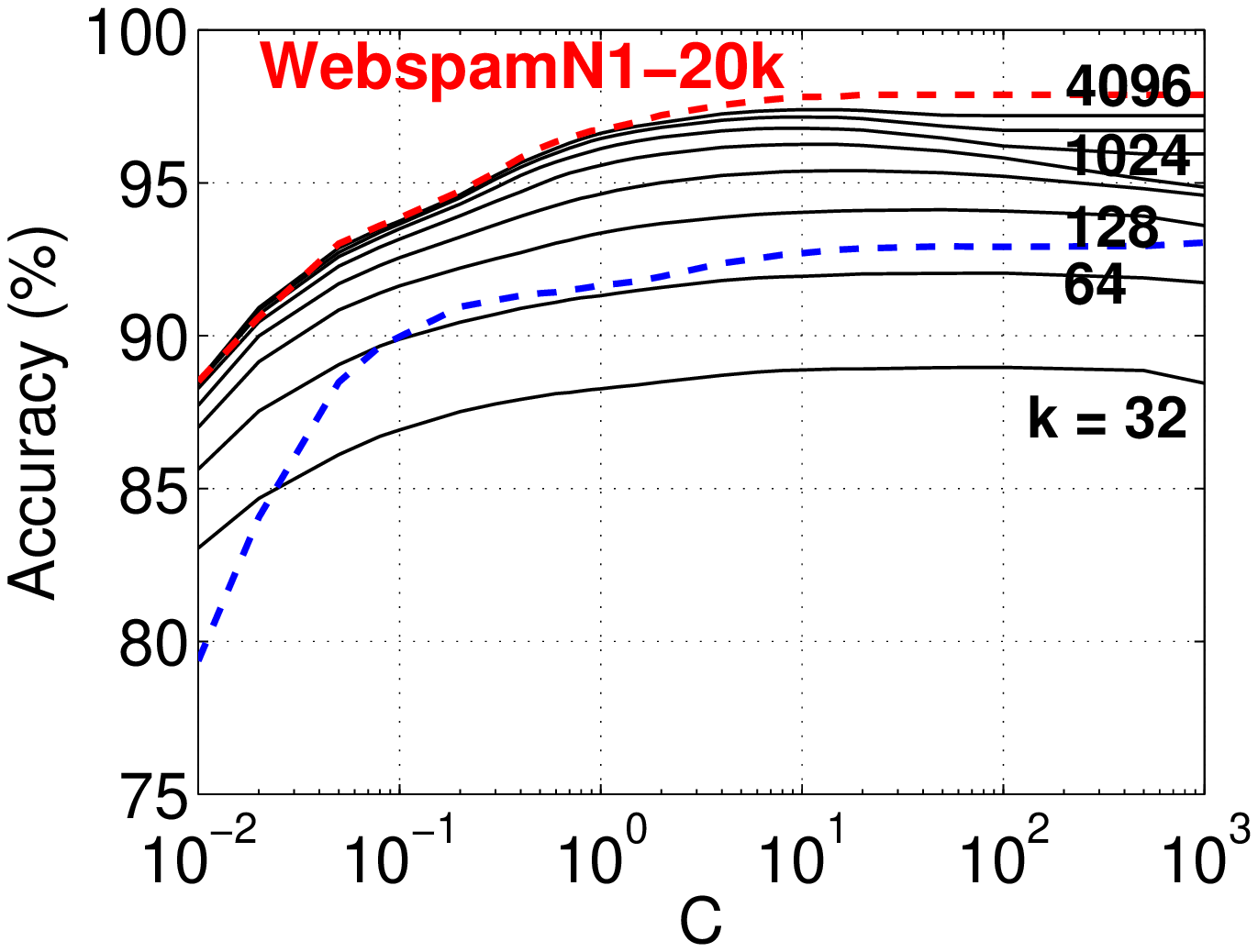}
}

\end{center}
\vspace{-0.3in}
\caption{\small Classification accuracies  by using 0-bit CWS hashing and linear SVM. The original CWS algorithm produces samples in the form of $(i^*,t^*)$. The 0-bit scheme  discards $t^*$. From left to right, the four columns represents the results for coding $i^*$ using 1 bit, 2 bits, 4 bits, and 8 bits, respectively. In each panel, the two dashed curves represent the original classification results using  min-max kernel (top and red) and  linear kernel (bottom and blue). The solid curves are the results of linear SVM and 0-bit CWS  with $k = 32$, 64, 128, 256, 512, 1024, 2048, 4096 (from bottom to top).}\label{fig_HashSVM}
\end{figure*}

\begin{figure*}
\begin{center}

\mbox{
\includegraphics[width=1.65in]{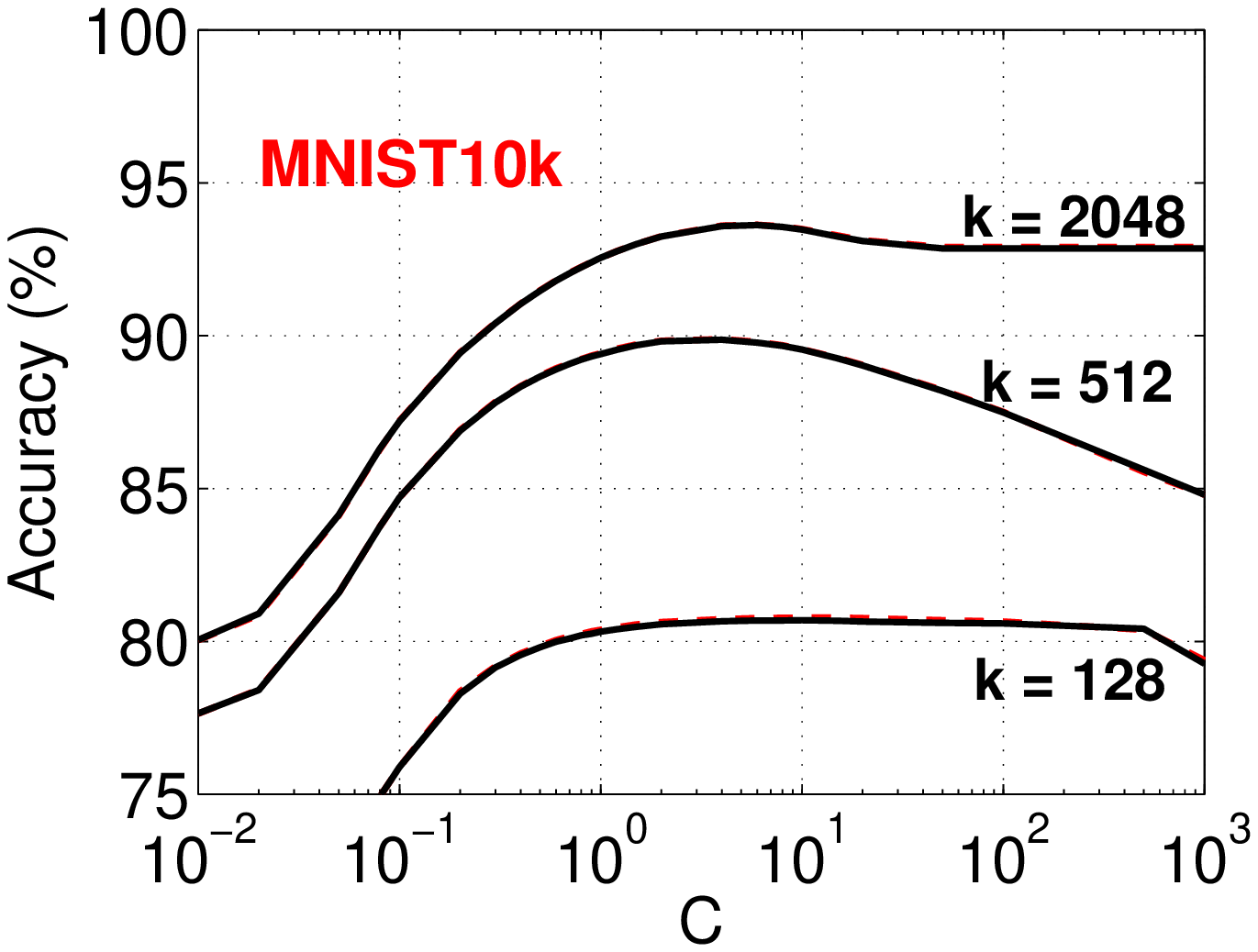}\hspace{-0.05in}
\includegraphics[width=1.65in]{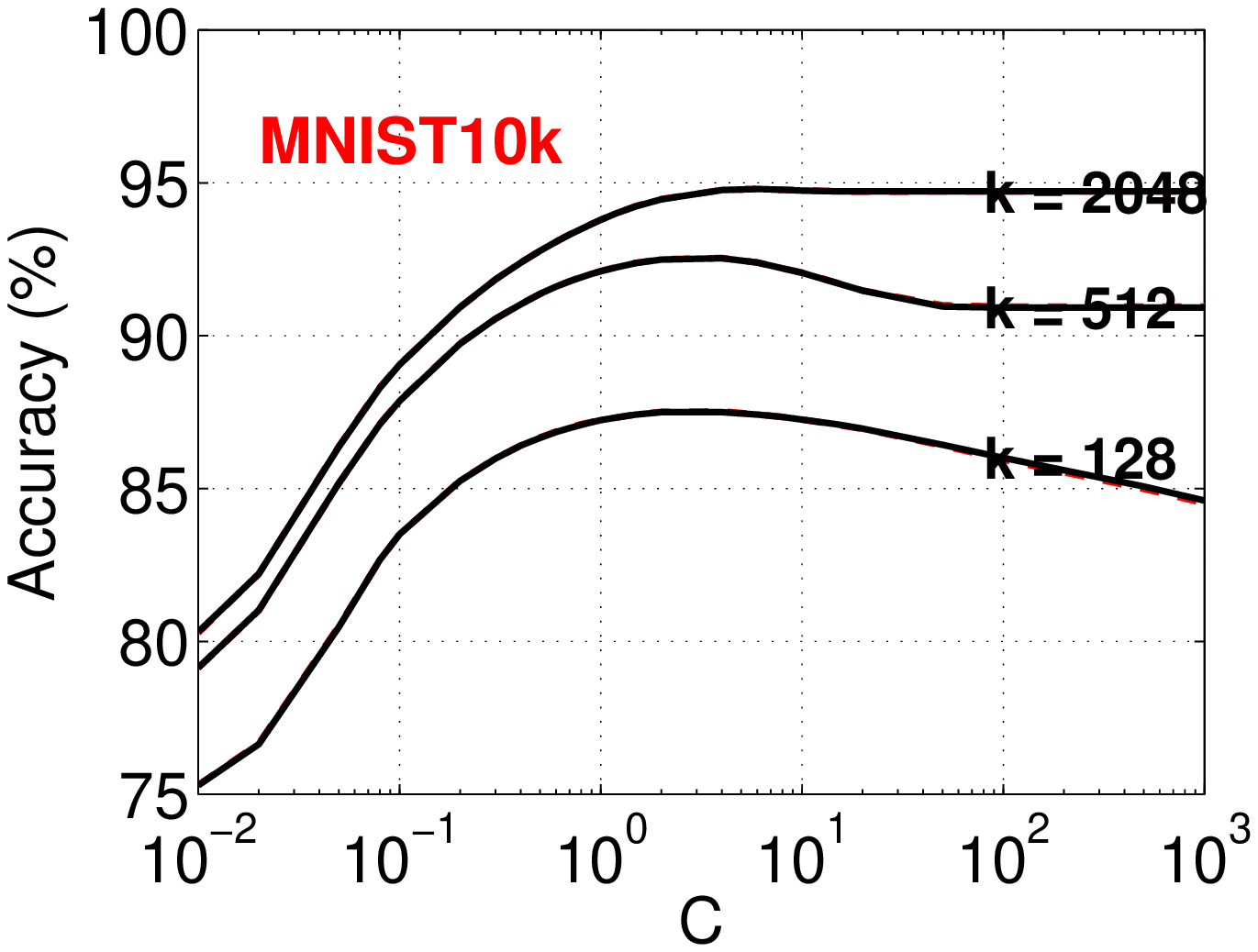}\hspace{-0.05in}
\includegraphics[width=1.65in]{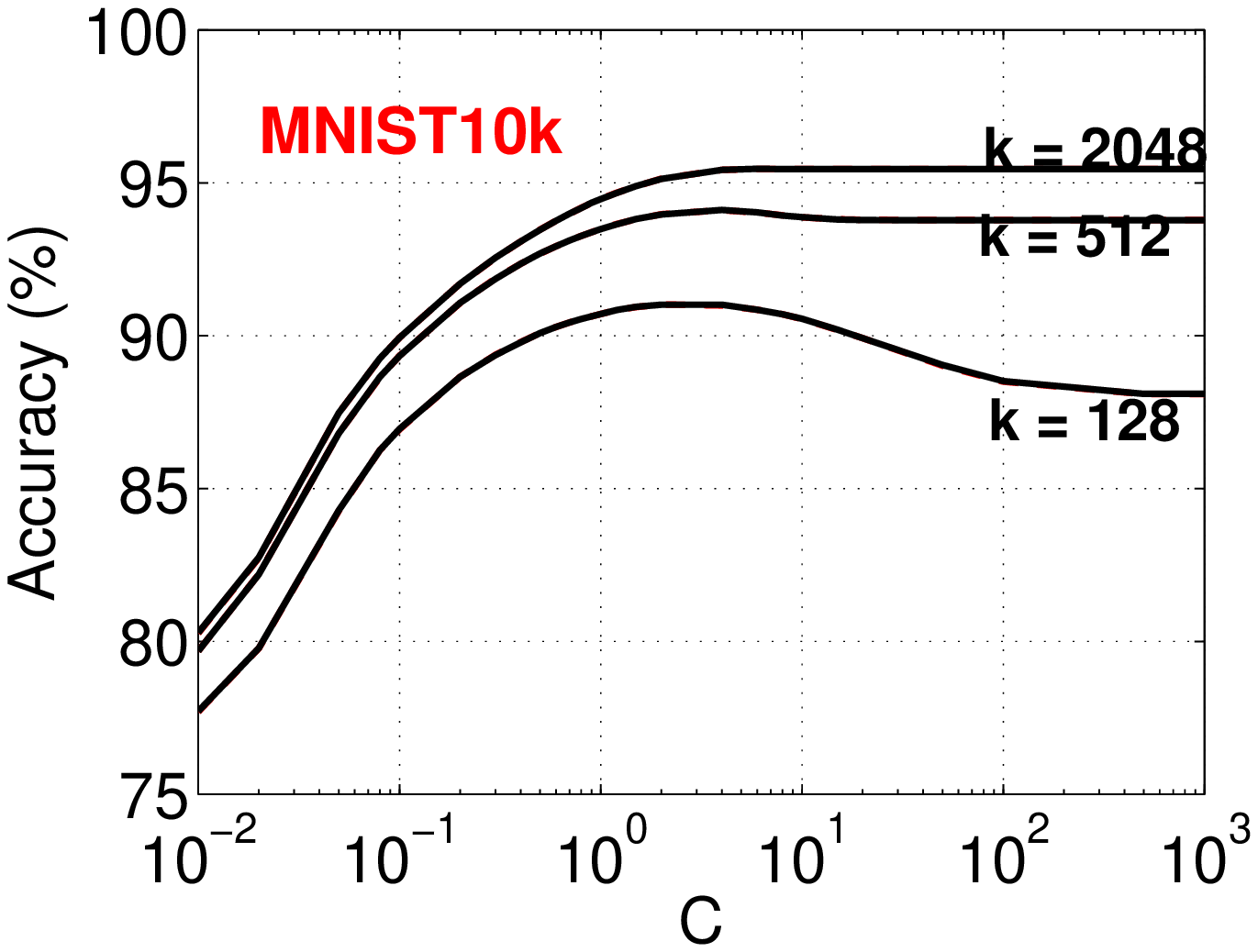}\hspace{-0.05in}
\includegraphics[width=1.65in]{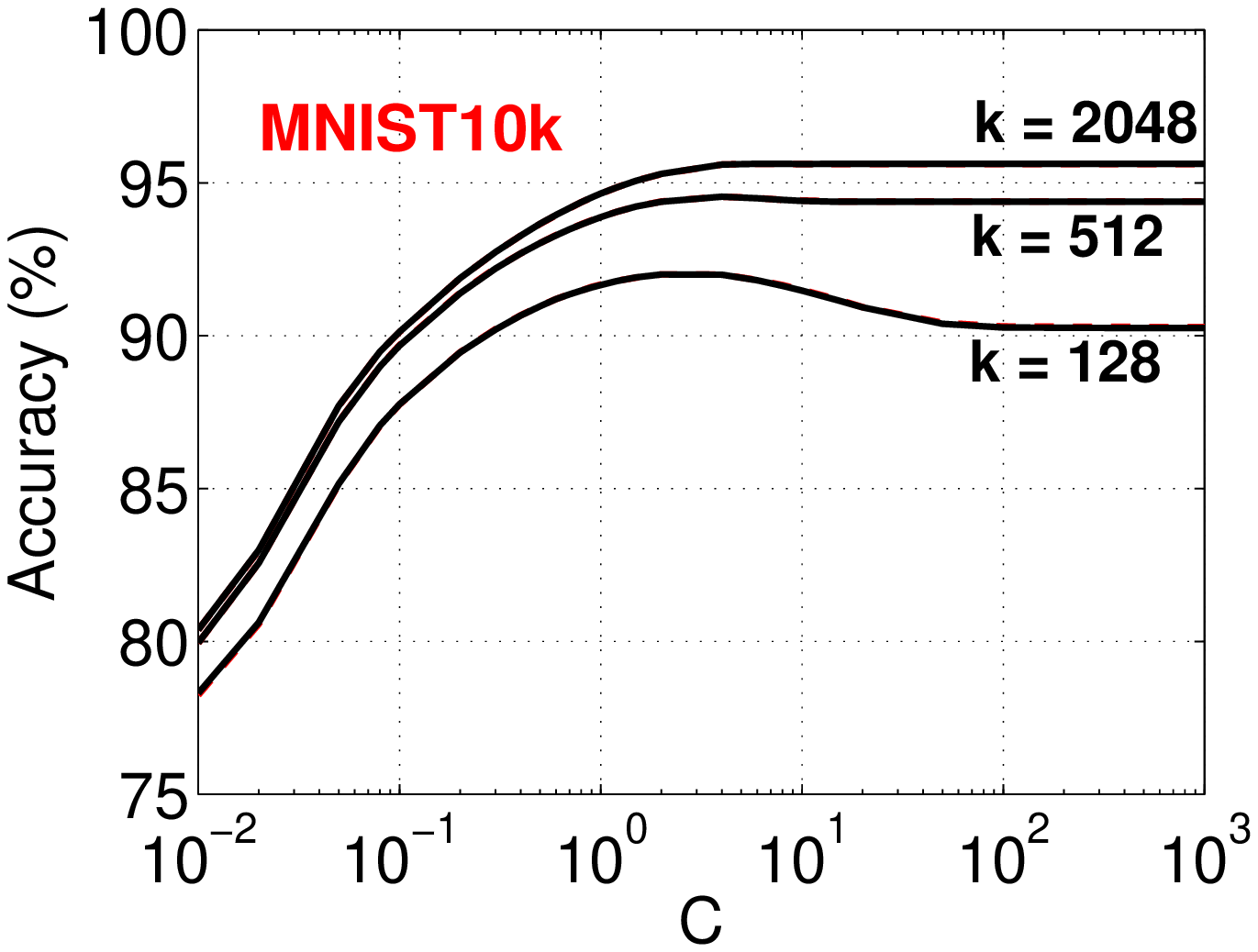}
}

\vspace{-0.11in}

\mbox{
\includegraphics[width=1.65in]{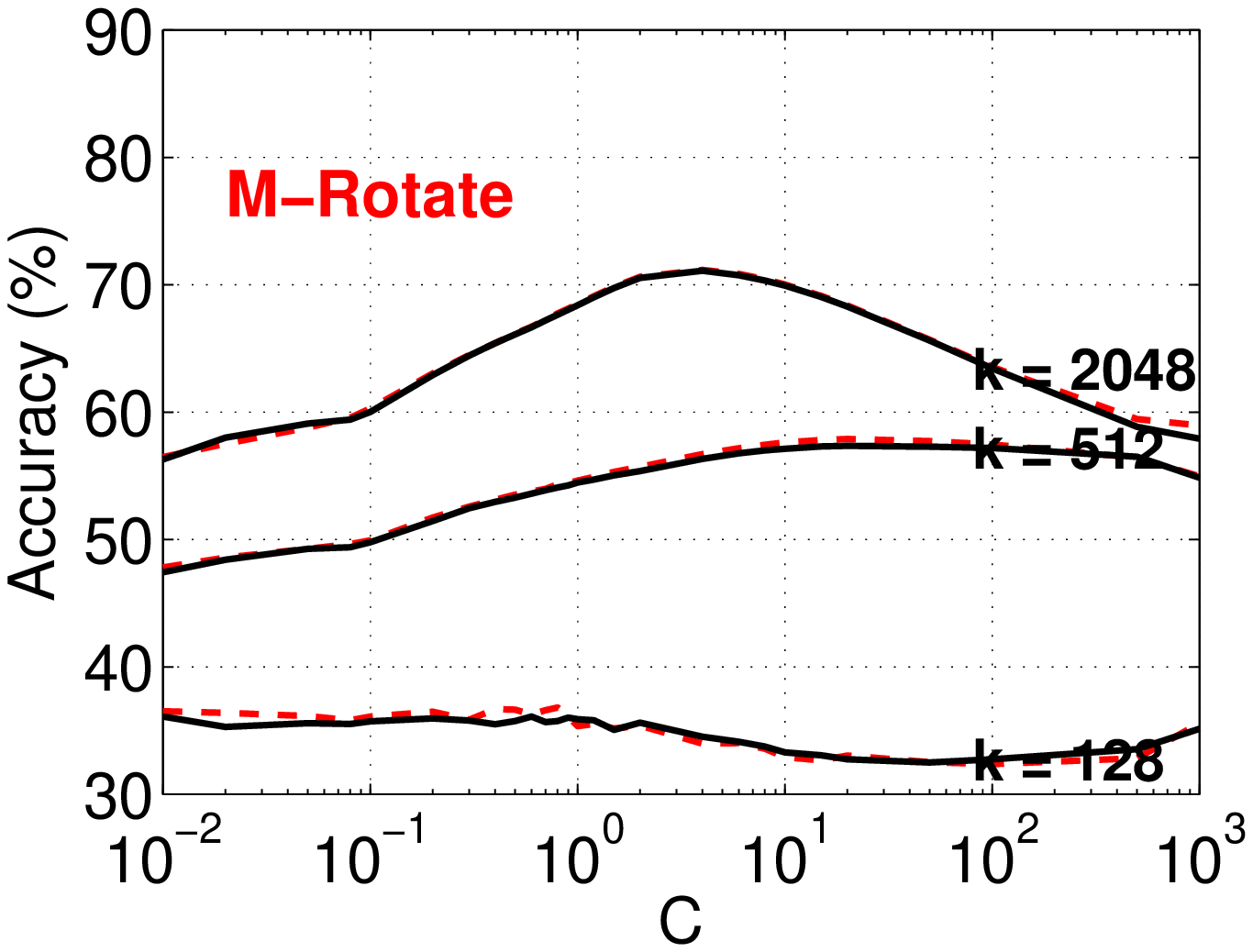}\hspace{-0.05in}
\includegraphics[width=1.65in]{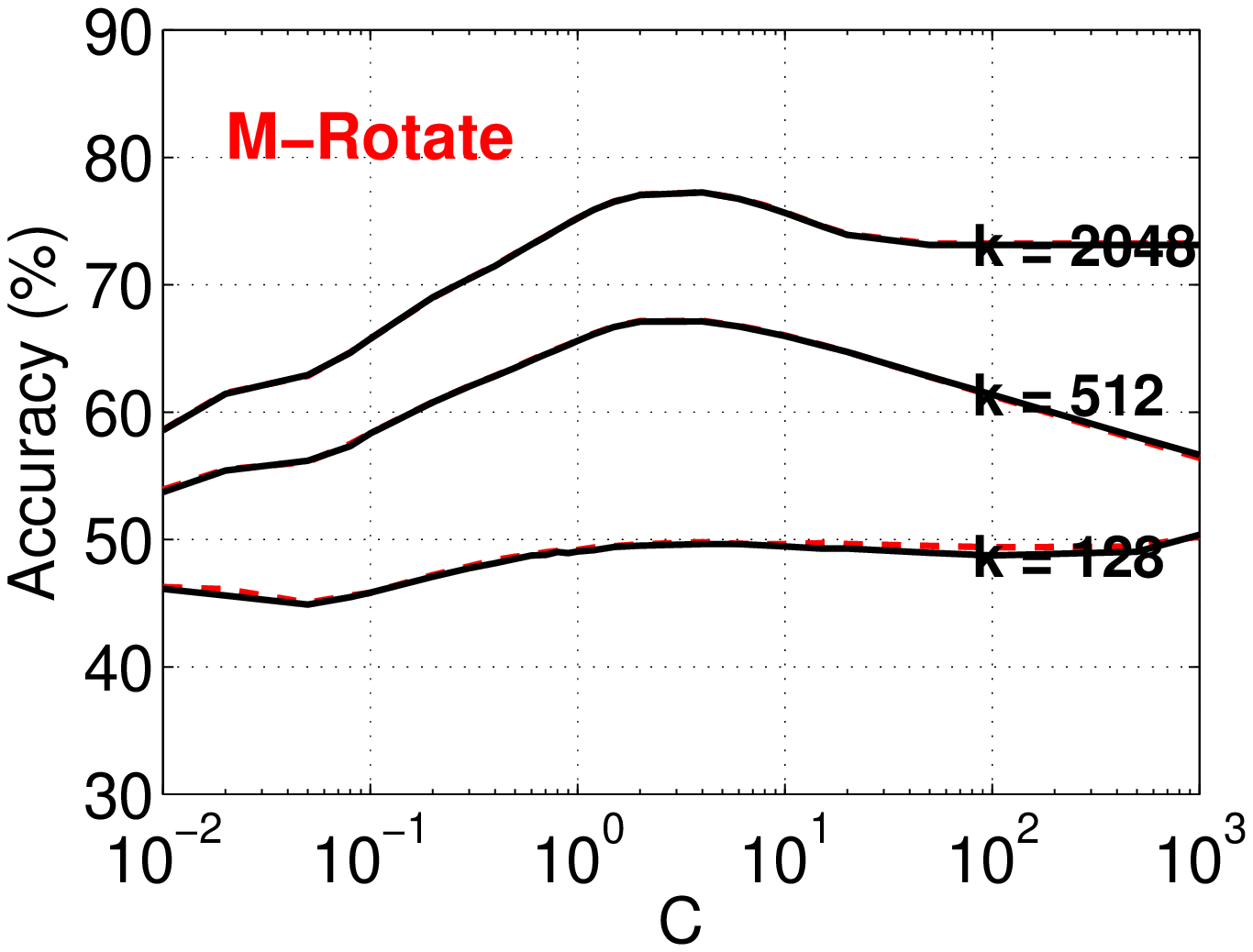}\hspace{-0.05in}
\includegraphics[width=1.65in]{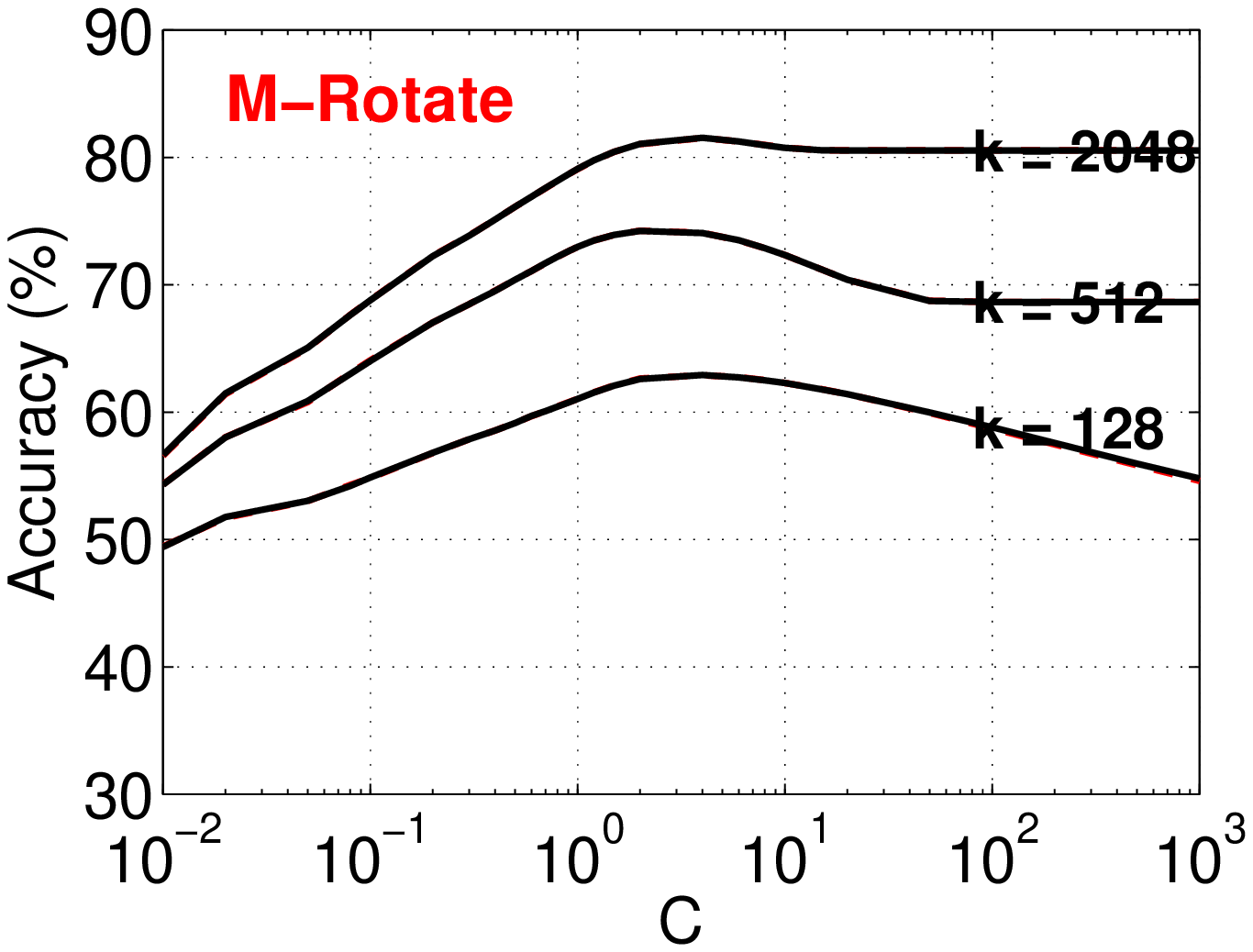}\hspace{-0.05in}
\includegraphics[width=1.65in]{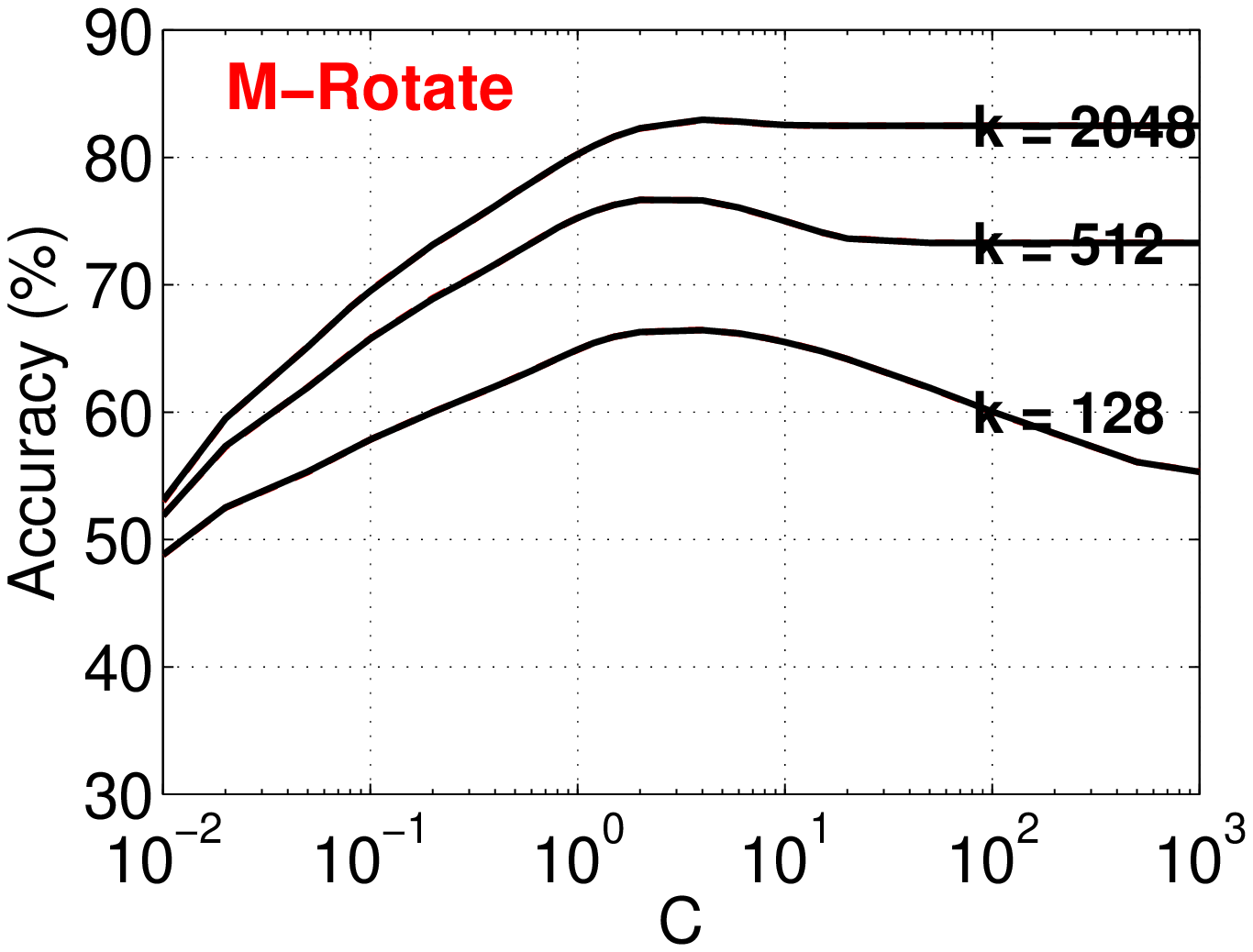}
}

\vspace{-0.11in}

\mbox{
\includegraphics[width=1.65in]{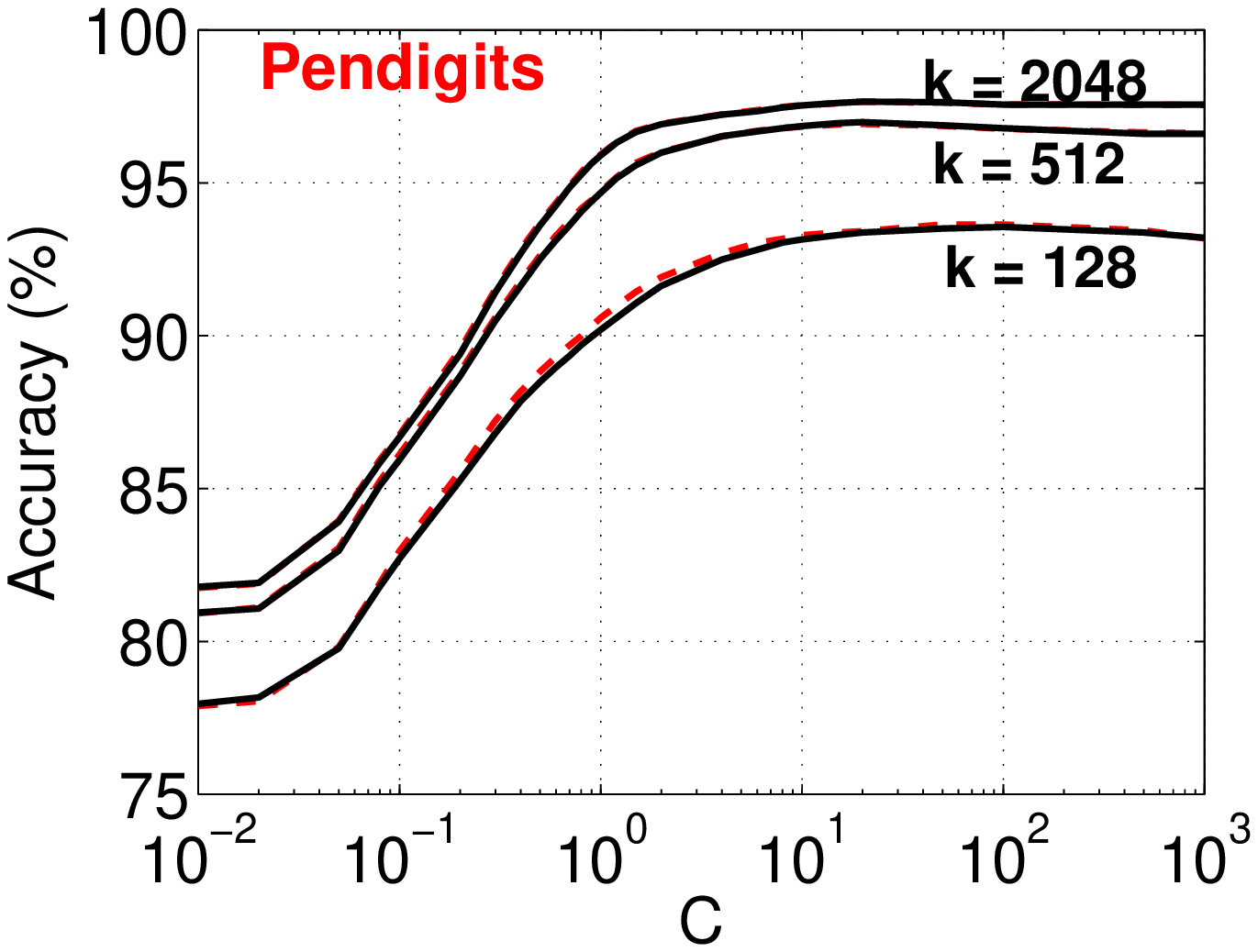}\hspace{-0.05in}
\includegraphics[width=1.65in]{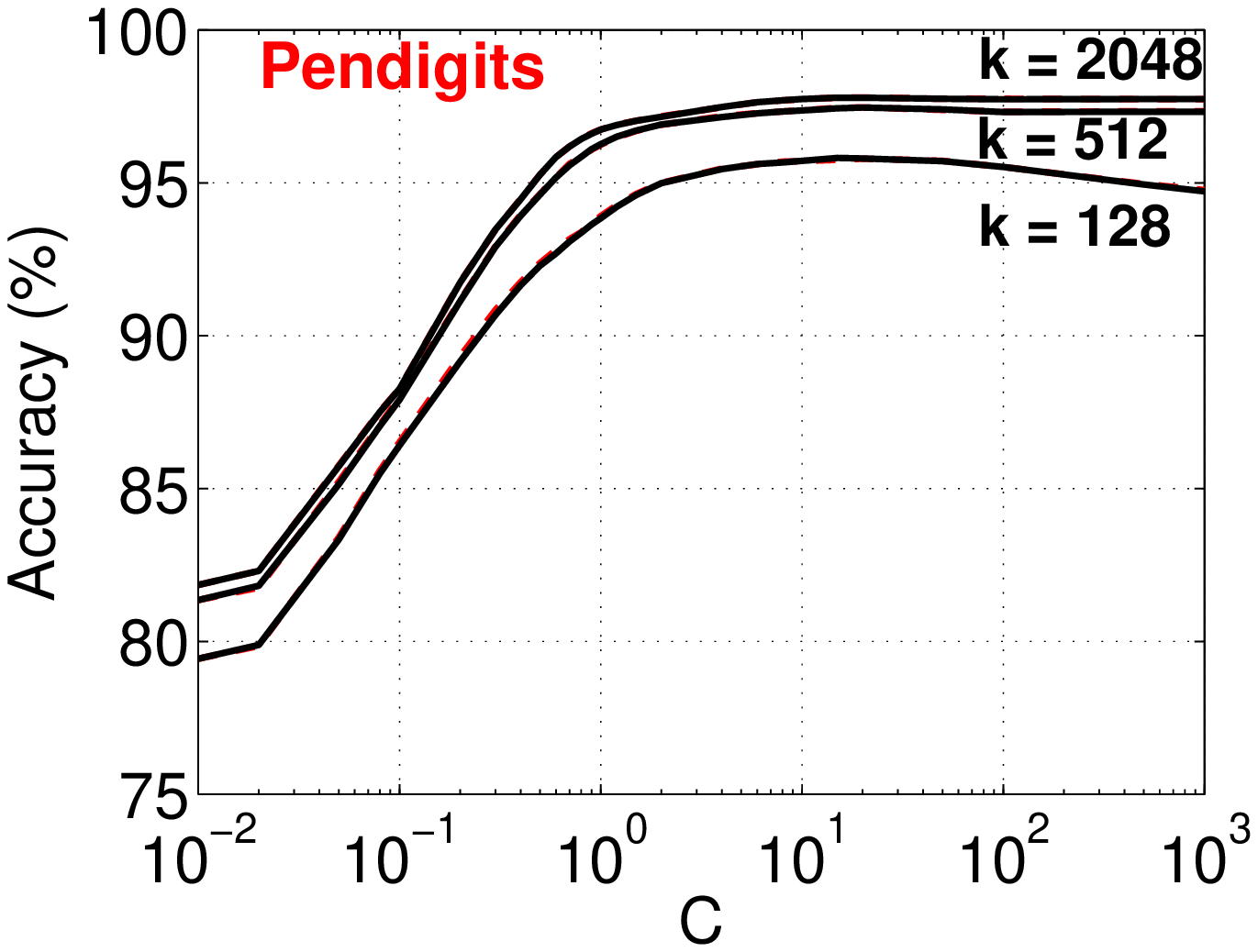}\hspace{-0.05in}
\includegraphics[width=1.65in]{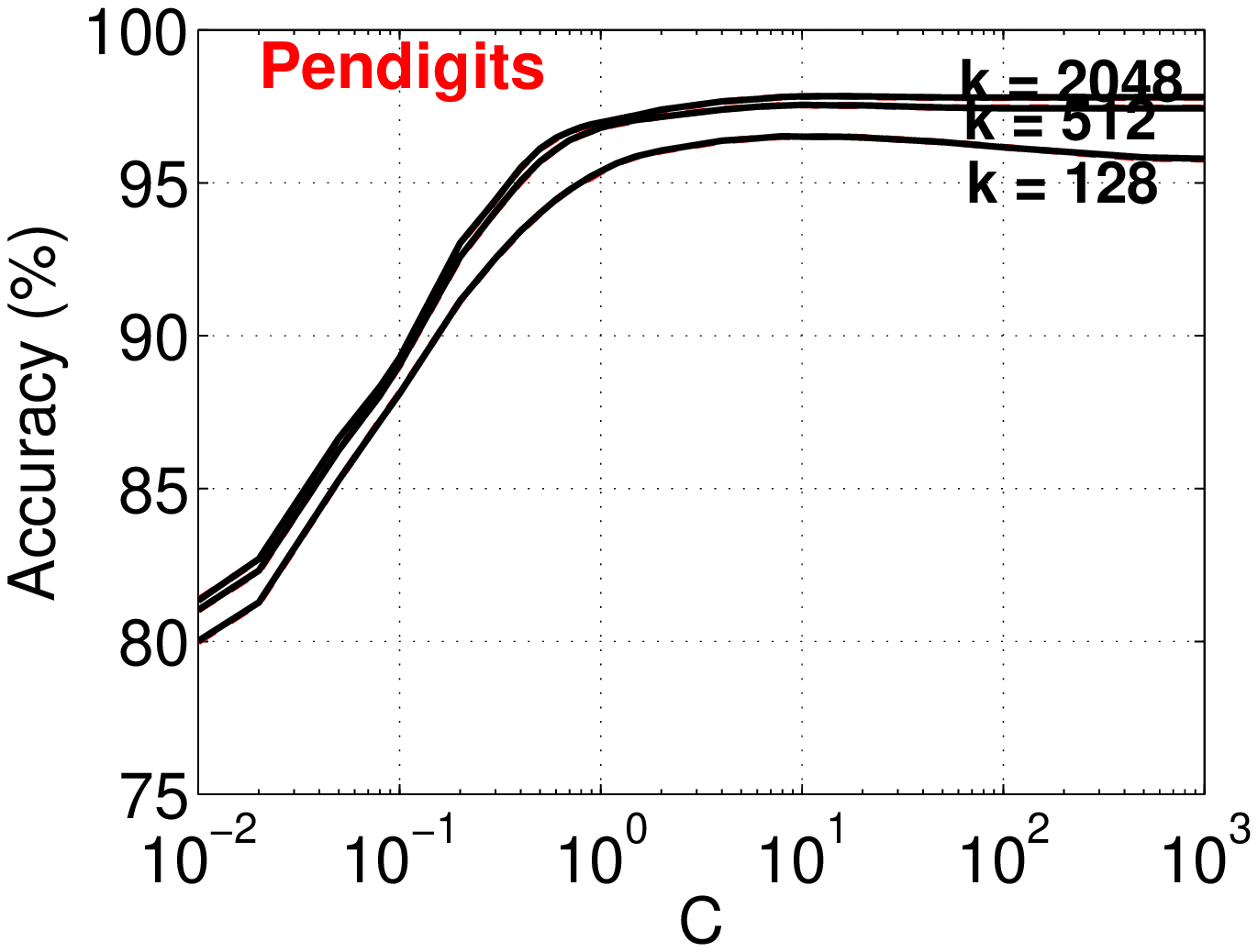}\hspace{-0.05in}
\includegraphics[width=1.65in]{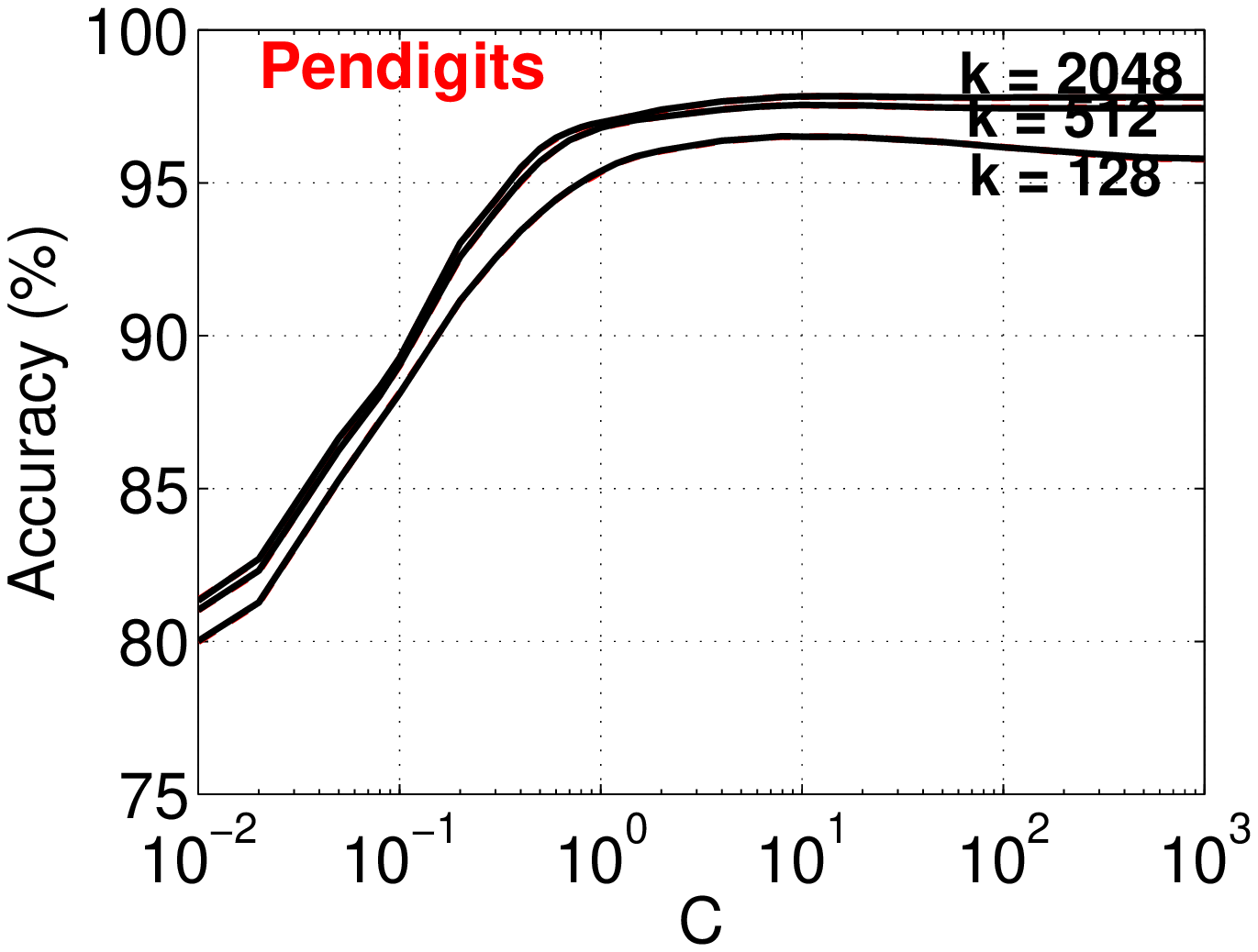}
}

\vspace{-0.11in}

\mbox{
\includegraphics[width=1.65in]{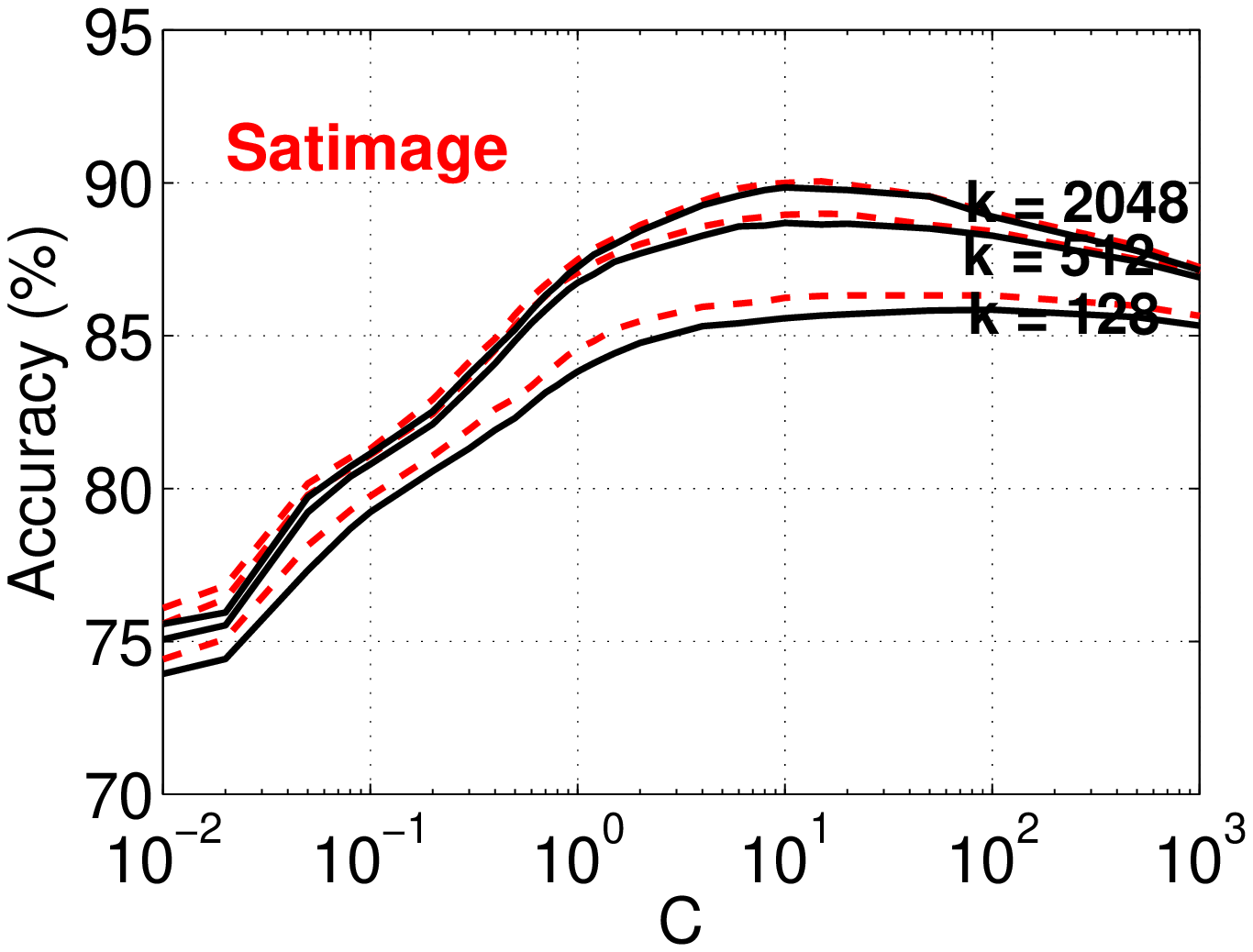}\hspace{-0.05in}
\includegraphics[width=1.65in]{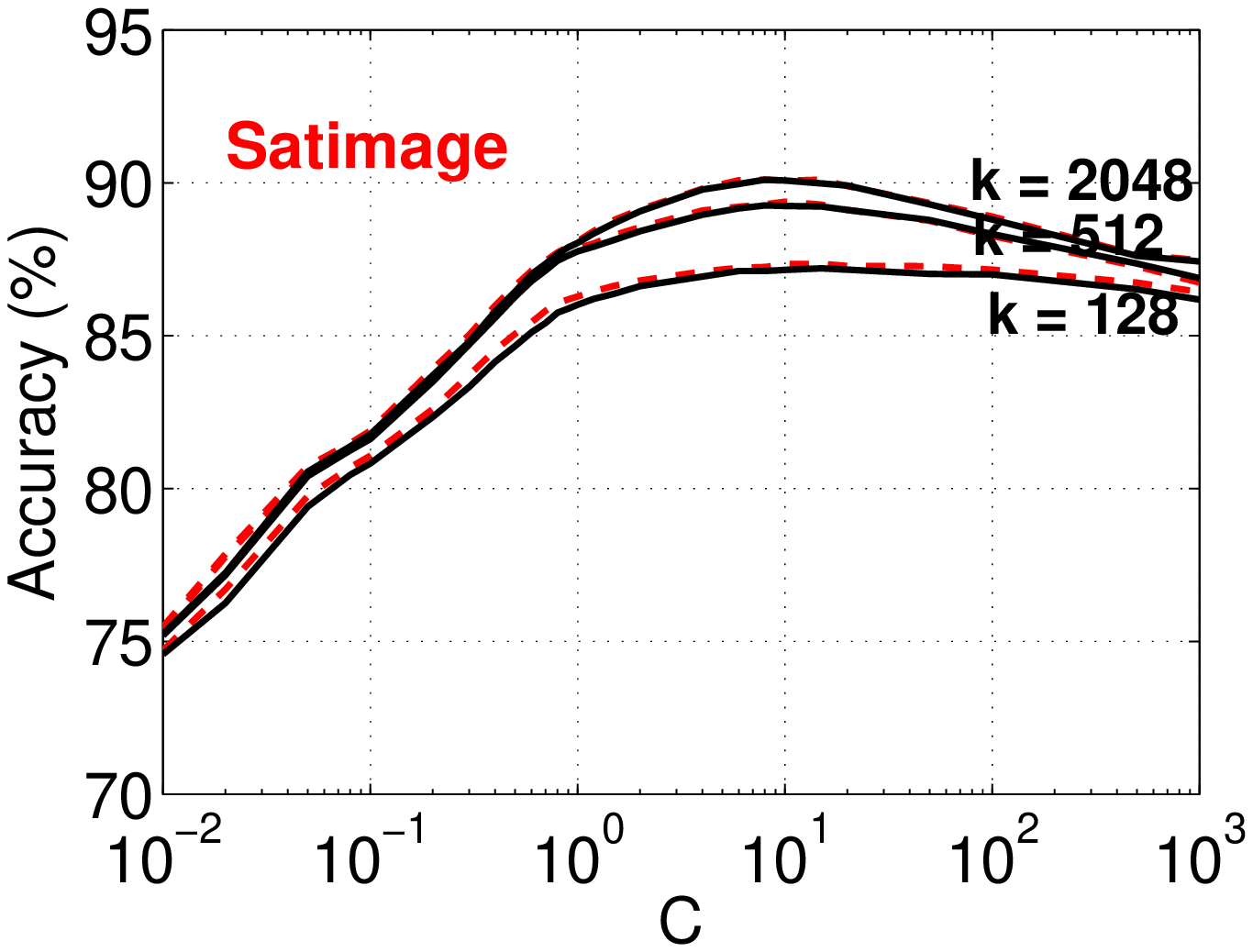}\hspace{-0.05in}
\includegraphics[width=1.65in]{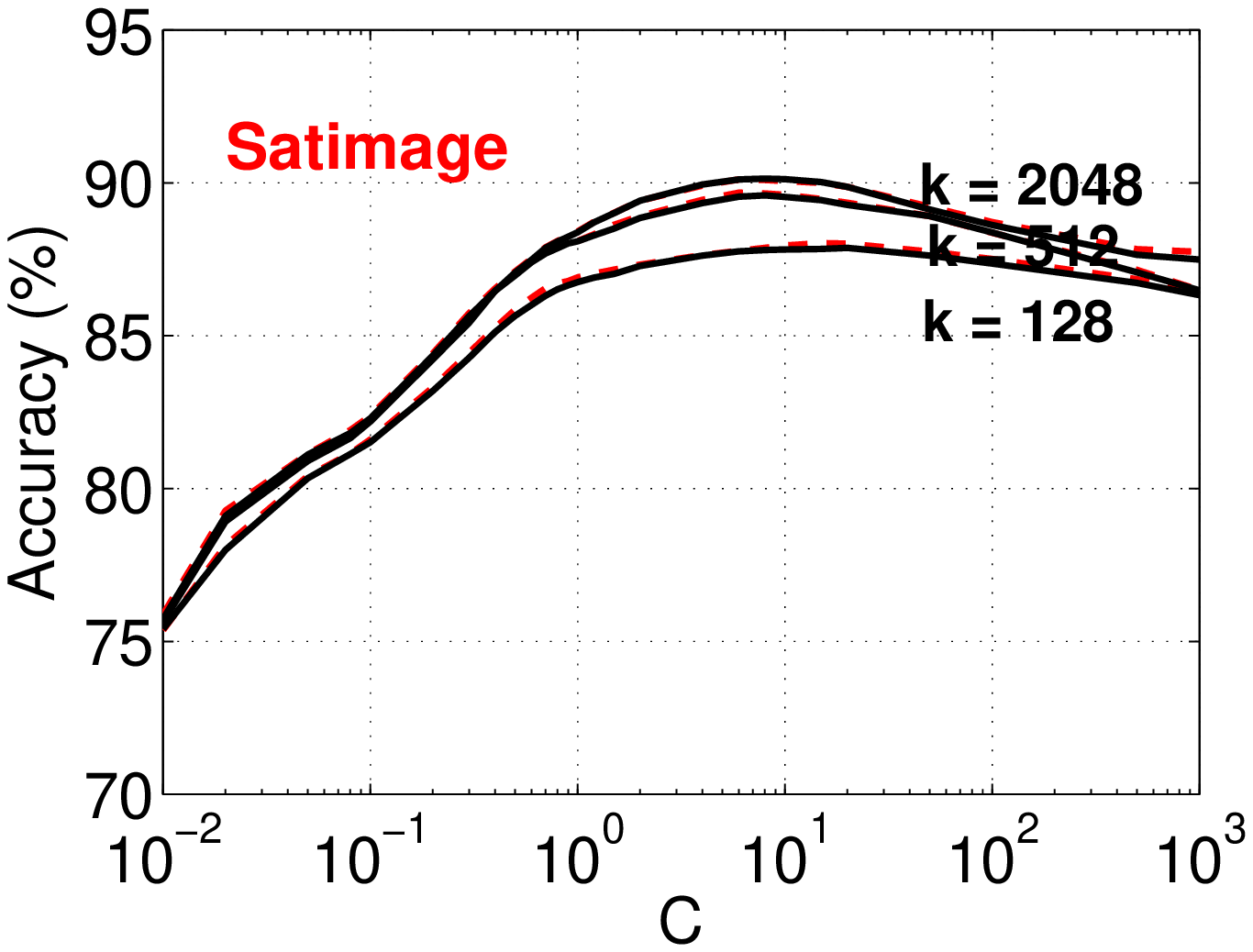}\hspace{-0.05in}
\includegraphics[width=1.65in]{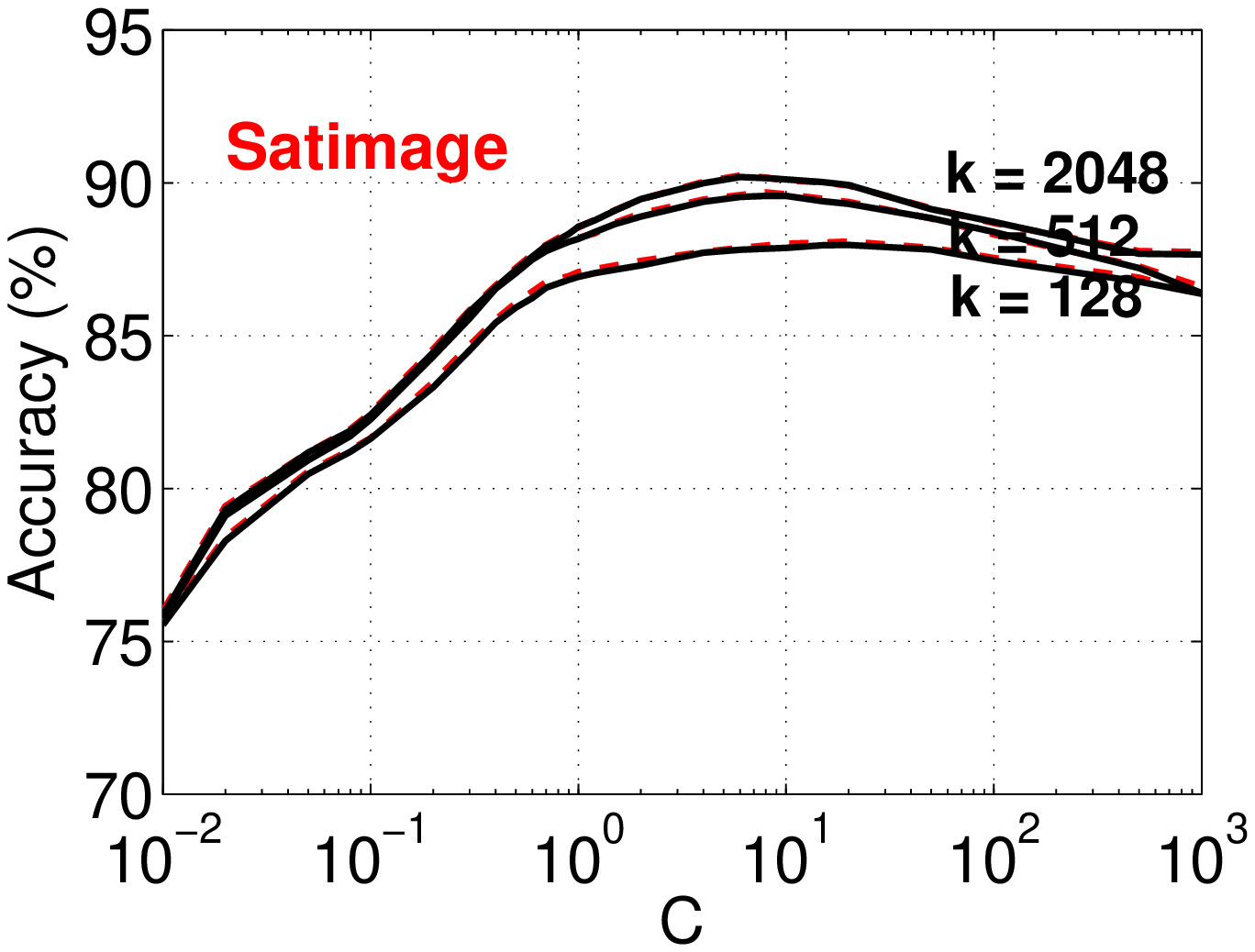}
}

\vspace{-0.11in}

\mbox{
\includegraphics[width=1.65in]{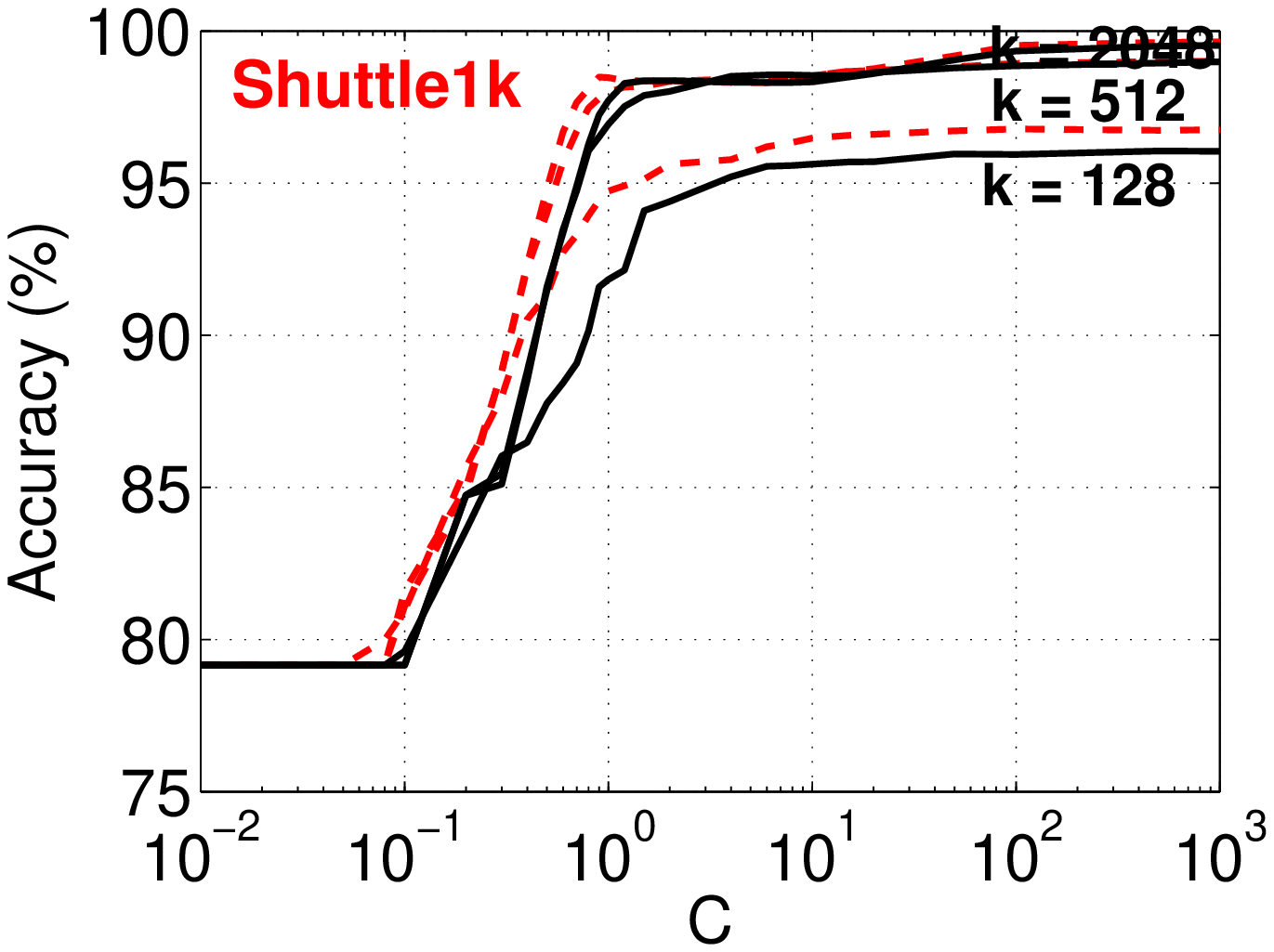}\hspace{-0.05in}
\includegraphics[width=1.65in]{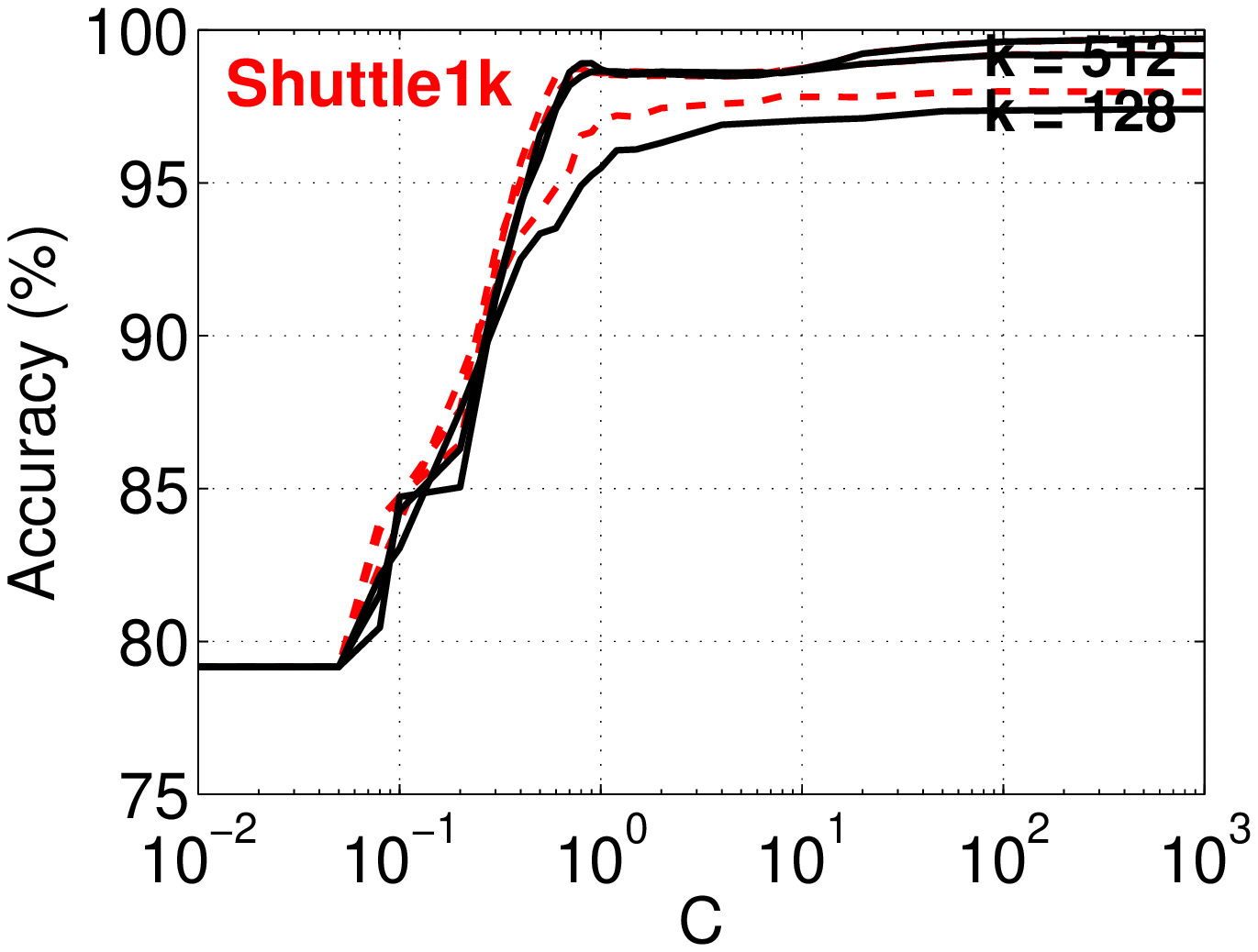}\hspace{-0.05in}
\includegraphics[width=1.65in]{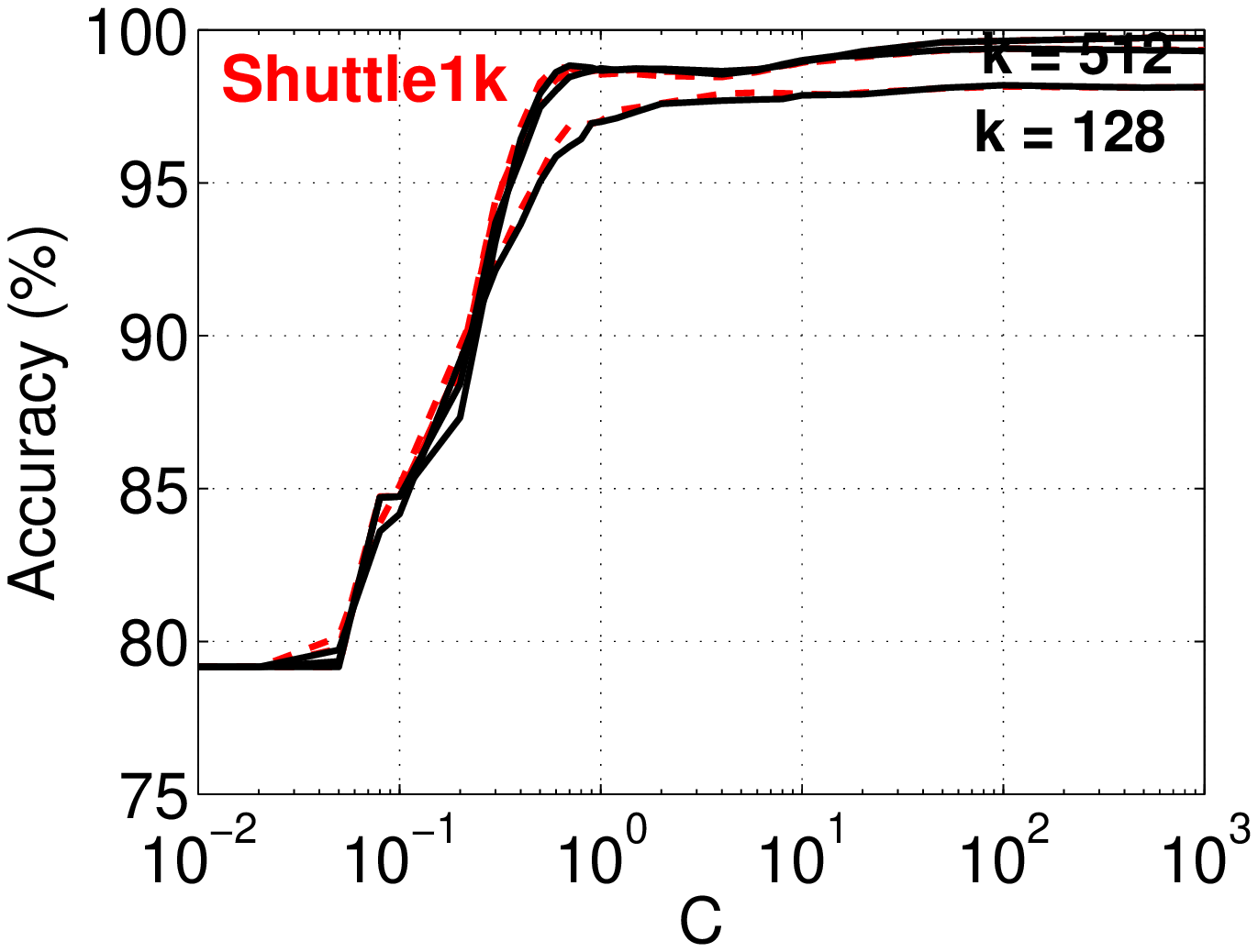}\hspace{-0.05in}
\includegraphics[width=1.65in]{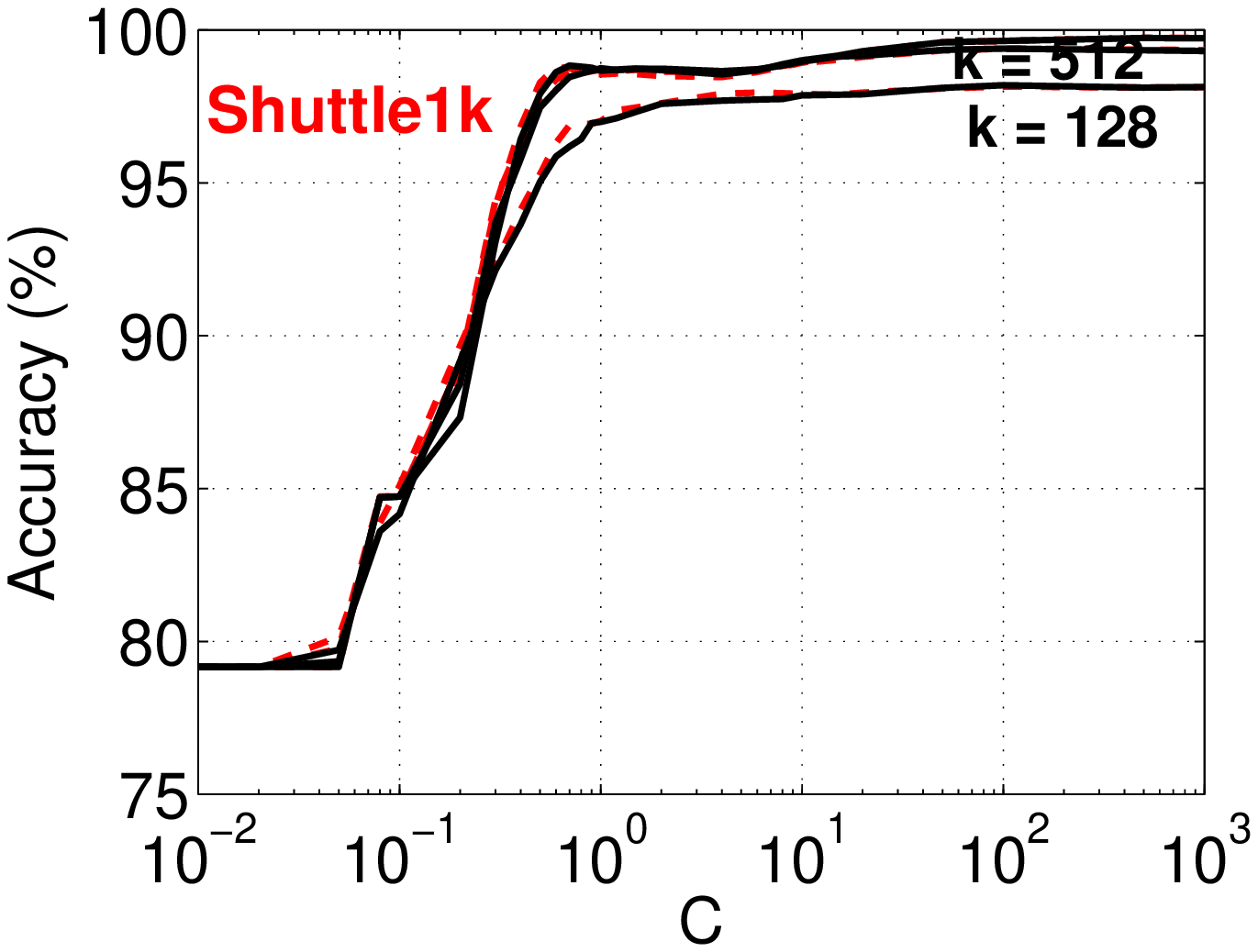}
}

\vspace{-0.11in}

\mbox{
\includegraphics[width=1.65in]{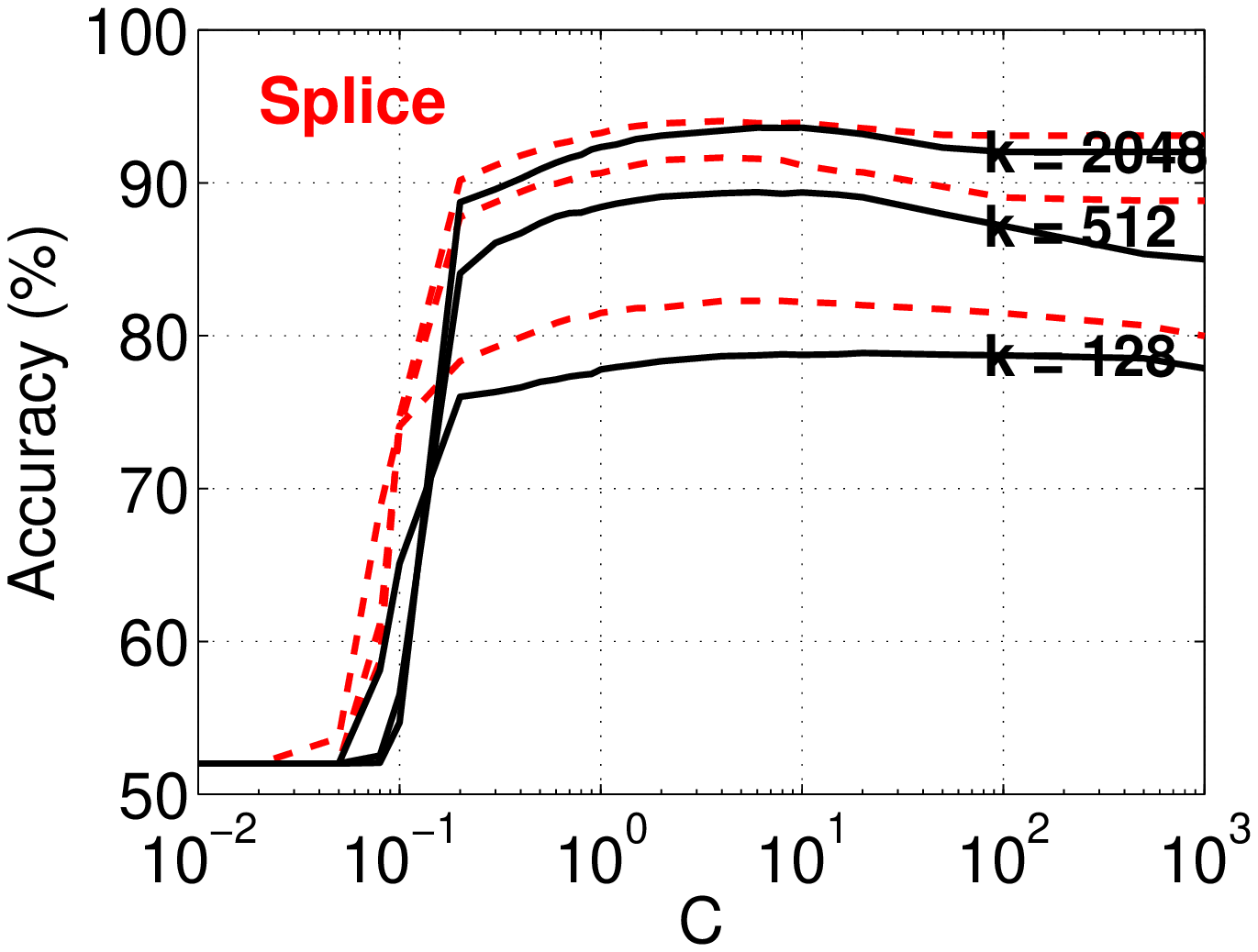}\hspace{-0.05in}
\includegraphics[width=1.65in]{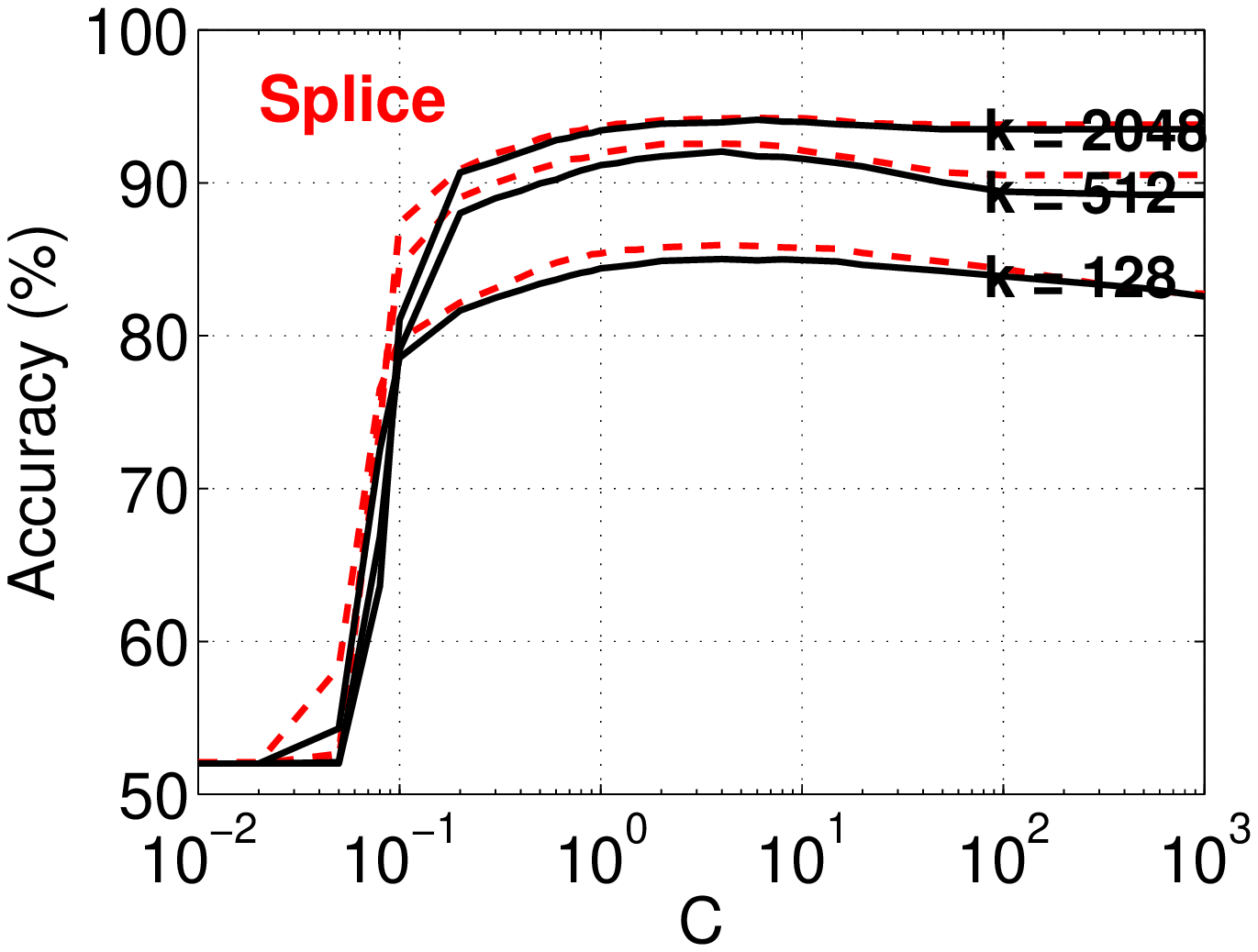}\hspace{-0.05in}
\includegraphics[width=1.65in]{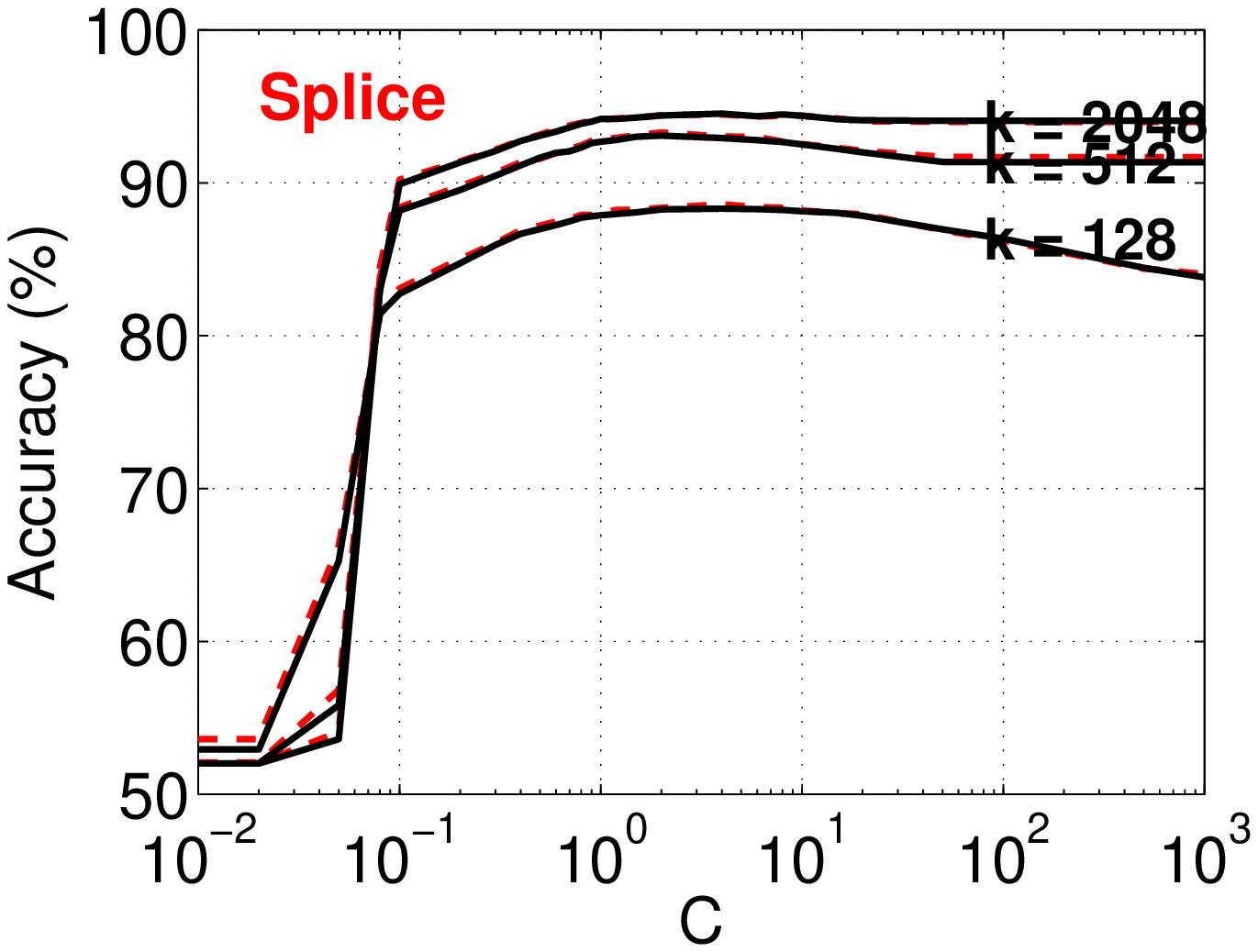}\hspace{-0.05in}
\includegraphics[width=1.65in]{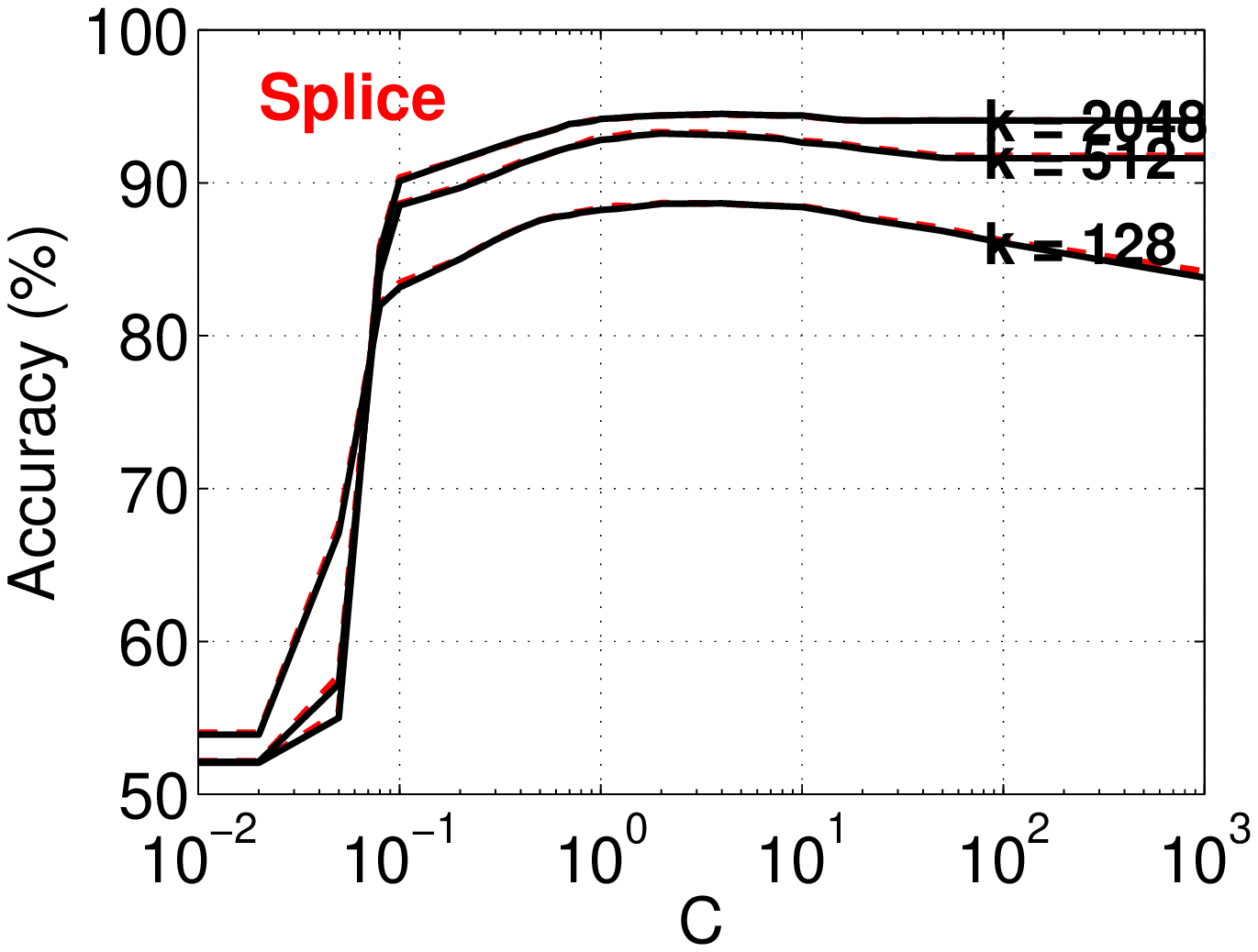}
}

\vspace{-0.11in}

\mbox{
\includegraphics[width=1.65in]{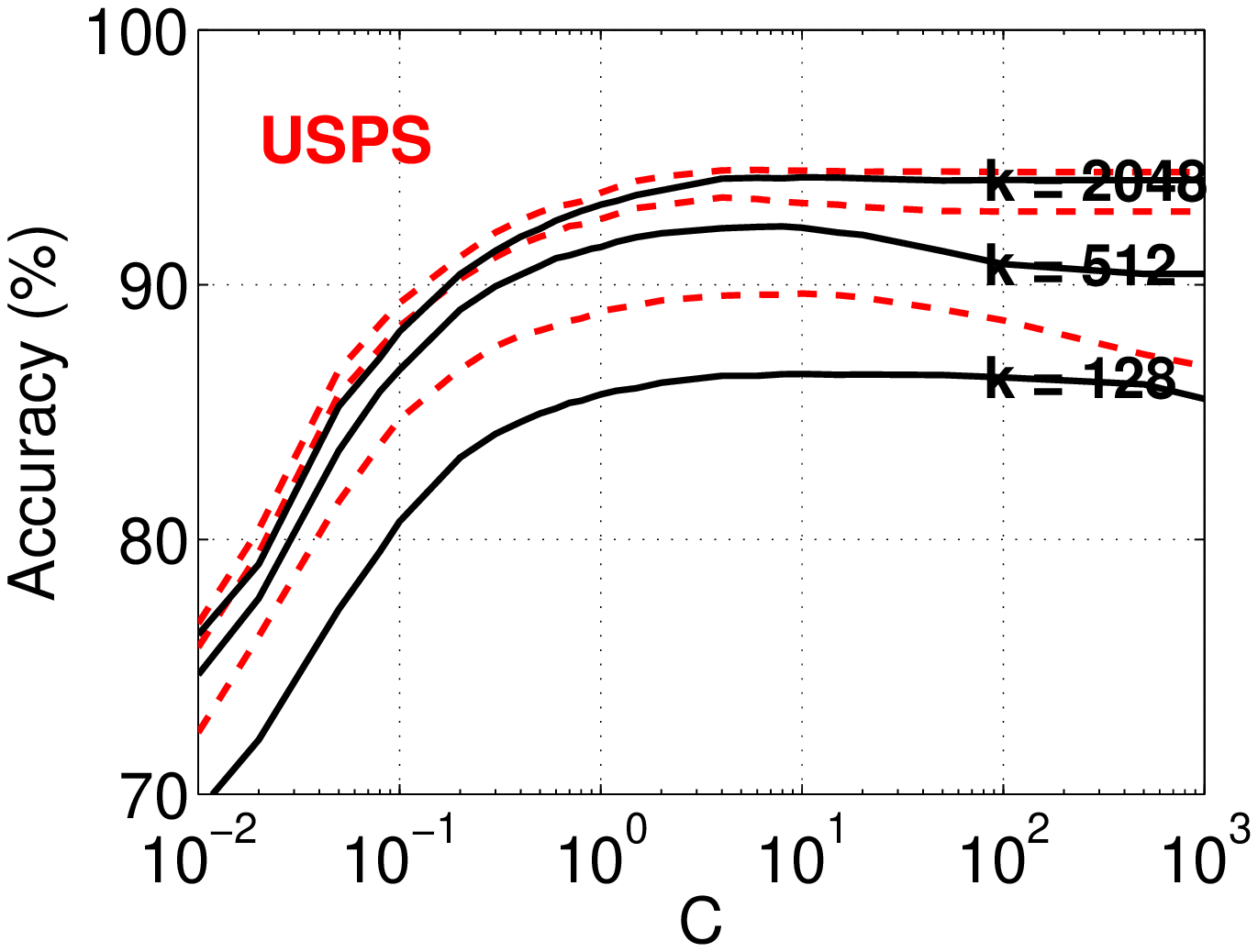}\hspace{-0.05in}
\includegraphics[width=1.65in]{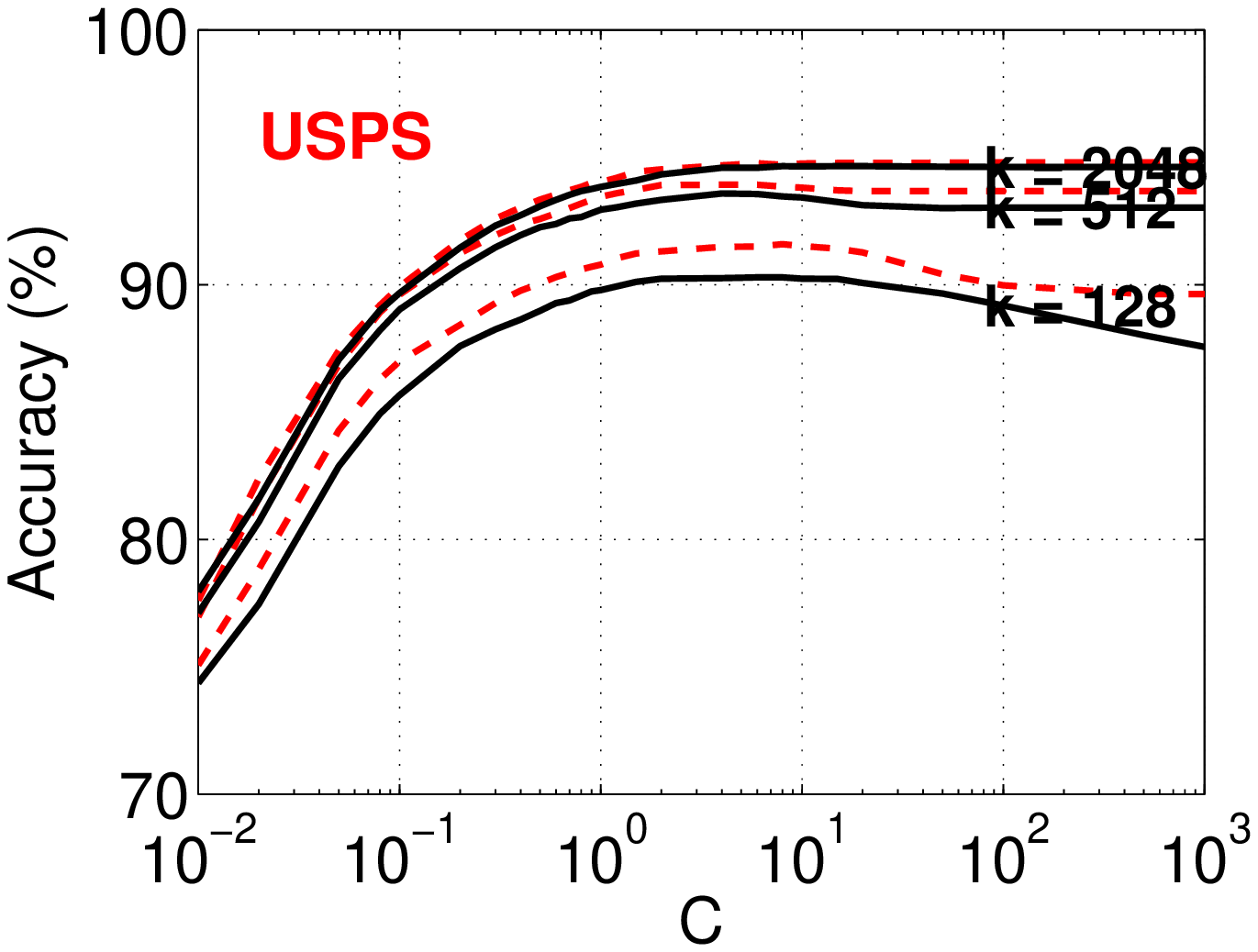}\hspace{-0.05in}
\includegraphics[width=1.65in]{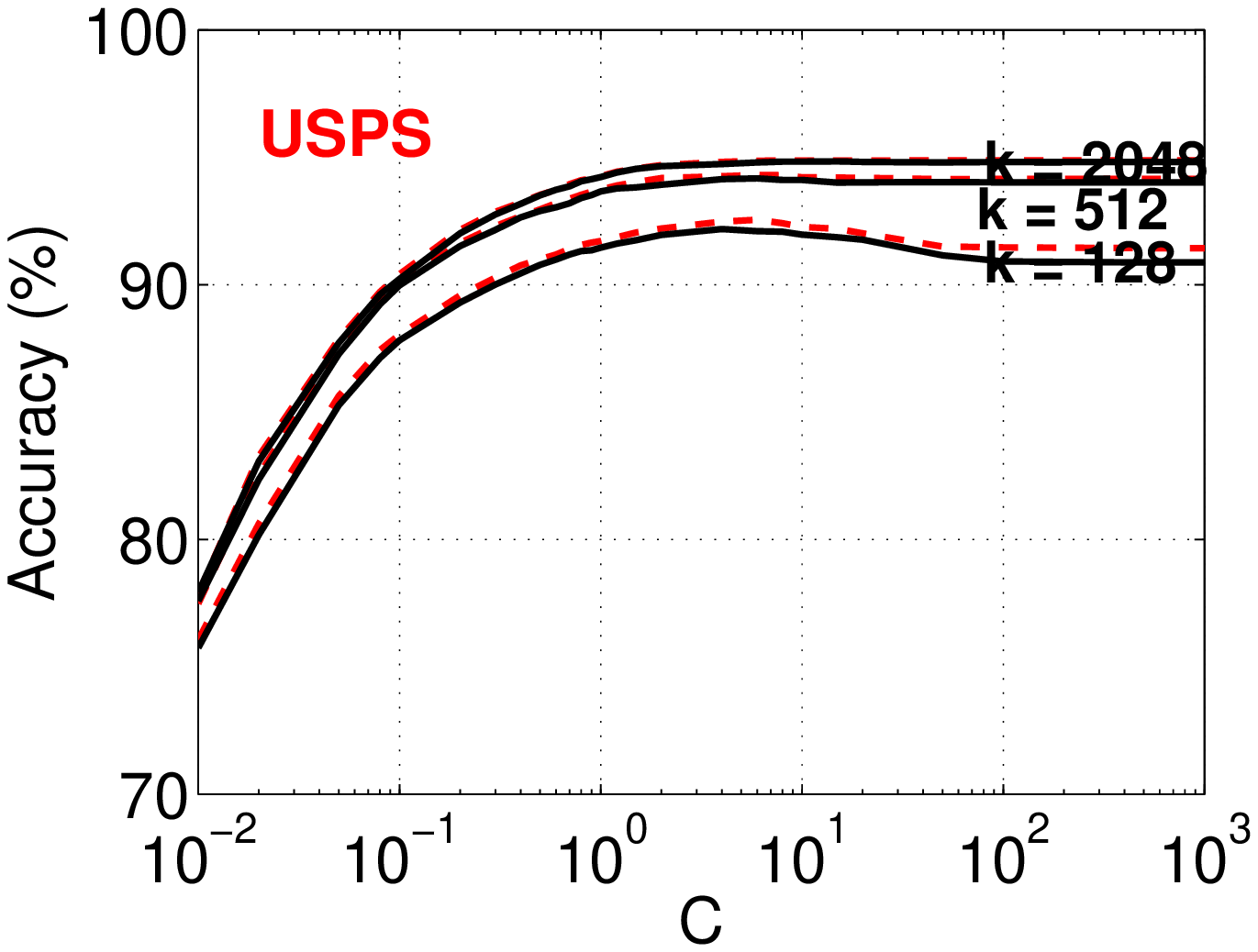}\hspace{-0.05in}
\includegraphics[width=1.65in]{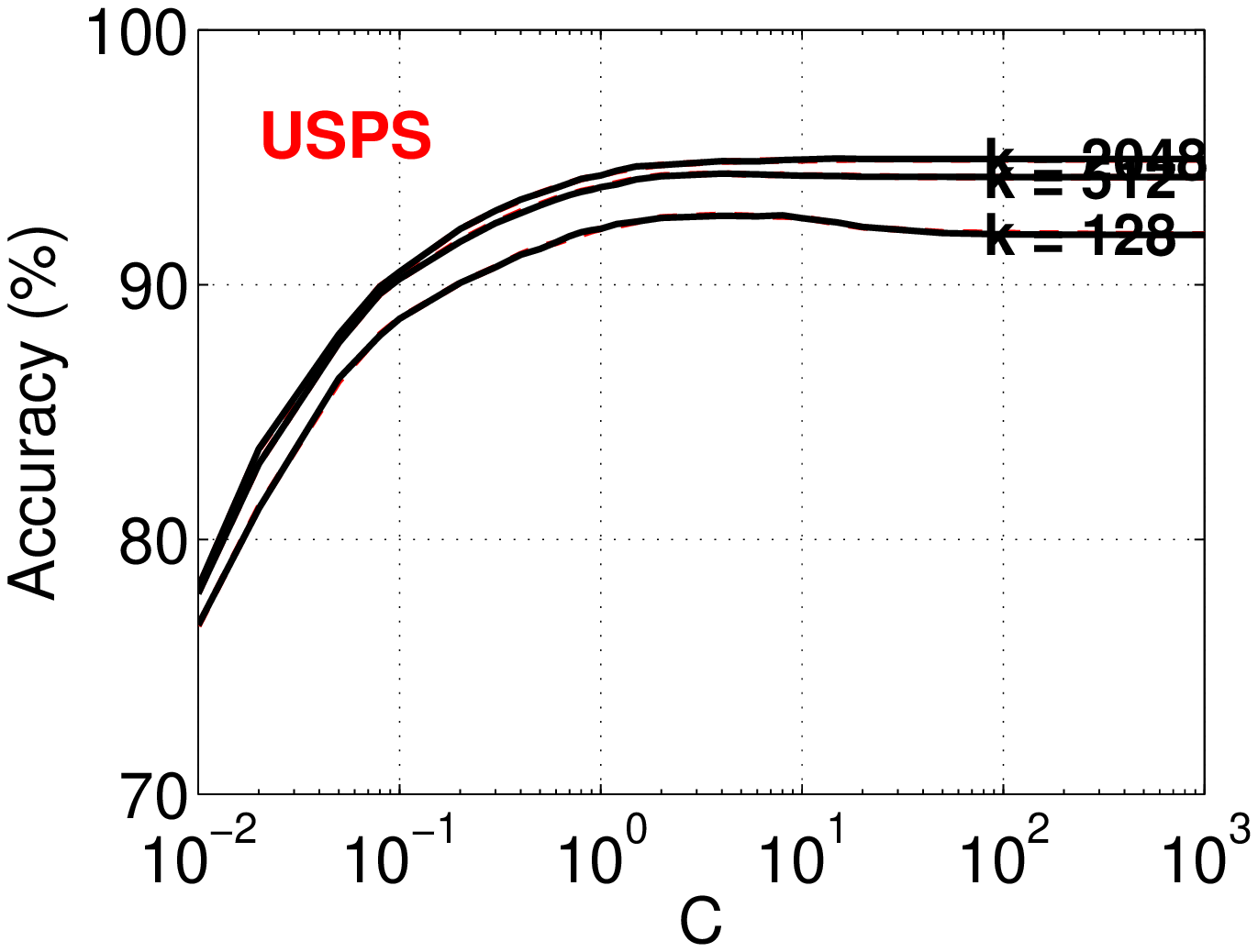}
}

\vspace{-0.11in}

\mbox{
\includegraphics[width=1.65in]{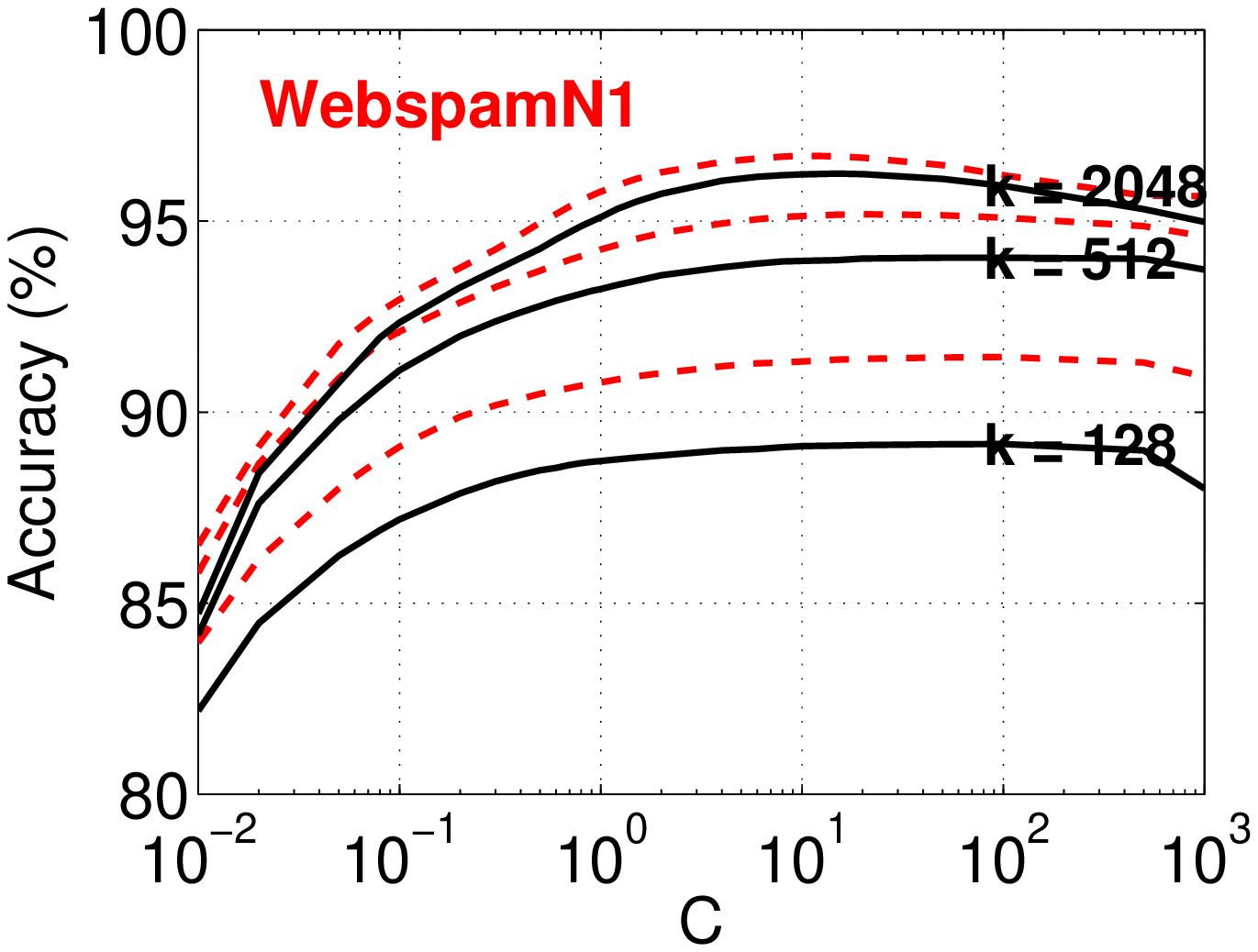}\hspace{-0.05in}
\includegraphics[width=1.65in]{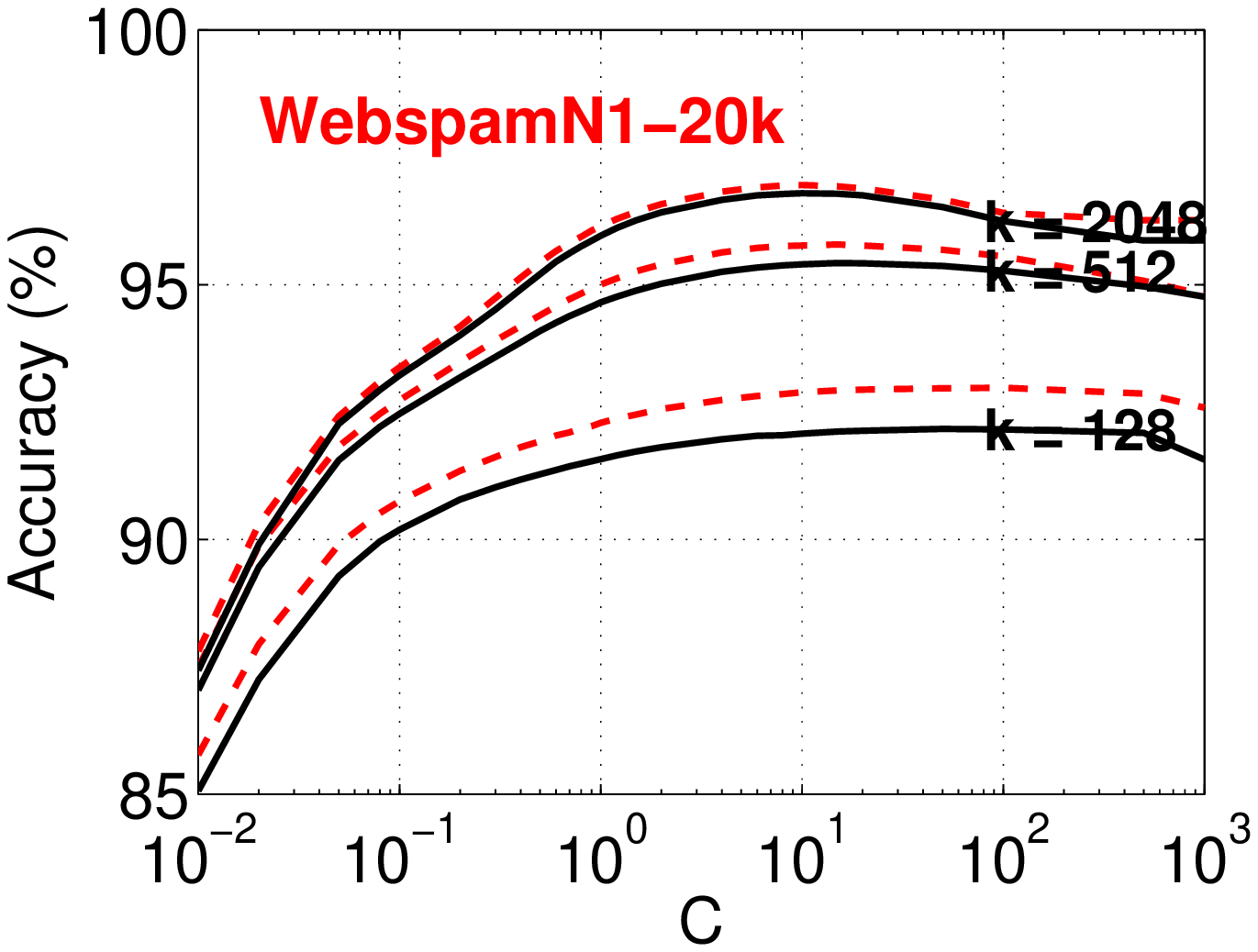}\hspace{-0.05in}
\includegraphics[width=1.65in]{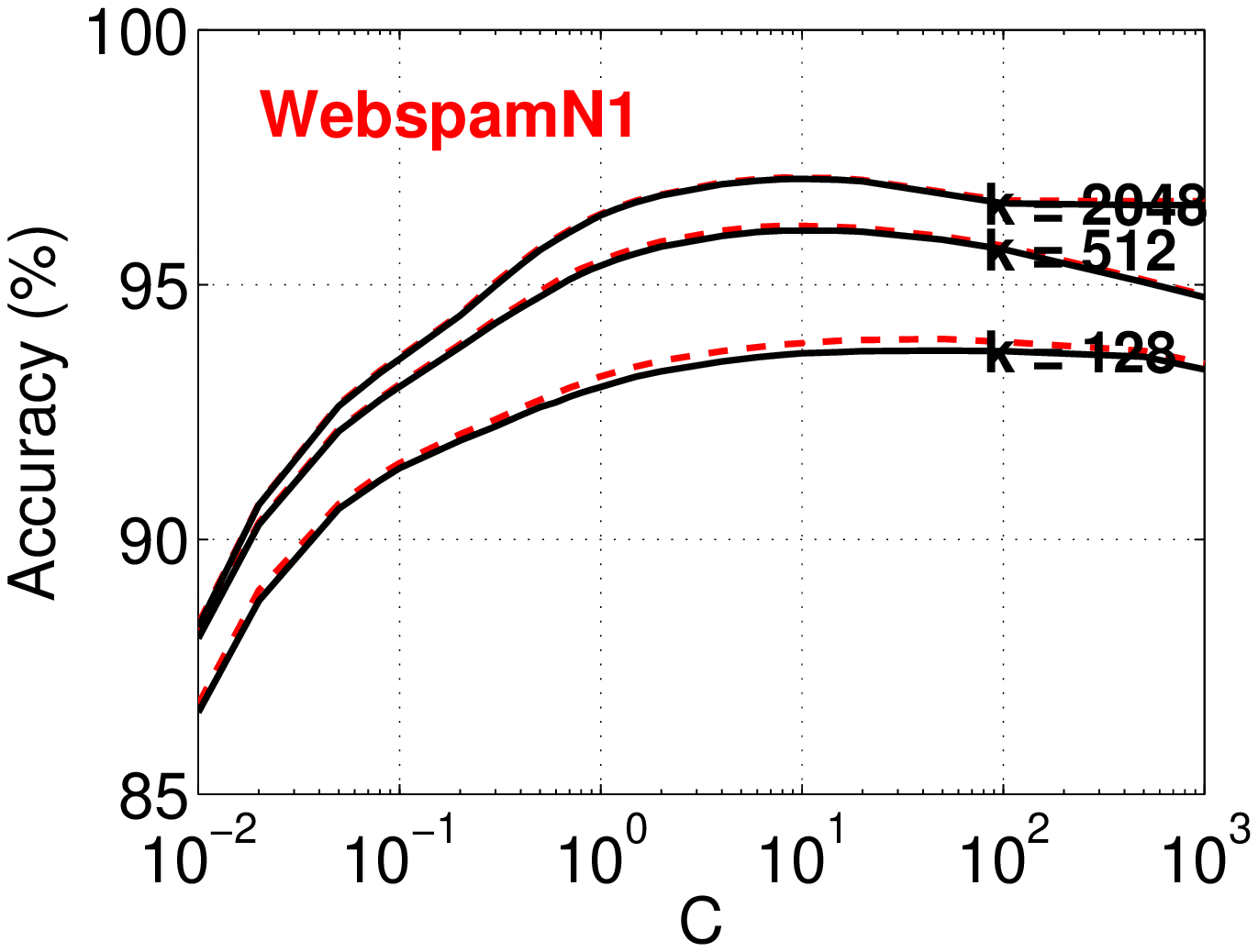}\hspace{-0.05in}
\includegraphics[width=1.65in]{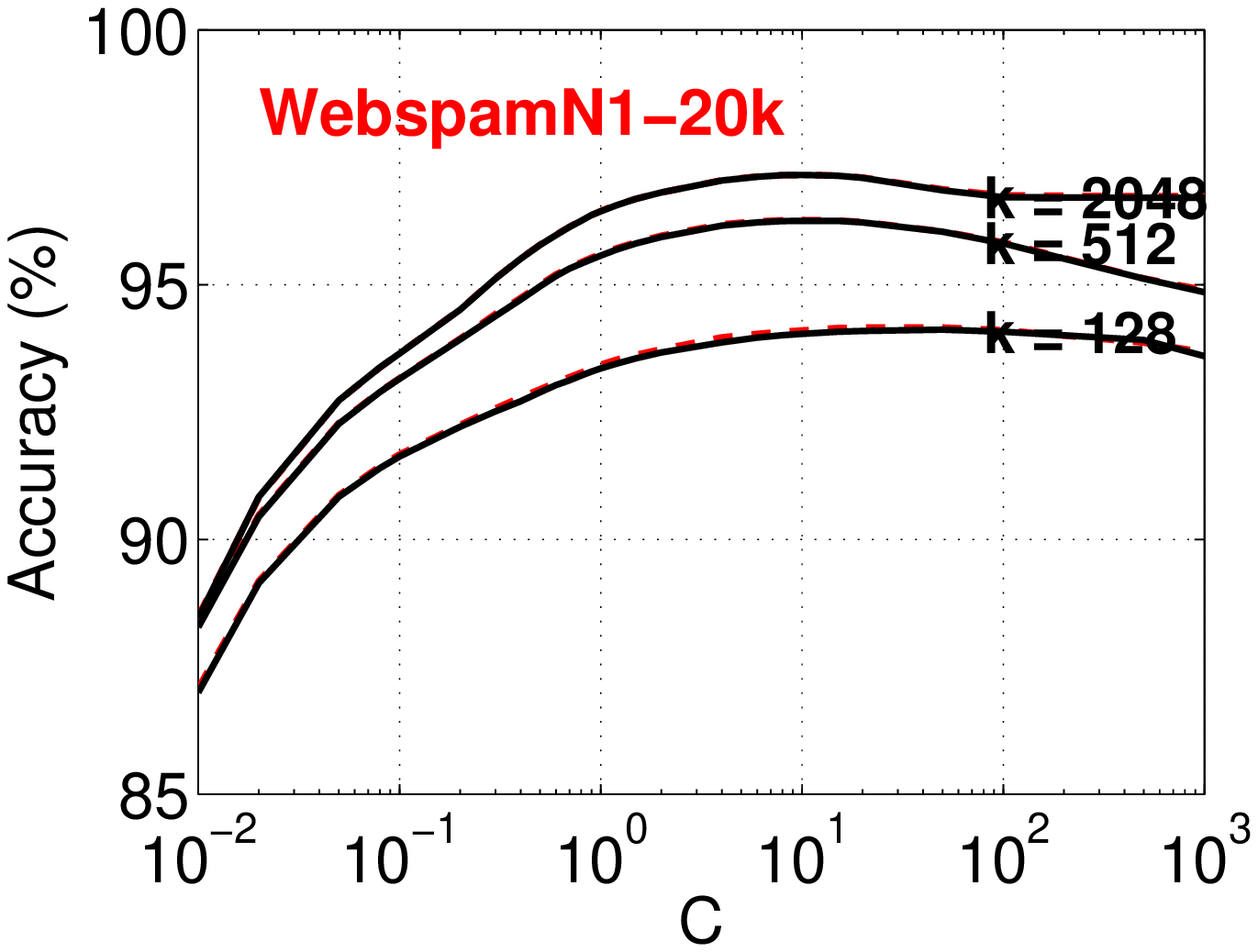}
}

\end{center}
\vspace{-0.3in}
\caption{\small Classification accuracies  by using linear SVM with 0-bit CWS (solid and black curves) and 2-bit CWS (dashed and red curves). The original CWS algorithm produces samples in the form of $(i^*,t^*)$. The 0-bit scheme  discards $t^*$ while the 2-bit scheme keeps 2 bits for each $t^*$.  From left to right, the four columns represent the results for coding $i^*$ using 1 bit, 2 bits, 4 bits, and 8 bits, respectively. In each panel, the 3 solid curves (0-bit scheme for $k=$128, 512, 2048) and the 3 dashed curves (2-bit scheme) essentially overlap especially when we use $\geq 4$ bits for coding $i^*$.  }\label{fig_HashSVM2}
\end{figure*}

There is one important detail. In practice, since the space $D$ is typically large, we often need to choose to store only a few  (say $b_i$) bits of $i^*$. In other words, after we obtain sample $(i^*, t^*)$, we will use $b_i$ bits for storing $i^*$ and 0 bit for storing $t^*$. The effective data matrix  will be $2^{b_i} \times k$  dimensions with exactly $k$ 1's in each row. In our experimental study, we always use four choices of $b_i\in\{1, 2, 4, 8\}$, corresponding to the four columns (from left to right) in Figures~\ref{fig_HashSVM} and~\ref{fig_HashSVM2}.

Figure~\ref{fig_HashSVM} presents the linear SVM experiments on a variety of datasets. In each panel, the two dashed curves (red/top and blue/bottom) correspond to the original test accuracies for the min-max kernel and the linear kernel (respectively).  In each panel, the solid curves are the results for feeding the 0-bit CWS hashed data to LIBLINEAR, for $k = 32, 64, 128, 256, 512, 1024, 2048, 4096$ (from bottom to top). For most of the datasets, we can see that the test accuracies approach the results of min-max kernels, when $k$ is large enough, especially if we use 8 bits to store each $i^*$.

Figure~\ref{fig_HashSVM2} presents an interesting study for comparing the 0-bit scheme (i.e., $b_t=0$ for $t^*$) with the   2-bit scheme (i.e., $b_t=2$ for $t^*$). We can see that once we use $\geq4$ bits for $i^*$, it makes no essential difference whether we use 0-bit or 2-bit scheme for $t^*$, i.e., the solid and dashed curves overlap.

\vspace{-0.05in}
\section{Discussion and Conclusion}

We can  view CWS as a tool for ``feature engineering'' in that it allows practitioners to  generate  data so that the inner products of the transformed  data approximate the  min-max kernel values of the original data. We can then utilize  extremely efficient and scalable (batch or online) linear methods to  equivalently train a nonlinear SVM. In other words, we  pay the price of linear learning for  nonlinear learning.

For certain applications, linear models based on the original data might be good enough. In that case, if there is a need for dimension reduction, we can use  well-known random projection methods. For many datasets (e.g.,  Table~\ref{tab_KernelSVM}), however, linear models are not sufficient and we often have to resort to  nonlinear models and computationally intensive procedures. Interestingly, min-max kernels are  suitable for many nonnegative datasets, and hence developing efficient  ways for approximating min-max kernels becomes  useful.

Our contributions  consist of three parts. Firstly, we conduct an extensive empirical study of training nonlinear kernel SVMs using min-max kernels, on a wide variety of public datasets. This study answers  why we should consider using min-max kernels instead of linear kernels. Secondly, we propose an efficient (and surprisingly simple) implementation of consistent weighted sample, called ``0-bit'' CWS, and we validate this proposal via an extensive simulation study using real text data. Finally, we show that the 0-bit CWS can be easily integrated into a linear learning system and we demonstrate, on a variety of datasets, that we can achieve the results of nonlinear SVMs by training linear SVMs.

\end{document}